\documentclass[oneside,letterpaper,11pt]{report}

\pdfoutput=1
\input{Preamble/preamble}

\graphicspath{ {Figs/} }

\begin{document}

\thispagestyle{empty}

\begin{titlepage}
    
\centering
\setstretch{1.0}
\null
\vspace*{0.25in}
\huge
Human Whole-Body Dynamics Estimation for Enhancing Physical Human-Robot Interaction \\
\normalfont\large
\Large
\vspace{2em}
Claudia Latella \\
\vspace{4em}
\begin{figure}[ht!]
    \centering
    \begin{subfigure}[b]{0.3\textwidth}
        \includegraphics[width=\textwidth]{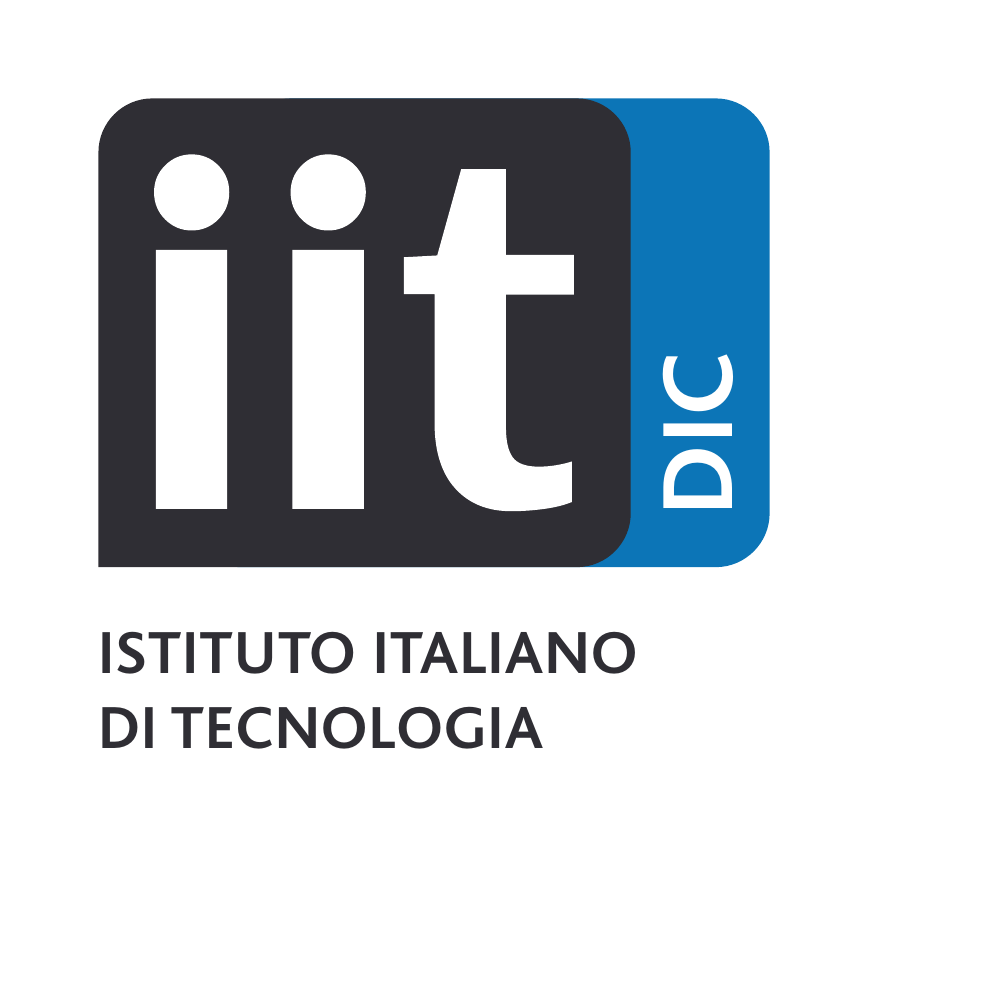}
    \end{subfigure}
    \begin{subfigure}[b]{0.28\textwidth}
        \includegraphics[width=\textwidth]{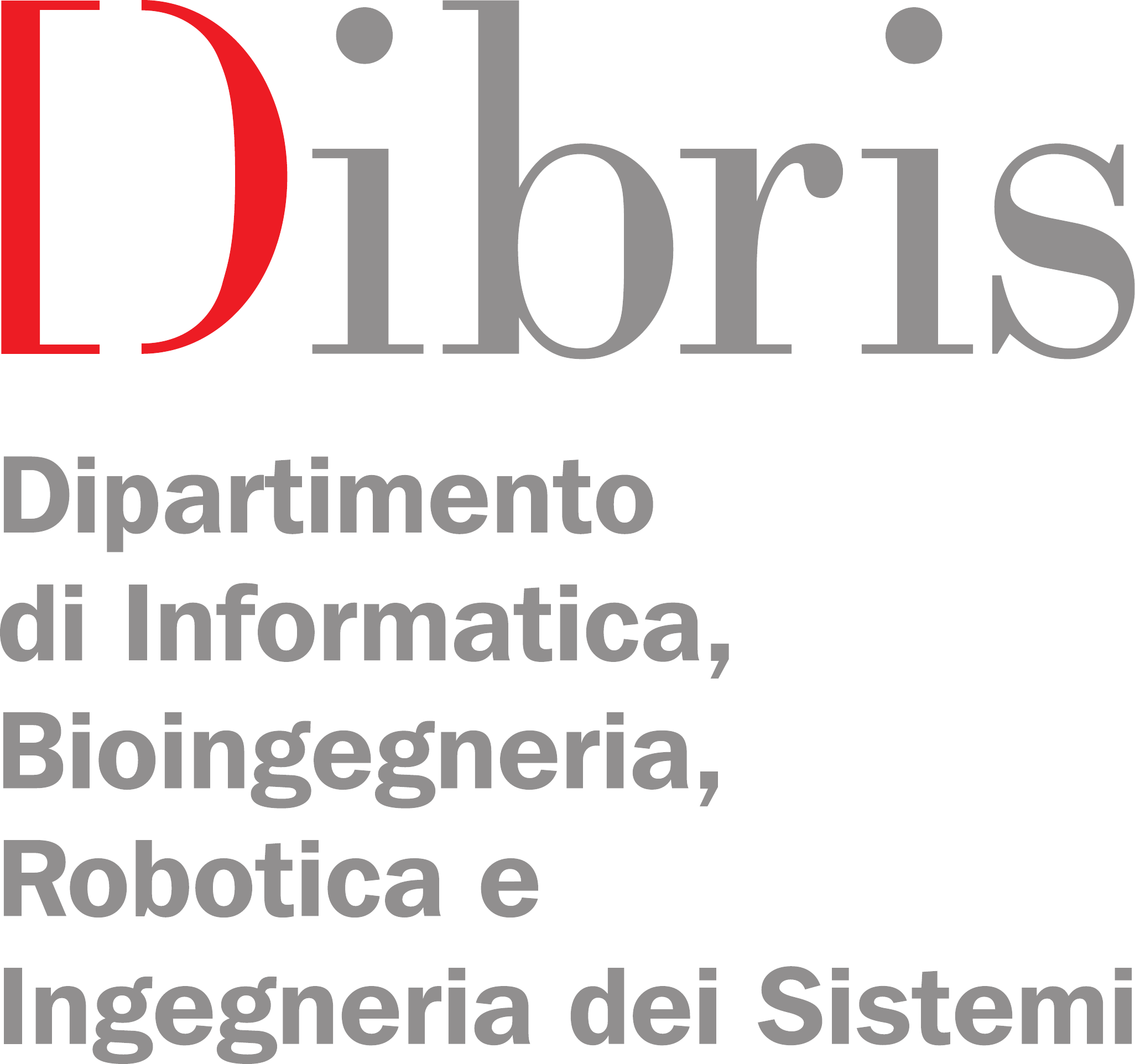}
    \end{subfigure}
\end{figure}
\vspace{2em}
Supervisor: Dott. Francesco Nori\\
\vspace{4em}
\large
Fondazione Istituto Italiano di Tecnologia\\
Dynamic Interaction Control Lab\\
Genova, Italy\\
\vspace{1em}
Dipartimento di Informatica, Bioingegneria, Robotica e Ingegneria dei Sistemi, Università di Genova
\\
\par
\end{titlepage}
\clearpage\null\pagenumbering{gobble}\newpage

\begin{abstract}

In the last two decades the scientific community has shown a great interest in
 understanding and shaping the interaction mechanisms between humans and
  robots.  The \emph{interaction} implies \emph{communication} between
   two dyadic agents and, if the type of interaction is `physical', the
     communication is represented by the set of forces exchanged during the
      interaction.
Within this context, the role of quantifying these forces becomes of pivotal
 importance for understanding the interaction mechanisms.
At the current scientific stage, classical robots are built to act \emph{for}
 humans, but the scientific demand is going towards the direction of robots
that will have to collaborate \emph{with} humans.  This will be possible by
 providing the robots with sufficient pieces of information of the agent they
  are interacting with (i.e., kinematic and dynamic model of the human).
In a modern age where humans need the help of robots $-$
 apparently in an opposite trend $-$ this thesis attempts to answer the following
 questions: \emph{``Do robots need humans? Should robots know their
  human partners?"}.
A tentative answer is provided here in the form of a novel framework for the
 simultaneous human whole-body motion tracking and dynamics estimation, in a
  real-time scenario.  The
  framework encompasses a set of body-mounted sensors and a probabilistic
   algorithm able of estimating physical quantities that in humans are not
    directly measurable (i.e., torques and internal forces).
This thesis is mainly focussed on the paramount role in retrieving the human
 dynamics estimation but straightforwardly it leaves the door open to the next
  development step: passing the human dynamics feedback to the robot
   controllers.  This step will enable the robot with the capability to observe
    and understand the human partner by generating an enhanced (intentional)
     interaction.
\end{abstract}

\clearpage\null\pagenumbering{gobble}\newpage

\chapter*{Acknowledgements}

First of all, I would like to thank Francesco (Nori) for the opportunity he
 gave me three years ago: they have been, professionally and personally
  speaking, very beautiful years.
To Francesco (Romans) for helping, supporting and tolerating me and to
 Daniele for being a great friend of mine before that of a colleague.
  To Silvio for his constant help and for what he has developed throughout
   these years: he gave us a solid foundation on which developing our work.

To the new An.Dy group.  To Luca (Tagliapietra), he was the last one to
 arrive but he is already a companion of a thousand misadventures (or ‘disagi’,
  as we prefer to call them).
  To Yeshi, our new control man, for his important contribution to the
   human-in-the-loop stuff!
And to Diego, our technician.
 But above all, to Marta and Maria, our M\&M’s: even if they are not currently
 part of our group, they collaborated in the past to build
   what the human research line is now .  For me, they will be always An.Dy
    girls!

To all my lab colleagues (of today and of the past): Gabry, Stefano,
 Aiko, Nuno, Luca (Fiorio), Jorh, Joan, Marie, Francisco, La Strama, Alessia,
  Boiello e Matte, and to the new comers!
 To the big purchases with Marianna and Angelica Rosboff.

To Dana for the opportunity she gave me to visit her labs in Waterloo and to
 meet the Waterloo guys.  To those of them that helped me with the setup
  (Jonathan, Vlad and Kevin) and to those ones that have been subject to the
   experiments.

To my loved ones for rooting for me: to my parents, my sister Katia and
 his boyfriend Silvio, my aunt Clelia, my husband's family (Mariarosa,
  Pierluigi, Giacomo, Sara and Filippo) for their support and all my friends.

And to Paolo, if I am here today I owe it to him, to that extreme fortitude to
 which he hanged up in a moment of extreme pain and difficulty.  Thanks for
  giving me this, too!

\vspace{0.8cm}
... And to those (non-scientific) filters that allow me to move forward!

\clearpage\null\pagenumbering{gobble}\newpage
\setcounter{page}{1}
\renewcommand{\thepage}{\roman{page}}%

\chapter*{Preliminaries}

The following lists represent reference tables for the entire manuscript.

\subsection*{Nomenclature}
\vspace{-0.5cm}
\rule{\textwidth}{0.3pt}
\begin{longtable}{l p{10cm}}
    \label{nomenclature_table}
$p$                     & Scalar (non-bold small letter)\\
\rowcolor{Gray}
$\bm p$                 & Vector (bold small letter)\\
$\bm A$                 & Matrix, tensor (bold capital letter)\\
\rowcolor{Gray}
$(\cdot)^{T}$           & Transpose matrix\\
$\times$                & Cross product for $6$D motion vectors\\
\rowcolor{Gray}
$\times^*$              & (Dual) cross product for $6$D force vectors\\
$|\cdot|$               & Matrix determinant\\
\rowcolor{Gray}
$\bm \skewOp(\cdot)$    & Skew-symmetric matrix\\
$\bm A$                 & Matrix, tensor (bold capital letter)\\
\rowcolor{Gray}
$\mathcal{A}$           & Coordinate reference frame (calligraphic letter)\\
$O_\mathcal{A}$         & Origin of coordinate frame $\mathcal{A}$\\
\rowcolor{Gray}
${}^\mathcal{A}\bm{p}$  & Vector expressed in frame $\mathcal{A}$\\
${}^\mathcal{A}(\cdot)_\mathcal{B}$    & Transformation operator from frame 
                        $\mathcal{B}$ to frame $\mathcal{A}$\\
\rowcolor{Gray}
$\dot {\bm{p}}$         & First-order time derivative\\
$\ddot {\bm{p}}$        & Second-order time derivative\\
\rowcolor{Gray}
$\bm {\underline p}$    & Spatial vector (underlined bold small letter)\\
$\bm{\mathrm{I}}$       & Inertia tensor\\
\rowcolor{Gray}
$\bm {\mathrm{J}}$      & Jacobian matrix\\
${\bar {\bm S}}$        & Motion freedom subspace\\
\rowcolor{Gray}
$\sum$                  & Summation operator\\
$(\cdot)^\dagger$       & Pseudoinverse of a matrix\\
\rowcolor{Gray}
$(\cdot)^{-1}$          & Inverse matrix\\
$rank(\cdot)$           & Rank of a matrix\\
\rowcolor{Gray}
$\mathcal{N}$           & Normal distribution\\
$E\llbracket\cdot\rrbracket$      & Expected value\\
\rowcolor{Gray}
$\mu$                   & Mean\\
$\bm \Sigma$, $cov \big\llbracket \cdot \big\rrbracket$   & Covariance matrix\\
\rowcolor{Gray}
$\arg \max(\cdot)$      & Maximizing argument\\
$\arg \min(\cdot)$      & Minimizing argument\\
\rowcolor{Gray}
$diag(\cdot)$           & Diagonal matrix\\
$\|\cdot\|$             & Norm\\
\rowcolor{Gray}
$Tr$                    & Trace\\
\end{longtable}

\subsection*{Abbreviations and Acronyms}
\vspace{-0.5cm}
\rule{\textwidth}{0.3pt}
\begin{longtable}{l p{10.6cm}}
    \label{abbrevationsAndAronims_table}
HRI          & Human-Robot Interaction\\
\rowcolor{Gray}
pHRI         & Physical Human-Robot Interaction\\
w.r.t.        & With respect to\\ 
\rowcolor{Gray}
$2$D         & Two-dimensional\\
DoF          & Degree of Freedom\\
\rowcolor{Gray}
ID           & Inverse Dynamics\\
IMU          & Inertial Measurement Unit\\
\rowcolor{Gray}
$3$D         & Three-dimensional\\
RGB          & Red-Green-Blue\\
\rowcolor{Gray}
$6$D         & Six-dimensional\\
CoM          & Centre of Mass\\
\rowcolor{Gray}
RNEA         & Recursive Newton-Euler Algorithm\\
URDF         & Unified Robot Description Format\\
\rowcolor{Gray}
fb           & Fixed base \\
PDF          & Probability Density Function\\
\rowcolor{Gray}
MAP          & Maximum-A-Posteriori\\
MSE          & Mean Square Error\\
\rowcolor{Gray}
MMSE         & Minimum Mean Square Error\\
LMMSE        & Linear Minimum Mean Square Error\\
\rowcolor{Gray}
GLS          & Generalized Least-Squares\\
IK           & Inverse Kinematics\\
\rowcolor{Gray}
F/T          & Force/Torque\\
RMSE         & Root Mean Square Error\\
\rowcolor{Gray}
API          & Application Programming Interface\\
YARP 		 & Yet Another Robot Platform\\
\rowcolor{Gray}
EM           & Expectation-Maximization\\
EKF          & Extended Kalman Filter\\
\end{longtable}

\tableofcontents

\clearpage\null\pagenumbering{gobble}\newpage
\setcounter{page}{1}
\renewcommand{\thepage}{\arabic{page}}%

\chapter{Introduction}  %
\label{chapter_introduction}
\begin{quotation}
\noindent\emph{
Any sufficiently advanced technology is indistinguishable from magic.
}
	\begin{flushright}
		Arthur C. Clarke
	\end{flushright}
\end{quotation}

\section{Rationale}

Human-Robot Interaction (HRI) is arousing a remarkable
 interest among the scientific community. The field, emerged in the
  early years
  of $2000$, addresses the design, the understanding and the evaluation of
   robotic systems which involve humans and robots interacting through
    cooperation.

\begin{figure}
  \centering
  \includegraphics[width=1\textwidth]{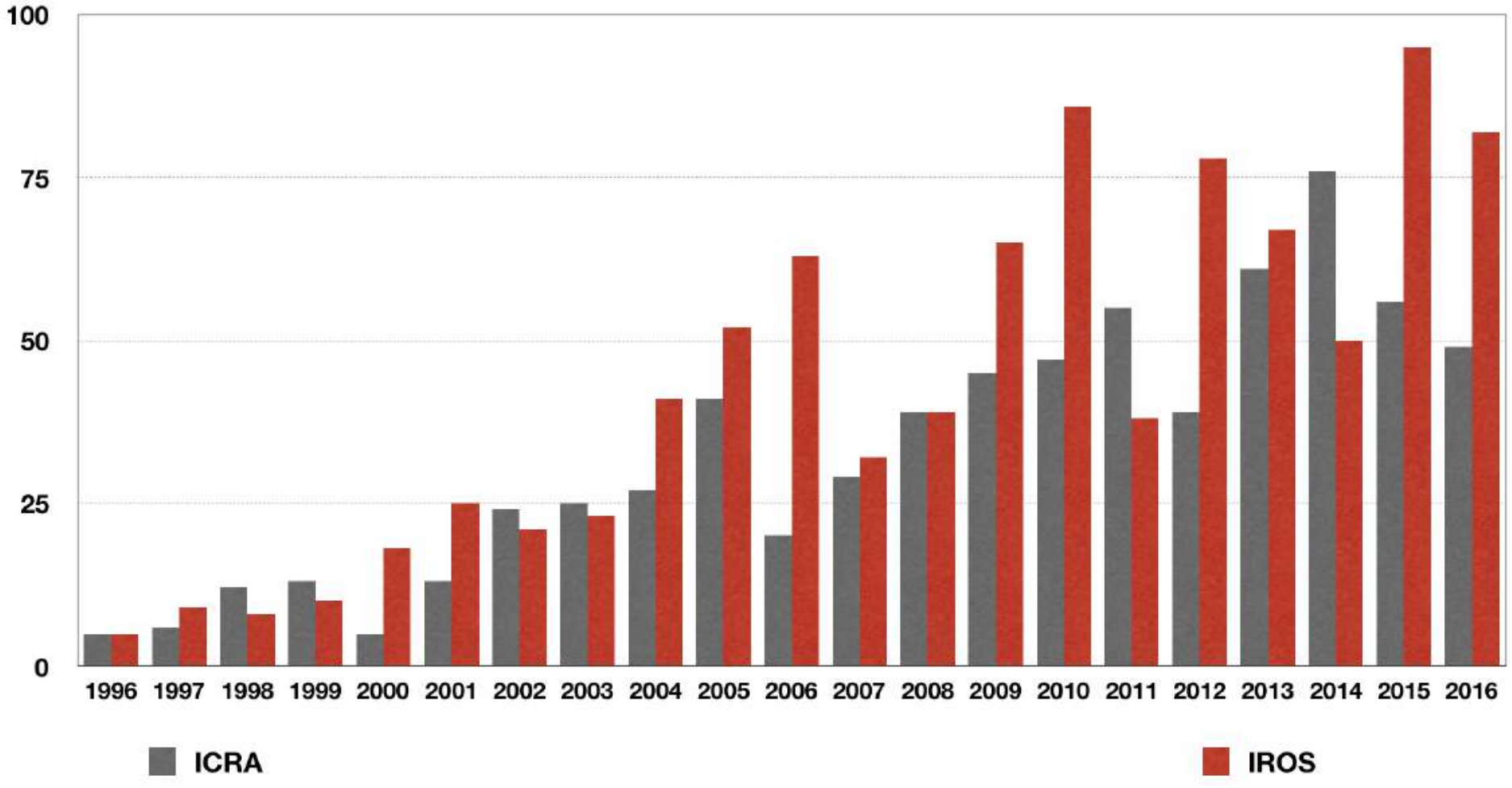}
  \caption{Number of papers on HRI topics published in the $1996$-$2016$
   decades, on the proceedings of the two major robotics conferences.}
  \label{fig:Figs_IEEE-ICRA_IROS_papersComparison}
\end{figure}

The ever-growing interest of the scientific community in studying coupled
 human-robot systems is visible in the thriving production of papers related to
  HRI topics in the last two decades.  Figure
   \ref{fig:Figs_IEEE-ICRA_IROS_papersComparison} shows the number of papers
    obtained by searching on the IEEE Explore website the keyword `human-robot'
     concerning the two main conferences of robotics: IEEE/RSJ
      International
Conference on Intelligent Robots and Systems (IROS) and IEEE International
 Conference on Robotics and Automation (ICRA).  Despite the existence of
  several fluctuations over the years, the trend has definitely increased
   straddling the two centuries (see also \citep{demircan_spotlight}).

The HRI problem is well-defined as the problem ``\emph{to understand and
 shape the interactions between one or more humans and one or more robots. [...]
  evaluating the capabilities of humans and robots, and designing the
   technologies and training that produce desirable interactions are essential
    components of HRI. Such work is inherently interdisciplinary in nature,
     requiring contributions from cognitive science, linguistics, and
      psychology; from engineering, mathematics, and computer science; and from
       human factors engineering and design}" \citep{Goodrich2007_survey}.
This definition clearly highlights the multidisciplinarity of the field and
 justifies the ever-increasing interest of the scientific community in dealing
  with it.

The word \emph{interaction} is synonym of \emph{communication}.  The taxonomy
 of an interaction is highly vast and therefore the concept of communication is
  different depending on whether it is a remotely-controlled or a proximate
   interaction.  If we consider those applications that require physical HRI
    (pHRI), this communication is not intended to be verbal (or visual)
     but it is the result of a set of forces exchanged between the interacting
      agents.

\begin{figure}[h]
    \centering
    \begin{subfigure}[b]{0.9\columnwidth}
        \includegraphics[width=\textwidth]{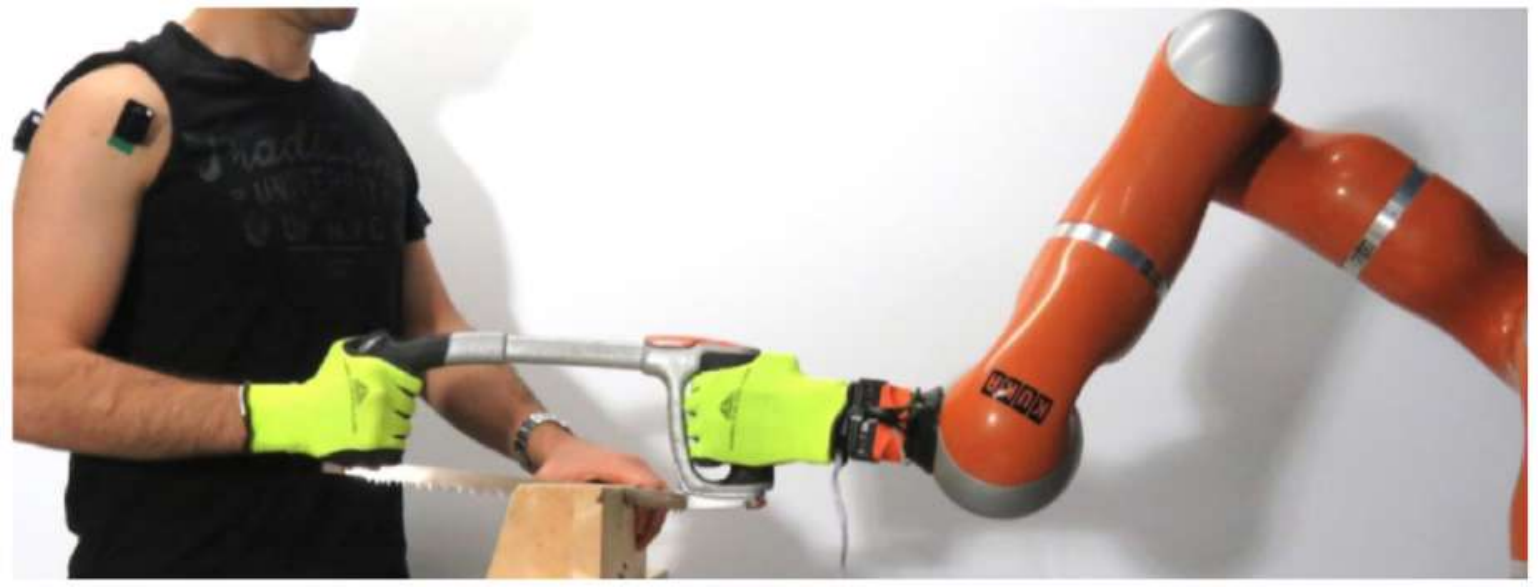}
        \caption{}
                 \label{fig:pHRI_example_pHRI_ADVR}
         \end{subfigure}
            \begin{subfigure}[b]{0.57\columnwidth}
                \includegraphics[width=\textwidth]{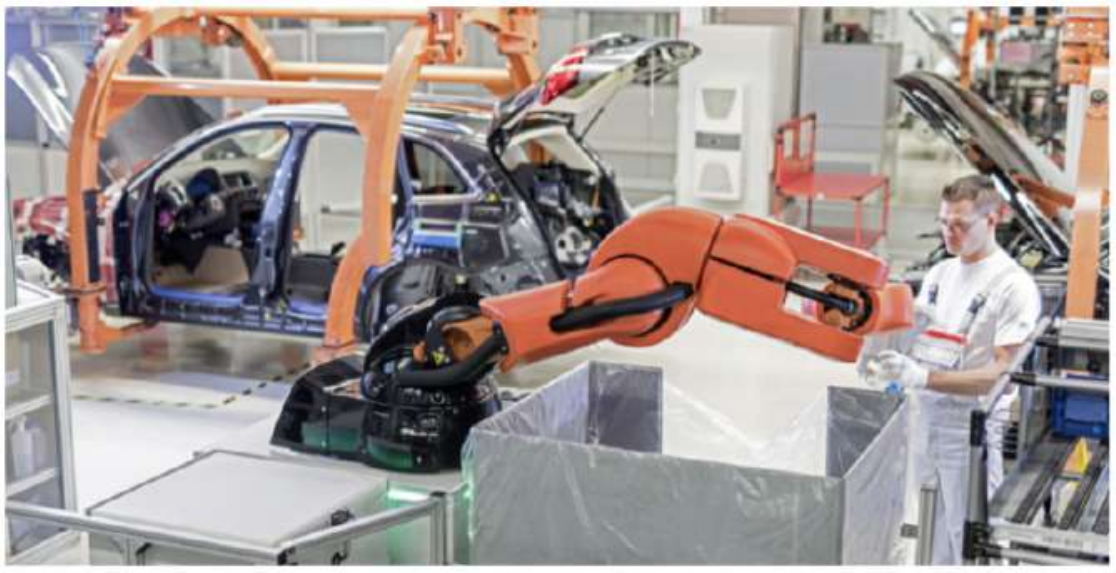} %
         \caption{}
        \label{fig:pHRI_example_automotive}
    \end{subfigure}
    \begin{subfigure}[b]{0.40\columnwidth}
        \includegraphics[width=\textwidth]{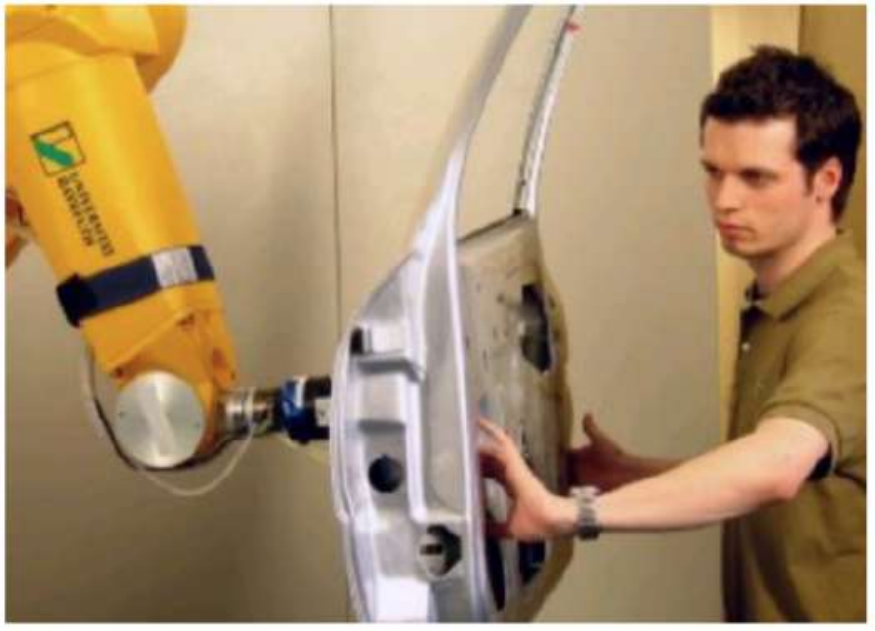}
        \caption{}
        \label{fig:pHRI_example_univBayreuth}
    \end{subfigure}
\caption{(\subref{fig:pHRI_example_pHRI_ADVR}) Human-KUKA cooperation.
 (\subref{fig:pHRI_example_automotive}) Human-robot cooperation in an
  automotive assembly line.
 (\subref{fig:pHRI_example_automotive}) An example of pHRI in a lab
  environment.}
\source{Websites of (\subref{fig:pHRI_example_pHRI_ADVR}) IIT-ADVR,
 (\subref{fig:pHRI_example_automotive}) beautomotive.be,
  (\subref{fig:pHRI_example_univBayreuth}) Universitat Bayreuth.}
  \label{pHRIscenesByWeb}
\end{figure}

Nowadays, the demand for physical robotic assistance to humans is
a widening practice.  Robotic devices are largely widespread in many fields
 for fulfilling practical, and very different, daily applications such as
  industrial manipulators for the assembly-lines context, assistive and
   rehabilitative technology in clinical scenarios.  By googling the keywords
    `human-robot physical interaction', it is easy to find on the web a huge
     amount of pictures where humans and robots are asked to share a common
      working space and to cooperate in order to accomplish a required task
      (Figure \ref{pHRIscenesByWeb}).

The modern and advanced robots are equipped with force and tactile sensors to
 measure contact forces and actuators to control them with reactive strategies.
   But physical collaboration also requires anticipatory strategies for
    predicting the posture (i.e., where the human is located with respect to
     (w.r.t.) the robot peripersonal space) and the forces applied by the
      collaborative partner.
At the current scientific stage, classical robots are built to act \emph{for}
 humans, but in order to adapt their functionality to the current technological
  demand, the new generation of robots will have to collaborate \emph{with}
   humans. The new frontiers in the pHRI field will deal with the development
    of robotic systems that will be able to react, interact and collaborate
     with people.
This implies that the robots will be endowed with the capability to control
physical collaboration through intentional interaction with humans.
To achieve an effective collaboration, the robots have to be aware of who the
human partner is (in terms of modelling), both in terms of positions (motion)
 and in terms of dynamics (contacts and exchanged forces, internal forces,
  joint torques).  However the current state of the robot knowledge in
   observing human whole-body dynamics yields to non-proficient and
    unadaptive interactions.

\begin{figure}[h]
\centering
 \includegraphics[width=.9\textwidth]{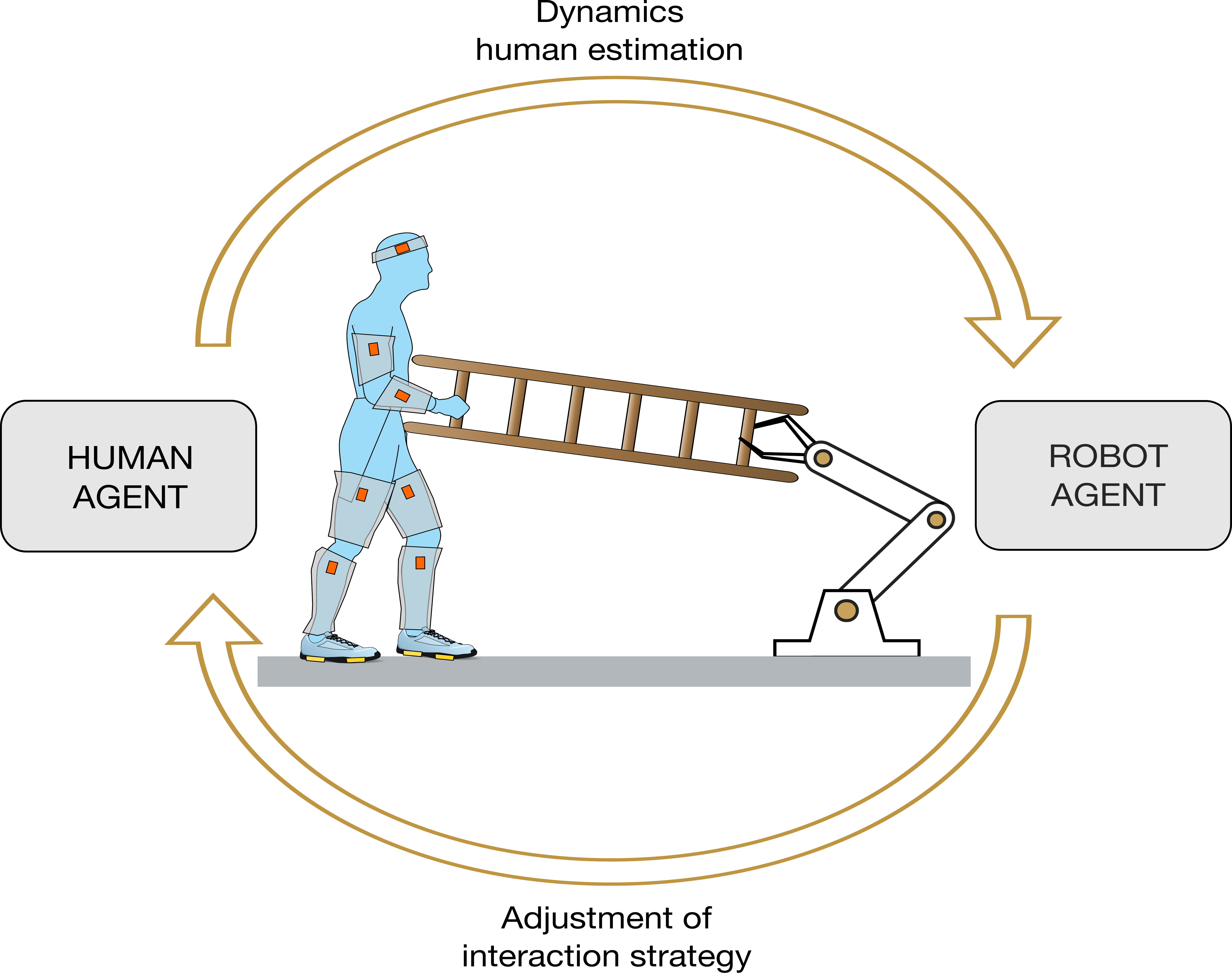}
\caption{A possible pHRI scenario: the human agent is provided with a
   wearable technology and an estimation algorithm allows to retrieve
    information about his dynamics.  By properly passing his dynamics
     feedback to the robot, the controller renders the robot compliant to the
      human-robot collaboration.}
\label{fig:Figs_pHRI_exampleWithLadder}
\end{figure}

To overcome this drawback, it is fundamental to understand what the response
 of the human body is while a physical interaction is occurring.  The
  importance in retrieving this information is exemplified in Figure
   \ref{fig:Figs_pHRI_exampleWithLadder}: once the human dynamics is completely
    known,  the human feedback may be provided to the robot controllers. As a
     straightforward consequence, the robot may accordingly adjust the strategy
      of the interaction to enhance in this way the common aim of the task.
This kind of \emph{online} awareness in the robot controllers will be the key
 to bridge the current limitation of robots in observing the human dynamics.

\vspace{1cm}

\section{State of the Art}

The aim of this Section is to provide an overview of what is the current state
 and the direction of the scientific community on the topics covered by this
  thesis.  To
 this purpose, it was necessary to split it into three main macro-areas
  concerning the state of the art on $1)$ the methodologies for estimating the
   human dynamics, $2)$ the current state of the sensor technology available on
    the market and $3)$ the research status on the pHRI topic.

\subsection{Human Body Dynamics Estimation} \label{ID_SoA}

The knowledge of the human body internal forces and joint torques is of pivotal
 importance in understanding the human dynamics.  The procedure for retrieving
  these variables starting from the measured motion quantities is referred to
   as the Inverse Dynamics (ID) estimation problem.

Given a human body model composed by interconnected segments,
the common mathematical approach iteratively solves the
Newton-Euler equations for each model segment.  The approach differs depending
 on the available input data,  following two different paths \citep{Winter1990}
  (see Appendix \ref{TopDown_BottomUp}).
\begin{itemize}
\item {A version that solves iteratively the
 Newtonian physics from the top-most to the bottom-most segment (i.e.,
  \emph{top-down} approach) by the means of only angular acceleration
   measurements.  However, the joint torques estimation is highly sensitive to
    uncertainties in acceleration data since they are retrieved by a
     double differentiation of the positions and, therefore, this
      method tends to produce noisy and non-reliable estimations
       \citep{Cappozzo1975}.}
\item{A modified version that encompasses additional measurements in the
 form of ground reaction forces.  The method, also called \emph{bottom-up},
  starts at the bottom-most segment where further boundary conditions are
   provided (typically ground reaction forces from force plates) and proceeds
    towards the top-most segment. Typically it produces more
    precise estimations but the introduction of additional data provides more
      equilibrium conditions to be satisfied: the problem arises at the
       top-most segment where the physics conditions are not satisfied anymore.}
\end{itemize}

In general, when an additional measurement is introduced into the computation, a
 further constraint has to be defined at the final segment (both for top-down or
  bottom-up methods) in order to satisfy the equations of motion.  This is due
   to the fact that the system of equations becomes overdetermined.
The solution is retrieved by discarding $6$ equations and it is strongly
 dependent on which equations are discarded, as recognized by \citep{Kuo98}
  that treats the problem of finding the joint torques
    as a least-squares problem.  The idea here is to find a set of joint torques
     that best agrees with the available data and to
      exploit the redundancy of the system when some measurements are missing
       or unreliable.  The main drawback is that it is applicable only to a $2$D
        fixed-base system.
In a subsequent study \citep{Bogert2008}, a further development of \citep{Kuo98}
 is presented, not restricted to $2$D fixed-base system but particularly
  tailored for gait analysis.  The method solves ID problems where the
  number of the unknown variables does not exceed the degrees of
  freedom (DoFs) of the system by
   employing a weighting covariance matrix derived via Monte Carlo simulation
    of measurement errors.

The least-squares approach became very popular for solving the human
 ID estimation since it is a versatile tool for addressing optimization problems
  solution.  Since it is historically well-known that ID accuracy is closely
   related to the reliability of the accelerations data \citep{Challis1996}, a
    weighted least-squares optimization approach is adopted in
     \citep{Cahouet2002} to retrieve optimum
     acceleration estimates from all the available imperfect and redundant
      measurements (positions by markers and forces by force plates).
Since the considered cost function does not contain information on the joint
 torques, the method is not so robust to errors in the joint torque
  estimation.  In \citep{Riemer2008}, a similar methodology is used but
   combining two important factors: $i)$ a cost function that minimizes the
    difference between the measured ground reaction forces and the forces
     estimated via a top-down approach; $ii)$ an optimization criterion for the
      measured joint angles. 

\subsection{Sensors Technology}

Several commercial solutions for the whole-body motion tracking are available on
 the market.  The standard approach for motion detecting is a marker-based
  technology for in-lab applications produced by \cite{vicon_site} or
  \cite{motionAnalysis_site}.  A more recent approach is to
  supplement or even replace marker-based technologies in several
   applications with wearable markerless technologies suitable for outdoor
    motion capturing produced by \cite{xsens_site} or 
    \cite{noraxon_site}, where markers are replaced by body-mounted Inertial
     Measurement Units (IMUs).
Another kind of solution is provided by the Microsoft \cite{kinect_site} depth
 camera system  which allows markerless low-cost whole-body motion
  tracking for indoor applications \citep{zhang2012microsoft}.

In biomechanics, soft wearable and stretchable systems have been proposed for
 measuring human body motion \citep{Menguc2013} \citep{Menguc2014}, as well as
  in health activity monitoring \citep{Paradiso2006} \citep{Gallego2011}.
    Combinations of different
   technologies have also been used for detecting human motion: in
    \citep{Wei2010},  a video-based motion technique has been adopted for
     capturing realistic human motion from video sequences.

Although these existing technologies provide a high level of accuracy in
 computing motion quantities, they have several limitations in measuring
  real-time dynamics quantities (i.e., forces and torques).  A key problem lies
  in the fact that motion capture methods typically employ only kinematic
   measurement modalities (positions, velocities and accelerations)
     \citep{Bonnet2013} and do not include sufficient information on the
      dynamics of the human movements.

The importance of including and exploiting dynamics information is a crucial
 point in several research areas such as ergonomics for industrial scenarios,
  rehabilitation monitoring, for developing prosthetic devices and exoskeleton
   systems or in pHRI scenarios.
For these reasons, the whole-body force tracking is not a new challenge for the
 scientific community, but the topic has been seldom explored \emph{in situ}
  and even more rarely analyzed in real-time modality (this step
  requires the development of estimation algorithms to be coupled to the
   sensors).

To overcome the drawbacks related to the in-lab force tracking acquisitions
 (usually attained through the combination of a motion capture system with
  commercial force plates),
several solutions of pressure sensing insoles are available on the market, such
 as \cite{moticon_site} or \cite{tekscan_site} used in
  \citep{Zhang2014}.  However, prototypes of sensing shoes have been proposed in
   the recent scientific scenario: a prototype of wearable shoes developed by
    Xsens \citep{xsens_shoesSchepers2007} and a sensory system for the foot
     pressure evaluation tailored for exoskeletons applications
      \citep{exoshoe2016}.

\subsection{Physical Human-Robot Interaction}

 Most of the pHRI studies take inspiration from the intrinsic
  behaviour of the human nature: the \emph{mutual adaptive nature} that
   automatically occurs when two humans are cooperating together to accomplish
    a common task.  To this purpose, the importance of understanding the human
     dynamics goes without saying and it is a crucial aspect of current
      scientific research.

Based on the pioneering study on the minimum-jerk model in human manipulation
 tasks \citep{Flash1985}, a method based on the minimization of the jerk is
  used as a suitable approximation for estimating the human partner motions in
   \citep{Maeda2001} during a human-robot cooperative manipulation.
Here the authors' attempt is to incorporate human characteristics into the
  control strategy of the robot.  The weakness in this type of approach lies
   in the pre-determination of the task and in the role that the robot has to
    play in the task execution. Furthermore, the minimum-jerk model
     reliability decreases considerably if the human partner decides to apply
      non-scheduled trajectory changes during the task \citep{Miossec2009}.

Another suitable route for pHRI is the \emph{imitation learning} approach,
 justified by a wide variety of computational algorithms that have been
  developed by the scientific community in the past decades (see an exhaustive
   overview in \citep{Schaal2003}).
This approach yields to the \emph{teacher-student} (or more pragmatic
 master-slave) interaction concept and consists in the following steps:
\begin{itemize}
\item \emph{Collection of human motions}: movements of a human agent (i.e., the
 teacher in the imitation process) are retrieved with motion capture techniques
  and clustered in motion databases.  Some publicly available examples are
   \citep{Guerra2011} \citep{Kuehne2011} \citep{Wojtusch2015}
    \citep{Tamim2015database}.
\item \emph{Classification for model abstraction}: human motions are classified
 via motion segmentation in order to abstract motion models.  The model
  abstraction is feasible by postulating that the human movements can be
   decomposed into a set of primitive actions, termed \emph{motion primitives}.
\item \emph{Models retargeting}: human motion models are mapped into a robot
 platforms, i.e., the student in the imitation process.  The retargeting
  implies the development of robot control algorithms capable of learning
   motion models
   from these primitives  \citep{Amor2014} \citep{Tamim2014}, \citep{Tamim2016}
    in order to emulate human-like movements.
\end{itemize}
Since the primitives have to be perceived and interpreted by a robot, a common
 `language' has to be mandatorily adopted.  The abstraction implies, therefore,
  to define the primitives in terms of positions, velocities and accelerations.
In a recent work \citep{Borras2017taxonomy}, the collection of human motions is
 performed when the human is interacting with the external environment and
  exploiting multi-contact support to enhance his stability.  Here, the
   classical characterization of a primitive (through the kinematic
    above-mentioned variables) is coupled with the important information on how
     the human is maintaining his stability.

In general, although in the imitation approach the task pre-determination does
 not represent a drawback anymore, a relevant problem (termed
  \emph{correspondence problem}) concernes the comparability of the
   models (teacher vs. student).  Human motion primitives are exploitable only
    if their knowledge could be properly mapped into a compatible robot
     kinematic/dynamic model.
This inevitably narrows down the potential student-role candidate to
 anthropomorphic robots and it requires an accurate analysis on the imitation
  mechanism assumptions.

\section{Contribution}

The thesis contribution lies in the development of a novel wearable technology
 for the simultaneous force and motion human tracking.  The attempt of this work
  is to bridge the current state-of-the-art gap on human dynamics analysis
   thanks to
\begin{itemize}
    \item{a combination of different sensors (body-mounted IMUs, force
     sensing)}
    \item{a whole-body force estimation algorithm, originally developed for the
     humanoid iCub robot \citep{NoriNav2015} and tailored here for the human
      body.}
\end{itemize}

The novelty of the approach consists in framing the dynamics computation in a
 probabilistic Gaussian framework when redundant measurements occur.
 In this way, sensors play an \emph{active} role in the computation since the
  classical boundary conditions in the recursive Newton-Euler algorithm are
   replaced with the sensors readings.
 The estimation algorithm solves a sort of weighted least-squares problem where
  the weights are given by the dynamic model and the sensors.

\subsection{Publications}

Here following the list of the related publications (at the time of writing)
 produced during my Ph.D. course period:
\begin{itemize}
\item{Romano, F.; Nava, G.; Azad, M.; Camernik, J.; Dafarra, S.; Dermy, O.;
 Latella, C.; Lazzaroni, M; Lober, R.; Lorenzini, M.; Pucci, D.; Sigaud, O.;
Traversaro, S.; Babic, J.; Ivaldi, S.; Mistry, M.; Padois, V.;  Nori, F. ``The
 CoDyCo Project achievements and beyond: Towards Human Aware Whole-body
  Controllers for Physical Human Robot Interaction", Special issue on Human
   Cooperative Wearable Robotic Systems in IEEE Robotics and Automation
    Letters, $3$:$516$-$523$, November 2017.
    (RA-L), $2017$, PP, $99$. doi: 10.1109/LRA.2017.2768126,
     \url{http://ieeexplore.ieee.org/document/8093992}.}
\item{Latella, C.; Kuppuswamy, N.; Romano, F.; Traversaro, S.; Nori,
 F. ``Whole-Body Human Inverse Dynamics with Distributed Micro-Accelerometers,
  Gyros and Force Sensing", Sensors $2016$, $16$, $727$. doi: 10.3390/s16050727,
   \url{http://www.mdpi.com/1424-8220/16/5/727}.}
\item{Latella, C.; Kuppuswamy, N.; Nori,F. ``WearDY: Wearable dynamics. A
 prototype for human whole-body force and motion estimation", AIP Conference
  Proceedings $1749$, $020011$ ($2016$). doi:
   \url{http://dx.doi.org/10.1063/1.4954494}.}
\end{itemize}

\section{Technological Outcome} 

The framework proposed in this thesis is the core of the Horizon-$2020$
 European project named An.Dy - Advancing
  Anticipatory  Behaviors in Dyadic Human-Robot Collaboration
   (H$2020$-ICT-$2016$-$2017$, No.$731540$) \citep{andy}.
The project aims at advancing the current state of the art in the pHRI field by
 providing a novel wearable technology that measures human whole-body dynamics
  and, in turns, endows robots with the ability to be more aware of the
   human partner they are interacting with.  An.Dy will enhance the intentional
    cooperation between humans and robots.  The results achieved in this thesis
     represent the entry point and the basis on which An.Dy will lay its
      foundations in the next years.

\section{Thesis Outline}

Hereafter, a brief summary of the topics of the thesis is provided.  Each
 chapter briefly introduces the topic and starts (if needed) with
  the notation useful for guiding the reader into its comprehension.
 
Chapter \ref{chapter_rigidMultiBodySystems} introduces a complete overview of
 the kinematics and the dynamics of a rigid body and, then, of a system of
  rigid bodies.

Given the notions from the previous chapter, Chapter
 \ref{Chapter_human_modelling} describes the human whole-body modelling as a
  system of articulated rigid bodies.

In Chapter \ref{Chapter_estimation_problem}, the human dynamics estimation
 problem is presented and it is provided a solution in a probabilistic domain.

Chapter \ref{chapter_implementationANDvalidation} describes the Matlab software
 implementation for the offline validation of the framework.  It shows the
  sensing technologies used for the algorithm and introduces also a new
   wearable prototype developed at Istituto Italiano di Tecnologia for
    measuring ground reaction forces.

The main contribution of Chapter \ref{chapter_towardsTheRealTime} is the design
 of a YARP tool for the real-time monitoring of the human dynamics.

Chapter \ref{chapter_theHumanInTheLoop} discusses the possibility to extend the
 human dynamics estimation fremework to a new framework where a physical interaction with the iCub robot is encompassed.

Chapter \ref{chapter_conclusion} concludes the thesis with a general overview
 on the obtained preliminary results and analyzes possible improvements as
  forthcoming works. %

\chapter{Rigid Multi-Body System}  %
\label{chapter_rigidMultiBodySystems}

\begin{quotation}
\noindent\emph{
Nothing happens until something moves.
}
    \begin{flushright}
        Albert Einstein
    \end{flushright}
\end{quotation}

\noindent
In this Chapter a complete overview of the kinematics and the
 dynamics of a rigid multi-body system is introduced.  For achieving this
  objective, a detailed description of the `sub-units' that describe a
   multi-body system is mandatorily needed.  The sub-units are the bodies and
    the joints which a system is made of.  The reader will be provided with
     some tools that describe the kinematics and the dynamics of a rigid body
      and, only then, with the multi-body system notions.

 \section{Notation}
 
Hereafter some important math preliminaries adopted through the thesis.

\begin{itemize}
\item Let $\mathbb{R}$ be the set of real numbers.  $\bm p \in
 \mathbb{R}^n$ is a $n$-dimensional column vector of real numbers.  Let $p$
  denote a scalar quantity.
\item A matrix of dimensions $m \times n$ $\in$ $\mathbb{R}^{m \times n}$.
\item Let $\bm{1}_n$ be the identity matrix of dimension $n$ $\in
 \mathbb{R}^{n \times n}$.
\item Let $\bm{0}_n$ be the zero matrix $\in \mathbb{R}^{n \times n}$ and
 $\bm{0}_{m \times n} \in \mathbb{R}^{m \times n}$.
\item {Given two generic vectors $\bm p$, $\bm u \in \mathbb{R}^3$, their inner
 product is denoted with $\bm p^T \bm u$.  The cross product $\in \mathbb{R}^3$}
 is denoted as $\bm p{\times}~ \bm u$, where

\begin{equation}\label{crossProcuct_inR3}
    \bm p{\times} :=
\begin{bmatrix}
   0    &  -p_z  &   p_y    \\
  p_z   &   0    &  -p_x     \\
 -p_y   &   p_x  &   0
\end{bmatrix} \in \mathbb{R}^{3\times 3}~.
\end{equation}
\item Given a vector $\bm{p}$ and a reference frame $\mathcal{B}$, the notation
 ${}^\mathcal{B}\bm{p}$ denotes the vector $\bm{p}$ expressed in $\mathcal{B}$.
\item Let $SO(3)$ be the set of $\mathbb{R}^{3 \times 3}$ orthogonal matrices
 with determinant equal to one, such that
\begin{equation}
    SO(3) := \{\bm R \in \mathbb{R}^{3 \times 3} |~~ \bm R^T \bm R = \bm 1_3
     ~,~~~|\bm R| = 1 \}~. 
\end{equation}
\item Let $so(3)$ be the set of the skew-symmetric matrices $\in \mathbb{R}^{3
 \times 3}$, such that
\begin{equation}
so(3) :=  \{ \bm\skewOp \in \mathbb{R}^{3 \times 3}  |~~ {\bm \skewOp}^T = -\bm
 \skewOp \}~.
\end{equation}

\item Let the set $SE(3)$ be defined as
\begin{equation}
SE(3) :=  \Big\{
\begin{bmatrix}
\bm R            & \bm p \\
\bm 0_{1\times3} & 1
\end{bmatrix} \in \mathbb{R}^{4 \times 4} |~~
\bm R \in SO(3)~,~~~\bm p \in \mathbb{R}^3
\Big\}~.
\end{equation}
\end{itemize}

\section{The Non-Deformability Assumption} \label{non_def_section}

In the physics formalism, a rigid body is represented as a body which does not
 deform under the application of an external force. This means that the
  distance between two generic points in the body does not vary when a force
   is exerted on it, i.e., the body is \emph{non-deformable}. 
In real life, this utopian condition does not exist. There are situations in
 which the body deforms depending , for example, on the magnitude or the point
  of application of a force.  However, under the hypothesis that the
   deformation is small, i.e., \emph{negligible}, the rigid body assumption
    allows to describe the dynamics of every physical object. 

\section{Rigid Body Kinematics}

The kinematics of a rigid body concerns the mechanism of its motion without
 considering the forces and torques that caused it. 
Among the robotics community, the most influential description of the rigid
 body kinematics \citep{handbook_robotics} \citep{Siciliano2009}
  is through its \emph{position} and its \emph{orientation} w.r.t. a known
   coordinate reference frame. The body velocity and
    acceleration depend on this representation.
 
\subsection{Pose Representation}

The rigid body kinematics can be described by the \emph{pose} representation,
 i.e., the position and orientation w.r.t. a 
  reference frame.  Let the frame $\mathcal{I}$ (i.e., `inertial')
   be a reference frame with the origin in a 3D point
    $O_\mathcal{I}$.
Let $\mathcal{B}$  (i.e., `body') be another frame with the origin
 in the 3D point $O_\mathcal{B}$ attached to the rigid body.  Consider now 
 that $\mathcal{B}$ moves w.r.t. $\mathcal{I}$.
Every type of motion of the body (translation, rotation or a combination of the
 two) will produce a variation in its pose.  This means that the pose
  representation becomes a powerful tool for describing the entire kinematics
   of the rigid body.

The coordinates of the point $O_\mathcal{B}$ w.r.t. $\mathcal{I}$ are
 clustered in the position vector ${}^\mathcal{I}\bm{o}_\mathcal{B}$ whose
  origins in $O_\mathcal{I}$ and points to
   $O_\mathcal{B}$, such that
\begin{equation}\label{coordinate_vector}
    {}^\mathcal{I}\bm{o}_\mathcal{B} = {\begin{bmatrix}
         {}^\mathcal{I} {o_x}\\
         {}^\mathcal{I} {o_y}\\
         {}^\mathcal{I} {o_z}
        \end{bmatrix}}_\mathcal{B} \in \mathbb{R}^3~.
    \end{equation}

\subsubsection{The Rotation Matrix}

\begin{figure}[H]
   \centering
     \includegraphics[width=.7\textwidth]{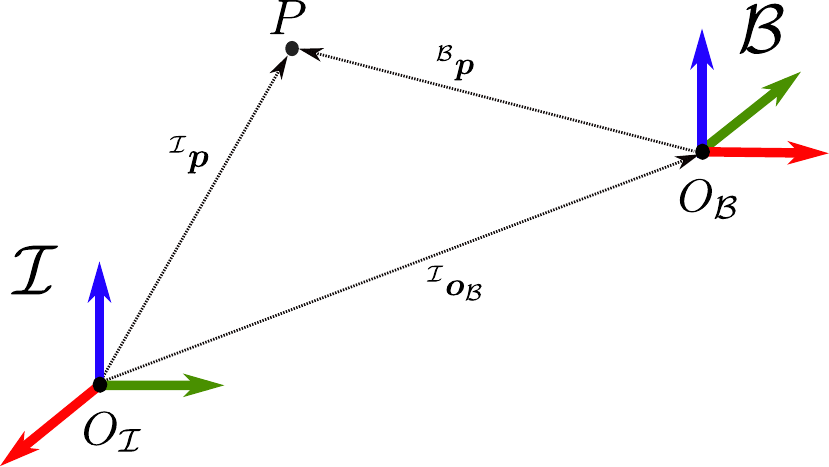}
     \caption{Standard representation of a point P in two different coordinate
      frames.  The figure introduces the RGB (Red-Green-Blue) convention for
       ${x}$-${y}$-${z}$ axes.}
   \label{fig:Pose_representation}
 \end{figure}

The orientation of $\mathcal{B}$ w.r.t. $\mathcal{I}$ is described by a
rotation matrix $\bm {R} \in SO(3)$, regardless of the positions of the origins
 $O_\mathcal{I}$ and $O_\mathcal{B}$.
 Additionally, the rotation matrix can be seen as a
  tool for operating coordinates transformation between two frames: given a
   point $P$ (in Figure \ref{fig:Pose_representation}) and its coordinates
    vector $\bm p$ expressed in $\mathcal{B}$, the rotation matrix
     ${}^\mathcal{I} \bm
     {R}_\mathcal{B}$ transforms its coordinates from the frame $\mathcal{B}$
      to the frame $\mathcal{I}$, such that
\begin{equation}\label{rotation_transform}
    {}^\mathcal{I} \bm p = {}^\mathcal{I}
     \bm {R}_\mathcal{B}{}^\mathcal{B} \bm p~.
\end{equation}

\subsubsection{The Homogeneous Transformation}

The formalism that best summarizes the pose of a rigid body is the homogeneous
 transformation matrix $\bm H \in SE(3)$.  Let $\bm {}^\mathcal{I} \tilde {\bm
  p}$, $\bm {}^\mathcal{B} \tilde {\bm p}$ be two vectors $\in \mathbb{R}^4$,
   i.e., $\bm {}^\mathcal{I} \tilde {\bm p} = [\bm {}^\mathcal{I} {\bm
    p}~~~1]^T$, $\bm {}^\mathcal{B} \tilde {\bm p} = [\bm {}^\mathcal{B} {\bm
     p}~~~1]^T$, such that
\begin{equation}\label{equation_with_H}
{}^\mathcal{I} \tilde {\bm p} = {}^\mathcal{I} \bm {H}
    _\mathcal{B}{}^\mathcal{B} \tilde {\bm p}~,
\end{equation}
being
\begin{equation} \label{homogeneous_matrix}
{}^\mathcal{I} \bm {H}_\mathcal{B} =
\begin{bmatrix}
    {}^\mathcal{I} \bm {R}_\mathcal{B} & {}^\mathcal{I}\bm{o}_\mathcal{B}\\
    \bm 0_{1\times3}                   & 1
    \end{bmatrix}~.
\end{equation}
The Equation \eqref{homogeneous_matrix} is a compact way to represent the
 position along with the rotational component of the motion. If $O_\mathcal{B}
  \equiv O_\mathcal{I}$ (i.e., null position vector
   ${}^\mathcal{I}\bm{o}_\mathcal{B}$), \eqref{equation_with_H} falls into the
    pure rotational case \eqref{rotation_transform}. If the frame $\mathcal{B}$
     is not rotated w.r.t. $\mathcal{I}$ (i.e., the rotation matrix
      ${}^\mathcal{I} \bm {R}_\mathcal{B} = \bm 1_3$) thus
       \eqref{equation_with_H} becomes ${}^\mathcal{I} {\bm p} =
    {}^\mathcal{B} {\bm p} + {}^\mathcal{I}\bm{o}_\mathcal{B}$.
 
It is worth noting that, throughout the thesis, the reference frames are shown
using a RGB (Red-Green-Blue) convention for ${x}$-${y}$-${z}$ axes, 
respectively.

\subsection{The Derivative of a Rotation Matrix}

In this framework, the derivative of the rotation matrix raises with particular
 emphasis since it allows to describe the velocity and the acceleration
 of a rigid body.  Even if not explicitly defined in the text, consider
  $\bm R$ a time-varying matrix.  It is shown (and
   the reader is addressed to \citep{Siciliano2009}, Section 3.1.1, for a
    detailed description) that there is a straightforward relation between $\bm
     R$ and the skew-symmetric operator $\bm \skewOp \in so(3)$, such that
\begin{equation}\label{def_skewOp}
    \bm \skewOp = \dot{\bm R} \bm R^T~.
\end{equation}

\subsubsection{Velocity}

\begin{figure}[H]
  \centering
    \includegraphics[width=.72\textwidth]{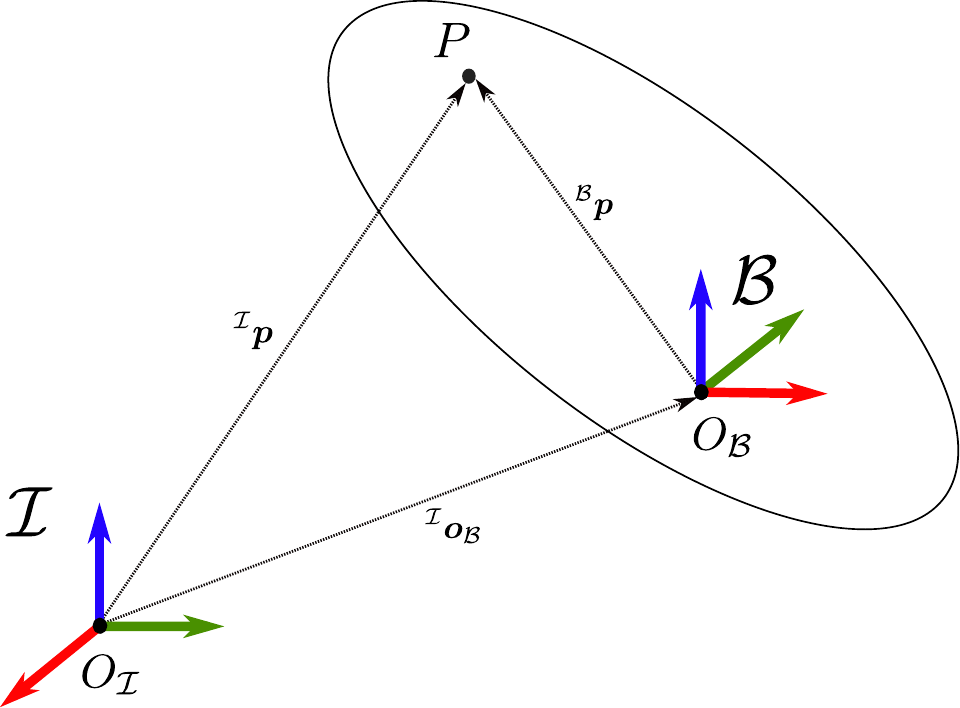}
  \caption{Rigid body with frame $\mathcal{B}$ and point P that is moving
   w.r.t. frame $\mathcal{I}$.}
  \label{fig:ref_frame_movingBody}
\end{figure}

Given an inertial frame $\mathcal{I}$, a body reference frame $\mathcal{B}$ and
 a point P of the body (Figure \ref{fig:ref_frame_movingBody}), for
  \eqref{equation_with_H} we can define the following equation
\begin{equation} \label{pos_p}
    {}^\mathcal{I} {\bm p} = {}^\mathcal{I}\bm{o}_\mathcal{B} + {}^\mathcal{I}
     \bm {R}_\mathcal{B}{}^\mathcal{B} {\bm p}~.
\end{equation}
The velocity of P w.r.t. $\mathcal{I}$ can be obtained by using the first-order
 time derivative of \eqref{pos_p}, such that
\begin{equation} \label{vel_pdot_step1}
    {}^\mathcal{I} \dot{\bm p} = \frac{d}{dt} \Big({}^\mathcal{I} {\bm
     p}\Big) ={}^\mathcal{I}\dot{\bm{o}}_\mathcal{B} +
     {}^\mathcal{I}\dot{\bm {R}}_\mathcal{B}{}^\mathcal{B} {\bm p}~.
\end{equation}
It is worth noting that ${}^\mathcal{B}\bm p$ is not a time-varying quantity as
 a direct consequence of the non-deformability assumption of the body (see
  Section \ref{non_def_section}). Equation \eqref{def_skewOp} yields to
   $\dot{\bm R} = \bm \skewOp \bm R$, thus in \eqref{vel_pdot_step1}
\begin{equation} \label{vel_pdot_step2}
    {}^\mathcal{I} \dot{\bm p} ={}^\mathcal{I}\dot{\bm{o}}_\mathcal{B} +
      \bm \skewOp {}^\mathcal{I}{\bm {R}}_\mathcal{B} {}^\mathcal{B} {\bm p}~.
\end{equation}
If $\bm \omega_B$ is the angular velocity of the frame $\mathcal{B}$ w.r.t.
 $\mathcal{I}$, thus the above equation can be written as
\begin{equation} \label{vel_pdot_step3}
    {}^\mathcal{I} \dot{\bm p} = {}^\mathcal{I}\dot{\bm{o}}_\mathcal{B} +
      \bm \omega_\mathcal{B}{\times}~{}^\mathcal{I}{\bm {R}}_\mathcal{B}
       {}^\mathcal{B} {\bm p}~.
\end{equation}
This implies that $\bm \skewOp$ is actually the skew-symmetric matrix for the
 angular acceleration, such that
\begin{equation}\label{skew_omega}
\bm \skewOp(\bm \omega_\mathcal{B}) := 
\bm \omega_\mathcal{B}{\times}~ = 
\begin{bmatrix}
   0        &  -\omega_z  &   \omega_y    \\
\omega_z    &      0      &  -\omega_x    \\
-\omega_y   &   \omega_x  &      0 
\end{bmatrix}_\mathcal{B} \in so(3)~.
\end{equation}

\subsubsection{Acceleration}

Similarly, the acceleration of P w.r.t. $\mathcal{I}$ can be obtained by using
 the second-order time derivative of \eqref{pos_p}, such that
\begin{eqnarray} \label{acc_ddp} \notag
    {}^\mathcal{I} \ddot{\bm p} &=&
    \frac{d^{2}}{d^{2}t} \Big({}^\mathcal{I} {{\bm p}}\Big) 
     = {}^\mathcal{I}\ddot{\bm{o}}_\mathcal{B} +
      \dot {\bm \omega}_\mathcal{B}{\times}~{}^\mathcal{I}{\bm {R}}_\mathcal{B}
       {}^\mathcal{B} {\bm p} + \bm \omega_\mathcal{B}{\times}~{}^\mathcal{I}{\dot {\bm R}}_\mathcal{B}
        {}^\mathcal{B} {\bm p} \\
      &=& {}^\mathcal{I}\ddot{\bm{o}}_\mathcal{B} +
      \dot {\bm \omega}_\mathcal{B}{\times}~{}^\mathcal{I}{\bm {R}}_\mathcal{B}
       {}^\mathcal{B} {\bm p} + \bm \omega_\mathcal{B}{\times}~\Big(\bm
        \omega_\mathcal{B}{\times}~{}^\mathcal{I}{{\bm R}}_\mathcal{B}
         {}^\mathcal{B} {\bm p} \Big)~.
\end{eqnarray}

\subsection{6D Motion Vectors}

A $6$D motion vector represents a \emph{spatial} vector of a quantity that
 describes the motion of a body (e.g., the velocity and the acceleration).
  Given a generic spatial vector ${\bm {\underline p}} \in \mathbb{R}^6$, the
   three first entries are the coordinates of its linear part $\in
    \mathbb{R}^3$, and the last three entries represent the angular part $\in
     \mathbb{R}^3$ of the same quantity, such that
\begin{equation} \label{spatial_vect}
    {\bm {\underline{p}}} = 
    \begin{bmatrix}
    {\bm p}_{lin} \\
    {\bm p}_{ang}
    \end{bmatrix} \in \mathbb{R}^6~.
\end{equation}

Within this new formalism, the velocity (i.e., the \emph{spatial velocity}) of
 the rigid body expressed w.r.t. $\mathcal{I}$ can be written as
\begin{equation} \label{spatial_vel}
    {}^\mathcal{I} {\bm {\underline v}}_\mathcal{B} = 
    \begin{bmatrix}
    {}^\mathcal{I} \dot{\bm p} \\
    \bm \omega_\mathcal{B}
    \end{bmatrix} \in \mathbb{R}^6~.
\end{equation}

Similarly, the spatial notation for the acceleration (i.e., the \emph{spatial
 acceleration}) of the rigid body  expressed w.r.t. $\mathcal{I}$ is 
\begin{equation} \label{spatial_acc}
    {}^\mathcal{I} {\bm {\underline a}}_\mathcal{B} = 
    \begin{bmatrix}
    {}^\mathcal{I} \ddot{\bm p} \\
    \dot {\bm \omega}_\mathcal{B}
    \end{bmatrix} \in \mathbb{R}^6~.
\end{equation}

\subsubsection{Adjoint Transformation for Motion Vectors}

When dealing with spatial vectors, the change of frame can not be performed by
 using an homogeneous transformation but it has to be introduced a new
  transformation matrix, i.e., the adjoint matrix $\bm X \in \mathbb{R}^{6
   \times 6}$ define as follows.

Let $\mathcal{A}$ and $\mathcal{B}$ be two generic frames and
 ${}^\mathcal{B}{\bm{o}}_\mathcal{A}$ the position vector of the origin of
  $\mathcal{A}$ w.r.t. $\mathcal{B}$, thus the adjoint transformation is
\begin{equation} \label{adjoint_motion_matrix}
{}^\mathcal{B} \bm {X}_\mathcal{A} =
\begin{bmatrix}
    {}^\mathcal{B} \bm {R}_\mathcal{A}   & \bm 0_{3}\\
     - {}^\mathcal{B} \bm {R}_\mathcal{A}
          \bm \skewOp\big({}^\mathcal{B}{\bm{o}}_\mathcal{A}\big) &
          {}^\mathcal{B} \bm
     {R}_\mathcal{A} 
    \end{bmatrix}~,
\end{equation}
such that 
\begin{equation} \label{adjoint_motion_changeOfFrame}
{}^\mathcal{B} {\bm {\underline v}_\mathcal{A}} = {}^\mathcal{B} \bm
 {X}_\mathcal{A}
\ {}^\mathcal{A}{\bm {\underline v}_\mathcal{A}}~.
\end{equation}

\subsubsection{The Cross Product $\times$ for Motion Vectors}

Within the framework of the motion vectors, the cross product operator as
 previously defined is not useful anymore.  In \eqref{crossProcuct_inR3} it is
  defined in $\mathbb{R}^3$ while now we are operating in the new
   $\mathbb{R}^6$ formalism.  Thus, it has to be properly modified for matching
    the spatial context.

Consider a rigid body with a body frame $\mathcal{B}$ that is
 moving with a spatial velocity ${\bm {\underline v}}_\mathcal{B}$, as defined
  in \eqref{spatial_vel}.  If the generic vector $\bm u$ is a motion vector
   (e.g., the spatial acceleration of the body), thus their cross product is
\begin{equation}\label{crossProduct_inR6}
    {{}^\mathcal{I}\bm {\underline v}}_\mathcal{B}{\times}~ {\bm {\underline
     u}} \in \mathbb{R}^6~.
\end{equation}
In particular, the cross product operator is such that
\begin{equation}\label{crossOperator_inR6}
    {{}^\mathcal{I} \bm {\underline v}}_\mathcal{B}{\times}~ :=
    \begin{bmatrix}
        \bm \omega_\mathcal{B}{\times}~  &   {}^\mathcal{I}\dot{\bm p}  \\
        \bm 0_{3}                        &  \bm \omega_\mathcal{B}{\times}~
        \end{bmatrix} \in \mathbb{R}^{6\times 6}~,
\end{equation}
where it is easy to identify in the term $\bm \omega_\mathcal{B}{\times}$ the
 skew-symmetric matrix defined as in Equation \eqref{skew_omega}.

\section{Rigid Body Dynamics}

In classical mechanics, the dynamics is that branch concerned with the study of
the forces and torques and their effect on the motion of a body. 
It is worth remarking that although sometimes it is referred to this branch as
 the \emph{kinetics} of the rigid body \citep{Winter1990}
  \citep{Robertson2014}, we prefer here to address to the \emph{dynamics} (as
   in \citep{handbook_robotics} \citep{Featherstone2008}).  Clearly the reader
    has to know that the two definitions are completely equivalent.

\subsection{6D Force Vectors}

Similarly to the motion vector case, also in the context of dynamics it is
 possible to define a
 specific class of $6$D vectors related to the force.  As previously, a force
  vector contains a pair of $3$D vectors: the first vector represents the
   resultant of the forces ${}^\mathcal{I}{\bm f}$ acting on the
    body, the last one is referred to the moments ${}^\mathcal{I}{\bm m}$
     produced by the resultant w.r.t. a given point.  This yields to the
     following notation:

\begin{equation} \label{spatial_force}
    {}^\mathcal{I} {\bm {\underline f}} = 
    \begin{bmatrix}
    {}^\mathcal{I}{\bm f} \\
    {}^\mathcal{I}{\bm m}
    \end{bmatrix} \in \mathbb{R}^6~.
\end{equation}

\vspace{0.1cm}
\subsubsection{Adjoint Transformation for Force Vectors}

The adjoint transformation for a $6$D force vector is
\begin{equation} \label{adjoint_force_matrix}
{}^\mathcal{B} \bm {X}_\mathcal{A}^* =
\begin{bmatrix}
    {}^\mathcal{B} \bm {R}_\mathcal{A}   &  - {}^\mathcal{B} \bm {R}_\mathcal{A}
          \bm \skewOp\big({}^\mathcal{B}{\bm{o}}_\mathcal{A}\big)\\
     \bm 0_{3} &  {}^\mathcal{B} \bm
     {R}_\mathcal{A} 
    \end{bmatrix}~,
\end{equation}
such that
\begin{equation} \label{adjoint_force_changeOfFrame}
{}^\mathcal{B} {\bm {\underline f}}_\mathcal{A} = {}^\mathcal{B} \bm
 {X}_\mathcal{A}^*
 {{}^\mathcal{A}\bm {\underline f}}_\mathcal{A}~,
\end{equation}
being $\mathcal{A}$, $\mathcal{B}$ and ${}^\mathcal{B}{\bm{o}}_\mathcal{A}$,
 two generic frames and the position vector of the origin of
  $\mathcal{A}$ w.r.t. $\mathcal{B}$, respectively.

\subsubsection{The (Dual) Cross Product $\times^*$ for Force Vectors}

Similarly to the cross product for the motion vectors, we have a cross product
 for the force vectors that is the dual version of \eqref{crossOperator_inR6}.
   Within the same framework explained for the motion case , we have

\begin{equation}\label{dualCrossOperator_inR6}
    {}^\mathcal{I}{\bm {\underline v}}_\mathcal{B}{\times}^*~ :=
    \begin{bmatrix}
        \bm \omega_\mathcal{B}{\times}~  &   \bm 0_{3}            \\
        {}^\mathcal{I}\dot{\bm p}        &  \bm \omega_\mathcal{B}{\times}~
        \end{bmatrix} \in \mathbb{R}^{6\times 6}~,
\end{equation}
such that
\begin{equation}\label{dualCrossProduct_inR6}
    {{}^\mathcal{I}\bm {\underline v}}_\mathcal{B}{\times}^*~ {\bm {\underline
     f}} \in \mathbb{R}^6~.
\end{equation}

\subsection{The Newton-Euler Equations of Motion}

The equations of motion describe the mathematical model for the dynamics of the
 rigid body.  Within the spatial formalism, it is possible to write the
  equations of motion as
\begin{equation} \label{equation_of_motion_1body}
    {\bm {\underline f}} = \frac{d}{dt}{\left(\bm {\mathrm{\underline I}} {\bm
     {\underline
     v}}\right)} = \bm {\mathrm{\underline I}} {\bm {\underline a}} + {\bm
      {\underline
      v}}{\times^*}~\bm {\mathrm{\underline I}}{\bm {\underline v}}~,
\end{equation}
where the force of a rigid body that is moving with velocity ${\bm
 {\underline v}}$ and acceleration ${\bm {\underline a}}$ is equal to the rate
  of change of its momentum $\bm {\mathrm{\underline I}}{\bm {\underline v}}$.
    In Equation \eqref{equation_of_motion_1body},
\begin{itemize}
\item{${\bm {\underline f}} \in \mathbb{R}^6$ in the net force acting on the
 rigid body;}
\item{$\bm {\mathrm{\underline I}}$ is the spatial inertia tensor $\in
 \mathbb{R}^{6 \times 6}$, such that}
\begin{equation}\label{spatial_inertia_tensor}
\bm {\mathrm{\underline I}} =
\begin{bmatrix} \bm {\mathrm{I}} + \textrm{m}~\bm c{\times}~\bm c^\top &
     \textrm{m}~\bm
     c{\times}~\\ \textrm{m}~\bm c{\times}^\top~& \textrm{m}~\bm {1}_{3}
\end{bmatrix}~,
\end{equation}
where $\bm {\mathrm{I}}$ is the inertia tensor w.r.t. the link center of
 mass (CoM), $\textrm{m}$ is the body mass, $\bm c$ is the position vector from
  the CoM
  of the body to the origin of the body frame $\mathcal{B}$;
\item{the term ${\bm {\underline v}}{\times^*}$ is the operator that maps $\bm
 {\mathrm{\underline I}}$ to its derivative $\dot{\bm {\mathrm{\underline
  I}}}$.}
\end{itemize}

\section{Rigid Multi-Body System}

A rigid multi-body system is a system composed of two or more interconnected
 rigid bodies. The resulting motion of a system is obtained by composing
  elementary motions of each body w.r.t. the coupled one.  The mobility of the
   system is ensured by the presence of the \emph{joints}: each joint
    interconnects two bodies and constraints their motion.
The type of motion is strongly related to the type of joint but the description
 of the different types of joints is out of the scope of the thesis.  Thus,
  they will be described in detailed only when required for the thesis
   understanding.

\subsection{Modelling}\label{multibodysystem_modelling}

Within a classical widespread formalism in robotics
 \citep{Featherstone2008} \citep{Siciliano2009}, an articulated rigid
  multi-body system is represented as a kinematic tree composed of
   rigid bodies.
The topology  of the system (i.e., the interconnectivity property) is such
 that it is often referred to it as a tree system.  Under the tree
  schematization formalism, the bodies are the nodes and the connection
   between them are the joints.  Each rigid body in the system is associated to
    a unique node in the tree.

Let $N_B$ be the number of moving rigid bodies of the system whose topological
 numbering is defined from $1$ to $N_B$ ( $0$ refers to
  the fixed base). Let $i$ be the index for a generic body in the tree such
   that $1<i<N_B$.  Node numbers can be always selected in a topological order
    so that the $i$-th node has a higher number than its unique parent
     $\lambda(i)$ and a smaller number than all the nodes in the set of its
      children $\mu(i)$.

\begin{figure}[ht!]
     \centering
         \includegraphics[width=0.9\textwidth]{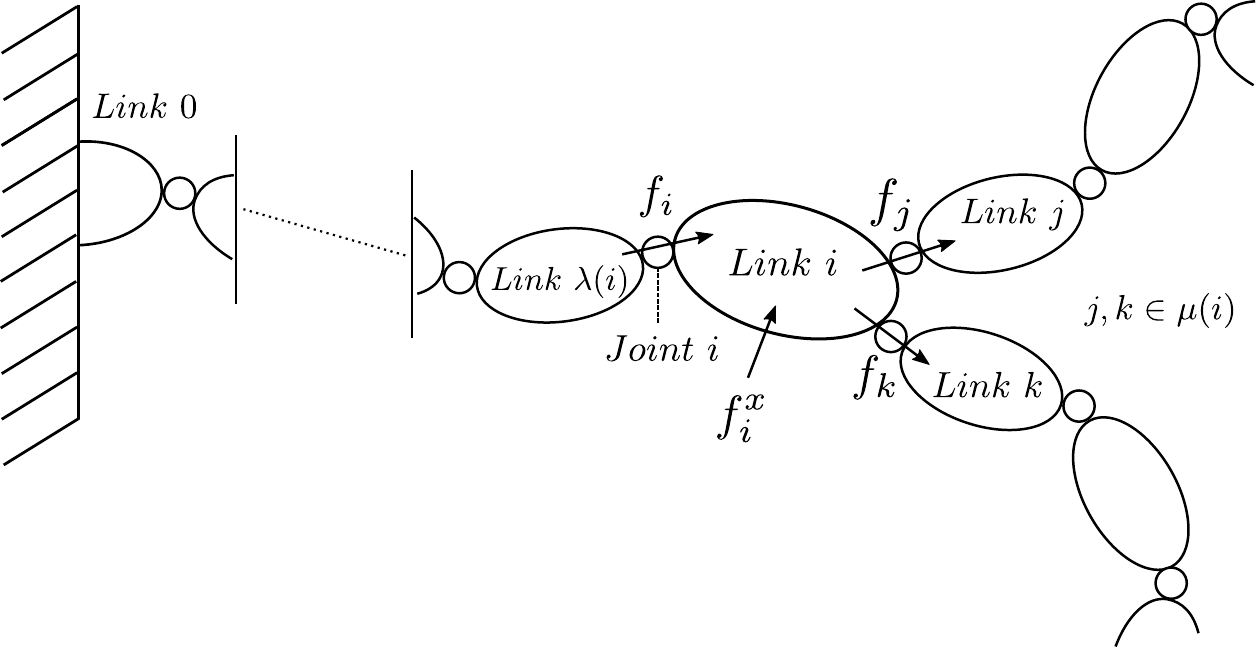}
         \caption{ Connectivity representation of an articulated rigid
          multi-body system 
         as a kinematic tree.}
     \label{figs:kinematicsTree}
 \end{figure}

The $i$-th body and its parent $\lambda(i)$ are coupled with joint $i$
 according to the Denavit-Hartenberg convention for joint numbering
  \citep{Denavit1955}.  The motion freedom subspace of the $i$-th joint is
   modelled with ${\bar {\bm S}}_i \in \mathbb R^{6 \times n_i}$, being $n_i$
    the DoFs number of the joint $i$. Within this notation, $n$ = $n_1$ + $...$
     + $n_{N_B}$ represents the internal DoFs of the system.
Figure \ref{figs:kinematicsTree} shows a connectivity system representation for
 the kinematic tree.

\subsection{Lagrangian Representation} \label{Lagrangian_definition}

A common representation of the dynamics of the system is derived from the
 Lagrangian
 formalism.  Given a rigid multi-body system with $n$ internal DoFs, by
  rearranging the Lagrange formulation (the reader could refer to several
   robotics textbooks like \citep{Siciliano2009}, Chapter 7), it is possible to
    define its floating-base dynamics representation\footnote{Representation in
     which none of the links have an a priori constant pose w.r.t. the inertial
      frame.}
      as follows:

\begin{eqnarray}\label{floating_Lagrangian_representation}
\bm {\mathrm{M}}(\bm {{q}}) \dot{\bm{\nu}} + \bm {\mathrm{C}}(\bm {{q}},
 \bm{\nu})
\bm{\nu} + \bm {\mathrm{G}}(\bm {{q}}) = \begin{bmatrix} \bm 0 \\ {\bm \tau}
 \end{bmatrix} + \bm {\mathrm{J}}^\top(\bm {{q}}) \bm{\mathrm{f}}~,
\end{eqnarray}
where $\bm {\mathrm{M}} \in \mathbb {R}^{({n+6})\times({n+6})}$ is the mass
 matrix, $\bm {\mathrm{C}} \in \mathbb {R}^{({n+6})\times({n+6})}$ is the
  Coriolis effects matrix, $\bm {\mathrm{G}}  \in \mathbb {R}^{n+6}$  is the
   gravity bias term.

In general, a floating-base system is defined by its configuration ${\bm q} =
\left({^{\mathcal I}}\bm {H}_{\mathcal B},\bm{q}_j\right)$ $\in
SE(3)\times\mathbb R^{n}$ and its velocity ${\bm \nu} = \left({^{\mathcal
I}}\bm {v}_{\mathcal B},\dot{\bm{q}}_j\right)\in \mathbb R^{n+6}$, where
  $\mathcal I$ is the inertial frame and $\mathcal B$ the base frame.
${^{\mathcal I}}\bm {H}_{\mathcal B}$ is the homogeneous transformation
from $\mathcal B$ to $\mathcal I$ and  ${^{\mathcal I}}\bm
{v}_{\mathcal B}$ is the base velocity w.r.t. $\mathcal I$.
The configuration and the velocity of the internal DoFs are denoted
with $\bm{q}_j$ and $\dot{\bm{q}}_j$, respectively.
The system can interact with the external environment and this results in the
 presence of $\bm{\mathrm{f}} \in \mathbb R^{6k}$, where $k$ is the number of
  the forces exchanged during the interaction. 
The Jacobian associated with the forces $\bm{\mathrm{f}}$ is denoted by
$\bm{\mathrm{J}}(\bm{\mathrm{q}})$ and the vector $\bm \tau \in \mathbb{R}^n$
 represents the joint torques of the system.

Throughout the thesis the human experiments will be performed with the feet fixed to the ground.  We limit to consider therefore 
the human system to be fixed base,
 i.e., ${^{\mathcal I}}\bm {H}_{\mathcal B} = const$ and known a priori and
   ${^{\mathcal I}}\bm {v}_{\mathcal B}=\bm 0$.  This choice straightforwardly
    implies $\bm{q} =
    \bm{q}_j$ and allows to reduce remarkably the framework complexity.  These
    assumptions yield to a simplified version of
     \eqref{floating_Lagrangian_representation}, such that

\begin{eqnarray}\label{fixed_Lagrangian_representation}
\bm {\mathrm{M}}(\bm {{q}}) \ddot{\bm{q}} + \bm {\mathrm{C}}(\bm {{q}}, \dot{
 \bm{q}})\dot{\bm{q}} + \bm {\mathrm{G}}(\bm {{q}}) = \begin{bmatrix} \bm 0 \\
  {\bm \tau} \end{bmatrix} + \bm {\mathrm{J}}^\top(\bm {{q}}) \bm{\mathrm{f}}~.
\end{eqnarray}

\subsection{Newton-Euler Representation} \label{RNEA}

The Newton-Euler formulation is an equivalent way to the Lagrangian formalism
 for representing the dynamics of a multi-body system.  It is based on the
  balance of the forces acting on
  each body composing the structure of the system and leads to a set
   of recursive equations for propagating the kinematics and dynamics
    quantities throughout the system. 

Consider a system of $N_B$ rigid bodies connected by joints.
  The topology and the body numbering are defined as in Section
   \ref{multibodysystem_modelling}.  Let $0$ be the fixed base.  If $1<i<N_B$,
    the velocity of the $i$-th link and the velocity through the $i$-th joint
     are defined recursively as follows:
\begin{eqnarray}
\label{eq:vJi}
\bm {\underline v}_{Ji} &=& \bm {\bar S}_i \dot {q_i}~,\\
\label{eq:vi}
\bm {\underline v}_i &=& \prescript{i}{} {\bm X_{\lambda(i)}} \bm {{\underline
 v}}_{\lambda(i)} + \bm {\underline v}_{Ji}~.
\end{eqnarray}
Given Equations \eqref{eq:vJi} and \eqref{eq:vi}, the recursive Newton-Euler
 algorithm (RNEA) consists of the following steps, expressed in body $i$
  coordinates\footnote{Except for the external forces that are expressed in the
   body $0$
   absolute coordinates.}:
\begin{eqnarray}
\label{eq:ai}
\bm {\underline a}_i &=& \prescript{i}{}{\bm X_{\lambda(i)}} \bm {\underline
 a}_{\lambda(i)} + \bm {\bar S}_i \ddot { q}_i + \bm {\underline
  v}_i{\times}~ \bm {\underline v}_{Ji}~,\\
\label{eq:fBi}
\bm {\underline f}^B_i & =&  \bm {\mathrm{\underline I}}_i \bm {\underline a}_i
 + \bm {\underline v}_i{\times^*}~ \bm {\mathrm{\underline I}}_i \bm
  {\underline v}_i~,\\
\label{eq:fi}
\bm {\underline f}_i & =&  \bm {\underline f}^B_i - \prescript{i}{} {\bm
 X_{0}^*} \bm {\underline f}_i^x + \sum_{\mu(i)} \prescript{i}{}{\bm
  X_{\mu(i)}^*} \bm {\underline f}_{\mu(i)}~,\\
\label{eq:taui}
\tau_i & =&  \bm {\bar S}^\top_i \bm {\underline f}_i~.
\end{eqnarray}

Equations \eqref{eq:vJi}, \eqref{eq:vi} and \eqref{eq:ai} are propagated
 throughout the kinematic tree with the initial boundary
  conditions $\bm {\underline v}_0 =\bm 0$ and $\bm {\underline a}_0 = -\bm
   {\underline g}$, which corresponds to the gravitational spatial
    acceleration vector expressed in the body frame $0$, such that
$\bm {\underline g} = 
\begin{bmatrix}
    0 & 0 & -9.81 & 0 & 0 & 0
\end{bmatrix}^T$.
Equation \eqref{eq:fBi} is exactly the equation of motion for the $i$-th body
 described in \eqref{equation_of_motion_1body} and represents here the net
  force acting on the body.  Equation \eqref{eq:fi} is the result of a balance
   of incoming forces (i.e., the internal force $\bm {\underline f}_i$
    transmitted through the $i$-th joint, the
    external force $\bm {\underline f}_i^x$) and
    outgoing forces (namely, the set of internal forces exchanged with the
     children bodies) w.r.t. the $i$-th link.
\clearpage\null

\chapter{Human Whole-Body Modelling}  %
\label{Chapter_human_modelling}

\begin{quotation}
\noindent\emph{
All models are wrong, but some are useful.
}
    \begin{flushright}
        George E. P. Box
    \end{flushright}
\end{quotation}

\noindent
In this Chapter a detailed description of the human whole-body
 modelling is introduced.  This is an important step for retrieving the
  estimation
  of the human dynamics.  The starting point for our analysis is recalled in
   \citep{LatellaSensors2016} where a simple 2-DoF model was employed for the
    whole-body modelling.
The rough approximation introduced by that model goes without saying and it 
was useful the `upgrade' to a more complex model able to describe a set of more
 complex movements.

\section{Human Whole-Body Modelling}

\begin{figure}[H]
    \centering
        \includegraphics[width=0.80\textwidth]{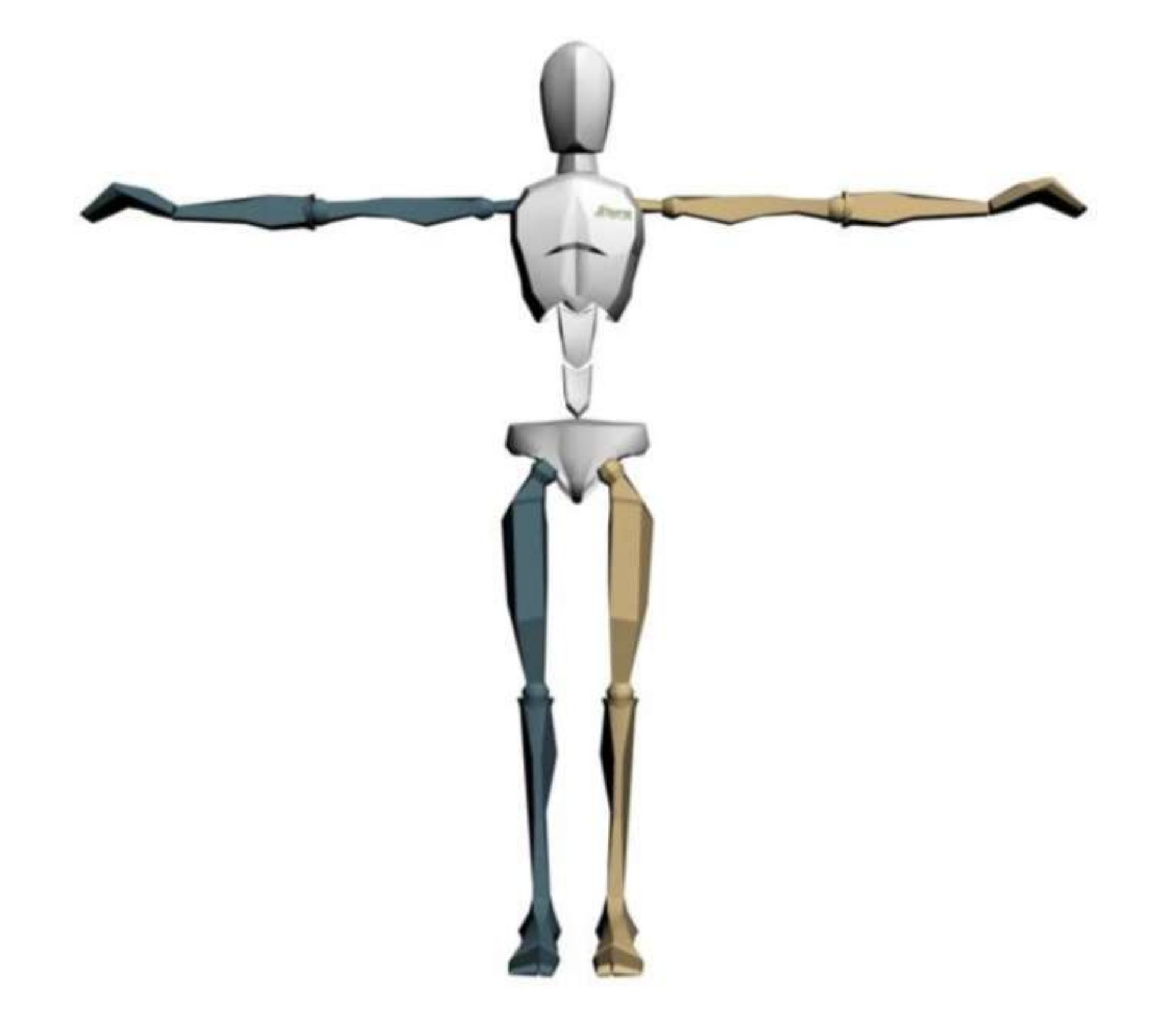}
        \caption{ Xsens MVN system representation for the human body model.}
        \source{Xsens, MVN User Manual, 2005.}
    \label{figs:Xsens_model}
\end{figure}

By inheriting the robotics-like formalism described in the Section
 \ref{multibodysystem_modelling}, we propose a human
 body model as an articulated multi-body system represented by a
  kinematic tree with $N_B$ = $23$ moving links\footnote{Throughout the thesis,
   it is often referred to a body as a \emph{link}.  The two entities are
    completely equivalent.} ($0$ is the fixed base, e.g.,
   the left foot\footnote{At the current stage, the algorithm used for the
    human dynamics computation requires a fixed-base model.  This motivated our
     choice to consider the left foot as the fixed base (the human subject
      performed the experimental tasks with the feet fixed on the ground).})
       and $n$ = $48$ internal DoFs.
The rationale behind the choice of this link number is that we have been guided
 by the model developed for the Xsens MVN motion capture system
  \citep{Roetenberg2009} (Figure \ref{figs:Xsens_model}).  Since the
  sensor setup included the Xsens motion capture system, we considered that
   having two comparable models could be an appropriate choice.  This choice is
   also justified by the fact that the Xsens sensors readings had to be
    associated to our model.

Several strong assumptions have been considered in designing our model:
\begin{itemize}
    \item[$i$)]  to represent the human model as a set of rigid links of
     simple geometric shape (parallelepipeds, cylinders and spheres);
    \item[$ii$)] to consider density isotropy for each link.
\end{itemize}

In the following sections, it is provided to the reader a detailed description
for each of the above-mentioned assumption by motivating through literature
 our choice.

\subsection{Inertial Properties Modelling}

Inertial properties of link segments, such as mass, centre of mass or moments
 of inertia, are important parameters in predicting realistic movement
  patterns. However, these quantities are not directly measurable.
More than a century of literature on this topic proves that the importance in
 retrieving this information has a pivotal role in studying human dynamics.
In \citep{Erdmann1999} the author claims that usually in geometric and inertial
 investigations, three approches are adopted, by involving living subjects,
  cadavers and models.

The first investigation (with living subjects) consists in the measurements of
 the body height and body mass.  This type of technique produced over the years
  well documented anthropometric tables, adopted in the most of inertial
studies \citep{Drillis1964} \citep{Winter1990} \citep{Herman2007}. 
   These tables correlate the total body mass/height with the mass/height of
    each link. 
The second investigation concerns dissections made on cadavers while the model
 technique is more common in simulation research towards prevention (e.g.,
  car-crash tests). 

The author of a more recent study \citep{Robertson2013} classifies into two
 macro-categories the methodologies for estimating body inertial parameters:
  regression equations models and geometric approximation models. The former
   technique is based on equations derived from cadaveric measurements,
    computed radiographs, or magnetic resonance imaging of a finite set of
     specimens. The author reveals the weakness of this approach since it
      extrapolates data from an average and has little accuracy for modelling
       a subject that falls outside the proportions of the initial dataset.
The other approach consists of a collection of \emph{in situ} measurements to be
 mapped into a geometric model to derive mass and properties from certain
  shape and density assumptions. Here too, the accuracy of the method depends
   on the complexity of the model and limited to the numbers of measurements.  
To this purpose, the Hatze model \citep{Hatze1980} consists of a $17$-segment
 geometrical model of the body but requires $242$ anthropometric measurements,
  which would imply a relatively substantial amount of time from each subject.

Even if the uniform/non-uniform density assumption is an important part of this
 analysis (there is an outstanding debate among the scientific community), the
  complexity and data collection time are the most likely reasons for the lack
   of the use of Hatze-like model in biomechanical research.

Our approach does not support neither the regression methodology nor the
 geometrical one.  With $i)$ we opt for a model of simple shapes and $ii)$
  offers the possibility of using some analytic formulas to compute the
   inertial proprieties of all the segments in the
model.  Under the assumption of density isotropy \citep{Hanavan1964}, the
 inertia tensor $\bm{\mathrm{I}}$ has been computed such that
\begin{equation}\label{inertiaTensor}
\bm{\mathrm{I}}  =  \begin{bmatrix}
        \mathrm{I}_{xx} & 0               & 0               \\
         0              & \mathrm{I}_{yy} & 0               \\
         0              & 0               & \mathrm{I}_{zz} \\
    \end{bmatrix}~,
\end{equation}
where $\mathrm{I}_{xx}$, $\mathrm{I}_{yy}$ and $\mathrm{I}_{zz}$ are the
 principal moments of inertia.  Table \ref{PrincipalMomentsInertia_shapes}
  lists analytical formulas for the principal moments of inertia computation.

\begin{table}[H]
\centering
\caption{Principal moments of inertia of three different shapes with mass $\textrm{m}$:
 (\emph{on left column}) a rectangular parallelepiped of width $\alpha$, height
  $\beta$ and depth $\gamma$; (\emph{on middle column}) a circular cylindrical
   of radius $r$ and height $h$; (\emph{on right column}) a sphere of radius
    $r$.}
\label{PrincipalMomentsInertia_shapes}
\centering
\scriptsize
\vspace{-0.1cm}
\begin{tabular}{c|cccc}
\\
\hline\hline
\\
\textbf{Inertia} & \textbf{Parallelepiped} & \textbf{Cylinder}
                 & \textbf{Sphere}\\
\\
\hline
\\
$\mathrm{I}_{xx}$      & $\frac{1}{12}\textrm{m} ~
 \big(\alpha^{2}+\beta^{2}\big)$
                       & $\frac{1}{12}\textrm{m} ~ \big(3r^{2}+h^{2}\big)$
                       & $\frac{2}{5}\textrm{m}r^{2}$ \\
                       \\
\rowcolor{Gray}
$\mathrm{I}_{yy}$      & $\frac{1}{12}\textrm{m} ~
 \big(\beta^{2}+\gamma^{2}\big)$
                       & $\frac{1}{2}\textrm{m} r^2$
                       & $\frac{2}{5}\textrm{m}r^{2}$ \\
                       \\
$\mathrm{I}_{zz}$      & $\frac{1}{12}\textrm{m} ~
 \big(\gamma^{2}+\alpha^{2}\big)$
                       & $\frac{1}{12}\textrm{m} ~ \big(3r^{2}+h^{2}\big)$
                       & $\frac{2}{5}\textrm{m}r^{2}$ \\
\\
\hline\hline
\\
\end{tabular}
\end{table}

\vspace{1cm}
\subsection{Link Modelling}

The labelling of the human model links is inherited by Xsens
 biomechanical model.  For each link, geometry, dimensions, inertial
  parameters, origin and orientation have to be mandatorily specified.
For the \emph{geometry}, the links are modelled as parallelepiped boxes,
 cylinders and spheres.
In order to make the model scalable with the subject proportions, the
 \emph{dimension} of each shape is obtained via the Xsens IMUs
  readings\footnote{Xsens motion capture system provides as inputs $64$ bony
   landmark points distributed all over the model.  We exploit the information
    of their position in the model for delimiting the predominant dimension of
    the link geometrical shape.}.
\emph{Inertial parameters} are computed by exploiting the anthropometric table
 \citep{Winter1990} \citep{Herman2007}: for retrieving the mass of
  each link the weight of the subject (and thus $\textrm{m}_{tot}$) is firstly
   measured and then the mass of each link is obtained by applying tabulated
    data.  Inertia moments are computed as described in Table
     \ref{PrincipalMomentsInertia_shapes}.  The \emph{origin} and
      \emph{orientation} of the frame associated to each link are defined in
       the following way:
\newpage
 \begin{longtable}{l p{9cm}}
 L5                 &  \textbf{origin: }midpoint on top-face of the Pelvis
  parallelepiped \\
                    &  \textbf{orientation: }$x$ pointing forward, $z$ pointing
                     upward (line jL5S1-jL4L3)  \vspace{0.3cm} \\
 L3                 &  \textbf{origin: }midpoint on top-face of the L5
  parallelepiped \\
                    &  \textbf{orientation: }$x$ pointing forward, $z$ pointing
                     upward (line jL4L3-jL1T12) \vspace{0.3cm} \\
 T12                &  \textbf{origin: }midpoint on top-face of the L3
  parallelepiped \\
                    &  \textbf{orientation: }$x$ pointing forward, $z$ pointing
                     upward (line jL1T12-jT9T8) \vspace{0.3cm} \\
 T8                 &  \textbf{origin: }midpoint on top-face of the T12
  parallelepiped \\
                    &  \textbf{orientation: }$x$ pointing forward, $z$ pointing
                     upward (line jT9T8-jT1C7) \vspace{0.3cm} \\
 Neck               &  \textbf{origin: }midpoint on top-face of the T8
  parallelepiped \\
                    &  \textbf{orientation: }$x$ pointing forward, $z$ pointing
                     upward (line jT1C7-jC1Head) \vspace{0.3cm} \\
 Head               &  \textbf{origin: }midpoint on top-circle of the Neck
  cylinder \\
                    &  \textbf{orientation: }$x$ pointing forward, $z$ pointing
                     upward (aligned with jT1C7 $z$ axis) \vspace{0.3cm} \\
 RightUpperLeg      &  \textbf{origin: }midpoint on top-circle of the
  RightUpperLeg cylinder \\
                    &  \textbf{orientation: }$x$ pointing forward, $z$ pointing
                     upward (line jRightKnee-jRightHip) \vspace{0.3cm} \\
 RightLowerLeg      &  \textbf{origin: }midpoint on top-circle of the
  RightLowerLeg cylinder \\
                    &  \textbf{orientation: }$x$ pointing forward, $z$ pointing
                     upward (line jRightAnkle-jRightKnee) \vspace{0.3cm} \\
 RightFoot          &  \textbf{origin: } point on top-face of the RightFoot
  parallelepiped \\
                    &  \textbf{orientation: }$x$ pointing forward, $z$ pointing
                     upward (aligned with jRightKnee $z$ axis) \vspace{0.3cm} \\
 RightToe           &  \textbf{origin: }midpoint on frontal-plane-face (attached
  to RightFoot) of the RightToe parallelepiped  \\
                    &  \textbf{orientation: }$x$ pointing forward, $z$ pointing
                     upward (aligned with jRightKnee $z$ axis) \vspace{0.3cm} \\
 LeftUpperLeg      &  \textbf{origin: }midpoint on top-cirle of the
  LeftUpperLeg cylinder \\
                    &  \textbf{orientation: }$x$ pointing forward, $z$ pointing
                     upward (line jLeftKnee-jLeftHip) \vspace{0.3cm} \\
 LeftLowerLeg      &  \textbf{origin: }midpoint on top-cirle of the
  LeftLowerLeg cylinder \\
                    &  \textbf{orientation: }$x$ pointing forward, $z$ pointing
                     upward (line jLeftAnkle-jLeftKnee) \vspace{0.3cm} \\
 LeftFoot          &  \textbf{origin: } point on top-face of the LeftFoot
  parallelepiped \\
                    &  \textbf{orientation: }$x$ pointing forward, $z$ pointing
                     upward (aligned with jLeftKnee $z$ axis) \vspace{0.3cm} \\
 LeftToe           &  \textbf{origin: }midpoint on frontal-plane-face (attached
  to LeftFoot) of the LeftToe parallelepiped  \\
                    &  \textbf{orientation: }$x$ pointing forward, $z$ pointing
                     upward (aligned with jLeftKnee $z$ axis) \vspace{0.3cm} \\
 RightShoulder{$^{\ast}$}    &  \textbf{origin: }midpoint on
  sagittal-plane-circle (attached to T8) of the RightShoulder cylinder  \\
                    &  \textbf{orientation: }$x$ pointing forward, $y$ pointing
                     right (line jRightC7Shoulder-jRightShoulder) \vspace{0.3cm} \\
 RightUpperArm{$^{\ast}$}    &  \textbf{origin: }midpoint on
  sagittal-plane-circle (attached to RightShoulder) of the RightUpperArm
   cylinder  \\
                    &  \textbf{orientation: }$x$ pointing forward, $y$ pointing
                     right (line jRightShoulder-jRightElbow) \vspace{0.3cm} \\
 RightForeArm{$^{\ast}$}    &  \textbf{origin: }midpoint on sagittal-plane-circle
  (attached to RightUpperArm) of the RightForeArm cylinder  \\
                    &  \textbf{orientation: }$x$ pointing forward, $y$ pointing
                     right (line jRightElbow-jRightWrist) \vspace{0.3cm} \\
 RightHand{$^{\ast}$}    &  \textbf{origin: }midpoint on sagittal-plane-face
  (attached to RightForeArm) of the RightHand parallelepiped  \\
                    &  \textbf{orientation: }$x$ pointing forward, $y$ pointing
                     right (aligned with jRightElbow $y$ axis) \vspace{0.3cm} \\
 LeftShoulder{$^{\ast}$}    &  \textbf{origin: }midpoint on
  sagittal-plane-circle (attached to T8) of the LeftShoulder cylinder  \\
                    &  \textbf{orientation: }$x$ pointing forward, $y$ pointing
                     right (line jLeftC7Shoulder-jLeftShoulder) \vspace{0.3cm} \\
 LeftUpperArm{$^{\ast}$}    &  \textbf{origin: }midpoint on
  sagittal-plane-circle (attached to LeftShoulder) of the LeftUpperArm
   cylinder  \\
                    &  \textbf{orientation: }$x$ pointing forward, $y$ pointing
                     right (line jLeftShoulder-jLeftElbow) \vspace{0.3cm} \\
 LeftForeArm{$^{\ast}$}    &  \textbf{origin: }midpoint on sagittal-plane-circle
  (attached to LeftUpperArm) of the LeftForeArm cylinder  \\
                    &  \textbf{orientation: }$x$ pointing forward, $y$ pointing
                     right (line jLeftElbow-jLeftWrist) \vspace{0.3cm} \\
 LeftHand{$^{\ast}$}    &  \textbf{origin: }midpoint on sagittal-plane-face
  (attached to LeftForeArm) of the LeftHand parallelepiped  \\
                    &  \textbf{orientation: }$x$ pointing forward, $y$ pointing
                     right (aligned with jLeftElbow $y$ axis) \vspace{0.3cm} \\
 \end{longtable}
 
 \noindent
It is worth noting that the frame orientation marked with
  ({$^{\ast}$}) is strongly dependent on the choice of the standard initial
   configuration of the model (T pose, as in Figure
    \ref{human_urdfModel_noSens}). Table \ref{table_linkModelling} synthesizes
     the properties for the link modelling.

\vspace{1cm}
\subsection{Joint Modelling}

The links are coupled by joints, whose labels are inherited by Xsens
 joint labelling.  Each joint requires information about the motion
  type (i.e., the joint DoFs), the origin of the frame associated to it and the
   links connected through it.
For the \emph{motion type}, the joints of the model are represented with $1$,
 $2$ or $3$ DoFs, depending on the anatomy of the relative biological joint. A
  biological joint is defined as the point where two or more bones
   articulate.
Its functional classification describes the degree of movement allowed between
 the bones into three anatomical planes: transversal, frontal and sagittal.
  Thus, the joint motion space is classified as uniaxial (for a movement in one
   plane), biaxial (for a movement in two planes) or multi-axial joints (for
     a movement involving three anatomical planes).
By classifying the joint motion type, a number of $48$ internal DoFs is defined
 for our modelling. 
  The \emph{origin} of the frame associated to each joint coincides with the 
  origin of the frames associated to the link.  Table
   \ref{table_jointModelling} synthetizes the properties for the joint
    modelling. 
\newpage

\begin{table}[H]
        \vspace{2.5cm}
\centering
\small
\caption{Link properties of the model. The labels
 reproduce faithfully the Xsens model labels. It is shown for each link the
  selected shape and the partial mass w.r.t. to
 the total mass of the subject (extracted from \citep{Winter1990}).}
\label{table_linkModelling}
\begin{tabular}{ccc}
\\
\hline\hline
\\
\textbf{Label} & \textbf{Shape} & \textbf{$\%$ Mass of} $\textrm{m}_{tot}$\\
\\
\hline
\\
Pelvis          &  parallelepiped  & $0.08$\\
\rowcolor{Gray}
L5              &  parallelepiped  & $0.102$\\
L3              &  parallelepiped  & $0.102$\\
\rowcolor{Gray}

T12             &  parallelepiped  & $0.102$\\
T8              &  parallelepiped  & $0.04$\\
\rowcolor{Gray}
Neck            &  cylinder        & $0.012$\\
Head            &  sphere          & $0.036$\\
\rowcolor{Gray}
RightShoulder   &  cylinder        & $0.031$\\
RightUpperArm   &  cylinder        & $0.030$\\
\rowcolor{Gray}
RightForeArm    &  cylinder        & $0.020$\\
RightHand       &  parallelepiped  & $0.006$\\
\rowcolor{Gray}
LeftShoulder    &  cylinder        & $0.031$\\
LeftUpperArm    &  cylinder        & $0.030$\\
\rowcolor{Gray}
LeftForeArm     &  cylinder        & $0.020$\\
LeftHand        &  parallelepiped  & $0.006$\\
\rowcolor{Gray}                            
RightUpperLeg   &  cylinder        & $0.125$\\
RightLowerLeg   &  cylinder        & $0.0365$\\
\rowcolor{Gray}
RightFoot       &  parallelepiped  & $0.013$ \\
RightToe        &  parallelepiped  & $0.015$ \\
\rowcolor{Gray}
LeftUpperLeg   &  cylinder        & $0.125$  \\
LeftLowerLeg   &  cylinder        & $0.0365$ \\
\rowcolor{Gray}
LeftFoot       &  parallelepiped  & $0.013$  \\
LeftToe        &  parallelepiped  & $0.0015$ \\
\\
\hline\hline
\\
\end{tabular}
\end{table}

\newpage

\begin{table}[H]
    \vspace{2cm}
\centering
\small
\caption{Joint properties  of the model: labels (inherited from Xsens
 labelling), DoFs per each joint and the links connected through it.}
\label{table_jointModelling}
\begin{tabular}{ccccc}
\\
\hline\hline
\\
\multicolumn{1}{c}{ \textbf{Label}} & \multicolumn{1}{c}{\textbf{DoF}} &
 \multicolumn{3}{c}{\textbf{Connected links}}\\
\\
\hline
\\
\multicolumn{1}{c}{jL5S1} & \multicolumn{1}{c}{$2$} &
 \multicolumn{1}{r}{Pelvis} & \multicolumn{1}{c}{$\longleftrightarrow$} &
 \multicolumn{1}{l}{L5} \\
\rowcolor{Gray}
\multicolumn{1}{c}{jL4L3} & \multicolumn{1}{c}{$2$} & \multicolumn{1}{r}{L5} &
 \multicolumn{1}{c}{$\longleftrightarrow$} & \multicolumn{1}{l}{L3} \\
\multicolumn{1}{c}{jL1T12} & \multicolumn{1}{c}{$2$} & \multicolumn{1}{r}{L3} &
 \multicolumn{1}{c}{$\longleftrightarrow$} & \multicolumn{1}{l}{T12} \\
\rowcolor{Gray}
\multicolumn{1}{c}{jT9T8} & \multicolumn{1}{c}{$3$} & \multicolumn{1}{r}{T12} &
 \multicolumn{1}{c}{$\longleftrightarrow$} & \multicolumn{1}{l}{T8} \\
\multicolumn{1}{c}{jT1C7} & \multicolumn{1}{c}{$3$} & \multicolumn{1}{r}{T8} &
 \multicolumn{1}{c}{$\longleftrightarrow$} & \multicolumn{1}{l}{Neck} \\
\rowcolor{Gray}
\rowcolor{Gray}
\multicolumn{1}{c}{jC1Head} & \multicolumn{1}{c}{$2$} &
 \multicolumn{1}{r}{Neck} & \multicolumn{1}{c}{$\longleftrightarrow$} &
  \multicolumn{1}{l}{Head} \\
\multicolumn{1}{c}{jRightHip} & \multicolumn{1}{c}{$3$} &
 \multicolumn{1}{r}{Pelvis} & \multicolumn{1}{c}{$\longleftrightarrow$} &
 \multicolumn{1}{l}{RightUpperLeg} \\
\rowcolor{Gray}
\multicolumn{1}{c}{jRightKnee} & \multicolumn{1}{c}{$2$} &
 \multicolumn{1}{r}{RightUpperLeg} & \multicolumn{1}{c}{$\longleftrightarrow$}
  & \multicolumn{1}{l}{RightLowerLeg} \\
\multicolumn{1}{c}{jRightAnkle} & \multicolumn{1}{c}{$3$} &
 \multicolumn{1}{r}{RightLowerLeg} & \multicolumn{1}{c}{$\longleftrightarrow$}
  & \multicolumn{1}{l}{RightFoot} \\
\rowcolor{Gray}
\multicolumn{1}{c}{jRightBallFoot} & \multicolumn{1}{c}{$1$} &
 \multicolumn{1}{r}{RightFoot} & \multicolumn{1}{c}{$\longleftrightarrow$} &
  \multicolumn{1}{l}{RightToe} \\
\multicolumn{1}{c}{jLeftHip} & \multicolumn{1}{c}{$3$} &
 \multicolumn{1}{r}{Pelvis} & \multicolumn{1}{c}{$\longleftrightarrow$} &
  \multicolumn{1}{l}{LeftUpperLeg} \\
\rowcolor{Gray}
\multicolumn{1}{c}{jLeftKnee} & \multicolumn{1}{c}{$2$} &
 \multicolumn{1}{r}{LeftUpperLeg} & \multicolumn{1}{c}{$\longleftrightarrow$} &
  \multicolumn{1}{l}{LeftLowerLeg} \\
\multicolumn{1}{c}{jLeftAnkle} & \multicolumn{1}{c}{$3$} &
 \multicolumn{1}{r}{LeftLowerLeg} & \multicolumn{1}{c}{$\longleftrightarrow$} &
  \multicolumn{1}{l}{LeftFoot} \\
\rowcolor{Gray}
\multicolumn{1}{c}{jLeftBallFoot} & \multicolumn{1}{c}{$1$} &
 \multicolumn{1}{r}{LeftFoot} & \multicolumn{1}{c}{$\longleftrightarrow$} &
  \multicolumn{1}{l}{LeftToe} \\
\multicolumn{1}{c}{jRightC7Shoulder} & \multicolumn{1}{c}{$1$} &
 \multicolumn{1}{r}{T8} & \multicolumn{1}{c}{$\longleftrightarrow$} &
  \multicolumn{1}{l}{RightShoulder} \\
\rowcolor{Gray}
\multicolumn{1}{c}{jRightShoulder} & \multicolumn{1}{c}{$3$} &
 \multicolumn{1}{r}{RightShoulder} & \multicolumn{1}{c}{$\longleftrightarrow$}
  & \multicolumn{1}{l}{RightUpperArm} \\
\multicolumn{1}{c}{jRightElbow} & \multicolumn{1}{c}{$2$} &
 \multicolumn{1}{r}{RightUpperArm} & \multicolumn{1}{c}{$\longleftrightarrow$}
  & \multicolumn{1}{l}{RightForeArm} \\
\rowcolor{Gray}
\multicolumn{1}{c}{jRightWrist} & \multicolumn{1}{c}{$2$} &
 \multicolumn{1}{r}{RightForeArm} & \multicolumn{1}{c}{$\longleftrightarrow$} &
  \multicolumn{1}{l}{RightHand} \\
\multicolumn{1}{c}{jLeftC7Shoulder} & \multicolumn{1}{c}{$1$} &
 \multicolumn{1}{r}{T8} & \multicolumn{1}{c}{$\longleftrightarrow$} &
  \multicolumn{1}{l}{LeftShoulder} \\
\rowcolor{Gray}
\multicolumn{1}{c}{jLeftShoulder} & \multicolumn{1}{c}{$3$} &
 \multicolumn{1}{r}{LeftShoulder} & \multicolumn{1}{c}{$\longleftrightarrow$} &
  \multicolumn{1}{l}{LeftUpperArm} \\
\multicolumn{1}{c}{jLeftElbow} & \multicolumn{1}{c}{$2$} &
 \multicolumn{1}{r}{LeftUpperArm} & \multicolumn{1}{c}{$\longleftrightarrow$} &
  \multicolumn{1}{l}{LeftForeArm} \\
\rowcolor{Gray}
\multicolumn{1}{c}{jLeftWrist} & \multicolumn{1}{c}{$2$} &
 \multicolumn{1}{r}{LeftForeArm} & \multicolumn{1}{c}{$\longleftrightarrow$} &
  \multicolumn{1}{l}{LeftHand} \\
\\
\hline\hline
\\
\end{tabular}
\end{table}

\newpage
\section{The Human URDF Model}

The Unified Robot Description Format (URDF) representation is very common among
 the robotics community but
  not widespread among that part of community that deals with human
   modelling.  The reason is quite clear: this format is particularly tailored
    for describing robots (e.g., the typical specification assumes rigid links)
     and it does not take into account the presence of muscles in the model (no
      flexible or shape-variable links are supported).  However, two main
       reasons led us to this choice:
\begin{itemize}
     \item[$1$)]  the possibility offered by this format of creating a valid
      representation for complex tree structures,
     \item[$2$)] the possibility of handling both the models for the human and
      for the robot with the same formalism.
\end{itemize}
See Appendix \ref{URDF_human_modelling_appendix} for a detailed description on
 how to build the human URDF model.

\subsection{Sensor Pose Estimation}

In order to perform the estimation of the human dynamics, the information about
 the Xsens IMUs has to be embedded in the URDF model. 
However, the pose of the sensor w.r.t. the attached link is not provided by
 Xsens as an output of the system.  Therefore, in this Section a
  procedure for estimating the sensor position by exploiting IMUs linear
   acceleration and angular velocity is proposed.
The procedure is very similar to \citep{Rotella2016} where it is used for a
 humanoid robot.

\subsubsection{Estimation Procedure}

\begin{figure}[H]
  \centering
    \includegraphics[width=.5\textwidth]{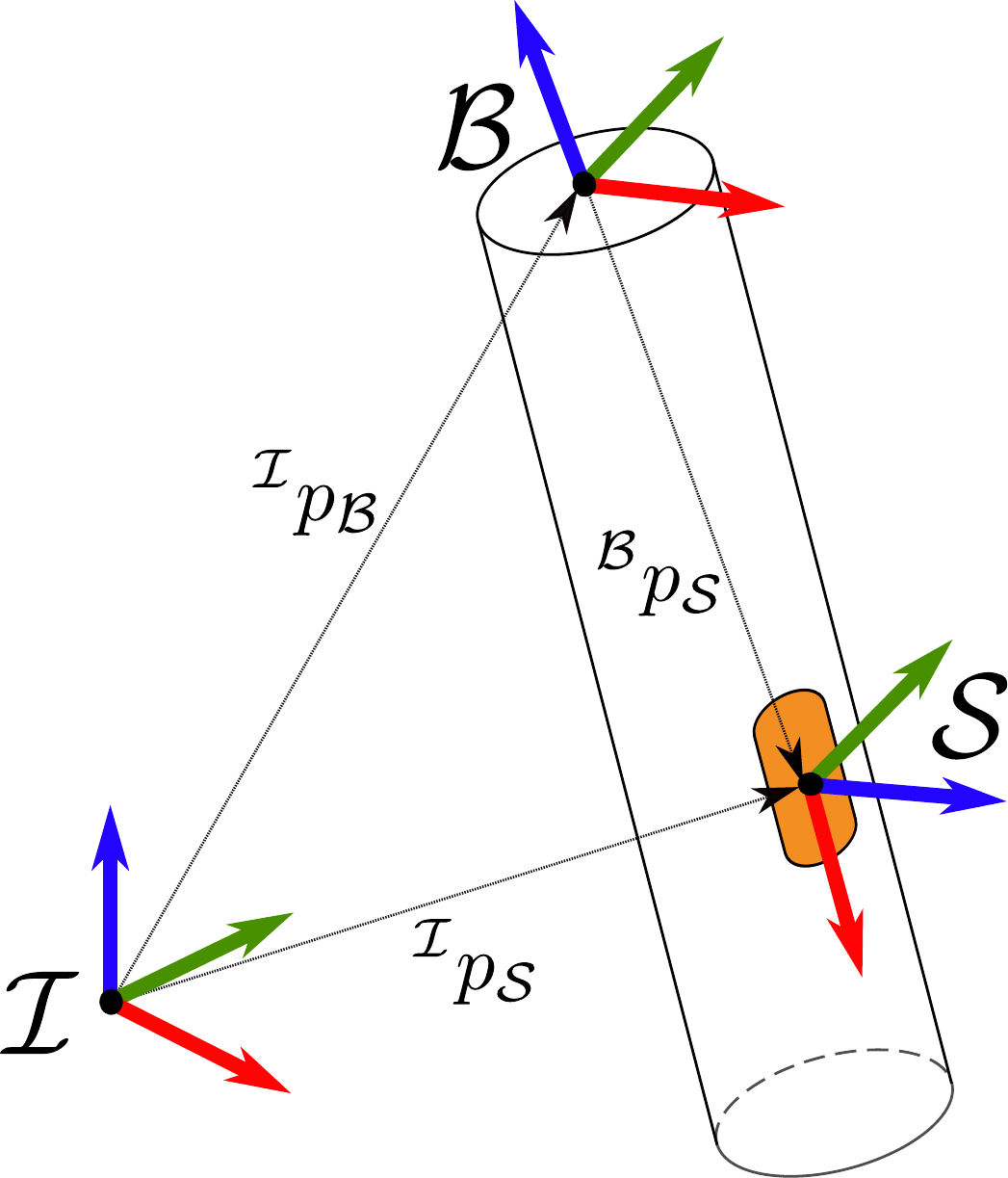}
  \caption{Rigid body identified by the frame $\mathcal{B}$ with a sensor
   (identified by the frame $\mathcal{S}$) rigidly attached on it.
       }
  \label{fig:Figs_framesIBS}
\end{figure}

Let $\mathcal{I}$ be an inertial frame and consider a generic rigid body.  We
 define the body reference frame $\mathcal{B}$ identifying its pose w.r.t.
  $\mathcal{I}$ (Figure \ref{fig:Figs_framesIBS}).
Consider also a sensor rigidly attached to the rigid body and its reference
 frame $\mathcal{S}$. Let ${}^\mathcal{I} {\bm g} \in \mathbb{R}^3$ be the
  gravity acceleration vector expressed in $\mathcal{I}$.  The estimation
   problem can be summarized as follows.
\vspace{0.5cm}
\begin{tcolorbox}[sharp corners, colback=white!30,
     colframe=white!20!black!30, 
     title=The problem]
\emph{Given the pose (${\bm p}_\mathcal{B}$, ${\bm R}_\mathcal{B}$), the
linear acceleration $\ddot{\bm p}_\mathcal{B}$, the angular velocity $\bm
\omega$ and acceleration $\dot{\bm \omega}$ of $\mathcal{B}$ w.r.t.
$\mathcal{I}$, the orientation ${\bm R}_\mathcal{S}$ and the proper acceleration
$\bm a$ of $\mathcal{S}$ w.r.t. $\mathcal{I}$, estimate the relative pose of
$\mathcal{S}$ w.r.t. $\mathcal{B}$, i.e., the position vector
${}^{\mathcal B} \bm p_{\mathcal S} \in
 \mathbb{R}^3$ and the orientation 
 ${}^\mathcal{B} \bm{R}_\mathcal{S} \in SO(3)$.}
\end{tcolorbox}

\noindent
The relative orientation between the sensor and the link can be easily obtained
 by the following equation:
 
\begin{equation}\label{eq:rel_orient}
    {}^\mathcal{B} {\bm R}_\mathcal{S} = {}^\mathcal{I} {\bm
     R}_\mathcal{B}^\top ~ {}^\mathcal{I} {\bm R}_\mathcal{S}~.
\end{equation}

\noindent
To estimate the relative position, consider the measurement equation:

\begin{align}\label{eq:acc_sens}
{}^{\mathcal{S}} \bm{a} &= {}^\mathcal{S} {\bm R}_\mathcal{I} \Big(
 {}^\mathcal{I} \ddot{\bm p}_\mathcal{S} - {}^\mathcal{I} {\bm g} \Big)
  \notag\\
&={}^\mathcal{S} {\bm R}_\mathcal{I} \left[{}^\mathcal{I} \ddot{\bm
 p}_\mathcal{B} + \dot{\bm{\omega}} \times {}^{\mathcal{I}} {\bm
  R}_{\mathcal{B}} ~{}^{\mathcal{B}} {\bm
   p}_{\mathcal{S}} + \bm{\omega} \times
    \Big(\bm{\omega} \times {}^{\mathcal{I}} {\bm R}_{\mathcal{B}}
     ~{}^{\mathcal B} \bm p_{\mathcal S}\Big)
      - {}^\mathcal{I} {\bm g}\right] \notag\\
&={}^\mathcal{S} {\bm R}_\mathcal{I} \left[{}^\mathcal{I} \ddot{\bm
 p}_\mathcal{B} + \left(\bm \skewOp{(\dot {\bm \omega})}  + \bm \skewOp{(\bm
  \omega)}^2
  \right) {}^{\mathcal{I}} {\bm R}_{\mathcal{B}} ~{}^{\mathcal B} \bm
   p_{\mathcal S} - {}^\mathcal{I} {\bm g}\right]~,
\end{align}

\noindent
where $\bm \skewOp{(\bm \omega)} = {}^\mathcal{I} \dot{\bm R}_\mathcal{B} ~
 {}^\mathcal{B} {\bm R}_\mathcal{I}$, from \eqref{def_skewOp}. 

To estimate ${}^{\mathcal B} \bm p_{\mathcal S}$ we have to
 solve Equation \eqref{eq:acc_sens}.  Given that measurements may be affected by
  errors, it is impractical to solve it with only one measurement. 
To estimate ${}^{\mathcal B} \bm p_{\mathcal S}$
 we thus collect multiple measurements and solve the resulting overdetermined
  system in the least-squares sense.
We first rewrite \eqref{eq:acc_sens} in the following form:
 
\vspace{-0.5cm}
\begin{eqnarray}
   \underbrace{\left(\bm \skewOp{(\dot {\bm \omega})}  + \bm \skewOp{(\bm
    \omega)}^2
    \right) {}^{\mathcal{I}} {\bm R}_{\mathcal{B}} }_{\bm A} {}^{\mathcal B}
     \bm p_{\mathcal S} =
      \underbrace{{}^\mathcal{I} {\bm R}_\mathcal{S} {}^{\mathcal{S}} \bm a -
       \Big({}^{\mathcal{I}}\ddot {\bm p}_{\mathcal{B}}- {}^{\mathcal{I}} \bm
        g \Big) }_{\bm b}~.
\end{eqnarray}

\noindent
By denoting with ${\bm A}_i$ and ${\bm b}_i$ the matrices $\bm A$ and $\bm b$
 associated with the $i$-th measurement, after collecting $N$ measurements we
  obtain the following linear system:
\begin{eqnarray}
\begin{bmatrix}
{\bm A}_1 \\
\vdots\\
{\bm A}_N
\end{bmatrix} 
{}^{\mathcal B} \bm p_{\mathcal S} =
 \begin{bmatrix}
{\bm b}_1 \\
\vdots\\
{\bm b}_N
\end{bmatrix}~,
\end{eqnarray}
thus
\vspace{-0.4cm}
\begin{eqnarray}
\xoverline{\bm A}~ {}^{\mathcal B} \bm p_{\mathcal S} &=&
\xoverline{\bm b}~,\\
{}^{\mathcal B} \bm p_{\mathcal S} &=&
 \xoverline{\bm A}^\dagger
  \xoverline{\bm b}\label{eq:rel_pos}~,
\end{eqnarray}
where $\xoverline{\bm A}^\dagger = \big(\xoverline{\bm A}^\top \xoverline{\bm
 A}\big)^{-1} \xoverline{\bm A}^\top$ denotes the Moore-Penrose pseudoinverse of
 $\xoverline{\bm A}$.
Equations \eqref{eq:rel_pos} and  \eqref{eq:rel_orient} identify the relative
 pose of $\mathcal{S}$ w.r.t. $\mathcal{B}$.  In the Algorithm
  \ref{algorithm_sensPos} the pseudocode for the procedure of sensor position
   estimation is shown.

\begin{figure}[ht!]
  \centering
  \vspace{0.3cm}
    \includegraphics[width=\columnwidth]{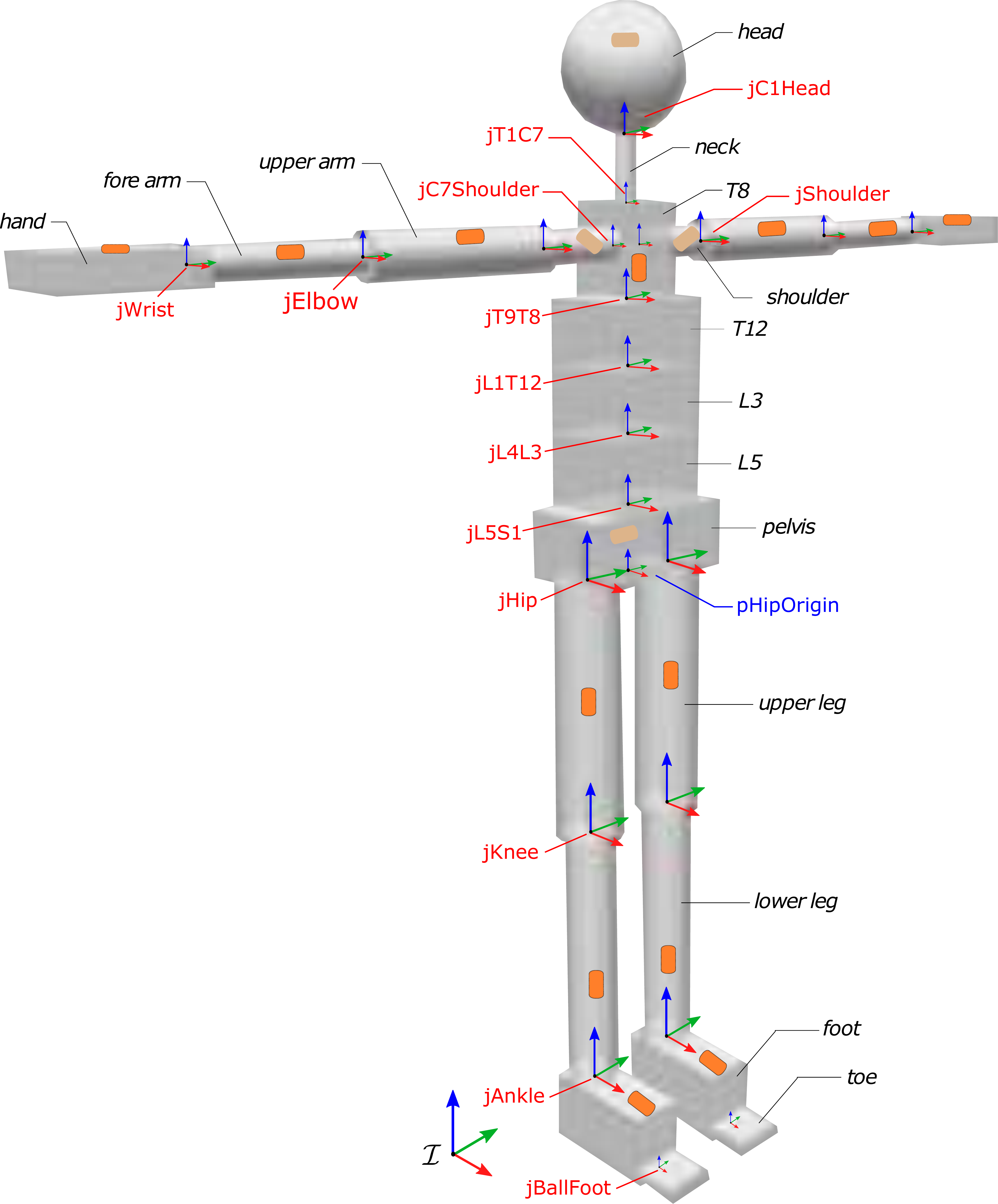}
     \caption{URDF human body model with $17$ IMUs distributed along the body.
       Their pose is estimated via IMUs readings by applying the estimation
        procedure of Algorithm \ref{algorithm_sensPos}.}
  \label{human_urdfModel_sens}
\end{figure}

\vspace{0.5cm}
\begin{algorithm}[ht!]
\caption{Compute sensor pose, Equation \eqref{eq:rel_pos}}
\label{algorithm_sensPos}
\begin{algorithmic}[1]
\Procedure{SensorPoseEstimation}{}
\State $N \gets \text{number of samples}$
\State $N_{IMU} \gets \text{number of IMUs, 17}$
\BState \emph{main loop}:
\For {$j = 1 \to N_{IMU}~$}
\BState \emph{nested loop}:
    \For {$i = 1 \to N~$}
    \State {\emph{$\Rightarrow$ compute ${\bm A}_i$} :}
    \State {${\bm A}_i = \left(\bm \skewOp{(\dot {\bm \omega})} +\bm \skewOp{(\bm
    \omega)}^2 \right) {}^{\mathcal{I}} {\bm R}_{\mathcal{B}} ~ \Big|_{i}$}
    \State {\emph{$\Rightarrow$ compute ${\bm b}_i$} :}
    \State {${\bm b}_i = {}^\mathcal{I} {\bm R}_\mathcal{S} {}^{\mathcal{S}}
     \bm a - \left({}^{\mathcal{I}}\ddot {\bm p}_{\mathcal{B}}-
      {}^{\mathcal{I}} \bm g \right)  ~ \Big|_{i}$}
    \State {\emph{$\Rightarrow$ compute RPY$_i$} :}
    \State{${}^\mathcal{B} {\bm R}_\mathcal{S} ~ \Big|_{i} = {}^\mathcal{I}
     {\bm R}_\mathcal{B}^\top ~ {}^\mathcal{I} {\bm R}_\mathcal{S}$ as
      RPY}
     \EndFor
     \State \textbf{goto} \emph{nested loop}.
     \State {\textbf{end}}
\State {\emph{$\Rightarrow$ compute ${}^{\mathcal B} \bm p_{\mathcal S}$} :}
\State {${}^{\mathcal B} \bm p_{\mathcal S} = {\bm A}\backslash{\bm b}$}
\State {\emph{$\Rightarrow$ compute RPY :}}
\State {${}^\mathcal{B} {\bm R}_\mathcal{S}$ = \emph{mean} (RPY$_i$)}
\EndFor
\State \textbf{goto} \emph{main loop}.
\State {\textbf{end}}
\EndProcedure
\end{algorithmic}
\end{algorithm}
\vspace{1cm}
\newpage
\noindent
A URDF model for the $48$-DoF human template is shown in Figure
 \ref{human_urdfModel_sens}.  The figure shows the $17$ IMUs distributed on the
  body.
\clearpage\null

\chapter{The Human Dynamics Estimation Problem: a Probabilistic Solution}  %
\label{Chapter_estimation_problem}

\begin{quotation}
\noindent\emph{
No great discovery was ever made without a bold guess.
}
    \begin{flushright}
        Isaac Newton
    \end{flushright}
\end{quotation}

\noindent
In this Chapter, the estimation problem for the human dynamics is
 investigated.  As recalled in Section \ref{ID_SoA}, the procedure for
  computing the joint torques $\bm \tau$  given the joint accelerations
   $\ddot{\bm q}$ and the external forces $\bm f^x$ acting on a kinematic tree
    model is the well-known identified problem of the inverse dynamics.
In \citep{Featherstone2008}, the problem is formulated as follows:
\begin{equation}\label{def_PB_ID}
\bm \tau = \mbox{ID}(model, \bm q, \dot{\bm q}, \ddot{\bm q}, \bm f^x)~.
\end{equation}
A classical and computationally efficient solution for the ID problem is the
 RNEA (presented in Section \ref{RNEA}) which also computes the internal forces
  $\bm f$ and the link accelerations $\bm a$.

We perform here a different choice which encompasses the computation of other
 human variables (in a new vector $\bm d$) in a probabilistic domain and
  includes the sensors measurements (defined in a new vector $\bm y$).  The
   problem in \eqref{def_PB_ID} could be therefore written in the form
 \begin{equation}\label{def_PB_MAP}
 \bm d = \mbox{MAPD}(model, \bm q, \dot{\bm q}, \ddot{\bm q}, \bm y)~,
 \end{equation}
where
\begin{itemize}
    \item{the torques $\bm \tau$ of \eqref{def_PB_ID} are embedded in a new
     vector $\bm d$,}
    \item{the external forces $\bm f^x$ used as input in \eqref{def_PB_ID}
  are embedded in a new vector of measurements $\bm y$.}
    \end{itemize}
A detailed explanation for the new quantities will follow in Sections
 \ref{explanation_d} and \ref{explanation_y}, respectively.
 
Clearly, there is an unquestionably strong parallelism with the problem
 \eqref{def_PB_ID} and over this Chapter it is explained how the classical
  robotics formalism has been adapted to fit the needs of the human dynamics
   framework.
The Chapter provides the reader with a detailed description of the
methodology used for solving the estimation problem whose novelty lies in
 considering simultaneously the fusion of different sensors information.

\section{The System Constraints Formulation}\label{explanation_d}

Given a suitable model for the human, as in Chapter
 \ref{Chapter_human_modelling}, it is convenient to define a new vector
  quantity $\bm d$ containing variables for describing both the kinematics and
   the dynamics of the model, with $N_B$ number of bodies and $n$  internal DoFs
    number.  If $i=1:N_B$, the vector $\bm d$ is such that
\begin{eqnarray} \label{vector_d}
\bm d = \begin{bmatrix} \bm d_{1}^\top  & \bm d_{2}^\top  &\hdots & \bm
 d_{N_B}^\top \end{bmatrix}^\top \in \mathbb R^{24N_B+2n}~,
\end{eqnarray}
where 
\begin{eqnarray} \label{vector_d_i}
\bm d_i = \begin{bmatrix} \bm
{\underline a}_{i}^\top & {\bm {\underline f}^B_i}^\top & \bm {\underline
 f}_{i}^\top & \tau_{i} & {\bm {\underline f}_{i}^x}^\top & \ddot {
  q_{i}} \end{bmatrix}^\top \in \mathbb R^{24+2n_i}~.
\end{eqnarray} 
It is worth remarking that $\bm d$ is a \emph{hybrid} vector that, by
 construction, contains quantity related to the $i$-th link (i.e., the
  acceleration $\bm {\underline a}_{i}$, the net force $\bm {\underline f}^B_i$
   and the external force $\bm {\underline f}_{i}^x$) and the $i$-th joint
    (i.e., the force exchanged through the joint $\bm {\underline f}_{i}$,
     the joint torque $\tau_{i}$ and the joint acceleration $\ddot {
      q_{i}}$).

If the Newton-Euler formalism is properly rearranged in a matrix form, thus
 Equations \eqref{eq:ai}-\eqref{eq:taui} can be considered as a set of
  linear constraint equations that $\bm d$ has to satisfy.
Given \eqref{vector_d} and \eqref{vector_d_i}, Equations
 \eqref{eq:ai}-\eqref{eq:taui} can be compactly written in the following matrix
  equation representing the constraints of the system:
\begin{equation} \label{eq:matRNEA} 
\bm D(\bm q, \dot {\bm q}) \bm d + \bm b_D
 (\bm q,\dot {\bm q})= \bm 0~,
\end{equation}
\vspace{0.2cm}
where $\bm D$ is a block matrix $\in \mathbb R^{(18 N_B+n) \times d}$ and $\bm
 b_D$ is a vector $\in \mathbb R^{18 N_B+n}$, as follows\footnote{Note on the
  notation: it is referred here to $\bm D \in \mathbb R^{(18 N_B+n) \times d}$.
    Within this notation $\bm D$ is a matrix with $(18 N_B+n)$ rows and $d$
     columns, i.e., number of rows of $\bm d$ \eqref{vector_d}, namely
      $(24N_B+2n)$.  This form will be recurring throughout the thesis.}:
\newpage
\begin{eqnarray}
 \bm D = \begin{bmatrix} \bm D_{1,1} & \dots & \bm D_{1,N_B} \\ \vdots &
  \ddots & \vdots \\
\bm D_{N_B,1} & \dots & \bm D_{N_B,N_B} \end{bmatrix}, \qquad \bm b_D =
\begin{bmatrix} \bm b_{1} \\ \vdots \\ \bm b_{N_B}
\end{bmatrix}.\notag
\end{eqnarray}
More in detail:
\begin{eqnarray}
\bm D_{i,i} =
\begin{bmatrix}
-\bm 1 & \bm 0 & \bm 0 &\bm 0 & \bm 0 & \bm {\bar S}_i\\ \bm
 {\mathrm{\underline I}}_i &
 -\bm 1 & \bm 0 & \bm 0 & \bm 0 & \bm 0\\ \bm 0 & \bm 1 & -\bm 1 & \bm 0 & \bm
  -\prescript{i}{}{\bm X_{0}^*} & \bm 0\\
\bm 0 & \bm 0 & \bm {\bar S}^\top_i &-\bm 1 & \bm 0 & \bm 0
\end{bmatrix}~,
\end{eqnarray}

\vspace{0.5cm}
\begin{equation}
\forall~ \mu(i) \quad \bm D_{i,\mu(i)} =
\begin{bmatrix}
\bm 0 &\bm 0 &\bm 0 & \bm 0 & \bm 0 & \bm 0\\ \bm 0 &\bm 0 &\bm 0 & \bm 0 & \bm
 0 & \bm 0\\ \bm 0 &\bm 0
&\prescript{i}{}{\bm X_{\mu(i)}^*}& \bm 0 & \bm 0 & \bm 0\\ \bm 0 &\bm 0 &\bm 0
 & \bm 0 & \bm 0 & \bm 0
\end{bmatrix}~,
\end{equation}

\vspace{0.5cm}
\begin{equation}
\forall~ \lambda(i) \quad \bm D_{i,\lambda(i)} =
\begin{bmatrix}
\prescript{i}{}{\bm X_{\lambda(i)}} & \bm 0 & \bm 0 & \bm 0 & \bm 0 & \bm 0 \\
\bm 0 & \bm 0 & \bm 0 & \bm 0 & \bm 0 & \bm 0 \\ \bm 0 & \bm 0 & \bm 0 & \bm 0
 & \bm 0 & \bm 0 \\ \bm 0 & \bm 0 & \bm 0 & \bm 0 & \bm 0 & \bm 0
\end{bmatrix}~,
\end{equation}

\vspace{0.5cm}
\begin{equation}
\mbox{if}~ \lambda(i)=0 \qquad \bm b_i =
\begin{bmatrix}
\prescript{i}{}{\bm X_{0}} \bm {\underline a}_0 + \bm {\underline v}_i{\times}~
 \bm {\bar S}_i \dot {q_i} \\ \bm {\underline v}_i{\times}^*~ \bm
  {\mathrm{\underline I}}_i \bm {\underline v}_i \\ \bm 0\\ \bm 0
\end{bmatrix}~,
\end{equation}

\vspace{0.5cm}
\begin{equation}
\mbox{if}~\lambda(i) \neq 0 \qquad \bm b_i =
\begin{bmatrix}
\bm {\underline v}_i{\times}~ \bm {\bar S}_i \dot {q_i} \\\bm {\underline
 v}_i{\times}^*~ \bm {\mathrm{\underline I}}_i \bm {\underline v}_i
\\ \bm 0\\ \bm 0
\end{bmatrix}~.
\end{equation}

\vspace{0.5cm}
\subsection{Considerations on the Representation}

Equation \eqref{eq:matRNEA} is an equivalent representation of
 \eqref{fixed_Lagrangian_representation}. The Lagrangian formulation can be
  obtained from Equation \eqref{eq:matRNEA} as hereafter described. First, the
vector $\bm d$ and the columns of $\bm D$ should be rearranged so that they
 respect the following order: $\bm {\underline a}_1 \hdots \bm {\underline
  a}_{N_B}$, $\bm {\underline f}_1^B \hdots \bm {\underline f}_{N_B}^B$, $\bm
   {\underline f}_1 \hdots \bm {\underline f}_{N_B}$, $\ddot {q}_1 \hdots
    \ddot { q}_{N_B}$, $\bm {\underline f}_{1}^x \hdots \bm {\underline
     f}_{N_B}^x$ and $ \tau_1 \hdots  \tau_{N_B}$.
The resulting $\bm D$ and $\bm b$ are:
\begin{equation}
\begin{tikzpicture}[decoration=brace]
    \matrix (m) [matrix of math nodes,left delimiter=[,right delimiter={]}]
    {
\bm D_{{a},{f},{f}_B} &
\bm D_{\ddot q} &
\bm D_{{f}^x} &
\bm D_{\tau} \\
\bm D_{{a},{f},{f}_B} &
\bm D_{\ddot q} &
\bm D_{{f}^x} &
\bm D_{\tau} \\
    };
    \draw[decorate,transform canvas={xshift=-1.5em, yshift=0.2em},thick]
     (m-1-1.south west) -- node[left=2pt] {\footnotesize{Equations
      \eqref{eq:ai}--\eqref{eq:fi}}} (m-1-1.north west);
    \draw[decorate,transform canvas={xshift=-1.5em, yshift=0.0em},thick]
     (m-2-1.south west) -- node[left=2pt] {\footnotesize{Equations
      \eqref{eq:taui}}} (m-2-1.north west);
    \draw[decorate,transform canvas={xshift=-0.0em,  yshift=0.5em},thick]
     (m-1-1.north west) -- node[above=2pt] {$\bm {a}, \bm
      {f}, \bm {f}_B$} (m-1-1.north east);
    \draw[decorate,transform canvas={xshift=0.2em,  yshift=0.5em},thick]
     (m-1-2.north west) -- node[above=2pt] {$\ddot{\bm q}, \bm {
      f}^x, \tau$} (m-1-4.north east);
\end{tikzpicture}
\end{equation}

\begin{equation}
\begin{tikzpicture}[decoration=brace]
    \matrix (m) [matrix of math nodes,left delimiter=[,right delimiter={]}]
    {
\bm b_D \\
\bm b_D \\
    };
    \draw[decorate,transform canvas={xshift=-1.5em, yshift=0.2em},thick]
     (m-1-1.south west) -- node[left=2pt] {\footnotesize{Equations
      \eqref{eq:ai}--\eqref{eq:fi}}} (m-1-1.north west);
    \draw[decorate,transform canvas={xshift=-1.5em, yshift=0.0em},thick]
     (m-2-1.south west) -- node[left=2pt] {\footnotesize{Equations
      \eqref{eq:taui}}} (m-2-1.north west);
\end{tikzpicture}
\end{equation}
Within this rearrangement, Equation \eqref{fixed_Lagrangian_representation} can
 be obtained as follows:
\begin{eqnarray}
&& \bm D^{\eqref*{eq:taui}}_{{a},{f},{f}_B}
\left\{~\left[ \bm D^{\eqref*{eq:ai}-\eqref*{eq:fi}}_{{
a},{f},{f}_B} \right]^{-1} \right. \notag
\\
&& \left. \left[ -\bm b_D^{\eqref*{eq:ai}-\eqref*{eq:fi}} - \bm
D^{\eqref*{eq:ai}-\eqref*{eq:fi}}_{\ddot q} \ddot {\bm q} - \bm
D^{\eqref*{eq:ai}-\eqref*{eq:fi}}_{{f}^x} \bm {f}^x -
\bm D^{\eqref*{eq:ai}-\eqref*{eq:fi}}_{\tau \phantom{j}} \bm \tau
\right]_{\phantom{j}} \right\} \notag
\\
&&+~ \bm D^{\eqref*{eq:taui}}_{\ddot q}\ddot{\bm q} + \bm
D^{\eqref*{eq:taui}}_{{f}^x}\bm { f}^x + \bm
D^{\eqref*{eq:taui}}_{\tau}\bm \tau + \bm b_D^{\eqref*{eq:taui}} = \bm 0~.
\end{eqnarray}

In the human dynamics estimation problem, the choice for preferring
 \eqref{eq:matRNEA} formulation  to \eqref{fixed_Lagrangian_representation} is
  two-fold.
On the one hand, Equation \eqref{eq:matRNEA} can be used to represent
 uncertainties that capture relevant modeling approximations. In particular,
  approximations result from the fact that human bones coupling is neither
   rigid nor purely rotational and these can be captured with additive noise on
    Equations \eqref{eq:ai}-\eqref{eq:taui}.
On the other hand, there are numerical advantages associated to
 Equation~\eqref{eq:matRNEA}. In the case of inverse and forward dynamics, the
  numerical advantages are exactly those obtained by the RNEA presented in
   \citep{Featherstone2008} and whose relations to Equation \eqref{eq:matRNEA}
    is discussed in \citep{NoriNav2015}.

\newpage
\section{The Measurements Equation}\label{explanation_y}

Let $\bm y$ $\in \mathbb R^y$ be the vector containing all the available
 sensors readings.  The explicit equation for the measurements is such that
\begin{equation} \label{eq:measEquation}
     \bm Y(\bm q, \dot {\bm q}) \bm d + \bm b_Y (\bm q,\dot {\bm q})= \bm y~.
\end{equation}
The structure of the $\bm Y$ matrix depends on how many sensors are considered
 in the analysis and it is independent from the number of link $N_B$ in the
  model (more sensors could be associated to the same link, e.g., a combination
   of an IMU + a force sensor).  If $i=1:N_S$ number of sensors, thus $\bm Y$
    is a block matrix $\in \mathbb R^{N_S \times d}$
\begin{eqnarray}
 \bm Y =
 \begin{bmatrix}
  \bm Y_1 & \bm Y_2 \quad \hdots \quad \bm Y_{N_S}
  \end{bmatrix}^\top \in \mathbb R^{N_S \times d}~,
\end{eqnarray}
where the dimension of each block depends from the type of sensor.
Similarly, the bias vector $\bm b_Y$ is such that
\begin{eqnarray}
 \bm b_Y =
 \begin{bmatrix}
  \bm b_{Y_1} & \bm b_{Y_2} \quad \hdots \quad \bm b_{Y_{N_S}}
  \end{bmatrix}^\top \in \mathbb R^{N_S}~.
\end{eqnarray}

\subsection{An Illustrative Example}

\begin{figure}[H]
  \centering
    \includegraphics[width=.5\textwidth]{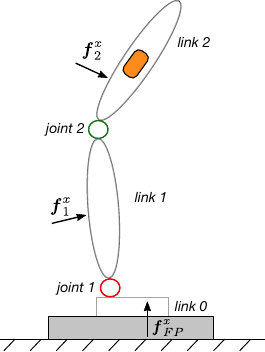}
  \caption{Representation of a $2$-DoF model standing on a force plate (in
   grey), with an IMU (in orange) positioned on link $2$, and external forces
    $\bm f_{FP}^x$, $\bm f_{1}^x$, $\bm f_{2}^x$ acting on it.}
  \label{illustrativeExample}
\end{figure}

Hereafter, an illustrative example on the measurements equation is provided.
Consider the generic $2$-DoF model depicted in Figure
 \ref{illustrativeExample}.  Suppose now that the model is standing on a force
  plate (in rigid contact with link $0$) that provides a measurement of the
   force that the model is exchanging with the ground.  Furthermore, an IMU is
    positioned on link $2$.  $\bm f_{1}^x$ and $\bm f_{2}^x$ represent possible
     external forces acting on the links.
In this specific example
\begin{eqnarray} \label{vector_d_2elements}
\bm d = \begin{bmatrix} \bm d_{1}^\top  & \bm d_{2}^\top
 \end{bmatrix}^\top \in \mathbb R^{52}~,
\end{eqnarray}
and the set of equations related to the IMU and the force plate sensors on the
 fixed base (i.e., $fb$) are, respectively, the following:
\begin{eqnarray}
\label{eqMeas:acc}
\bm y_{2,IMU} &=& \left( \prescript{IMU}{} {\bm X_{2}} \bm {\underline a}_2
 \right)_{lin} + \left( \prescript{IMU}{} {\bm X_{2}} \bm {\underline v}_2
  \right)_{ang}{\times}~\left(\prescript{IMU}{} {\bm X_{2}} \bm {\underline
   v}_2 \right)_{lin}~,\\
\label{eqMeas:fp}
{\bm y}_{{f}_{{FP,fb}}^x} &=&  \prescript{FP}{} {\bm X_{0}^*}
 \left(\prescript{0}{} {\bm
 X_{1}^*} \bm {\underline f}_1 - \bm {\mathrm{\underline I}}_0 {\bm
  {\underline g}}\right)~.
\end{eqnarray}
See Equation \eqref{A:TD_link0} in Appendix \ref{TopDown_BottomUp} for
 retrieving Equation \eqref{eqMeas:fp} , being $\bm
 {\underline a}_0 = - \bm {\underline g}$ and $\bm {\underline v}_0 =
 \bm 0$.

Consider now to define the measurement equation \eqref{eq:measEquation} for
 this example.  The algorithm that solves the human dynamics estimation
  (explained in the following sections) requires mandatorily to specify in the
   vector $\bm y$ the following measurements:
\begin{itemize}
    \item {the linear acceleration for those links where an IMU is
     attached (e.g., $\bm y_{2,IMU}$);}
    \item {the acceleration of all the joints of the model\footnote{Class of
     fictitious DoF-acceleration sensors.} (e.g., $y_{\ddot {q_{1}}}$,
      $y_{\ddot {q_{2}}}$);}
    \item {the force on the fixed base measured by the force plate or
     similar (e.g., ${\bm y}_{{f}_{{FP,fb}}^x}$);}
    \item {the external force acting on all the links of the
     model\footnote{If any external force is acting on a link, the external
      force measurement for that link is a vector $\in \mathbb R^{6}$ with
       entries equal to 0.} (e.g., ${\bm y}_{{f}_{1}^x}$, ${\bm y}_{{f}_{2}^x}$
        ).}
\end{itemize}
This yields to a vector $\bm y$ such that
\vspace{-0.1cm}
\begin{eqnarray} \label{vector_y_illustrativeExample}
\bm y = \begin{bmatrix}
\bm y_{2,IMU} &
y_{\ddot {q_{1}}} &
y_{\ddot {q_{2}}} &
{\bm y}_{{f}_{{FP,fb}}^x} &
{\bm y}_{{f}_{1}^x} &
{\bm y}_{{f}_{2}^x}
 \end{bmatrix}^\top \in \mathbb R^{23}~.
\end{eqnarray}
It is possible now to write down the measurements equation
 \eqref{eq:measEquation} as follows:

\begin{eqnarray} \label{eq:expanded_measEquation}
\begin{bmatrix}
\bm 0 & \bm 0 & \bm 0 & \bm 0 & \bm 0 & \bm 0 &
\left(\prescript{IMU}{}{\bm X_{2}}\right)_{lin} & \hdots & \bm 0 & \bm 0\\
\bm 0 & \bm 0 & \bm 0 & 0 & \bm 0 & 1 & \bm 0 & \hdots & \bm 0 & 0\\
\bm 0 & \bm 0 & \bm 0 & 0 & \bm 0 & 0 & \bm 0 & \hdots & \bm 0 & 1\\
\bm 0 & \bm 0 & \prescript{FP}{} {\bm X_{0}^*} \prescript{0}{} {\bm X_{1}^*} &
\bm 0 & \bm 0 & \bm 0 & \bm 0 & \hdots & \bm 0 & \bm 0\\
\bm 0 & \bm 0 & \bm 0 & \bm 0 & \bm 1_6 & \bm 0 & \bm 0 & \hdots & \bm 0 & \bm
 0\\
\bm 0 & \bm 0 & \bm 0 & \bm 0 & \bm 0 & \bm 0 & \bm 0 & \hdots & \bm 1_6 & \bm
 0\\
\end{bmatrix}
\begin{bmatrix}
\bm {\underline a}_1 \\ \bm {\underline f}_1^B \\ \bm {\underline f}_1 \\
 \tau_{1} \\ \bm {\underline f}_1^x \\  \ddot q_{1} \\ \bm {\underline a}_2
  \\ \bm {\underline f}_2^B \\ \bm {\underline f}_2 \\  \tau_{2} \\ \bm
   {\underline f}_2^x \\  \ddot q_{2}
\end{bmatrix}
+ \notag
\\
+
\begin{bmatrix}
\left( \prescript{IMU}{} {\bm X_{2}} \bm {\underline v}_2
\right)_{ang}{\times}~\left( \prescript{IMU}{} {\bm X_{2}} \bm {\underline
 v}_2 \right)_{lin} \\
0 \\
0 \\
-\prescript{FP}{} {\bm X_{0}^*}~\bm {\mathrm{\underline I}}_0 ~{\bm {\underline
 g}}\\
\bm 0 \\
\bm 0 
\end{bmatrix}
=
\begin{bmatrix}
\bm y_{2,IMU} \\
y_{\ddot {q_{1}}} \\
y_{\ddot {q_{2}}} \\
{\bm y}_{{f}_{{FP,fb}}^x} \\
{\bm y}_{{f}_{1}^x} \\
{\bm y}_{{f}_{2}^x}
\end{bmatrix}~.
\end{eqnarray}
\begin{figure}[H]
  \centering
    \includegraphics[width=.75\textwidth]{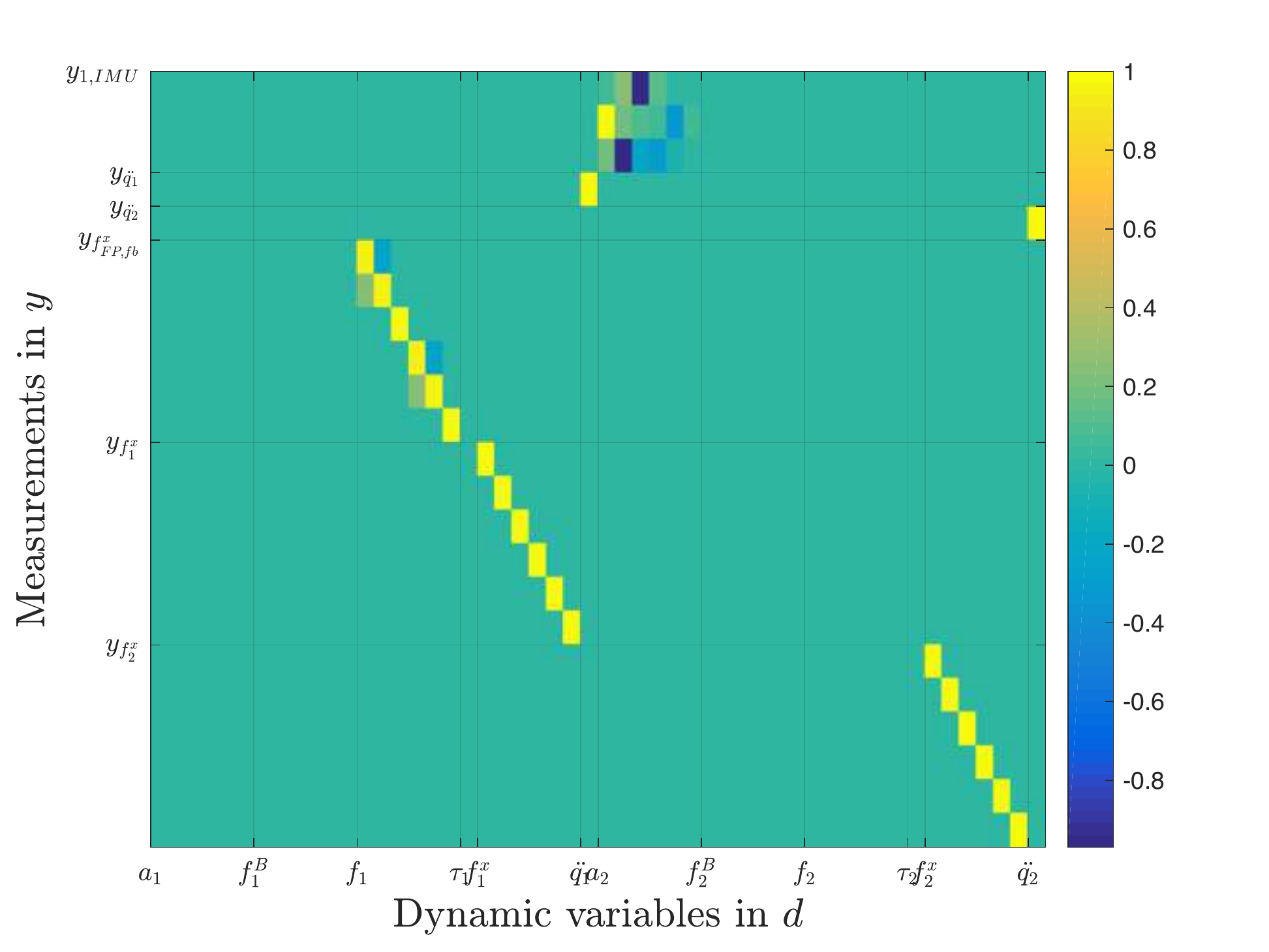}
  \caption{Example of $\bm Y$ matrix for the Equation \eqref{eq:expanded_measEquation}.}
  \label{fig:Figs_plots_dummyYforExample}
\end{figure}

\vspace{0.3cm}
\section{The Estimation Problem Formulation}

By stacking together the Equations \eqref{eq:matRNEA} and
 \eqref{eq:measEquation}, we obtain the following system of equations:

\begin{eqnarray} \label{eq:systemEq}
    \begin{bmatrix}
    \bm Y(\bm{q}, \dot{\bm{q}}) \\ \bm D(\bm{q}, \dot{\bm{q}}) \\
     \end{bmatrix} \bm d +
     \begin{bmatrix} \bm b_Y(\bm{q}, \dot{\bm{q}})
    \\ \bm b_D(\bm{q}, \dot{\bm{q}})
    \end{bmatrix} =
    \begin{bmatrix} \bm y\\
     \bm 0
     \end{bmatrix}~, \qquad
\end{eqnarray}

\begin{eqnarray}\label{rankAssumption}
rank\left(\begin{bmatrix}
    \bm Y(\bm{q}, \dot{\bm{q}}) \\ \bm D(\bm{q}, \dot{\bm{q}}) \\
     \end{bmatrix}\right) = d~.
\end{eqnarray}

\vspace{0.3cm}
\noindent
The bottom part of \eqref{eq:systemEq} represents the constraints equation in
 \eqref{eq:matRNEA}, while the upper part contains the information coming from
  the, possibly noisy or redundant, sensors.  It is possible to compute the
   whole-body dynamics estimation by solving the system in $\bm d$.  The
    assumption on the rank in \eqref{rankAssumption}
    guarantees that the available measurements $\bm y$ give enough constraints
     on $\bm d$.

The novelty of the framework lies in moving away from the classical approach by
 replacing RNEA boundary conditions with measurements coming from sensors.
Within this new framework, there are situations in which the system is
 overdetermined and an exact solution does not exist.  An `approximated'
  solution can be obtained only by making some additional hypothesis.
If there is a valid reason to assume that all the constraints have equal
 relevance and all measurements the same accuracy, one possible approach is to
  solve \eqref{eq:systemEq} in the ordinary least-squares sense, by using a
   Moore-Penrose
   pseudoinverse.  Otherwise, if we have good reasons for weighting
    differently the constraints and the sensors readings, we can use the
     weighted pseudoinverse to obtain a weighted least-squares solution.
      However, finding proper weights might be not an easy task.

In this thesis a different choice is performed: given the prior knowledge of
 the measurements $\bm y$ and the prior
  information about the constraints of the model, the estimation of $\bm d$ is
   framed in a Gaussian domain by means of a maximum-a-posteriori estimator.

\vspace{0.5cm}
\section{Probabilistic Preliminaries}

Since the estimation problem solution is framed in a Gaussian domain, some
 basic notations related to the probability theory are here recalled.

\begin{itemize}
\item Given a stochastic variable $\bm x$, let $p(\bm x)$ be its probability
 density distribution and $p(\bm x | \bm y)$ the conditional probability
  distribution of $\bm x$ given the assumption that another stochastic
   variable $\bm y$ has occurred.
\item Given $\bm x$ and $\bm y$, let $ p(\bm x, \bm y)= p(\bm x) p(\bm y | \bm
 x)$ be their joint probability distribution.
\item  If $\bm y$ is associated to a deterministic function $f(\bm x)$, let
 $E_{\bm x}\big\llbracket f(\bm x)\big\rrbracket$ be the expected value of
  $f(\bm x)$ w.r.t. $p(\bm x)$.
Let $\bm \mu_{\bm x}$ be the mean of $\bm x$, i.e., $\bm \mu_{\bm x} =
 E\big\llbracket\bm x \big\rrbracket$
and $\bm{\Sigma_{\bm x}}$ the covariance associated to $\bm x$, such that $\bm
 \Sigma_{\bm x} = cov \big\llbracket\bm x \big\rrbracket = E \big\llbracket\bm
  x \bm x^\top \big\rrbracket$.
\item Given a multivariate Gaussian distribution $\bm x \in \mathbb R^n$,
 expressed with the notation
\begin{equation}
\bm x \sim \mathcal N(\bm \mu_{\bm x}, \bm \Sigma_{\bm x})~,
\end{equation}
its probability density function (PDF) is
\begin{equation}
p(\bm x)={(2 \pi)^{-\frac{n}{2}} \left | \bm \Sigma_{\bm x} \right
 |^{-\frac{1}{2}} } \exp -\frac{1}{2}\left\{ { \left( \bm x{-}\bm \mu_{\bm x}
  \right)^\top \bm \Sigma_{\bm x}^{-1} \left( \bm x{-}\bm \mu_{\bm x} \right)}
   \right\}~,
\end{equation}
where $\left |\bm \Sigma_{\bm x} \right|$ denotes the determinant of the matrix
 $\bm \Sigma_{\bm x} \in \mathbb R^{n \times n}$.  It is worth noting that when
  in a multivariate normal distribution the covariance $\bm {\Sigma}$ is not a
   full-rank matrix, then the distribution is degenerate and does not have a
    density.  In order to avoid the drawback, it can be useful to restrict the
     problem on a subset of $\bm {\Sigma}$ such that the covariance matrix for
      this subset is positive definite.
\end{itemize}

\newpage
\section{The Maximum-A-Posteriori Solution}

The Maximum-A-Posteriori (MAP) estimator is here adopted as a tool for
 computing an estimation of the vector $\bm d$ and therefore the solution of
  \eqref{eq:systemEq}.  The MAP algorithm can be interpreted as a way for
   choosing a weighted pseudoinverse according to the reliability of the
    available measurements.

The first assumption for adopting the MAP approach is to consider the vector of
 dynamics variables $\bm d$ and the vector of the measurements $\bm y$
as two stochastic variables with Gaussian distributions. The estimation problem
 can be summarized as follows.
\vspace{0.5cm}
\begin{tcolorbox}[sharp corners, colback=white!30,
     colframe=white!20!black!30, 
     title=The problem]
\emph{Given the measurements $\bm y$ and the prior knowledge of the model
 constraints, estimate the vector $\bm d$ by maximizing the conditional
  probability distribution of $\bm d$ given $\bm y$, such that}
\begin{equation}\label{d_map}
   \bm d^{\mbox{\tiny{MAP}}}=\arg \max_{\bm d}~p(\bm d| \bm y) \propto \arg
\max_{\bm d}~p(\bm d, \bm y)~.
\end{equation}
\end{tcolorbox}
\noindent
Since the normal distributions $\bm x$ and $\bm y$ are jointly Gaussian, the
 conditional probability distribution  $p(\bm d|\bm y)$ is such that
\begin{equation}\label{conditional_prob_distribution}
p(\bm d|\bm y) = \frac{p(\bm d, \bm y)}{p(\bm y)} = \frac{ p(\bm d) p(\bm y|\bm
 d)} {p(\bm y)}~.
\end{equation} 
In the following computation, the term $p(\bm y)$ is negligible since it does
 not depend on $\bm d$.  This is the reason of the proportionality between the
  conditional probability distribution and the joint distribution in
   \eqref{d_map}.  Hereafter each term in \eqref{conditional_prob_distribution}
    is computed separately to obtain the final analytical solution.  For the
     sake of simplicity,  $(\bm q, \dot {\bm q})$ dependencies are omitted in
      the computations.

\vspace{0.5cm}
\noindent
\textbf{Computation of $p(\bm y|\bm d)$}
\vspace{0.2cm}

\noindent
Let us first give an expression for the conditional PDF $p (\bm y|\bm d)$:
\begin{eqnarray}\label{pdf:y_given_d}
p(\bm y | \bm d) &\propto& \exp{{-}\frac{1}{2} \left\{ \left(\bm y-\bm
 {\mu}_{y|d}\right)^\top \bm {\Sigma}_{y|d}^{-1} \left(\bm y-\bm
  {\mu}_{y|d}\right)\right\}} \notag \\
& =&  \exp{{-}\frac{1}{2} \left\{\big[\bm y- \left(\bm Y\bm d + \bm b_Y \right) \big]^\top \bm
 {\Sigma}_{y|d}^{-1} \big[\bm y- \left(\bm Y\bm d + \bm b_Y \right) \big]\right\}},
\end{eqnarray} 
which implicitly makes the assumption that the measurements equation
 \eqref{eq:measEquation} is affected by a Gaussian noise with zero mean and
  covariance $ \bm {\Sigma}_{y|d}$. 

\vspace{0.5cm}
\noindent
\textbf{Computation of $p(\bm d)$}
\vspace{0.2cm}

\noindent
Define now a PDF for the normal distribution $\bm d$.  By pursuing the same
 methodology, we would like to write  its distribution in the following form
\begin{equation} \label{distr:dmoddyn_short}
\bm d \sim \mathcal N ({\bm {\mu}}_D,{\bm {\Sigma}}_D)~,
\end{equation}
such that the PDF
\begin{equation}\label{pdf:dmoddyn}
p(\bm d) \propto \exp  -\frac{1}{2}{\left\{\bm e(\bm d)^\top \bm
 {\Sigma}_D^{-1} \bm e(\bm d)\right\}}~,
\end{equation}
taking into account constraints of Equation \eqref{eq:matRNEA} with $\bm e (\bm
 d)=\bm D \bm d+\bm b_D$.

However, this intuitive choice leads to a degenerate normal distribution and a
 regularization term is needed. For example, if we have a Gaussian prior
  knowledge on $\bm d$ in the form of $\bm d \sim \mathcal N \left({\bm
{\mu}}_d,{\bm {\Sigma}}_d\right)$ distribution, we can reformulate Equation
 \eqref{distr:dmoddyn_short} as follows:
\begin{equation} \label{distr:d_short}
\bm d \sim \mathcal N
(\xoverline {\bm {\mu}}_D, \xoverline {\bm {\Sigma}}_D)~,
\end{equation}
such that \eqref{pdf:dmoddyn} becomes
\begin{eqnarray} \label{pdf:d}
p(\bm d) &\propto& \exp -\frac{1}{2}{\left\{\bm e(\bm d)^\top \bm
{\Sigma}_D^{-1} \bm e(\bm d)+(\bm d{-}\bm {\mu}_d)^\top \bm {\Sigma}_d^{-1}
 (\bm d-\bm {\mu}_d) \right\}} \notag\\
&=& \exp{-\frac{1}{2} \left\{ (\bm D\bm d + \bm b_D)^\top \bm {\Sigma}_D^{-1}
 \bm (\bm D\bm d + \bm b_D){+}(\bm d-\bm {\mu}_d)^\top \bm {\Sigma}_d^{-1} (\bm
  d-\bm {\mu}_d)\right\}} \notag \\
&=& \exp{{-}\frac{1}{2} \left\{ \big (\bm d-\xoverline {\bm {\mu}}_D \big )^\top
 \xoverline {\bm {\Sigma}}_D^{-1} \big (\bm d-\xoverline {\bm
  {\mu}}_D \big )\right\}}~,
\end{eqnarray}
where the covariance and the mean are, respectively,
\begin{subequations} 
\begin{eqnarray} 
\label{eq:sigmaDbar}
\xoverline {\bm {\Sigma}}_D &=& \left(\bm D^\top
\bm {\Sigma}_D^{-1} \bm D + \bm {\Sigma}_d^{-1}\right)^{-1}~,
\\
\label{eq:muDbar}
\xoverline {\bm {\mu}}_D &=& \xoverline {\bm {\Sigma}}_D
\left(\bm {\Sigma}_d^{-1} \bm {\mu}_d - \bm D^\top \bm {\Sigma}_D^{-1} \bm b_D
 \right)~. 
\end{eqnarray} 
\end{subequations}
The role of $\bm {\Sigma}_D$ is to establish how much the dynamic model
 \eqref{eq:matRNEA} should be considered correct. The quantities $\bm {\mu}_d$
  and $\bm {\Sigma}_d$, instead, define the Gaussian prior distribution on $\bm
   d$ (namely, the regularization term).

\vspace{0.5cm}
\noindent
\textbf{Computation of $p(\bm d | \bm y)$}
\vspace{0.2cm}

\noindent
By combining Equations \eqref{pdf:y_given_d} and \eqref{pdf:d} we are now ready
 to give a new formulation of the estimation problem for the conditional PDF of
  $\bm d$ given $\bm y$, i.e.,
\begin{eqnarray}
p(\bm d|\bm y) &\propto& \exp {-\frac{1}{2}} \left\{ \big (\bm d-\xoverline {\bm
 {\mu}}_D \big)^\top \xoverline {\bm {\Sigma}}_D^{-1} \big (\bm d-\xoverline
  {\bm {\mu}}_D \big)~+ \right. \notag
\\
&+& \left. \Big[\bm y-(\bm {Yd} + \bm b_Y)\Big]^\top \bm {\Sigma}_{y|d}^{-1}
 \Big [\bm y-(\bm {Yd} + \bm b_Y)\Big] \right\}~,
\end{eqnarray}
with covariance matrix and mean as follows:
\begin{subequations}\label{MAP_solution}
\begin{eqnarray}
\label{eq:sigma_dgiveny}
\bm {\Sigma}_{d|y} &=& \left(\xoverline{ \bm {\Sigma}}_D^{-1} + \bm Y^\top \bm
 {\Sigma}_{y|d}^{-1}\bm Y\right)^{-1}~, 
\\
\label{eq:mu_dgiveny} 
\bm {\mu}_{d|y} &=& \bm {\Sigma}_{d|y} \left[ \bm Y^\top \bm
 {\Sigma}_{y|d}^{-1} (\bm y-\bm b_Y) + \xoverline{\bm {\Sigma}}_D^{-1}
  \xoverline{\bm {\mu}}_D\right]~.
\end{eqnarray} 
\end{subequations}
Moreover, in the Gaussian case the MAP solution coincides with the mean of the
 PDF $p(\bm d|\bm y)$ yielding to:
\begin{eqnarray} \label{eq:d_gaussian}
\bm d^{\mbox{\tiny{MAP}}} = \bm {\mu}_{d|y}~.
\end{eqnarray} 

\subsection{The Cholesky Decomposition} \label{Cholesky_sec}

Computations for \eqref{eq:d_gaussian} can be performed in several ways but
 computationally cost-efficient solutions need to exploit the sparsity in the
  matrices $\bm D$ and $\bm Y$.   The
   solution is here retrieved by applying the
   Cholesky decomposition.  The Cholesky decomposition is mainly used for the
    numerical solution of systems expressed in the $\bm A \bm x = \bm b$ form.
This approach allows to decompose the generic symmetric\footnote{The Cholesky
 decomposition embraces the widest case of Hermitian matrices.  For the MAP
  case all the entries of the matrix $\in \mathbb{R}$, thus the Hermitian
   coincides exactly with its transpose.}, positive-definite matrix $\bm A \in
    \mathbb{R}^{n \times n}$ into a product of a couple of triangular matrices
     (a lower triangular and its transpose), such that $\bm A = \bm L \bm L^T$.

When $\bm A$ is a sparse matrix (i.e., the most of its entries are zero), it is
 convenient to compute a heuristic permutation matrix $\bm P$ for sparsing
  $\bm L$, such that,  $\bm A = \bm P \bm L \bm L^T \bm P^T$.
 This speeds up remarkably the time of
   the execution for computing the system solution and the computational
    cost\footnote{The computational cost is considered as the amount of the
     operations performed during an algorithm execution.
It is a parameter to define how the cost scales with the size $n$ of the
 problem.  For example, in the case of a $n \times n$ matrix related operation,
  the syntax $O(n^3)$ means that the algorithm has a computational complexity
that grows proportionally to $n^3$ (e.g., the complexity of a matrix
 inversion).} decreases passing from a Cholesky to a sparse Cholesky
decomposition ($O(n^3)_{sparse} < O(n^3)_{nonSparse}$).

In the Algorithms \ref{algorithm_cholSol_muDbar} and
 \ref{algorithm_cholSol_mudgiveny}, the pseudocodes for computing the solutions
  \eqref{eq:muDbar} and \eqref{eq:mu_dgiveny} via sparse Cholesky method are
   shown, respectively.  For the sake of simplicity, they were here split into
    two different algorithms.  Clearly, in the code computation they represent
     a whole algorithm since the solution \eqref{eq:muDbar} is mandatorily
      required for computing \eqref{eq:mu_dgiveny}.

\begin{algorithm}[H]
\caption{Compute $\xoverline {\bm {\mu}}_D$, Equation \eqref{eq:muDbar}}
\label{algorithm_cholSol_muDbar}
\begin{algorithmic}[1]
\Require {Pre-computation of the permutation matrix $\xoverline {\bm
 {P}}_D$ }
\Procedure{SolutionWithSparseCholesky}{}
\State $N \gets \text{number of samples}$
\State $cholSolver \gets \textit{Matlab function}$
\BState \emph{loop}:
    \For {$i = 1 \to N~$}
    \State {\emph{$\Rightarrow$ consider $\bm b_i = \left(\bm {\Sigma}_d^{-1}
     \bm{\mu}_d - \bm D^\top \bm {\Sigma}_D^{-1} \bm b_D \right)  ~ \Big|_{i}$}}
    \State {\emph{$\Rightarrow$ solve  $\bm A_i\bm x = \bm b_i$} :}
    \State {$\xoverline {\bm {\mu}}_D   ~ \Big|_{i}= cholSolver
     \left(\xoverline {\bm {\Sigma}}_D^{-1}  ~ \Big|_{i}, \bm b_i,~\xoverline
      {\bm {P}}_D \right)$}
    \State \textbf{goto} \emph{loop}.
    \EndFor
\State {\textbf{end}}
\EndProcedure
\end{algorithmic}
\end{algorithm}
\vspace{1cm}
\begin{algorithm}[H]
\caption{Compute $\bm d^{\mbox{\tiny{MAP}}}$, Equation \eqref{eq:mu_dgiveny}}
\label{algorithm_cholSol_mudgiveny}
\begin{algorithmic}[1]
    \Require {Pre-computation of the permutation matrix $\bm
     {P}_{d|y}$ }
\Procedure{SolutionWithSparseCholesky}{}
\State $N \gets \text{number of samples}$
\State $cholSolver \gets \textit{Matlab function}$
\BState \emph{loop}:
    \For {$i = 1 \to N~$}
    \State {\emph{$\Rightarrow$ consider $\bm b_i = \left[ \bm Y^\top \bm
    {\Sigma}_{y|d}^{-1} (\bm y-\bm b_Y) + \xoverline{\bm {\Sigma}}_D^{-1}
    \xoverline{\bm {\mu}}_D\right]  ~ \Big|_{i}$}}
    \State {\emph{$\Rightarrow$ solve  $\bm A_i\bm x = \bm b_i$} :}
    \State {$\bm {\mu}_{d|y}   ~ \Big|_{i}= cholSolver \left(\bm
    {\Sigma}_{d|y}^{-1}  ~ \Big|_{i}, \bm b_i,~{\bm {P}}_{d|y} \right)$}
    \State \textbf{goto} \emph{loop}.
    \EndFor
\State {\textbf{end}}
\EndProcedure
\end{algorithmic}
\end{algorithm}
\vspace{1cm}

Both the algorithms require the pre-computation of the permutation matrix $\bm
 P$ through the Matlab \texttt{chol} built-in function.  Moreover, it is
  important to note that, even though the algorithm currently does not work in
   this way, in lines $8$, it would be useful to re-order the matrix $\bm
    \Sigma$ (accordingly to the permutation matrix structure) before the
     $cholSolver$ step.

\newpage
\section{Considerations on the Estimator Choice}

In this Section some considerations on the choice of the MAP estimator are
 made.  The accuracy of the estimation heavily depends on the adopted
  estimator and therefore a detailed description is needed.

\subsection{Linear Estimator for Jointly Gaussian Vectors}
 \label{estimatorForGaussianJointlyVectors}

 Let $\bm x$, $\bm y$ be two random Gaussian vectors with distributions
\begin{equation}
\bm x \sim \mathcal N (\bm \mu_x,\bm {\Sigma}_{x})~,
\qquad
\bm y \sim \mathcal N (\bm \mu_y,{\bm {\Sigma}}_{y})~,
\end{equation}
respectively, and jointly Gaussian, i.e.,
\begin{equation} \label{jointly_Gaussian}
\begin{bmatrix}
\bm x\\
\bm y
\end{bmatrix} \sim \mathcal N 
\left(\begin{bmatrix}
\bm \mu_x\\
\bm \mu_y
\end{bmatrix}, 
\begin{bmatrix}
\bm \Sigma_x       &    \bm \Sigma_{xy}\\
\bm \Sigma^T_{xy}  &    \bm \Sigma_y
\end{bmatrix}\right)~,
\end{equation}
where the terms $\bm \Sigma_{xy}$ and $\bm \Sigma^T_{xy}$ represent the jointly
 covariance matrices.

Consider now the following estimation problem.
\vspace{0.2cm}
\begin{tcolorbox}[sharp corners, colback=white!30,
     colframe=white!20!black!30, 
     title=The problem]
\emph{Given two Gaussian vectors $\bm x$ and $\bm y$ jointly Gaussian, estimate
 the value of (the unobserved) $\bm x$ given that we have observed $\bm y$.
  The estimate of $\bm x$ is a function of $\bm y$, i.e., $\hat {\bm x}(\bm y)$
   and the associated estimation error is defined as $\bm e =\bm x - \hat{\bm
    x}(\bm y)$.}
\end{tcolorbox}

\noindent
The time has now come to choose the proper form of the estimator $\hat{\bm
 x}(\bm y)$.
Let us start with considering a minimum mean square error (MMSE) estimator
 $\hat{\bm x}^{\mbox{\tiny{MMSE}}}(\bm y)$ of $\bm x$ that is a function which
  minimizes the mean square error (MSE), i.e., the trace of the error
   covariance matrix, such that
\begin{eqnarray}\label{def_minVarLinEst}
\hat {\bm x}^{\mbox{\tiny{MMSE}}} (\bm y) &=& \arg \min_{\hat{\bm x}}
 \underbrace{Tr \left\{
 E\left\llbracket\big(\bm x - \hat{\bm x}(\bm y) \big) \big(\bm x - \hat{\bm
  x}(\bm y) \big)^\top \right\rrbracket\right\}}_{MSE}  \notag
\\
& =& \arg \min_{\hat{\bm x}} ~ E\left\llbracket{Tr} \big(\bm e \bm e^\top\big)
 \right\rrbracket \notag
\\
& =& \arg \min_{\hat{\bm x}} ~ E\left\llbracket\bm e^\top \bm e
 \right\rrbracket~.
\end{eqnarray} 
See Equation ($17$) of \citep{matrixcookbook} for retrieving Equation
 \eqref{def_minVarLinEst}.

It is shown in \citep{Oppenheim2016} (as well as in many other textbooks of
 probability theory) that the estimator as defined in
  \eqref{def_minVarLinEst} coincides with the expectation of the
   conditional distribution
\begin{equation}\label{distr:x_given_y}
\bm x|\bm y \sim \mathcal N (\bm \mu_{x|y},\bm {\Sigma}_{x|y})~,
\end{equation}
such as
\begin{eqnarray} \label{eq:minVar_equals_condExp}
\hat {\bm x}^{\mbox{\tiny{MMSE}}} (\bm y) = \bm {\mu}_{x|y}~.
\end{eqnarray}
For the Gauss-Markov theorem \citep{Nowak2011}, if $\bm x$ and $\bm y$ are
 jointly Gaussian random vectors with the distribution in
  \eqref{jointly_Gaussian}, then the conditional distribution of $\bm x$ given
   $\bm y$ is 
\begin{equation}\label{distr:x_given_y_GM}
\bm x|\bm y \sim \mathcal N (\bm \mu_x + \bm \Sigma_{xy}\bm \Sigma^{-1}_{y}
 \big(\bm y - \bm \mu_y\big),\bm \Sigma_x - \bm \Sigma_{xy}\bm \Sigma^{-1}_y
  \bm\Sigma^T_{xy})~.
\end{equation}
This yields to a new formulation that identifies the estimator , such that
\begin{subequations}\label{eq:newDef_estimator_MMSE}
\begin{eqnarray}
\label{mean_estimator}
\hat {\bm x}^{\mbox{\tiny{MMSE}}} (\bm y) &=& \bm \mu_x + \bm \Sigma_{xy}\bm
 \Sigma^{-1}_{y}(\bm y - \bm \mu_y)~,
\\
\label{variance_estimator}
cov \Big\llbracket\bm x - \hat{\bm x}^{\mbox{\tiny{MMSE}}} (\bm y)
 \Big\rrbracket &=& \bm
 \Sigma_x - \bm \Sigma_{xy}\bm \Sigma^{-1}_y \bm\Sigma^T_{xy}~,
\end{eqnarray}
\end{subequations}
where \eqref{mean_estimator} is the mean and \eqref{variance_estimator} is the
covariance associated to the estimator.
According to the Gauss-Markov theorem, Equation \eqref{mean_estimator} is a
 linear function of $\bm y$ and in this case the MMSE estimator  is said to be
  linear.
In many cases, it is difficult to determine the analytical expression of the
 MMSE estimator.  A widely used tradeoff is to restrict the estimator to be a
  linear (i.e., linear plus a constant) function of $\bm y$, and to choose the
   linear relationship in order to minimize the MSE \citep{Oppenheim2010}. The
    resulting estimator is called the linear minimum mean square error (LMMSE)
     estimator.

\subsection{Linear Estimator for Linear Regressors}
 \label{estimatorForLinearRegressor}

Let us suppose now to consider the following linear regressor model:
\begin{eqnarray} \label{eq:genericRegressorModel}
{\bm y} = \bm C \bm x + \bm e~,
\end{eqnarray}
where
\begin{equation}
\bm x \sim \mathcal N \left(\bm \mu_x,\bm C \bm {\Sigma}_{x} \bm C^T \right)~,
\qquad
\bm e \sim \mathcal N (\bm 0,{\bm {\Sigma}}_{e})~,
\end{equation}
are the distributions of the Gaussian variable $\bm x$ and of the error $\bm
 e$, respectively.  This straightforwardly yields to
\begin{equation}
\bm y \sim \mathcal N \left(\bm C \bm \mu_x, \bm C \bm {\Sigma}_{x} \bm C^T +
 {\bm {\Sigma}}_{e} \right)~,
\end{equation}
being $\bm \mu_y = \bm C \bm \mu_x$ and $\bm {\Sigma}_{y} = \bm C \bm
 {\Sigma}_{x} \bm C^T + {\bm {\Sigma}}_{e}$ the mean and the covariance of the
  regressor model, respectively.  Also in this case we can consider the same
   jointly Gaussian situation as in \eqref{jointly_Gaussian}, where now:
\begin{equation}
\bm {\Sigma}_{xy} = cov \Big\llbracket\bm x, \bm C \bm x \Big\rrbracket
= cov \Big\llbracket\bm x, \bm x \Big\rrbracket \bm C^T
= E \Big\llbracket \left(\bm x \bm x^T \right) - \bm \mu_x \bm \mu_x^T
 \Big\rrbracket\bm C^T = \bm {\Sigma}_{x} \bm C^T~.
\end{equation}
Given that, the set of Equations \eqref{eq:newDef_estimator_MMSE} become
\begin{subequations}\label{eq:newDef_estimator_LMMSE}
\begin{eqnarray}
\label{mean_estimator_LMMSE}
\hat {\bm x}^{\mbox{\tiny{LMMSE}}} (\bm y)&=&~~\bm \mu_x + \bm {\Sigma}_{x} \bm
 C^T \big( \bm C
 \bm {\Sigma}_{x} \bm C^T + {\bm {\Sigma}}_{e} \big)^{-1}
\big (\bm y - \bm C\bm \mu_x \big)~,\notag \\
\\
\label{covariance_estimator_LMMSE}
cov \Big\llbracket\bm x - \hat{\bm x}^{\mbox{\tiny{LMMSE}}} (\bm y)
 \Big\rrbracket&=&~~\bm
 \Sigma_x - \bm {\Sigma}_{x} \bm C^T  \big( \bm C \bm {\Sigma}_{x} \bm C^T +
  {\bm {\Sigma}}_{e} \big)^{-1} \bm C \bm\Sigma_{x}~.
\end{eqnarray} 
\end{subequations}
This formulation is useful to understand what happens to the solution when the
 variable $\bm x$ (i.e., the prior in our algorithm) is very reliable.  This
  yields to $ \bm {\Sigma}_{x} \approx \bm 0$ and therefore to
\begin{subequations}\label{eq:solution_priorReliable}
\begin{eqnarray}
\label{mean_estimator__priorReliable}
\hat {\bm x}^{\mbox{\tiny{LMMSE}}} (\bm y) &\approx& \bm \mu_x~,
\\
\label{variance_estimator__priorReliable}
cov \Big\llbracket\bm x - \hat{\bm x}^{\mbox{\tiny{LMMSE}}} (\bm y)
 \Big\rrbracket &\approx& \bm 0~.
\end{eqnarray}
\end{subequations}

Another equivalent representation can be provided by applying the Woodbury
 identities:
\begin{itemize}
\item to \eqref{mean_estimator_LMMSE} (see Equation ($156$) in
 \citep{matrixcookbook} or in
 \citep{Golub_book}, Chapter $2.1.4$) 
\begin{equation} \label{application_WoodburyID_cookbook158}
\bm {\Sigma}_{x} \bm C^T \big( \bm C \bm {\Sigma}_{x} \bm C^T + {\bm
 {\Sigma}}_{e} \big)^{-1} = \big(\bm {\Sigma}_{x}^{-1} + \bm C^T \bm
  {\Sigma}_{e}^{-1} \bm C\big)^{-1} \bm C^T \bm {\Sigma}_{e}^{-1}~,
\end{equation}
\item to \eqref{covariance_estimator_LMMSE} (see Equation ($158$) in
 \citep{matrixcookbook}) 
\begin{equation} \label{application_WoodburyID_cookbook156}
\bm \Sigma_x - \bm {\Sigma}_{x} \bm C^T  \big( \bm C \bm {\Sigma}_{x} \bm C^T +
 {\bm {\Sigma}}_{e} \big)^{-1} \bm C \bm\Sigma_{x} = \big(\bm {\Sigma}_{x}^{-1}
  + \bm C^T \bm {\Sigma}_{e}^{-1} \bm C\big)^{-1}~.
\end{equation}
\end{itemize}
Given \eqref{application_WoodburyID_cookbook158} and
 \eqref{application_WoodburyID_cookbook156}, Equations
 \eqref{eq:newDef_estimator_LMMSE} can be written as
\begin{subequations}\label{eq:newDef_estimator_lin_regressorModel_eqiv}
\begin{eqnarray}
\label{mean_estimator_lin_regressorModel_equiv}
\hat {\bm x}^{\mbox{\tiny{LMMSE}}} (\bm y) &=&~~ \bm \mu_x + \big(\bm
 {\Sigma}_{x}^{-1} + \bm
 C^T \bm {\Sigma}_{e}^{-1} \bm C\big)^{-1} \bm C^T \bm {\Sigma}_{e}^{-1}
\big(\bm y - \bm C\bm \mu_x \big) \notag
\\
&=&\left[ \big(\bm {\Sigma}_{x}^{-1} + \bm C^T \bm {\Sigma}_{e}^{-1} \bm
 C\big)^{-1} \bm C^T \bm {\Sigma}_{e}^{-1}\right]\bm y~+ \notag
\\
&+& \left[ \bm 1 -  \big(\bm {\Sigma}_{x}^{-1} + \bm C^T \bm {\Sigma}_{e}^{-1}
 \bm C\big)^{-1} \bm C^T \bm {\Sigma}_{e}^{-1} \bm C \right]\bm \mu_x \notag
\\
&=&~\big(\bm {\Sigma}_{x}^{-1} + \bm C^T \bm {\Sigma}_{e}^{-1} \bm C\big)^{-1}
\big(\bm C^T \bm {\Sigma}_{e}^{-1} \bm y + \bm {\Sigma}_{x}^{-1} \bm \mu_x
 \big)~,\notag \\
\\
\label{variance_estimator_lin_regressorModel_equiv}
cov \Big\llbracket\bm x - \hat{\bm x}^{\mbox{\tiny{LMMSE}}} (\bm y) \Big\rrbracket &=&~~
 \big(\bm {\Sigma}_{x}^{-1}
  + \bm C^T \bm {\Sigma}_{e}^{-1} \bm C\big)^{-1}~.
\end{eqnarray} 
\end{subequations}
Equation \eqref{mean_estimator_lin_regressorModel_equiv} clearly shows that the
 solution is the weighted sum of $\bm \mu_x$ (i.e., the mean of the prior in
  our algorithm) and the variable $\bm y$ (i.e., the measurements vector in our
   algorithm).
The second formulation is convenient to understand what happens in case of $\bm
 x$ is very unreliable ($\bm {\Sigma}_{x}^{-1} \approx \bm 0$). In this case,
  the solution is mainly relying on $\bm y$, such that
\begin{subequations}\label{eq:solution_priorNotReliable}
\begin{eqnarray}
\label{mean_estimator__priorNotReliable}
\hat {\bm x}^{\mbox{\tiny{LMMSE}}} (\bm y) &\approx& \big(\bm C^T \bm
 {\Sigma}_{e}^{-1}\bm C\big)^{-1} \bm C^T \bm {\Sigma}_{e}^{-1}\bm y~,
\\
\label{variance_estimator__priorNotReliable}
cov \left\llbracket\bm x - \hat{\bm x}^{\mbox{\tiny{LMMSE}}} (\bm y)
 \right\rrbracket &\approx&
\big(\bm C^T \bm {\Sigma}_{e}^{-1} \bm C\big)^{-1} ~.
\end{eqnarray}
\end{subequations}

\subsection{Generalized Least-Squares Estimator}

In this Section it is shown that the linear estimator
solution of the problem in Section \ref{estimatorForGaussianJointlyVectors}
 coincides with the solution of the generalized\footnote{The generalized
  least-square problem is a weighted least-squares where $\bm \Sigma$ is a
   non-diagonal matrix, but still symmetric and positive-definite.}
    least-squares (GLS) problem.

Define the following GLS problem for the system in
 the form  $\bm A \bm x = \bm b$:

\begin{eqnarray}\label{argmin_gen_xLS}
\hat{\bm x}^{\mbox{\tiny{GLS}}}(\bm y) &=&~\arg \min_x \left\| \bm A \bm x - \bm
 b \right\|^2_{\bm \Sigma}\notag \\
&=&~\arg \min_x (\bm A \bm x - \bm b)^T \bm \Sigma (\bm A \bm x - \bm b)~.
\end{eqnarray}
Thanks to Cholesky factorization (see Section \ref{Cholesky_sec}), if $\bm
 \Sigma = \bm \Sigma^T$ (symmetric\footnote{Indeed here it is required an
 Hertian matrix $\bm \Sigma = \bm \Sigma^\ast$, but the case of only real
  entries yields to $\bm \Sigma^\ast = \bm \Sigma^T$.}) and $\bm \Sigma
   \textgreater 0$ (positive-definite), it is always possible the following
    decomposition:
\begin{equation}
\bm \Sigma = \bm U^T \bm U~,
\end{equation}
\noindent
where $\bm U$ is an upper triangular matrix.  Thus in
 \eqref{argmin_gen_xLS},
\begin{eqnarray}\label{genLSproblem_solution}
\hat{\bm x}^{\mbox{\tiny{GLS}}}(\bm y) &=& \arg \min_x (\bm A \bm x - \bm b)^T
 \bm U^T \bm U (\bm A \bm x - \bm b) \notag\\
&=& \arg \min_x (\bm U \bm A \bm x - \bm U \bm b))^T (\bm U \bm A \bm x - \bm U
 \bm b)) \notag\\
&=& (\bm A^T \bm U^T \bm U \bm A)^{-1} \bm A^T \bm U^T \bm U \bm b \notag\\
&=& (\bm A^T \bm \Sigma \bm A)^{-1} \bm A^T \bm \Sigma \bm b~.
\end{eqnarray}
Consider in our specific case
$\bm A = \begin{bmatrix}
\bm C \\ \bm 1
\end{bmatrix}$ and
$\bm b = \begin{bmatrix}
\bm y \\ \bm \mu_x
\end{bmatrix}$ , such that in \eqref{argmin_gen_xLS}
\begin{eqnarray} \label{x_LMMSE=x_GLS}
\hat{\bm x}^{\mbox{\tiny{GLS}}}(\bm y) &=&~\arg \min_x \left\| \begin{bmatrix}
\bm C \\ \bm 1
\end{bmatrix} \bm x - \begin{bmatrix}
\bm y \\ \bm \mu_x
\end{bmatrix} \right\|^2_{diag \left({\bm {\Sigma}_{e}^{-1},~ \bm
 {\Sigma}_{x}^{-1}}\right)} \notag
\\
&=&~ \arg \min_x \Big(\left\| \bm C \bm x - \bm y  \right\|^2_{\bm
 \Sigma_e^{-1}} + \left\| \bm x - \bm \mu_x  \right\|^2_{\bm \Sigma_x^{-1}}\Big)
\notag
\\
&=&~\left(\begin{bmatrix} \bm C & \bm 1 \end{bmatrix}
\begin{bmatrix} \bm \Sigma_e^{-1} & \bm 0 \\ \bm 0 & \bm \Sigma_x^{-1}
 \end{bmatrix}
\begin{bmatrix} \bm C \\ \bm 1 \end{bmatrix} \right)^{-1}
\begin{bmatrix} \bm C & \bm 1 \end{bmatrix}
\begin{bmatrix} \bm \Sigma_e^{-1} & \bm 0 \\ \bm 0 & \bm \Sigma_x^{-1}
\end{bmatrix} \begin{bmatrix} \bm y \\ \bm \mu_x \end{bmatrix} \notag
\\
&=&~~ \big(\bm {\Sigma}_{x}^{-1} + \bm C^T \bm {\Sigma}_{e}^{-1} \bm C\big)^{-1}
\big(\bm C^T \bm {\Sigma}_{e}^{-1} \bm y + \bm {\Sigma}_{x}^{-1} \bm \mu_x \big)
\notag \\
&{\equiv}&~~ \hat {\bm x}^{\mbox{\tiny{LMMSE}}}(\bm y)~.
\end{eqnarray}

\vspace{-0.5cm}
\subsection{The MAP Estimator}
\vspace{-0.2cm}
The solution \eqref{x_LMMSE=x_GLS} is written in the form of the MAP solution
 \eqref{MAP_solution}.  Thus solving the problem in \eqref{d_map} is equivalent
  to solve a generalized least-squares problem \eqref{x_LMMSE=x_GLS}.  Because
   of this, the MAP problem can be written in the following form:
\begin{eqnarray} \label{MAP_as_LS_1stFormulation}
\bm d^{\mbox{\tiny{MAP}}} &=&~\arg \min_d  \left\|\begin{bmatrix}
\bm D \\ \bm Y \\ \bm 1
\end{bmatrix} \bm d -
\begin{bmatrix}
\bm b_D \\ \bm y - \bm b_Y \\ \bm \mu_d
\end{bmatrix}
\right\|^2_{diag\left(\bm \Sigma_D^{-1},~ \bm \Sigma_{y|d}^{-1},~
{\bm \Sigma_d^{-1}} \right)} \notag 
\\
&=&~ \arg \min_d \Big(\left\| \bm D \bm d - \bm b_D \right\|^2_{\bm
 \Sigma_D^{-1}} + \left\| \bm Y \bm d - (\bm y - \bm b_Y)  \right\|^2_{\bm
  \Sigma_{y|d}^{-1}} + \left\| \bm d - \bm \mu_d  \right\|^2_{\bm
   \Sigma_d^{-1}} \Big)~.\notag \\
\end{eqnarray}
The importance of the Equation \eqref{MAP_as_LS_1stFormulation} is two-fold:
\begin{itemize}
\item[$i$)] it shows clearly the role of each part in the estimation of $\bm
 d$: the contribution of the dynamic constraints weighted by $\bm
  \Sigma_D^{-1}$, the contribution coming from the sensor readings weighted by
   $\bm \Sigma_{y|d}^{-1}$ and the contribution due to the prior on $\bm d$
    weighted by $\bm \Sigma_d^{-1}$;
\item[$ii$)] it provides the form of $\bm A$ and $\bm b$ for the solution in
 \eqref{argmin_gen_xLS}:
\begin{eqnarray}\label{AandB_form}
\bm A = \begin{bmatrix}
\bm D \\ \bm Y\\\bm 1
\end{bmatrix},
\quad
\bm b = \begin{bmatrix}
\bm b_D \\ \bm y -\bm b_Y \\ \bm \mu_d
\end{bmatrix}~.
\end{eqnarray}
\end{itemize}
If we prefer, Equation \eqref{MAP_as_LS_1stFormulation} can be written in the
 following form:
\begin{eqnarray}\label{MAP_as_LS_2ndFormulation}
\bm d^{\mbox{\tiny{MAP}}} =\arg \min_{d} \left\{\left\| \bm Y\bm d - (\bm y -
 \bm b_Y)  \right\|^2_{\bm \Sigma_{y|d}^{-1}}+\left\| \bm d - \xoverline{\bm
  \mu}_D  \right\|^2_{\xoverline{ {\bm \Sigma}}_D^{-1}} \right\}~,
\end{eqnarray}
where $\xoverline{\bm \mu}_D$ can be seen as a regularization term in
 \emph{shaping} the measurements equation.

\chapter{Software Implementation and Algorithm Validation}  %
\label{chapter_implementationANDvalidation}
\begin{quotation}
\noindent\emph{
\begin{itemize}
\item[-]{If an experiment works, something has gone wrong.}
\item[-]{In any collection of data, the figure most obviously correct, beyond
 all need of checking, is the mistake.}
\end{itemize}
}
    \begin{flushright}
        Finagle's laws, Arthur Block
    \end{flushright}
\end{quotation}

\vspace{0.2cm}
\noindent
This Chapter presents the software implementation of the MAP solution
 and several offline validation procedures to assess the effectiveness of the
 presented estimation algorithm.  Since the sensors play a paramount role in the
  analysis, they will be detailed described in a dedicated section.
A new type of wearable force sensing is also introduced in this Chapter and an
 investigation for its validation w.r.t. the force gold standard is
  provided to the reader.

\vspace{-0.2cm}
\section{MAP Software Implementation}

\begin{figure}[h]
  \centering
    \includegraphics[width=0.95\textwidth]{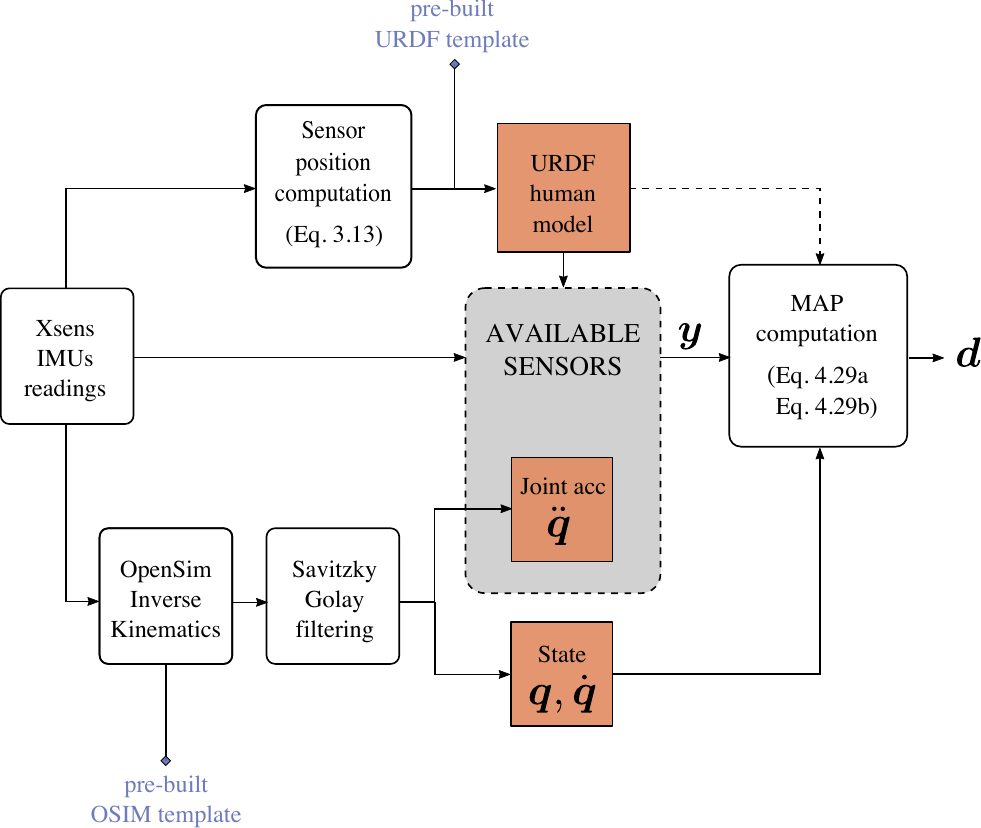}
  \caption{Pipeline for the MAP offline estimation framework.}
  \label{fig:Figs_schemeAlgorithm_generic_pipeline}
\end{figure}

In this Section the software architecture of the MAP framework is presented
 (see pipeline in Figure \ref{fig:Figs_schemeAlgorithm_generic_pipeline}).  It
  was implemented in Matlab and created specifically for offline validations.
The code related to the MAP offline computation is hosted on GitHub in a
 dedicated public repository \citep{MAPest_repo}.
The starting point of the analysis consists in the IMUs readings.  Kinematic
 data are parsed into a compatible Matlab format and then processed
 as in the following steps.

\begin{itemize}
\item{Computation of the position of each sensor in the suit w.r.t. the
attached link.  Xsens does not provide directly this information as output,
therefore it is necessary to compute it through the estimation Algorithm
\ref{algorithm_sensPos} (Chapter \ref{Chapter_human_modelling}).  This step is
 mandatory since sensors position
has to be encoded in the non-standard extension of the URDF (see Section
 \ref{nonStandardURDF} of Appendix \ref{URDF_human_modelling_appendix}).
It is worth
remarking that, at this stage, a URDF template exists, built according to
the human whole-body modelling in Figure \ref{human_urdfModel_noSens}.
The dimension of the URDF elements (cylinders, parallelepipeds and
spheres) is estimated from IMUs data by making in this way the model
scalable with the subject proportions.  This is possible thanks to the Xsens
trajectories acquisition of `fictitious' markers positioned at several
 anatomical known bony landmarks.}
\item{Computation of the Inverse Kinematics (IK) by means of the OpenSim
 specific toolbox \citep{opensim_site} \citep{Delp2007} that solves a
 weighted least-squares problem for obtaining the joint angles $\bm{q}$.  This
  is possible thanks to $i)$ an OpenSim (OSIM) template previously built in
   order to match with the URDF model and $ii)$ a .trc file containing the
    markers trajectories acquired by the Xsens system, (see Appendix \ref{IK}).
      Joint velocities $\dot{\bm{q}}$ and
accelerations $\ddot{\bm{q}}$ are computed by using a weighted sum of moving
windows of elements with a third-order polynomial Savitzky-Golay filtering
\citep{SavitzkyGolay1964}}.
\item{Extraction of the IMUs linear acceleration readings  for filling the
 measurements vector $\bm y$.}
\end{itemize}

All the above-listed points are used to obtain
  Equation \eqref{eq:measEquation} for the measurements.  They are typically
   kinematic readings,
   force sensing data (e.g., the ground reaction forces but in general the
    external forces acting on the model links) and joint accelerations that are
     here considered as acquired from a class of `fictitious' DoF-acceleration
      sensors.
By exploiting the information coming from the model and the state, the pipeline
 block of the MAP computation provides the estimation of $\bm d$ given $\bm y$
  (see Algorithm \ref{algorithm_cholSol_mudgiveny}, Chapter
   \ref{Chapter_estimation_problem}).
It is worth to highlight that Figure
 \ref{fig:Figs_schemeAlgorithm_generic_pipeline}
 does not refer to the temporal aspect of the data flow.  For a better
  understanding of the steps sequentiality refer to the Algorithm
   \ref{algorithm_MAPoffline}, where a situation with $S$ subjects performing
    $T$ different tasks is considered.

\vspace{0.7cm}
\begin{algorithm}[H]
\caption{MAP offline estimation, Figure
 \ref{fig:Figs_schemeAlgorithm_generic_pipeline}}
\label{algorithm_MAPoffline}
\begin{algorithmic}[1]
\Require {IMUs Xsens acquisition, URDF and OSIM templates}
\Procedure{MAPofflineProcedure}{}
\State $S \gets \text{number of subjects}$
\State $T \gets \text{number of tasks}$
\BState \emph{main loop}:
\For {$s = 1 \to S~$}
\BState \emph{nested loop}:
    \For {$t = 1 \to T~$}
    \State {\emph{$\Rightarrow$ parse Xsens data} : suit$^s\{$t$\}$}
    \State {\emph{$\Rightarrow$ compute sensor position} :
     suit$^s\{$t$\}$.sensor $~$\textbf{goto} Algorithm \ref{algorithm_sensPos}}
     \State {\emph{$\Rightarrow$ create URDF model} : model$^s\{$t$\}$.urdf}
     \State {\emph{$\Rightarrow$ create OSIM model} : model$^s\{$t$\}$.osim}
    \State {\emph{$\Rightarrow$ compute IK} : ${\bm{q}}^{s}\{$t$\}$}
    \State {\emph{$\Rightarrow$ compute Savitzky-Golay filtering} :
    ${\dot{\bm{q}}}^{s}\{$t$\}$, ${\ddot{\bm{q}}}^{s}\{$t$\}$}
    \State {\emph{$\Rightarrow$ wrap available measurements in} $\bm
     y^s\{$t$\}$}
    \State {\emph{$\Rightarrow$ compute MAP} :
     $\bm d^s\{$t$\}$ $~$\textbf{goto} Algorithm
      \ref{algorithm_cholSol_mudgiveny}}
     \EndFor
     \State \textbf{goto} \emph{nested loop}.
     \State {\textbf{end}}
\EndFor
\State \textbf{goto} \emph{main loop}.
\State {\textbf{end}}
\EndProcedure
\end{algorithmic}
\end{algorithm}

\section{Experimental Sensor Setup}

\subsection{Motion Capture}

\begin{figure}[h]
  \centering
    \begin{subfigure}[b]{0.18\textwidth}
      \includegraphics[width=\textwidth]{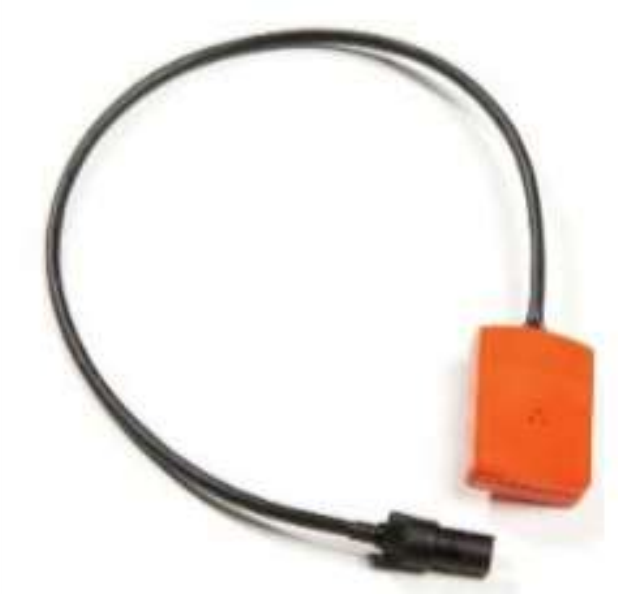}
    \caption{}
    \label{fig:wiredIMU}
          \end{subfigure}
    \qquad
    \begin{subfigure}[b]{0.18\textwidth}
      \includegraphics[width=\textwidth]{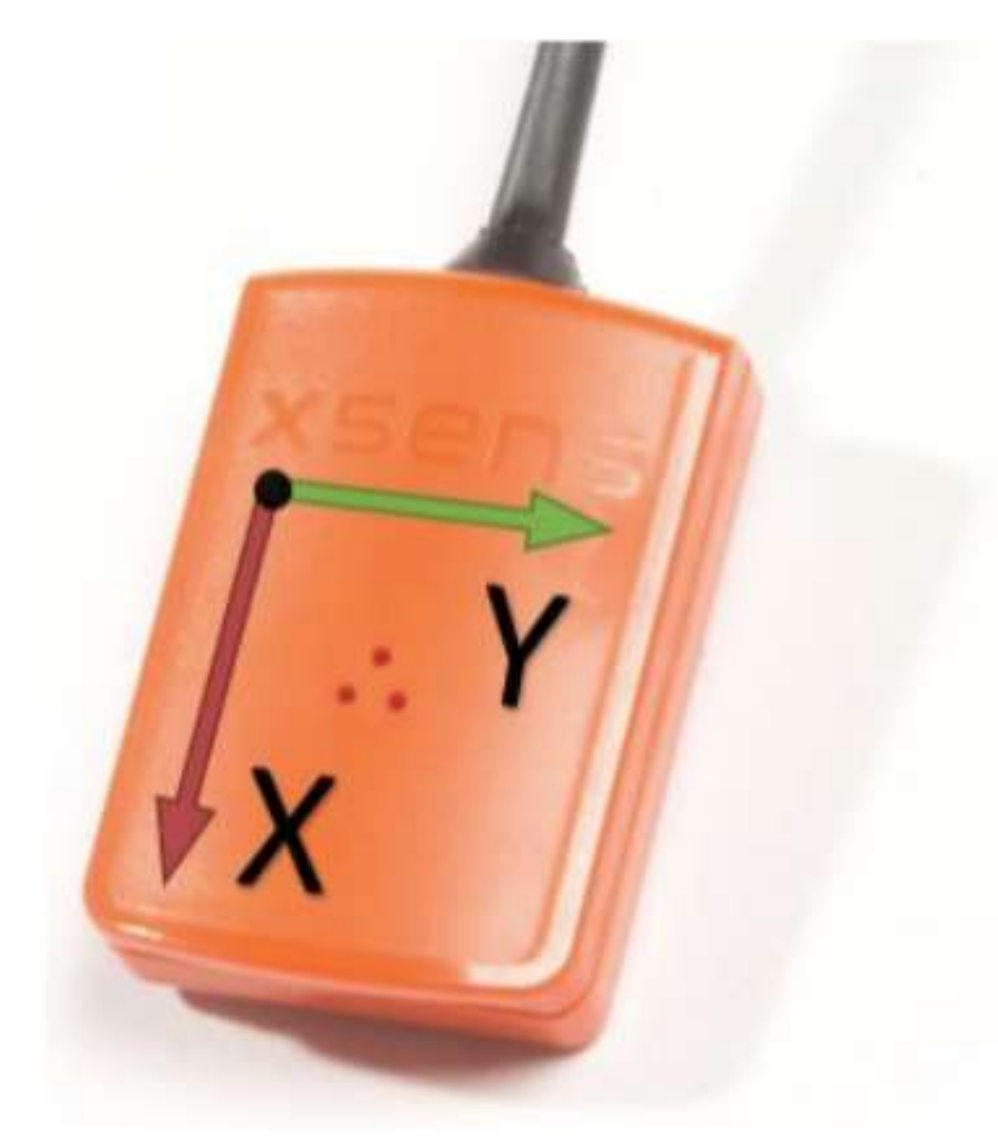}
    \caption{}
    \label{fig:IMUref}
          \end{subfigure}
    \begin{subfigure}[b]{0.76\textwidth}
    \includegraphics[width=\textwidth]{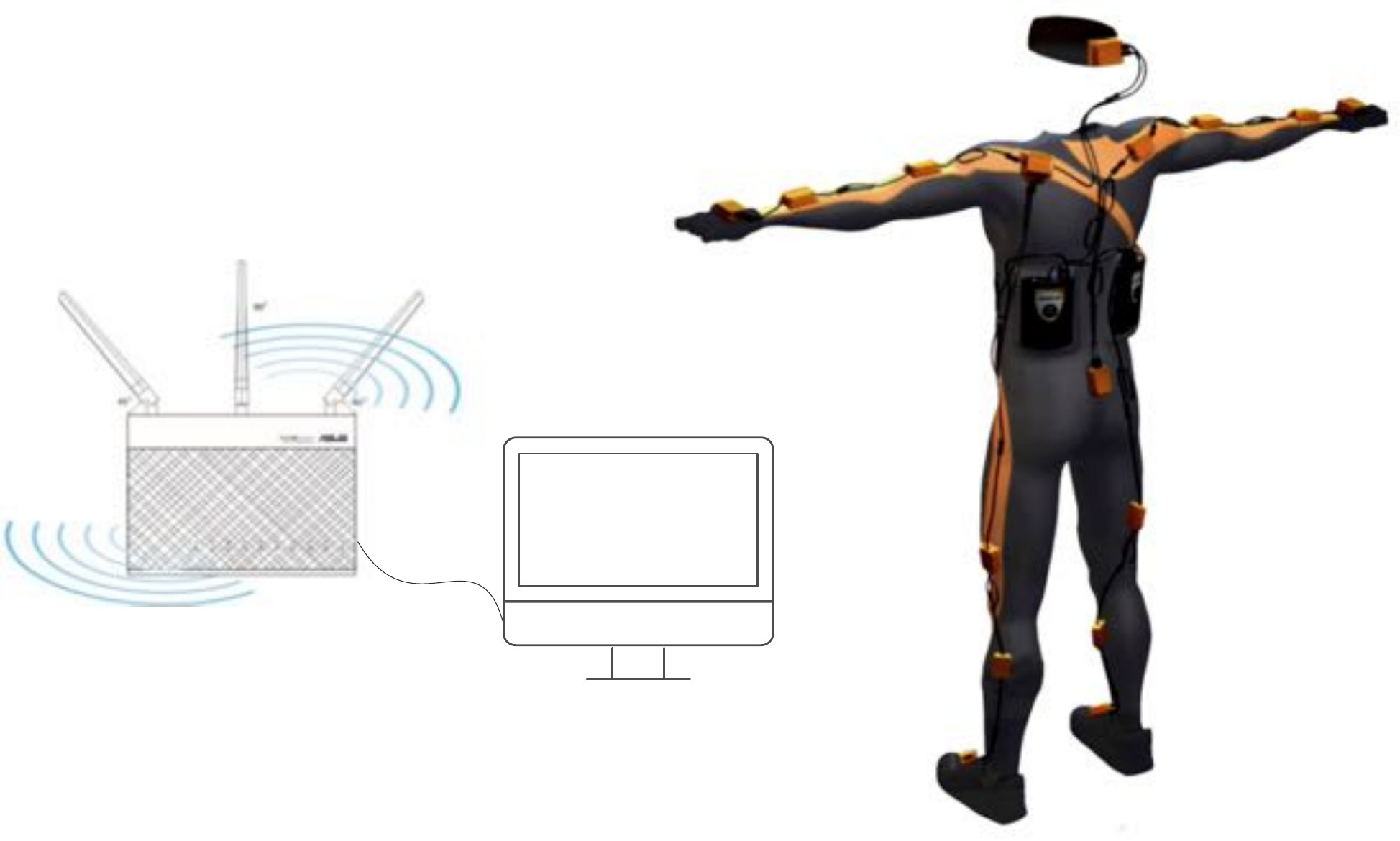}
    \caption{}
  \label{fig:Figs_Xsens_setup}
          \end{subfigure}
\caption{(\subref{fig:wiredIMU}) An example of a wired IMU.
 (\subref{fig:IMUref}) Reference frame of the IMU.
 (\subref{fig:Figs_Xsens_setup}) Xsens motion capture setup: $17$-wired IMUs
 embedded in a wearable
   lycra suit, an Access Point connected to a laptop used for the data
    acquisition.}
    \source{Xsens.}
\end{figure}

The full-body motion capture tracking of the human is retrieved by a Xsens
 wearable lycra suit in which $17$ wired body-mounted IMUs (Figures
  \ref{fig:wiredIMU}, \ref{fig:IMUref}) are located in the upper legs, lower
   legs, upper arms, lower arms, the pelvis, the sternum, the shoulders and the
   head.
All the IMUs are connected to the Xbus Masters (mounted on the back of the
 subject) that is in charge of $i)$ synchronizing the sensor readings, $ii)$
  providing them with power and $iii)$ handling the wireless communication
   with a laptop.
The wireless connection is guaranteed by an external Access Point connected via
 Ethernet to a laptop. See Figure \ref{fig:Figs_Xsens_setup} for the Xsens
  motion capture tracking setup.
The output of this system is a XML-like file with the kinematic variables of
 the subject who is wearing the suit, according with the Xsens body modelling
  (Figure \ref{figs:Xsens_model}).

\subsection{Ground Reaction Forces Tracking}

\begin{figure}[H]
  \centering
  \begin{subfigure}[b]{0.21\textwidth}
    \includegraphics[width=\textwidth]{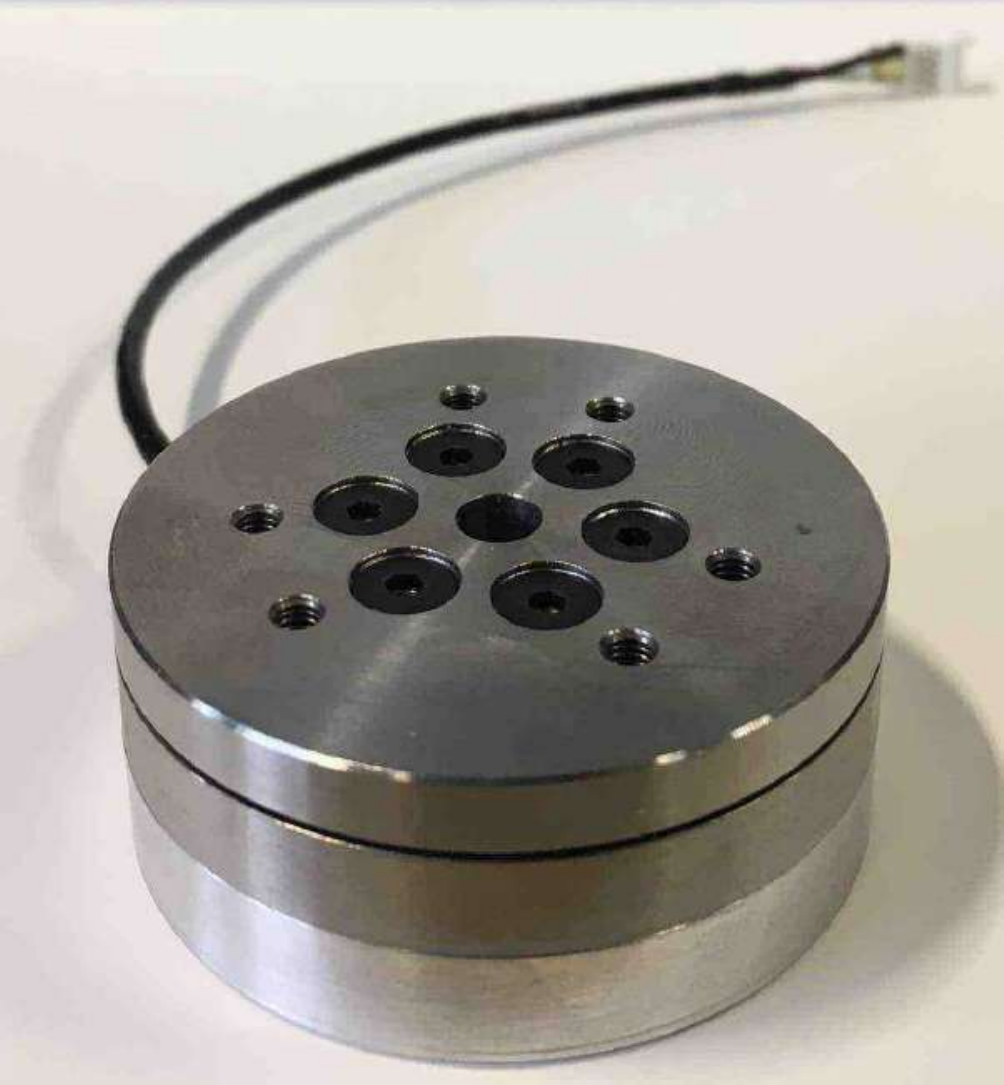}
  \caption{}
  \label{fig:iCubFT}
        \end{subfigure}
~\quad
  \begin{subfigure}[b]{0.28\textwidth}
    \includegraphics[width=\textwidth]{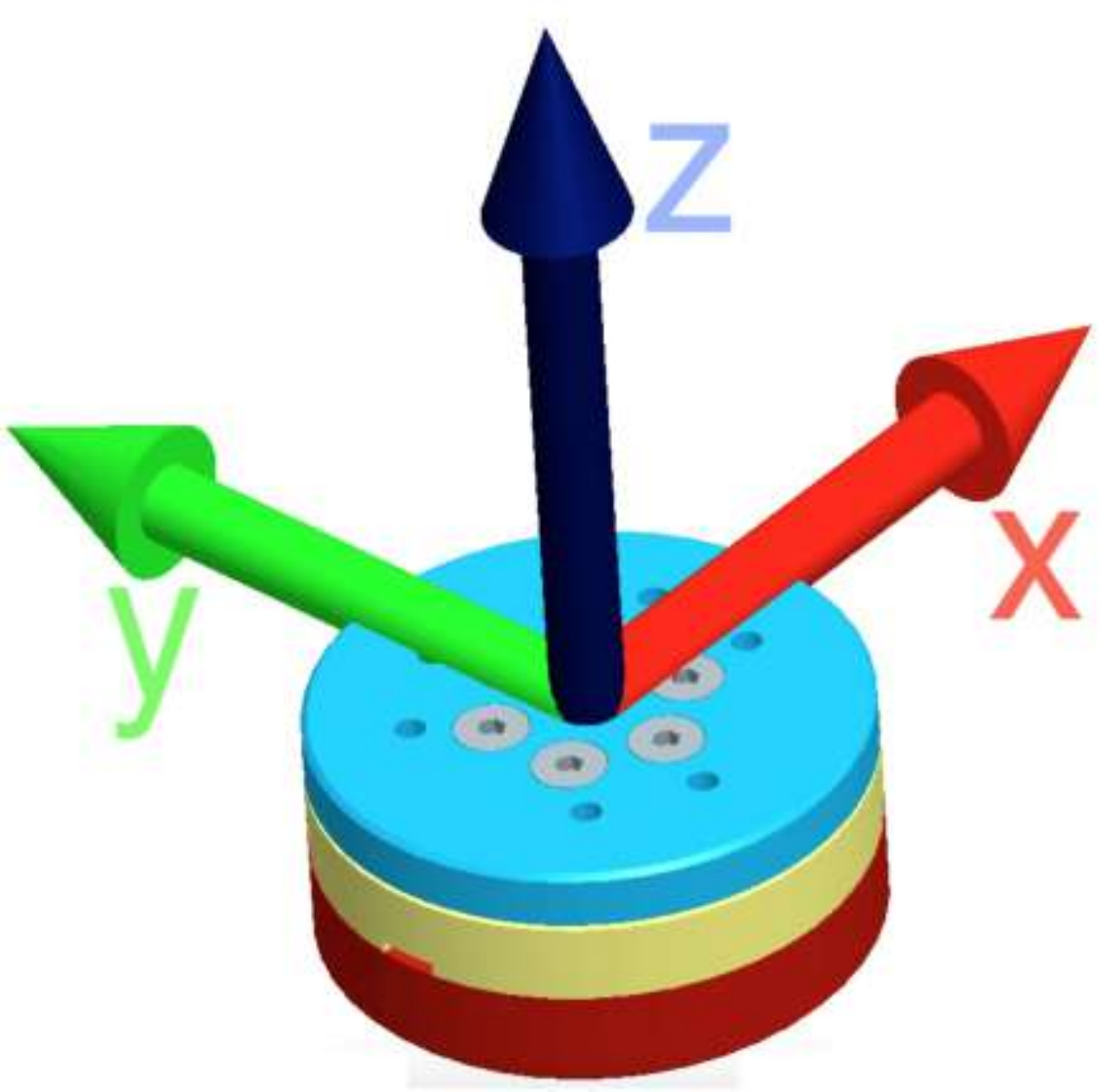}
  \caption{}
  \label{fig:FTref}
        \end{subfigure}
~\quad
  \begin{subfigure}[b]{0.80\textwidth}
    \includegraphics[width=\textwidth]{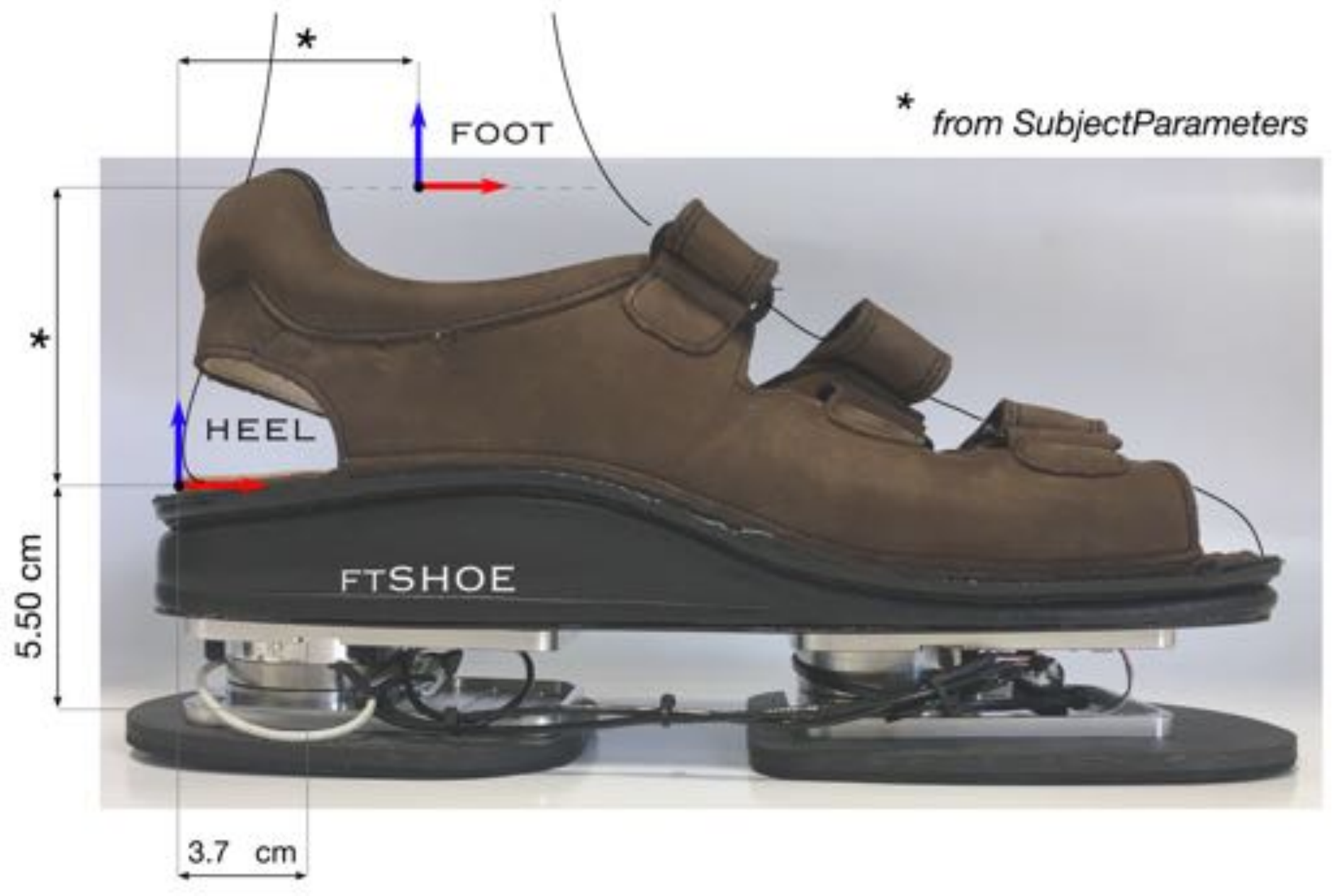}
  \caption{}
  \label{fig:Figs_footInShoe}
      \end{subfigure}
\caption{A F/T sensor developed for the iCub robot (\subref{fig:iCubFT}) with
 reference frame (\subref{fig:FTref}).
      Each shoe is provided of two F/T sensors. (\subref{fig:Figs_footInShoe})
      Sketch of human foot located into the ftShoe with reference
       frames used for transforming the forces from the shoe frame
        into the human related foot frame.}
\end{figure}

For tracking the human ground reaction forces a new sensing shoes technology
 was used.  By starting from a Xsens original design
  \citep{xsens_shoesSchepers2007}, the \emph{ftShoes} wearable
 technology is a prototype recently developed at Istituto Italiano di Tecnologia
 for measuring the human ground reaction forces.  It consists of a pair of
  shoes instrumented with four Force/Torque (F/T) sensors (Figures
   \ref{fig:iCubFT}, \ref{fig:FTref}) applied at the bottom of both soles.
Each shoe has two F/T sensors positioned at the heel and the forefoot,
   respectively, and in turn each sensor is comprised in between two
    metallic thin plates, thus they can be considered independent w.r.t. each
     other.  Each human foot is placed in the ftShoe as in Figure
      \ref{fig:Figs_footInShoe}.

\section{Data Analysis} \label{dataAnalysis}

\begin{figure}[H]
  \centering
  \begin{subfigure}[b]{0.62\textwidth}
    \includegraphics[width=\textwidth]{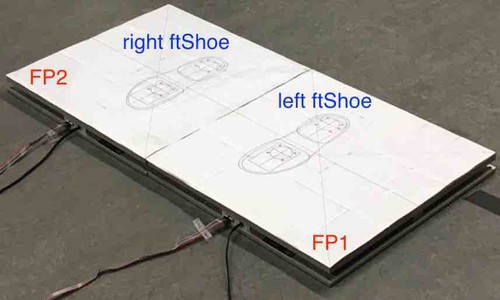}
  \caption{}
  \label{fig:sketchesOnFP}
      \end{subfigure}
~
  \begin{subfigure}[b]{0.34\textwidth}
    \includegraphics[width=\textwidth]{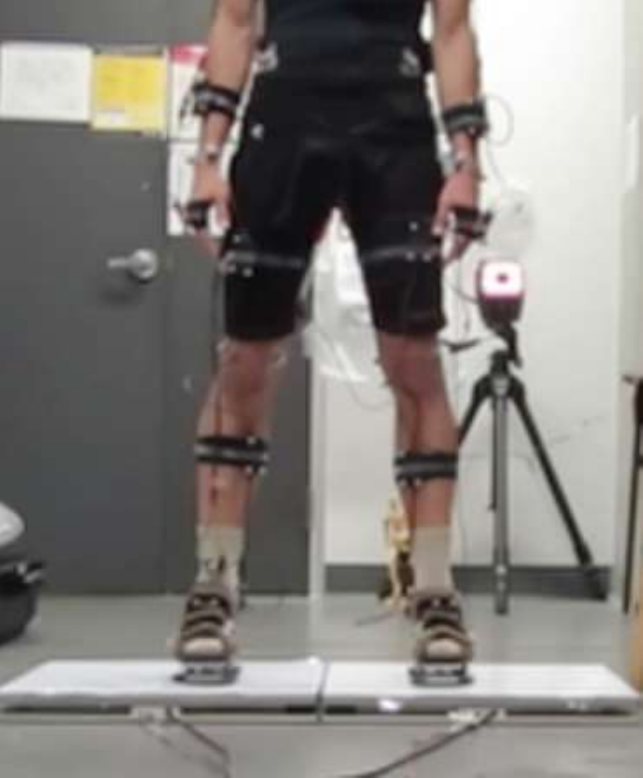}
  \caption{}
  \label{fig:subj_shoes_FP}
        \end{subfigure}
        \vspace{-0.1cm}
    \caption{(\subref{fig:sketchesOnFP}) ftShoes fixture
     applied on the top surface of two force plates.  The left ftShoe is
      coupled with the force plate FP1 and the right ftShoe with 
       FP2. (\subref{fig:subj_shoes_FP}). A subject equipped stands on the force
        plates with the ftShoes, on the position defined in
         (\subref{fig:sketchesOnFP}).}
\end{figure}

Data for the algorithm validation were collected at University of
 Waterloo (ON, Canada).  The setup encompassed $i)$ the Xsens wearable suit (in
  the look-and-hoop strap version) for the human motion tracking , $ii)$ the
   ftShoes prototype for the ground reaction computation.
Two portable AMTI force plates and a printed fixture of the shoes to
 be applied on the force plates (Figure \ref{fig:sketchesOnFP}) were
  additionally required for the shoes validation.

\begin{figure} [H]
    \centering
    \begin{subfigure}[b]{1\textwidth}
        \includegraphics[width=\textwidth]{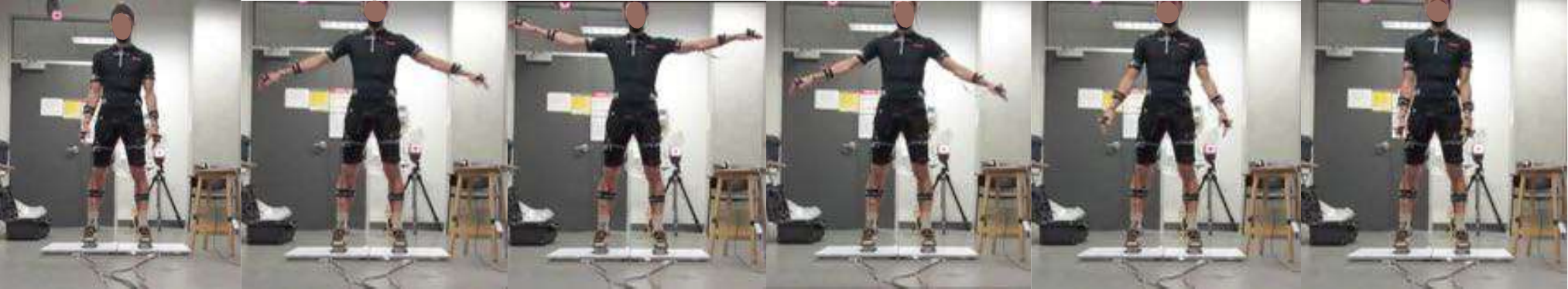}
        \caption{Task T1}
        \label{fig:Figs_butterflyTask}
    \end{subfigure}
    ~ %
    \begin{subfigure}[b]{1\textwidth}
        \vspace{0.3cm}
        \includegraphics[width=\textwidth]{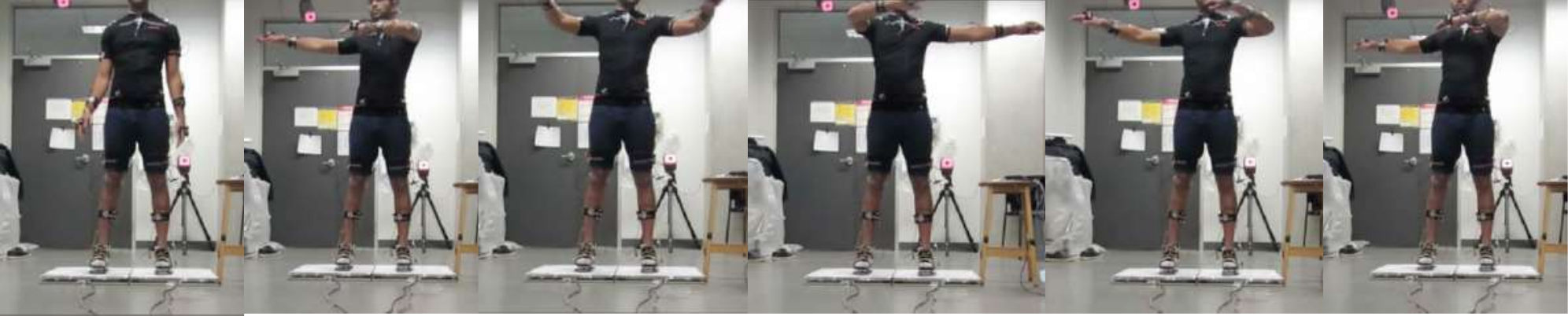}
        \caption{Task T2}
        \label{fig:Figs_torsoTwisting}
    \end{subfigure}
\caption{Human subjects while performing (\subref{fig:Figs_butterflyTask}) the
up-and-down arms movement (task T1) and (\subref{fig:Figs_torsoTwisting}) the
right-left-right torso twisting (task T2).}
\label{fig:tasks_overview}
\end{figure}

Five healthy subjects (Table \ref{Subjects_for_analysis_UW}) were asked to wear
 the sensorized suit and the ftShoes and to perform on the force plates
 different tasks, with the feet aligned to the printed fixture (as Figure
   \ref{fig:subj_shoes_FP}). The tasks selected for this analysis are listed
    here as follows and shown in Figure \ref{fig:tasks_overview}:
\begin{itemize}
\item   $10$ repetitions of up-and-down arm movements (T1), Figure
        \ref{fig:Figs_butterflyTask};
\item   $10$ repetitions of right-left-right torso twisting (T2), Figure
        \ref{fig:Figs_torsoTwisting}.
\end{itemize}

Kinematic data were acquired at a frequency of $240$\unit{}{\hertz}, force
 plates and ftShoes data at $100$\unit{}{\hertz}.  The synchronization between
  the motion capture system and the force plates was guaranteed by a synch
   station provided by Xsens.  Readings from the shoes\footnote{ftShoes
    readings are obtained through YARP, see in chapter
     Section \ref{YARP_description}.} were synchronized by code.
Data with different acquisition rates were linearly interpolated before being
 processed.

\begin{table}[H]
\centering
\caption{Subjects eligible for the analysis.  Each subject was provided of a
 written informative consent before starting the experiment.}
\label{Subjects_for_analysis_UW}
\centering
\scriptsize
\begin{tabular}{c|cccc}
\\
\hline\hline
\\
\textbf{Subject} & \textbf{Gender} & \textbf{Age} 
                 & \textbf{Height} [\unit{}{\centi\meter}]
                 & \textbf{Mass} [\unit{}{\kilo\gram}]\\
\\
\hline
\\
\textbf{S1}      & M    & $24$  & $179$  & $75.9$    \\
\rowcolor{Gray}
\textbf{S2}      & F    & $26$  & $163$  & $63.9$    \\
\textbf{S3}      & M    & $27$  & $179$  & $83.0$    \\
\rowcolor{Gray}
\textbf{S4}      & M    & $26$  & $187$  & $93.6$    \\
\textbf{S5}      & M    & $27$  & $175$  & $72.0$  \\
\\                              
\hline\hline
\\
\end{tabular}
\end{table}

\vspace{1cm}
\subsection{\emph{ftShoes} Validation}

The Section reports the analysis for the validation of the new technology
 w.r.t. the gold standard in ground reaction force measurements (i.e., the
  force plates).
The analysis required to express both readings in a common reference
 frame (e.g., the human foot frame\footnote{IMUs readings have been used here
  for retrieving the position of each foot w.r.t. the reference frame of the
   related ftShoe/force plate.}).

\begin{figure}[H]
    \vspace{2.5cm}
  \centering
  \begin{subfigure}[b]{0.49\textwidth}
\includegraphics[width=\textwidth]
{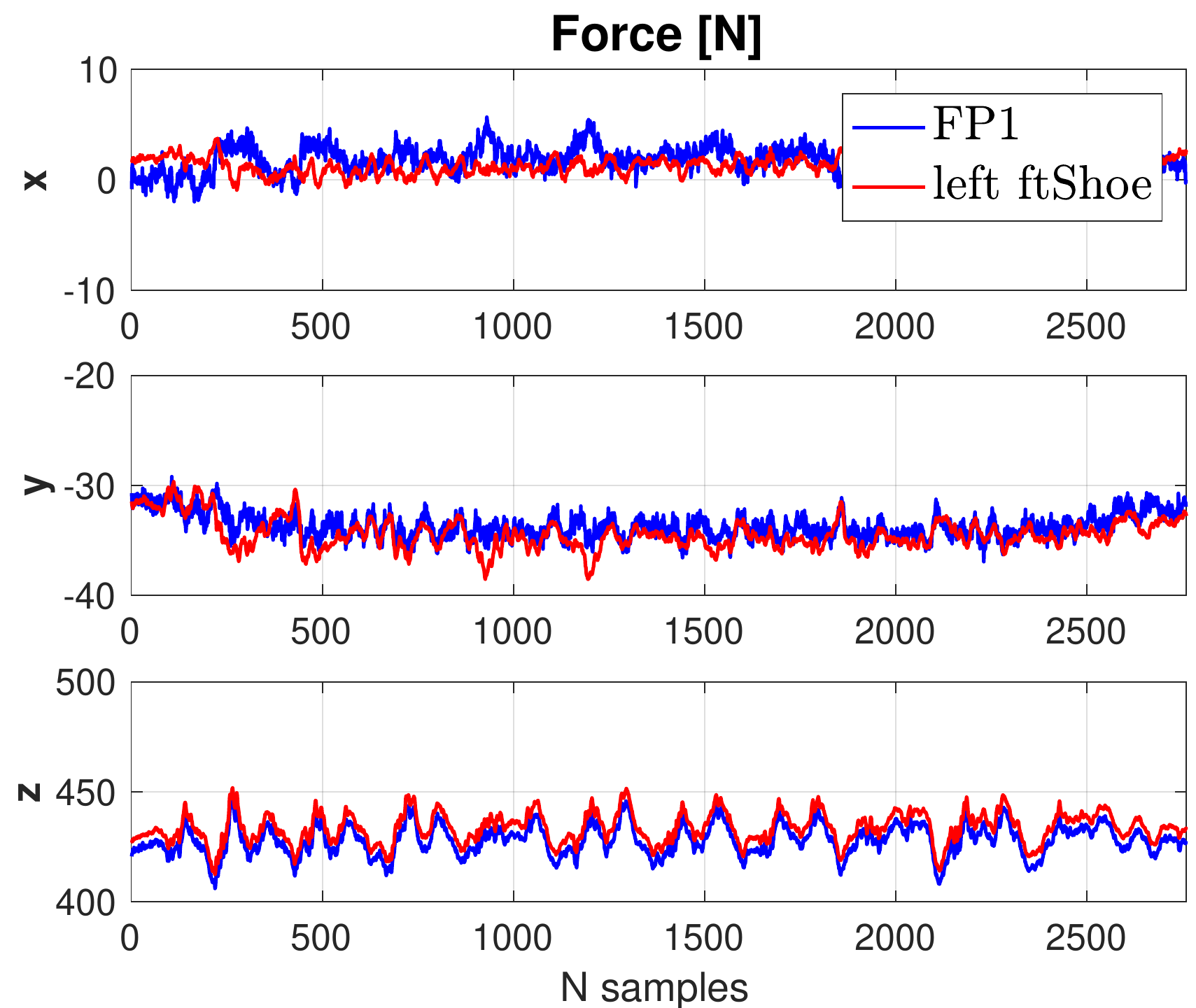}
  \caption{}
  \label{fig:fp1_leftShoe_FORCE_val_butterfly}
      \end{subfigure}
  \begin{subfigure}[b]{0.49\textwidth}
\includegraphics[width=\textwidth]
{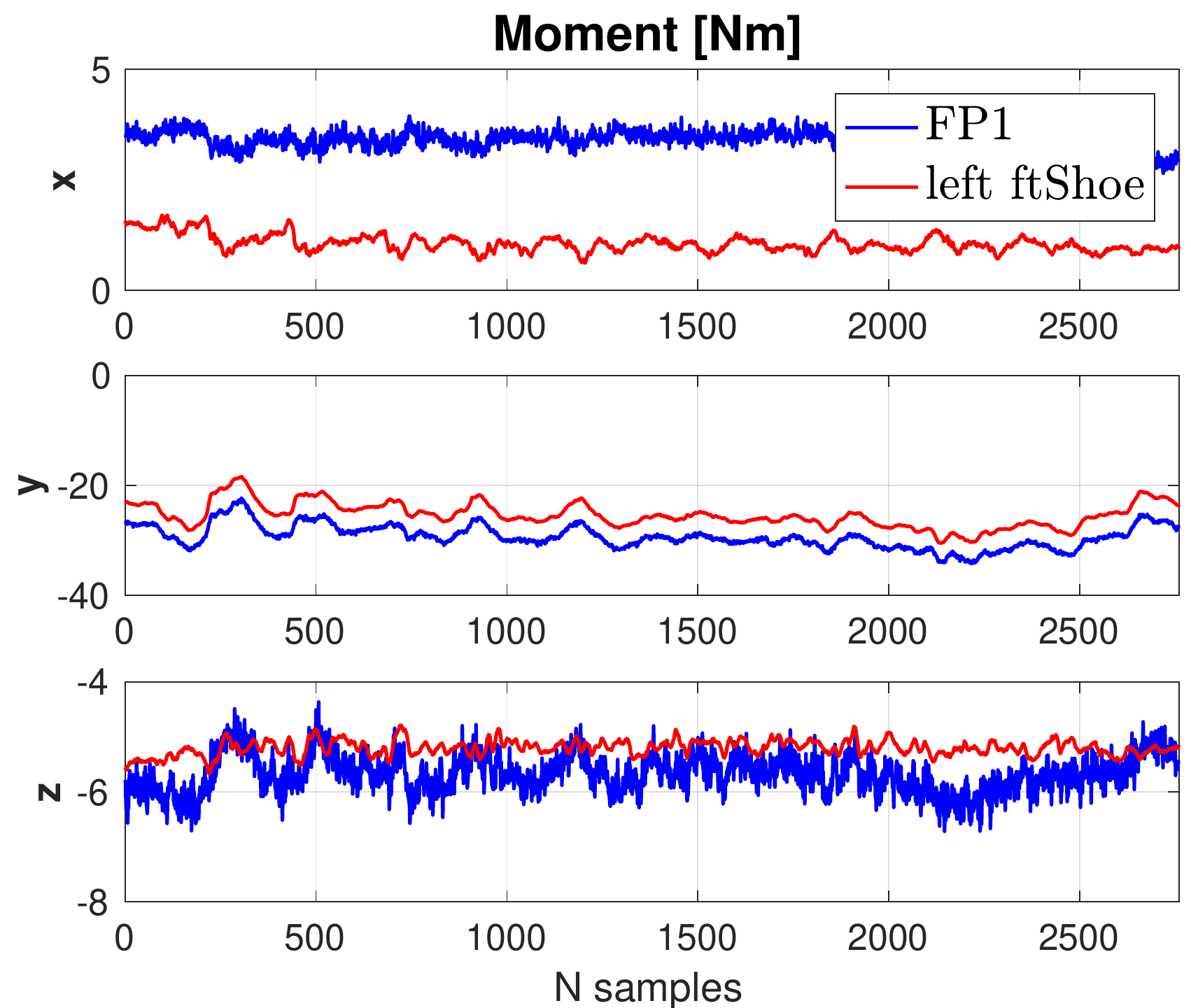}
  \caption{}
  \label{fig:fp1_leftShoe_MOMENT_val_butterfly}
        \end{subfigure}
  \centering
  \begin{subfigure}[b]{0.49\textwidth}
\includegraphics[width=\textwidth]
{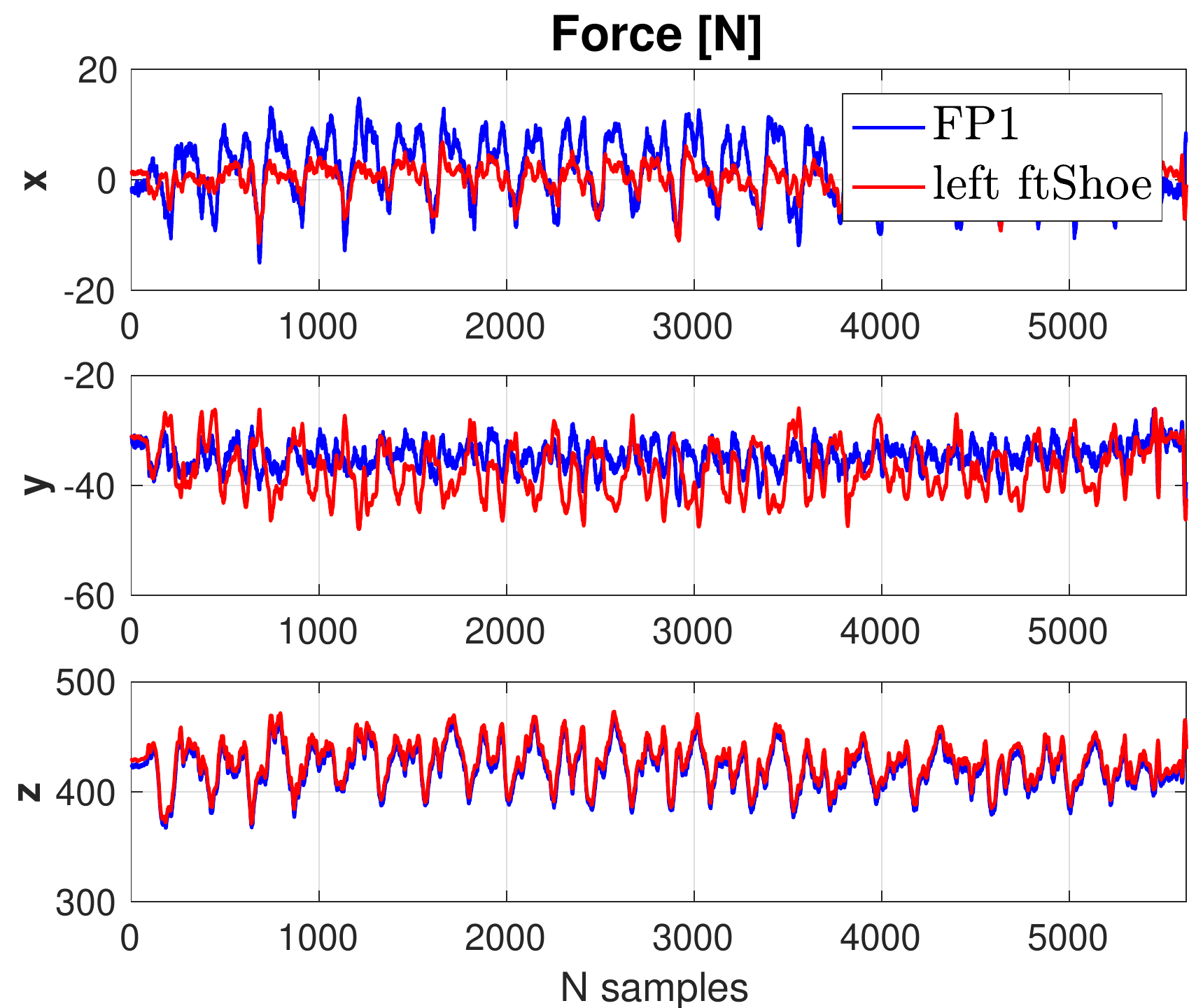}
  \caption{}
  \label{fig:fp1_leftShoe_FORCE_val_torsoTwist}
      \end{subfigure}
  \begin{subfigure}[b]{0.49\textwidth}
\includegraphics[width=\textwidth]
{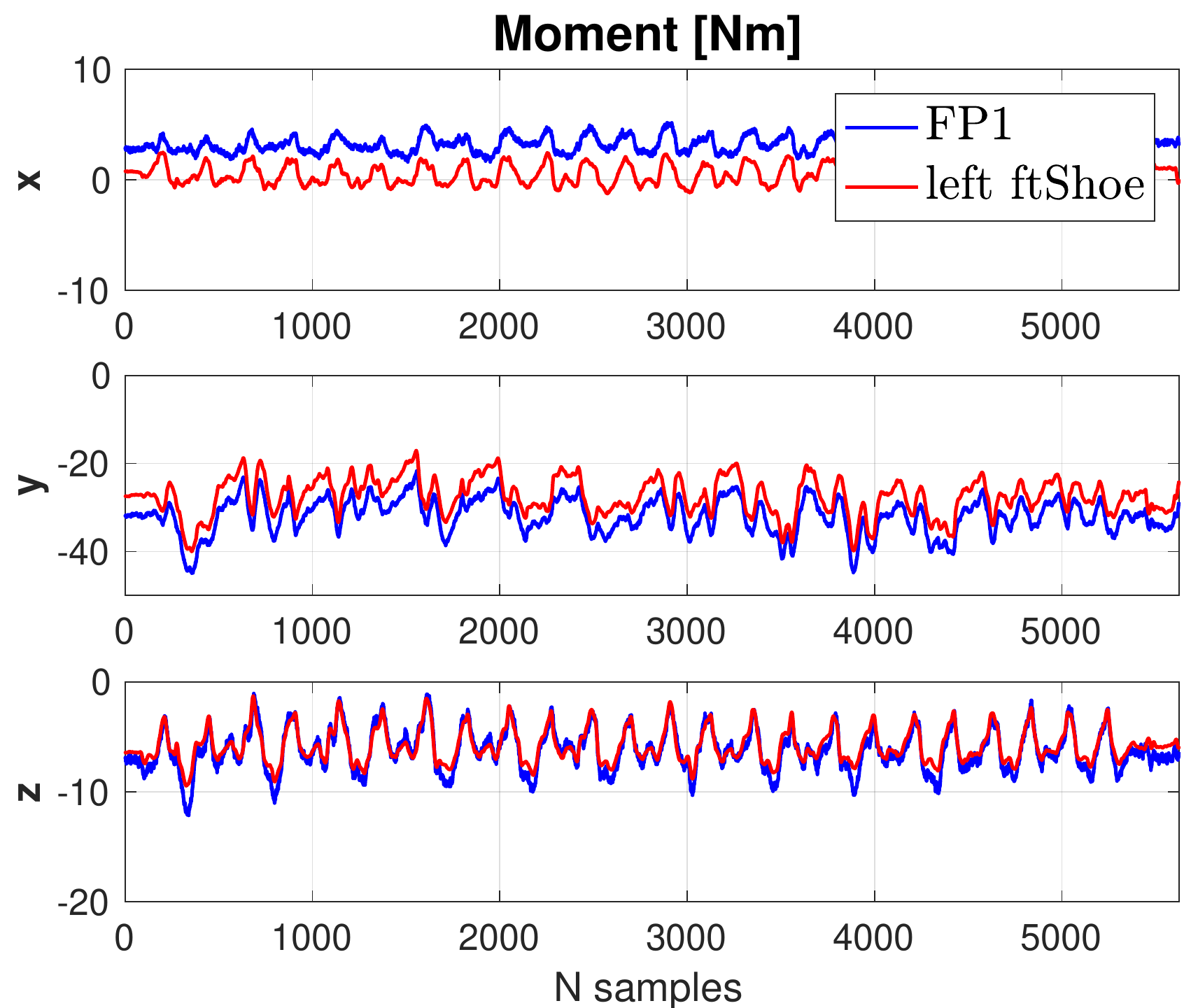}
  \caption{}
  \label{fig:fp1_leftShoe_MOMENT_val_torsoTwist}
        \end{subfigure}
\vspace{-0.1cm}
\caption{Subject S1 validation of the left ftShoe (in red) with the related
 force plate FP1 (in blue) for the forces (on the left column) and
  the moment (on the right column) for the task T1
   (\subref{fig:fp1_leftShoe_FORCE_val_butterfly})-(\subref{fig:fp1_leftShoe_MOMENT_val_butterfly}) and task T2
    (\subref{fig:fp1_leftShoe_FORCE_val_torsoTwist})-(\subref{fig:fp1_leftShoe_MOMENT_val_torsoTwist}).}
\label{complete_fp1_leftShoe_validation}
\end{figure}

\begin{figure}[H]
        \vspace{2.5cm}
  \centering
  \begin{subfigure}[b]{0.49\textwidth}
       \vspace{0.3cm}
\includegraphics[width=\textwidth]
{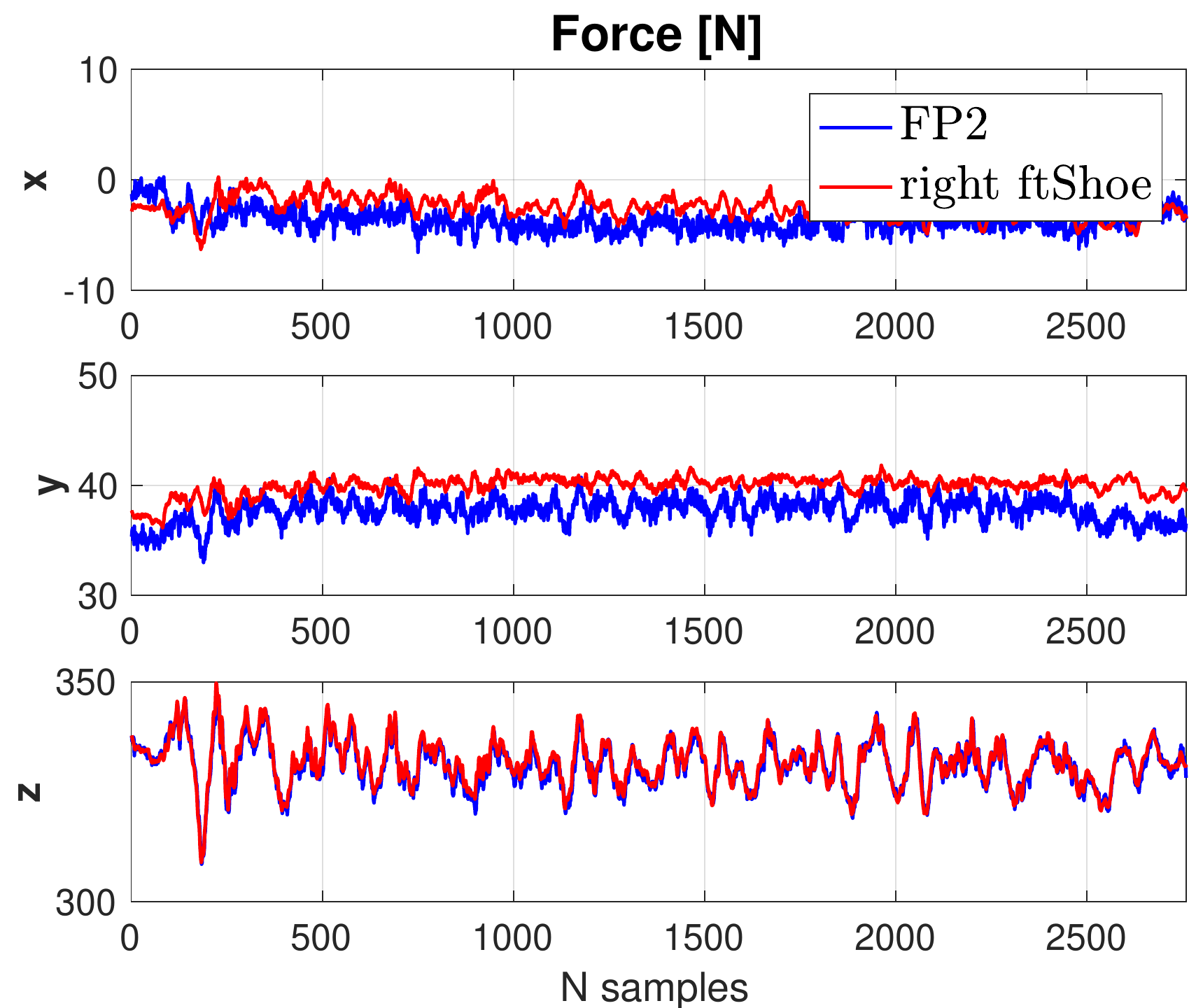}
  \caption{}
  \label{fig:fp2_rightShoe_FORCE_val_butterfly}
      \end{subfigure}
  \begin{subfigure}[b]{0.49\textwidth}
\includegraphics[width=\textwidth]
{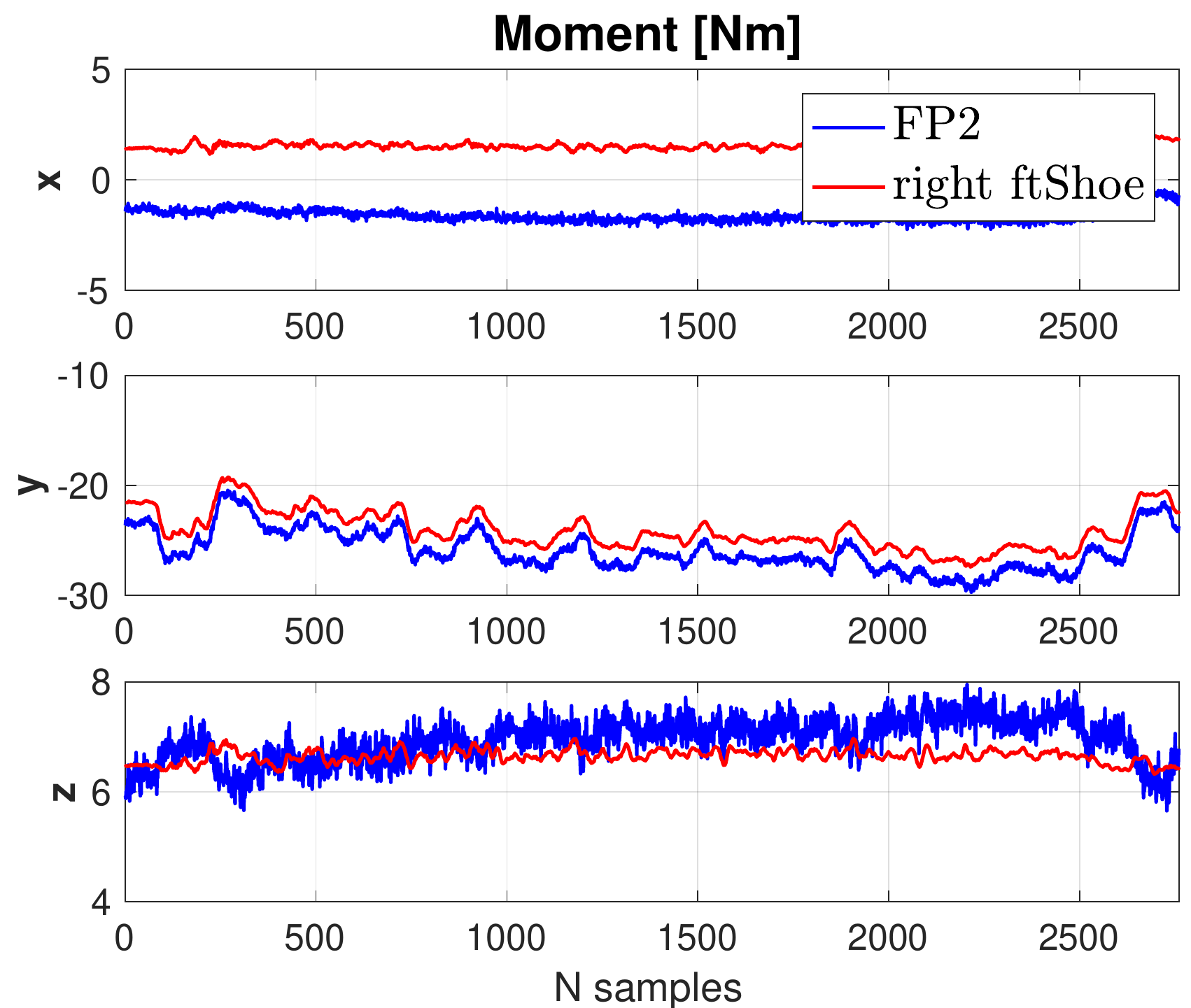}
  \caption{}
  \label{fig:fp2_rightShoe_MOMENT_val_butterfly}
        \end{subfigure}
  \centering
  \begin{subfigure}[b]{0.49\textwidth}
\includegraphics[width=\textwidth]
{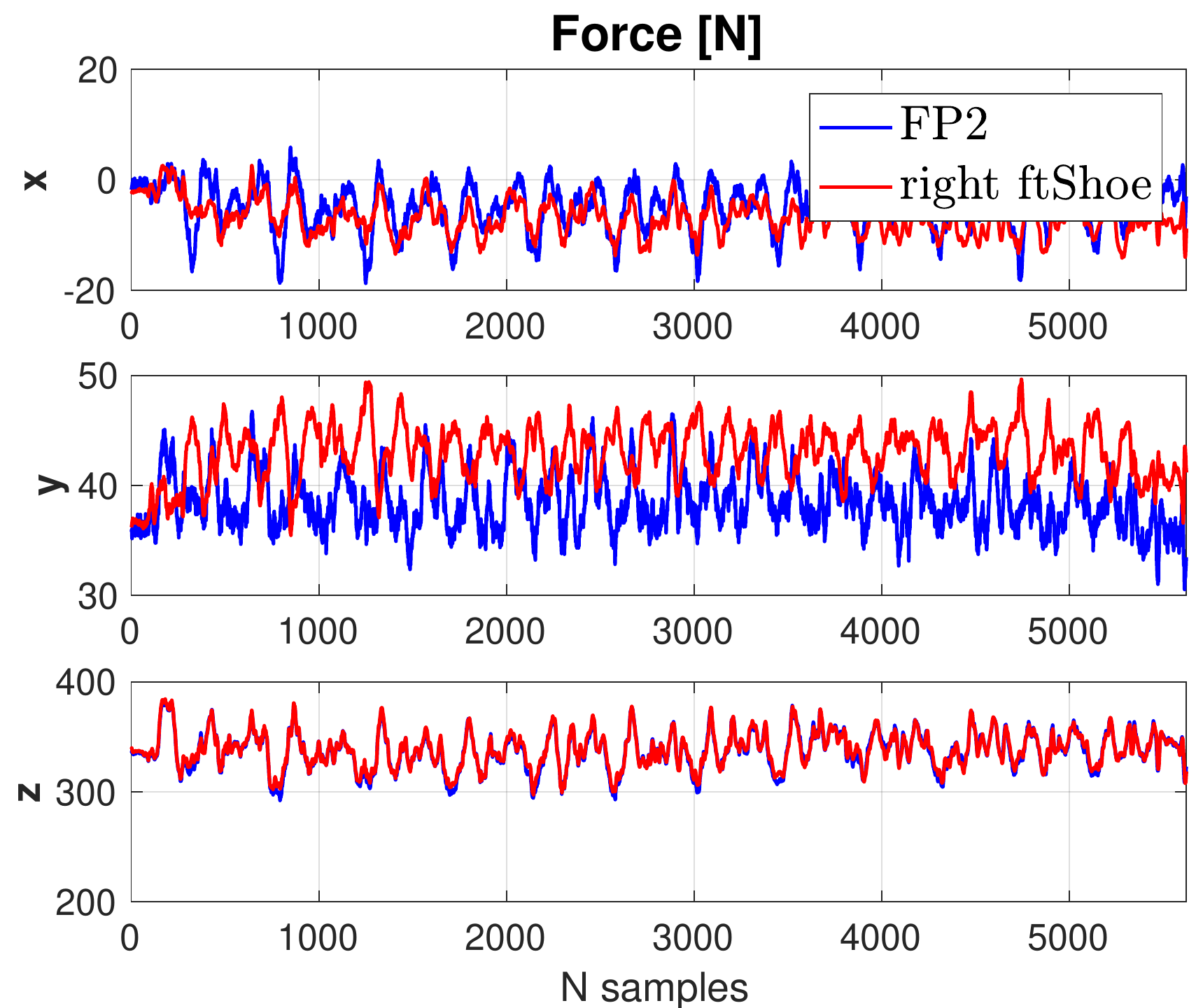}
  \caption{}
  \label{fig:fp2_rightShoe_FORCE_val_torsoTwist}
      \end{subfigure}
  \begin{subfigure}[b]{0.49\textwidth}
\includegraphics[width=\textwidth]
{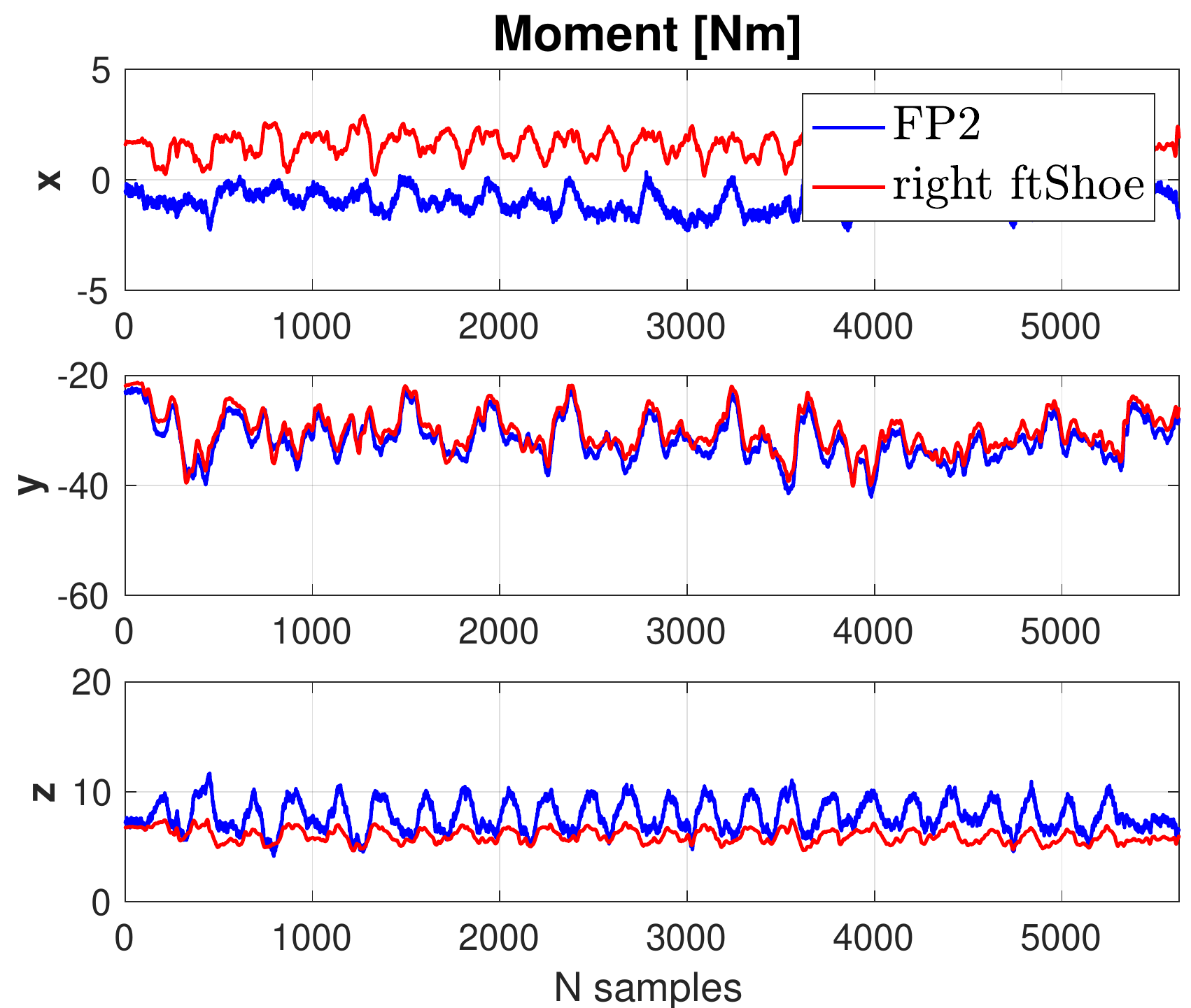}
  \caption{}
  \label{fig:fp2_rightShoe_MOMENT_val_torsoTwist}
        \end{subfigure}
\vspace{-0.1cm}
\caption{Subject S1 validation of the right ftShoe (in red) with the
 related force plate FP2 (in blue) for the forces (on the left column) and
  the moment (on the right column) for the task T1
   (\subref{fig:fp1_leftShoe_FORCE_val_butterfly})-(\subref{fig:fp1_leftShoe_MOMENT_val_butterfly}) and task T2
    (\subref{fig:fp1_leftShoe_FORCE_val_torsoTwist})-(\subref{fig:fp1_leftShoe_MOMENT_val_torsoTwist}).}
\label{complete_fp2_rightShoe_validation}
\end{figure}

A Root Mean Square Error (RMSE) analysis (Table \ref{RMSE_shoesVal}) has been
 performed for the above described dataset, for validating the three components
  of the forces ($f_x$, $f_y$,$f_z$) and moments ($m_x$, $m_y$, $m_z$).
Figure \ref{complete_fp1_leftShoe_validation} shows the forces (on the left
 column) and moments  (on the right column) for validating the left ftShoe
  reading (in human left foot frame) w.r.t. the gold standard provided by its
   coupled force plate FP1, for subject S1, tasks T1-T2.
Similarly, Figure \ref{complete_fp2_rightShoe_validation} validates the right
 ftShoe w.r.t. the force plate FP2, for the same subject, in the same tasks.

The RMSE values are in general very low denoting a good matching of each shoe
 signal w.r.t. the related force plate.  Despite some values (highlighted in
 red) seem to provide a worst estimation of the forces, they are anyway
  considered acceptable from the force plate datasheet\footnote{From AMTI
   datasheet: by considering an error of $0.4\%$ on the full scale due to a
   combination of hysteresis and non-linearity noises and a subject standard
    weight of $700$ [\unit{}{\newton}], the error $\bm e_{f}$ on the forces
     could be quantified as $\bm e_{f}=[23,~23,~18]$ [\unit{}{\newton}].
       Similarly for the moment error $\bm e_{m}=[4.5,~4.5,~2.25]$
        [\unit{}{\newton\meter}].}.
Another important source of error could be attributed to fact that the shoes
are not rigidly attached to the force plates.  It is worth noting that, even if
explicitly required, the subject did not hold his feet fixed on the
 fixture (e.g., some tasks implicitly caused slippery conditions), while
  the analysis is completely based on this strong assumption
  (i.e., constant transformation between the ftShoes/force plates origins).

  \vspace{1cm}
\begin{table}[H]
\centering
  \vspace{1cm}
\caption{RMSE analysis of the forces [\unit{}{\newton}] and
 the moments [\unit{}{\newton\meter}] for validating the left ftShoe w.r.t.
  the forceplate FP1 and the right ftShoe w.r.t. the force plate FP2,
   respectively. (Subjects S1, S2, S3, S4, S5; tasks T1, T2).}
\label{RMSE_shoesVal}
\centering
\tiny
\begin{tabular}{c|c|cccccc|cccccc}
    \\
    \hline\hline
    \\
\multirow{2}{*}{\textbf{Subject}} & \multirow{2}{*}{\textbf{Task}} & \multicolumn{6}{c|}{\textbf{FP1 - left ftShoe}}  &
\multicolumn{6}{c}{\textbf{FP2 - right ftShoe}}  \\ 
&                          & $f_x$ & $f_y$ & $f_z$ & $m_x$ & $m_y$ & $m_z$
                           & $f_x$ & $f_y$ & $f_z$ & $m_x$ & $m_y$ & $m_z$ \\
\\
\hline
\\
\multirow{2}{*}{\textbf{S1}}
& {\textbf{T1}}     & $1.52$ & $1.30$ & $5.53$ & $2.39$ & $3.93$ & $0.57$
                    & $1.74$ & $2.42$ & $1.52$ & $3.13$ & $1.77$ & $0.48$\\
& \cellcolor{Gray}{\textbf{T2}}     & \cellcolor{Gray}$4.50$ & \cellcolor{Gray}$4.72$ & \cellcolor{Gray}$5.86$ & \cellcolor{Gray}$2.79$ &
 \cellcolor{Gray}$4.44$ & \cellcolor{Gray}$0.74$
                    & \cellcolor{Gray}$4.19$ & \cellcolor{Gray}$5.80$ &
                     \cellcolor{Gray}$2.98$ & \cellcolor{Gray}$2.60$ &
                      \cellcolor{Gray}$1.73$ &
                     \cellcolor{Gray}$2.03$\\
                    \hline 
\multirow{2}{*}{\textbf{S2}}
& {\textbf{T1}}     & $1.76$ & $2.65$ & $0.96$ & $1.81$ & $0.29$ & $1.53$
                    & $1.92$ & $3.34$ & $3.37$ & $1.14$ & $0.49$ & $0.33$\\
& \cellcolor{Gray}{\textbf{T2}}     & \cellcolor{Gray}$4.13$ & \cellcolor{Gray}$4.59$ & \cellcolor{Gray}$1.22$ & \cellcolor{Gray}$1.92$ &
 \cellcolor{Gray}$0.46$ & \cellcolor{Gray}$1.74$
                    & \cellcolor{Gray}$4.97$ & \cellcolor{Gray}$4.95$ &
                     \cellcolor{Gray}$6.02$ & \cellcolor{Gray}$0.70$ &
                      \cellcolor{Gray}$1.30$ &
                     \cellcolor{Gray}$0.79$\\
                     \hline
\multirow{2}{*}{\textbf{S3}}
& {\textbf{T1}}     & $3.28$ & $1.60$ & \cellcolor{red!30}$11.09$ & $2.31$ &
 $2.24$ & $0.58$
                    & $7.61$ & $1.03$ & $1.99 $ & $2.03$ & $3.26$ & $1.13$\\
& \cellcolor{Gray}{\textbf{T2}}     & \cellcolor{Gray}$4.96 $ & \cellcolor{Gray}$6.08$ & \cellcolor{red!30}$11.20$ & \cellcolor{Gray}$3.55$ &
 \cellcolor{Gray}$2.23$ & \cellcolor{Gray}$1.12$
                    & \cellcolor{Gray}$8.99 $ & \cellcolor{Gray}$4.69$ &
                     \cellcolor{Gray}$4.31 $ & \cellcolor{Gray}$1.19$ &
                      \cellcolor{Gray}$5.01$ &
                     \cellcolor{Gray}$2.50$\\
                     \hline
\multirow{2}{*}{\textbf{S4}}
& {\textbf{T1}}     & $1.88 $ & $1.68 $ & $2.73$ & $2.89$ & $4.05$ & $0.53$
                    & $1.51 $ & $1.60 $ & $3.65$ & $3.20$ & $5.84$ & $0.61$\\
& \cellcolor{Gray}{\textbf{T2}}     & \cellcolor{Gray} $9.55 $ &
 \cellcolor{red!30}$11.66$ & \cellcolor{Gray}$3.44$ & \cellcolor{Gray}$4.71$ &
  \cellcolor{Gray}$5.66$ & \cellcolor{Gray}$1.03$
                    & \cellcolor{red!30} $10.58$ & \cellcolor{Gray}$8.17 $ &
                     \cellcolor{Gray}$5.99$ & \cellcolor{Gray}$1.68$ &
                      \cellcolor{Gray}$3.75$ & \cellcolor{Gray}$2.60$\\
                     \hline
\multirow{2}{*}{\textbf{S5}}
& {\textbf{T1}}     & $2.89$ & $4.90$ & $1.38$ & $2.01$ & $0.97$ & $1.67$
                    & $1.40$ & $3.48$ & $1.33$ & $2.74$ & $1.63$ & $1.57$\\
& \cellcolor{Gray}{\textbf{T2}}     & \cellcolor{Gray}$4.50 $ &
 \cellcolor{Gray}$4.29 $ & \cellcolor{Gray}$1.53$ & \cellcolor{Gray}$1.72$ &
  \cellcolor{Gray}$1.16$ & \cellcolor{Gray}$1.63$
                    & \cellcolor{Gray}$3.16 $ & \cellcolor{Gray}$4.09 $ &
                     \cellcolor{Gray}$3.41$ & \cellcolor{Gray}$3.24$ &
                      \cellcolor{Gray}$2.01$ & \cellcolor{Gray}$0.88$\\
\\
\hline\hline
\\
\end{tabular}
\end{table}

\newpage

\newpage
\subsection{Estimation of Human Variables}
\label{MAPanalysis_UW}

The MAP algorithm is a tool for estimating quantities related to the joints and
 links of the human model.  It is worth remarking that the vector $\bm d$ (the
  solution \eqref{eq:mu_dgiveny}) contains variables that can be directly
   measured ($\bm {\underline a}$, ${\bm {\underline f}^x}$) and variables that
    can not be directly measured in humans (${\bm {\underline f}^B}$, ${\bm
     {\underline f}}$, $\bm \tau$) but only estimated through the algorithm.

The MAP algorithm represents, in a sense, the probabilistic way to estimate
 those quantities for which a direct measure does not exist.  We consider the
  goodness in the estimation of the measurable quantities as a
   parameter of reliability of the method in estimating those variables that
    are no  directly measurable.  The leading idea is to compare the same
     variables (measured and then estimated) in order to prove the
      quality of the proposed algorithm.

Subjects of Table \ref{Subjects_for_analysis_UW} were asked to perform tasks T1
 and T2 equipped with the Xsens suit for the motion tracking and the ftShoes
  prototype for computing the ground reaction forces.

Xsens data were acquired at $240$\unit{}{\hertz} through the MVN Studio
 software, ftShoes data at $100$\unit{}{\hertz} through the YARP driver.  Data
  were synchnonized and downsampled in post-processing phase by means of Matlab.
A $48$-DoF model template was used both for the URDF and for the OpenSim
 model, ($\bm d \in \mathbb{R}^{1248}$).
The IK was computed by using the OpenSim API for Matlab and a Savitzky-Golay
 filtering ($3$-th order, $57$-elements moving window) was used for retrieving
  joint velocities $\dot{\bm{q}}$ and accelerations $\ddot{\bm{q}}$.
The MAP settings were tuned as follows: $\bm{\Sigma}_d = 10^{4}$ (i.e., no
 reliable prior on vector $\bm d$), $\bm{\Sigma}_D = 10^{-4}$ (i.e., high
  reliable
  prior on the dynamic model), $\bm{\Sigma}_y$ composed of each sensor variance
   submatrix (IMUs $\approx 10^{-3}$, joint acceleration $\approx 10^{-3}$,
    external forces on the feet $\approx 10^{-3}$ and on the other links of the
     model $\approx 10^{-6}$).

The analysis shows the comparison between the measure and the estimation for
 the external forces, linear accelerations and joint accelerations of certain
  links/joints of the model.  Their choice was task-driven by considering those
   links/joints mainly involved in the tasks.  The analysis has been performed
    for all the subjects in Table
 \ref{Subjects_for_analysis_UW} along with a RMSE investigation for the
  external forces (Table \ref{RMSE_MAPestim_FEXT}), the linear accelerations
   (Table \ref{RMSE_MAPestim_ACC}) and the joint accelerations
  (Table \ref{RMSE_MAPestim_DDQ}).
However, the following figures refer
   to one subject (i.e., S2) and they have to be interpreted in the following
    way: the three columns are the \emph{x-y-z} components; the rows are
     related to the
  link/joint chosen for the evaluation.  Blank plots mean that there is not a
   related quantity for that component (depending on the joint DoFs).

\begin{figure}[H]
  \centering
\vspace{-5.8cm}
\includegraphics[width=.9\textwidth]
{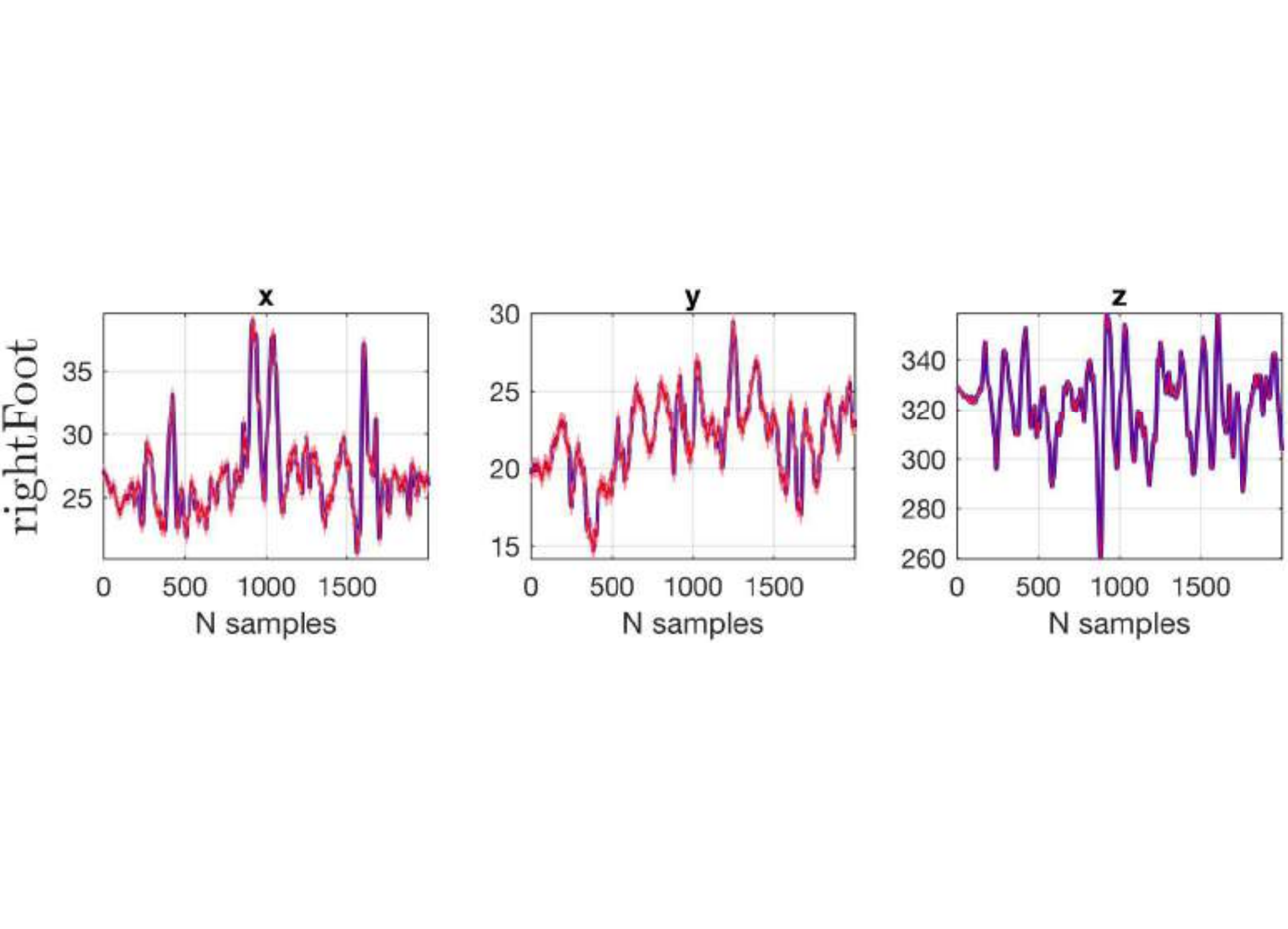}
\vspace{-2.5cm}
\caption{External force [\unit{}{\newton}] comparison.  The
 plots show the measured quantity (with $2\sigma$
  standard deviation, in red) and the MAP estimation (in blue) and  for the S2
   right foot link in the tasks T1 (top) and T2 (bottom).}
  \label{fig:Figs_fext_dx_comparison}
\end{figure}

\begin{figure}[H]
  \centering
\vspace{-0.4cm}
\includegraphics[width=.8\textwidth]
{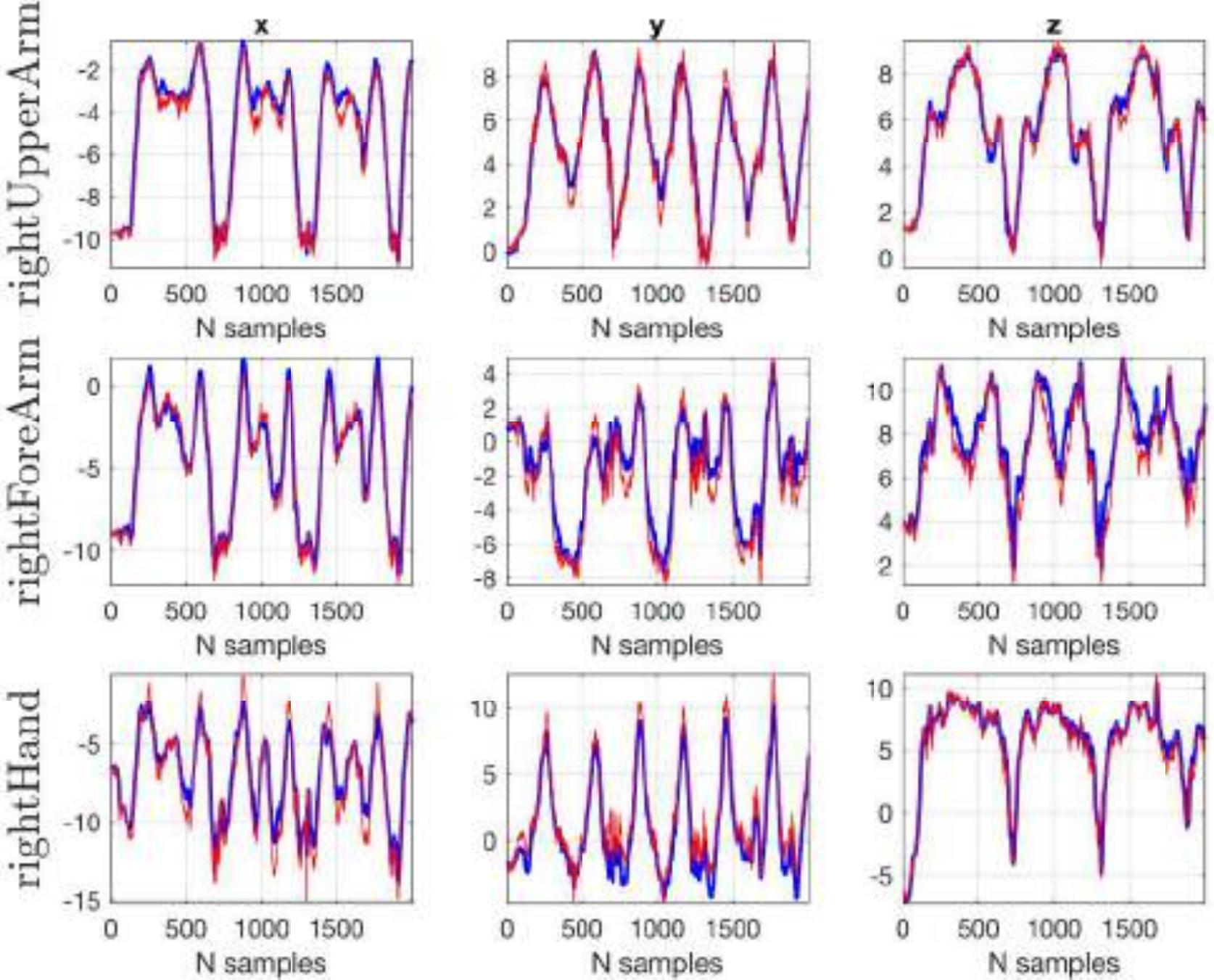}
\caption{Linear acceleration [\unit{}{\meter\per\second^2}] comparison.  The
 plots show the estimate (in blue) and the measured quantity (with $2\sigma$
  standard deviation, in red) for the S2 right upper arm, forearm and hand
   links, respectively, in the tasks T1 (top) and T2 (bottom).}
  \label{fig:Figs_acc_dx_comparison}
\end{figure}

\begin{figure}[H]
  \centering
\vspace{-0.4cm}
\includegraphics[width=.8\textwidth]
{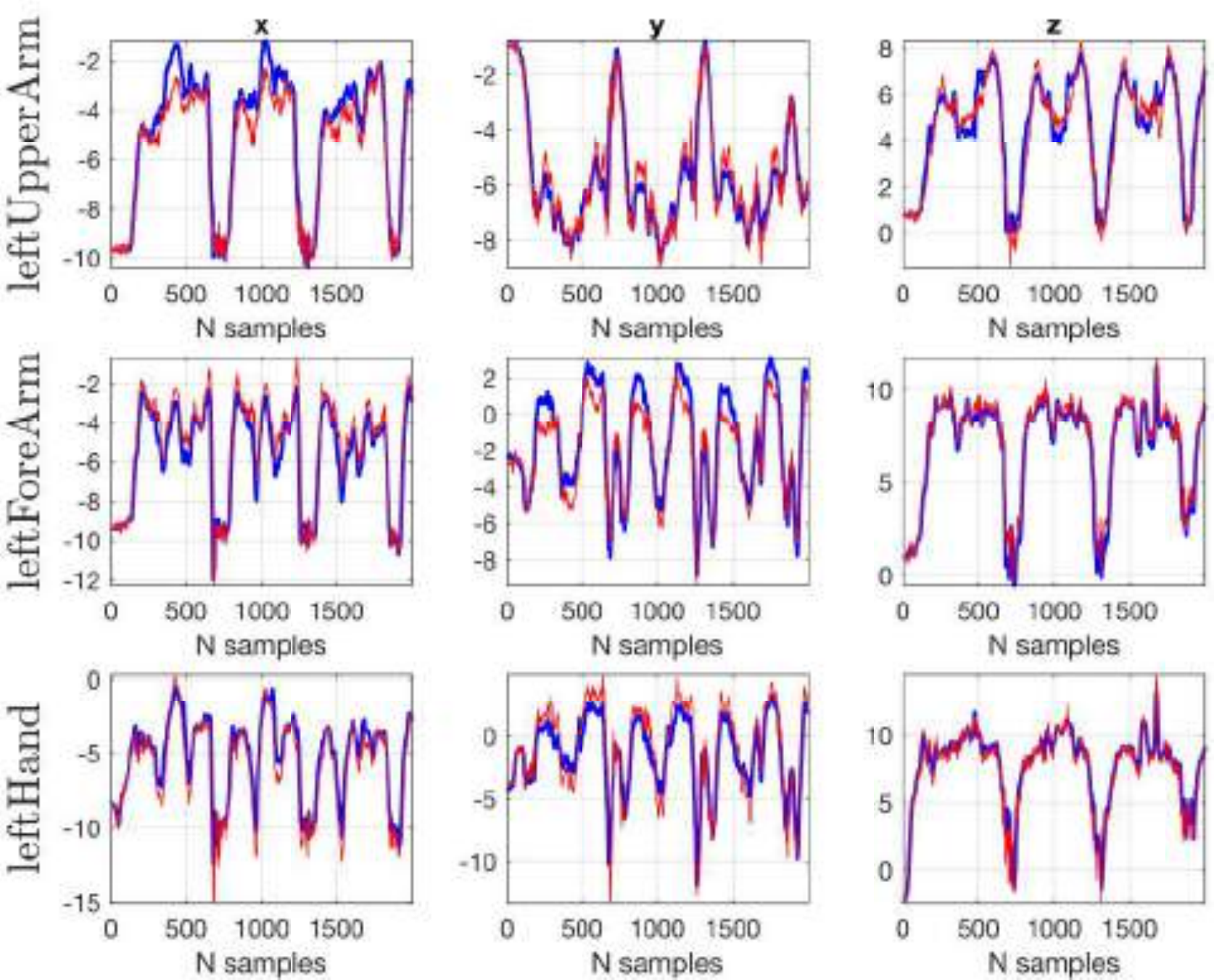}
\caption{Linear acceleration [\unit{}{\meter\per\second^2}] comparison.  The
 plots show the estimate (in blue) and the measured quantity (with $2\sigma$
  standard deviation, in red) for the S2 left upper arm, forearm and hand
   links, respectively, in the tasks T1 (top) and T2 (bottom).}
  \label{fig:Figs_acc_sx_comparison}
\end{figure}

\begin{figure}[H]
  \centering
\vspace{-0.2cm}
\includegraphics[width=.9\textwidth]
{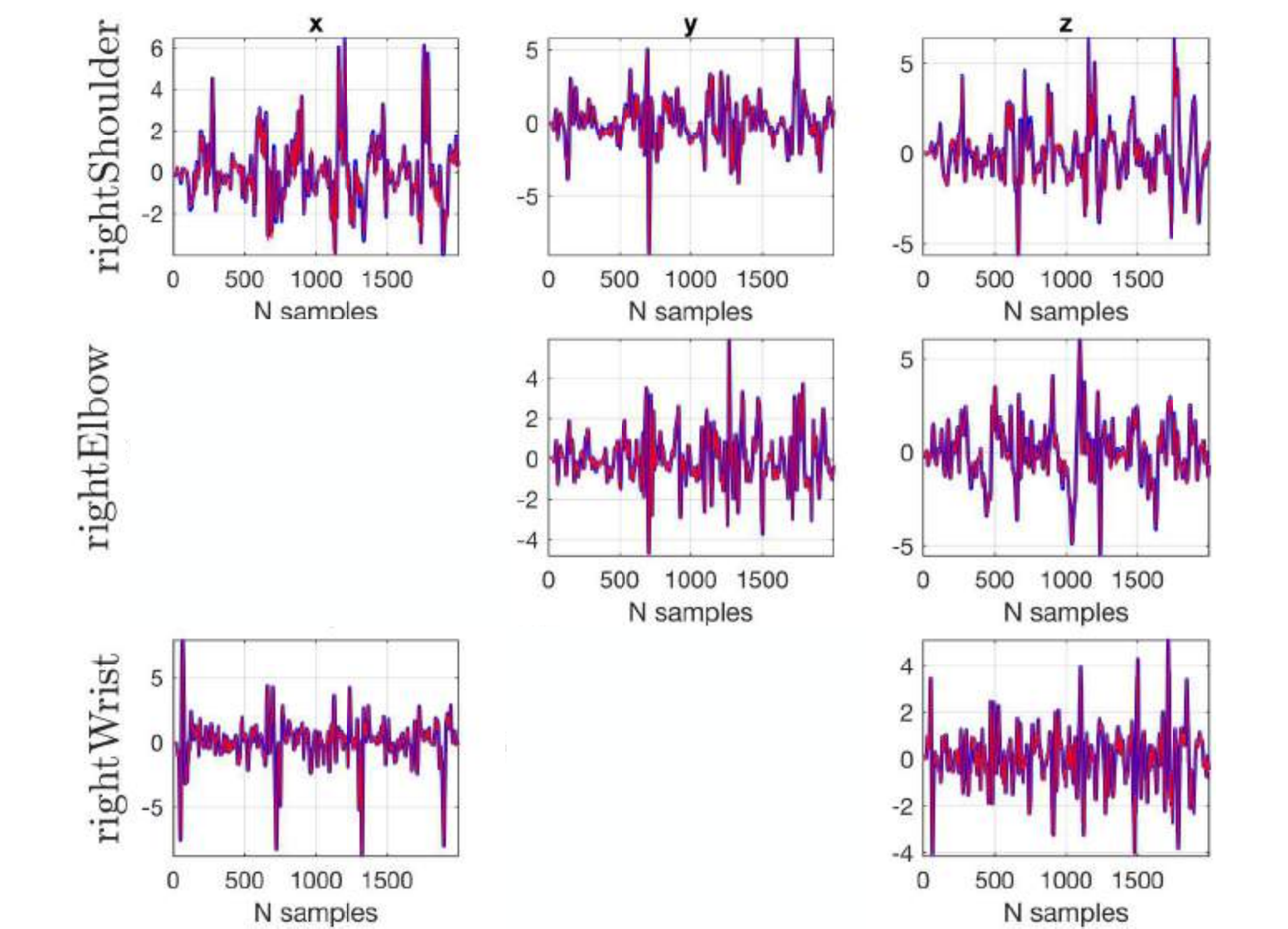}
\caption{Joint acceleration [\unit{}{\rad\per\second^2}] comparison.  The
 plots show the estimate (in blue) and the measured quantity (with $2\sigma$
  standard deviation, in red) for the S2 right shoulder, elbow and wrist
   joints, respectively, in the tasks T1 (top) and T2 (bottom).}
  \label{fig:Figs_ddq_dx_comparison}
\end{figure}

\begin{figure}[H]
  \centering
\vspace{-0.2cm}
\includegraphics[width=.9\textwidth]
{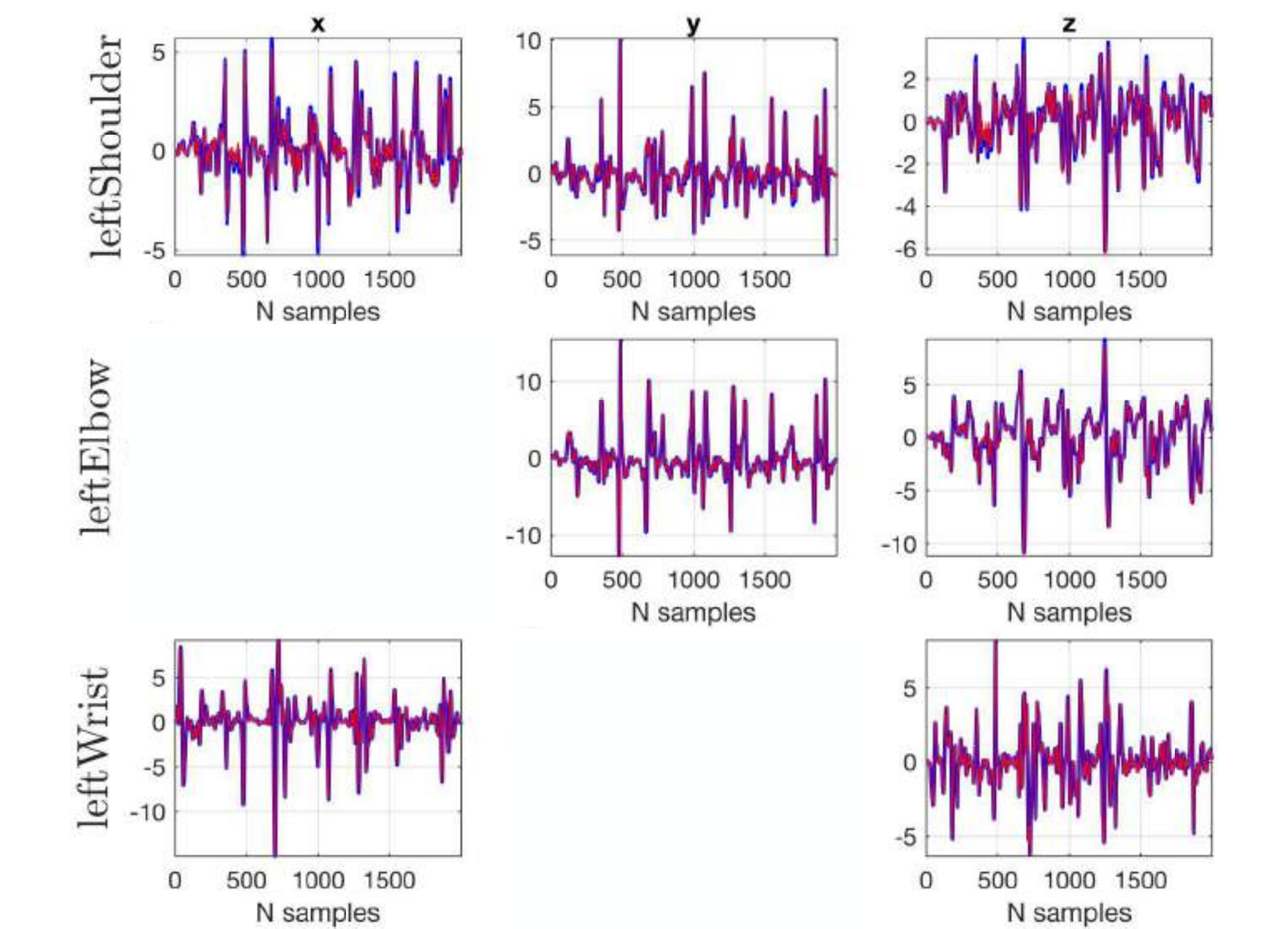}
\caption{Joint acceleration [\unit{}{\rad\per\second^2}] comparison.  The
 plots show the estimate (in blue) and the measured quantity (with $2\sigma$
  standard deviation, in red) for the S2 left shoulder, elbow and wrist joints,
   respectively, in the tasks T1 (top) and T2 (bottom).}
  \label{fig:Figs_ddq_sx_comparison}
\end{figure}

Figure \ref{fig:Figs_fext_dx_comparison} shows the comparison between the
 external force [\unit{}{\newton}] measured (in red) and estimated by the
  algorithm (in blue) for the right foot\footnote{The similar
  analysis on the left foot is not retrieved since the left foot is the fixed
   base of the model.  In general, the iDynTree library \citep{Nori2015icub}
    used here for describing the dynamics of the model does not support the
     application of any kind of sensor on the fixed base and therefore neither
      the estimation of its dynamics.}.
Figure \ref{fig:Figs_acc_dx_comparison} shows the comparison between the
 linear acceleration [\unit{}{\meter\per\second^2}] measured by sensors (in
  red) and estimated quantity (in blue) by the algorithm for right upper
   arm, forearm and hand links, respectively, in the tasks T1 (on top) and T2
    (on bottom).  Figure \ref{fig:Figs_acc_sx_comparison} shows the same
     analysis for the related left links.
Similarly, the comparison between the joint acceleration
 [\unit{}{\rad\per\second^2}] (assumed to be) measured by
  the class of `fictitious' DoF-acceleration sensors (in red) and the related
   estimation (in blue) via MAP is shown for the right (in Figure
    \ref{fig:Figs_ddq_dx_comparison}) and for the left (in Figure
     \ref{fig:Figs_ddq_sx_comparison}) shoulder, elbow and
    wrist joints, respectively, for T1 (on top) and T2 (on bottom).

\vspace{1cm}
\begin{figure}[h]
  \centering
    \includegraphics[width=.28\textwidth]{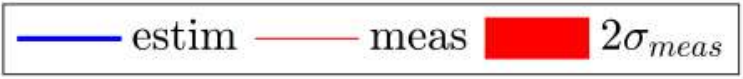}
\end{figure}
\begin{figure}[H]
  \centering
\vspace{-3.2cm}
\includegraphics[width=.9\textwidth]
{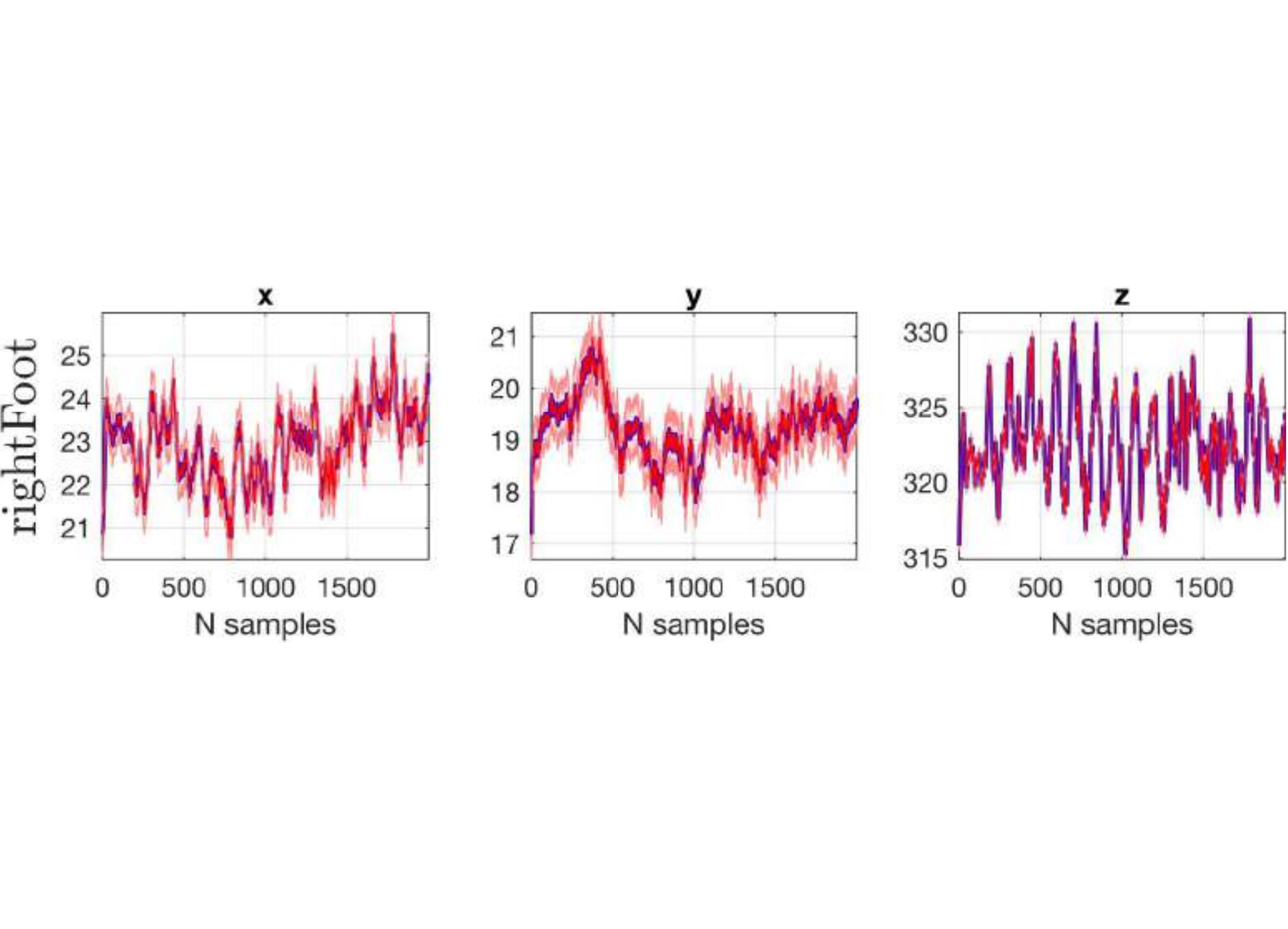}
\end{figure}

\begin{table}[H]
\centering
\caption{RMSE analysis of the force [\unit{}{\newton}] estimation
   w.r.t. the related measured quantity. (Subjects S1, S2, S3, S4, S5; tasks T1,
    T2).}
\label{RMSE_MAPestim_FEXT}
\centering
\scriptsize
\begin{tabular}{c|c|cc}
    \\
    \hline\hline
    \\
 {\textbf{Subject}} & {\textbf{Task}} &   & \textbf{rightFoot} \\
\\
\hline
\\
\multirow{6}{*}{\textbf{S1}} & \multirow{3}{*}{\textbf{T1}} & $f_x$ & $0.0024$ \\
                    &                     & \cellcolor{Gray}$f_y$ & \cellcolor{Gray}$0.0010$ \\
                    &                     & $f_z$ & $0.0211$\\ \cline{2-4}
                    & \multirow{3}{*}{\textbf{T2}} & $f_x$ & $0.0028$ \\
                    &                     & \cellcolor{Gray}$f_y$ & \cellcolor{Gray}$0.0029$ \\
                    &                     & $f_z$ & $0.0215$ \\ \cline{1-4}
\multirow{6}{*}{\textbf{S2}} & \multirow{3}{*}{\textbf{T1}} & $f_x$ & $6.2968~10^{-4}$ \\
                    &                     &\cellcolor{Gray}$f_y$ & \cellcolor{Gray}$3.4981~10^{-4}$ \\
                    &                     & $f_z$ & $0.0185$ \\ \cline{2-4}
                    & \multirow{3}{*}{\textbf{T2}} & $f_x$ & $0.0033$\\
                    &                     & \cellcolor{Gray}$f_y$ & \cellcolor{Gray}$0.0058$ \\
                    &                     & $f_z$ & $0.0188$ \\ \cline{1-4}
\multirow{6}{*}{\textbf{S3}} & \multirow{3}{*}{\textbf{T1}} & $f_x$ & $0.0079$ \\
                    &                     &\cellcolor{Gray}$f_y$ & \cellcolor{Gray}$0.0065$\\
                    &                     & $f_z$ & $0.0246$ \\ \cline{2-4}
                    & \multirow{3}{*}{\textbf{T2}} & $f_x$ & $0.0095$ \\
                    &                     & \cellcolor{Gray}$f_y$ & \cellcolor{Gray}$0.0080$ \\
                    &                     & $f_z$ & $0.0247$\\ \cline{1-4}
\multirow{6}{*}{\textbf{S4}} & \multirow{3}{*}{\textbf{T1}} & $f_x$ & $0.0141$ \\
                    &                     &\cellcolor{Gray}$f_y$ & \cellcolor{Gray}$0.0116$ \\
                    &                     & $f_z$ & $0.0272$ \\ \cline{2-4}
                    & \multirow{3}{*}{\textbf{T2}} & $f_x$ & $0.0208$\\
                    &                     & \cellcolor{Gray}$f_y$ & \cellcolor{Gray}$0.0184$\\
                    &                     & $f_z$ & $0.0282$ \\ \cline{1-4}
\multirow{6}{*}{\textbf{S5}} & \multirow{3}{*}{\textbf{T1}} & $f_x$ & $0.0063$\\
                    &                     &\cellcolor{Gray}$f_y$ & \cellcolor{Gray}$0.0051$ \\
                    &                     & $f_z$ & $0.0206$ \\ \cline{2-4}
                    & \multirow{3}{*}{\textbf{T2}} & $f_x$ & $0.0072$\\
                    &                     & \cellcolor{Gray}$f_y$ & \cellcolor{Gray}$0.0076$ \\
                    &                     & $f_z$ & $0.0211$ \\ 
\\
\hline\hline
\\
\end{tabular}
\end{table}

\newpage
\begin{figure}[h]
  \centering
    \includegraphics[width=.3\textwidth]{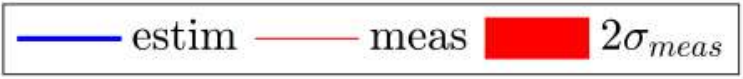}
\end{figure}
\begin{figure}[H]
  \centering
    \vspace{-0.7cm}
\includegraphics[width=.8\textwidth]
{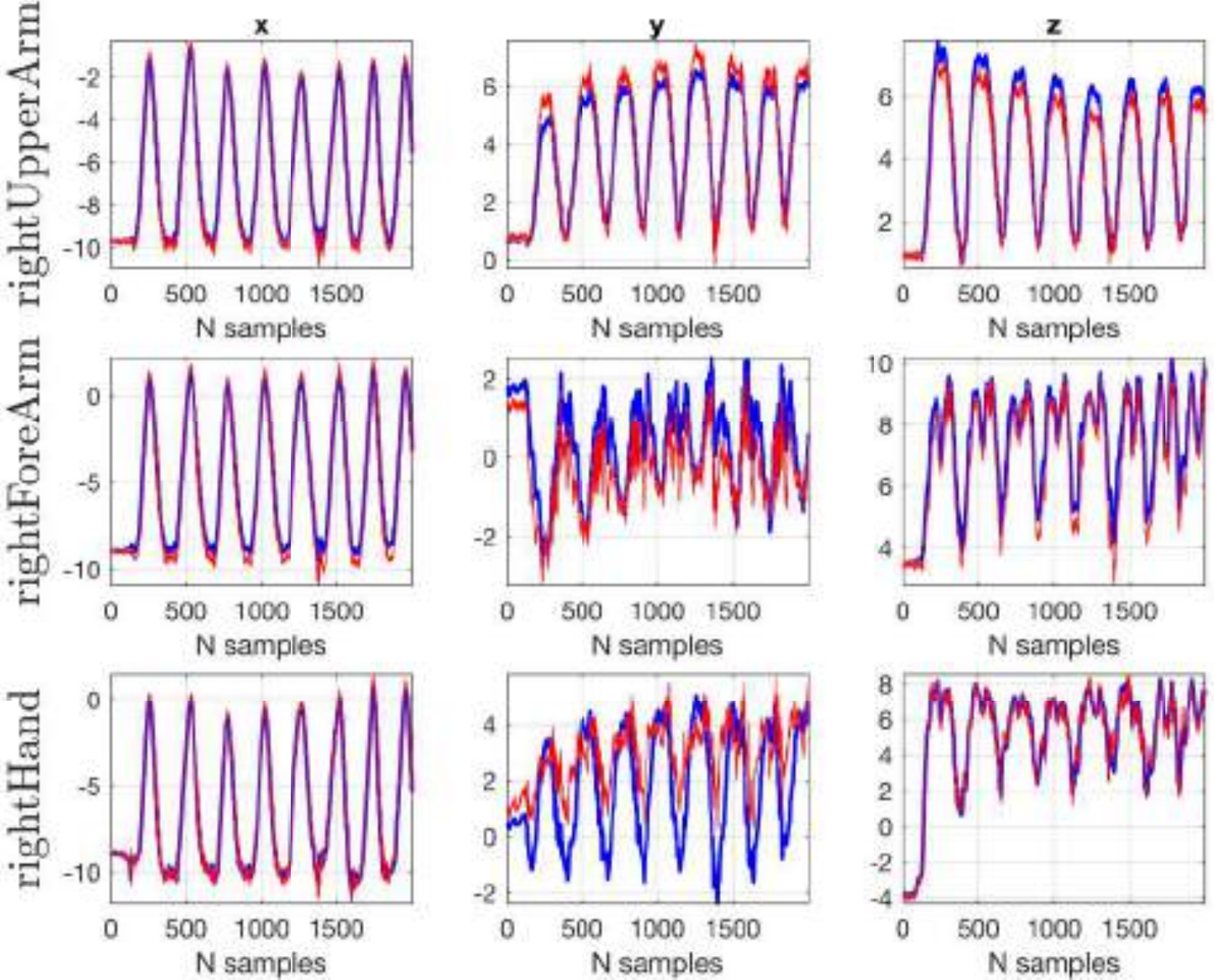}
\end{figure}

\newpage
\begin{figure}[h]
  \centering
    \includegraphics[width=.3\textwidth]{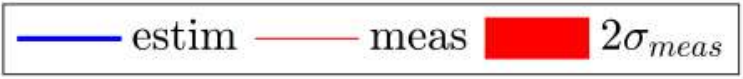}
\end{figure}
\begin{figure}[H]
  \centering
    \vspace{-0.7cm}
\includegraphics[width=.8\textwidth]
{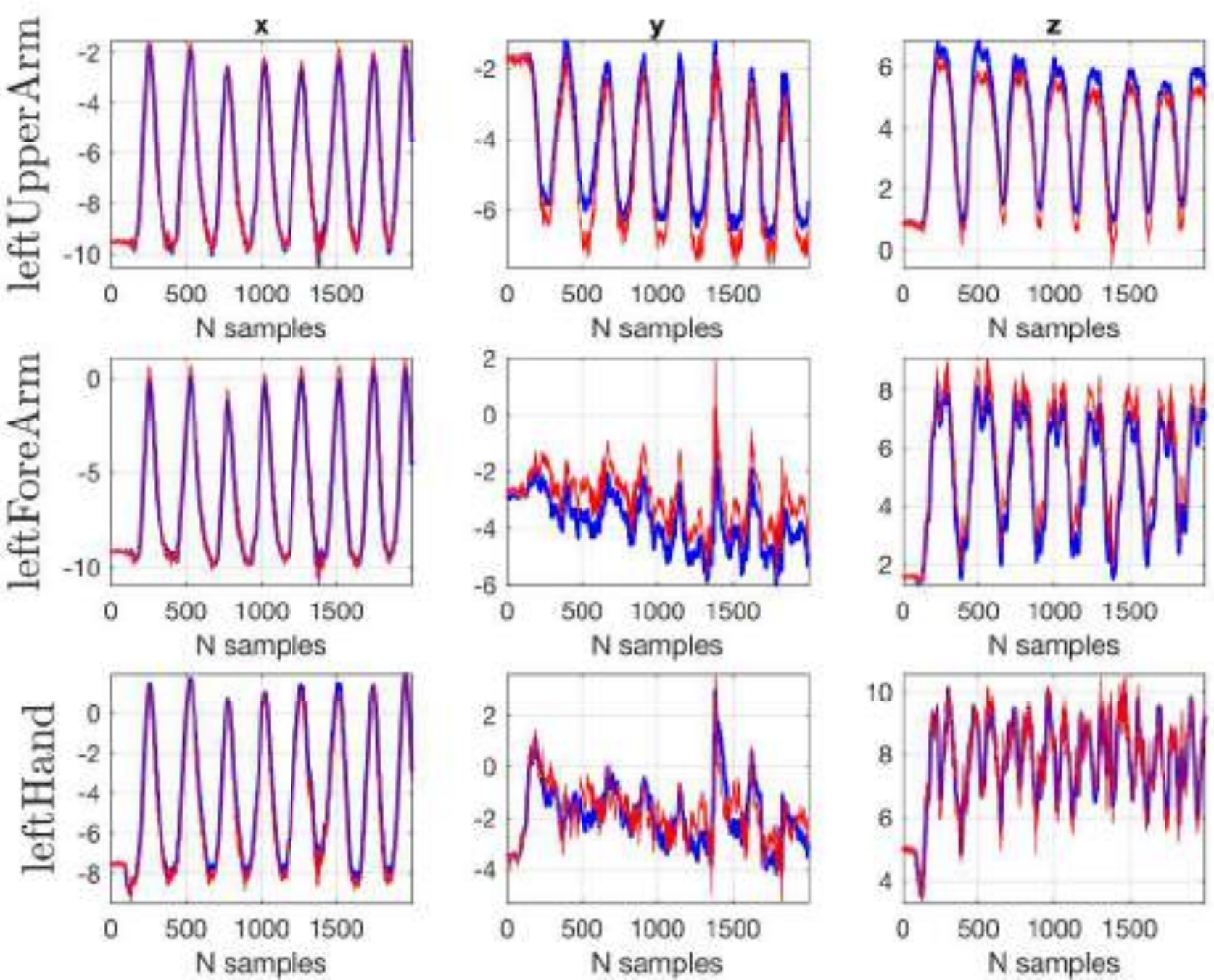}
\end{figure}

\begin{table}[H]
\centering
\caption{RMSE analysis of the linear acceleration
 [\unit{}{\meter\per\second^2}] estimation w.r.t. the related measured
  quantity. (Subjects S1, S2, S3, S4, S5; tasks T1, T2).}
\label{RMSE_MAPestim_ACC}
\centering
\scriptsize
\begin{tabular}{c|c|ccccccc}
    \\
    \hline\hline
    \\
 {\textbf{Subject}} & {\textbf{Task}}  &   & \mcrot{1}{l}{30}{\textbf{rightUpperArm}} & \mcrot{1}{l}{30}{\textbf{rightForeArm}} &
           \mcrot{1}{l}{30}{\textbf{rightHand}}
        & \mcrot{1}{l}{30}{\textbf{leftUpperArm}}  & \mcrot{1}{l}{30}{\textbf{leftForeArm}}  & \mcrot{1}{l}{30}{\textbf{leftHand}} \\
\\
\hline
\\
\multirow{6}{*}{\textbf{S1}} & \multirow{3}{*}{\textbf{T1}}
    & $a_x$ & $1.23$ & $1.67$ & $0.49$ & $0.24$ & $0.20$ & $0.37$ \\
&   & \cellcolor{Gray}$a_y$ & \cellcolor{Gray}$0.68$ & \cellcolor{Gray}$1.07$ & \cellcolor{Gray}$1.01$ & \cellcolor{Gray}$0.58$ & \cellcolor{Gray}$1.03$ & \cellcolor{Gray}$0.45$ \\
&   & $a_z$ & $0.79$ & $1.17$ & $0.92$ & $0.34$ & $0.57$ & $0.75$ \\ \cline{2-9}
& \multirow{3}{*}{\textbf{T2}}
    & $a_x$ & $1.65$ & $1.83$ & $1.17$ & $1.03$ & $1.01$ & $1.09$\\
&   & \cellcolor{Gray}$a_y$ & \cellcolor{Gray}$1.82$ & \cellcolor{Gray}$1.26$ & \cellcolor{Gray}$1.73$ & \cellcolor{Gray}$0.87$ & \cellcolor{Gray}$1.80$ & \cellcolor{Gray}$1.86$\\
&   & $a_z$ & $1.41$ & $1.11$ & $2.55$ & $0.87$ & $1.16$ & $1.45$\\ \cline{1-9}
\multirow{6}{*}{\textbf{S2}} & \multirow{3}{*}{\textbf{T1}}
    & $a_x$ & $0.28$ & $0.49$ & $0.26$ & $0.18$ & $0.34$ & $0.46$\\
&   & \cellcolor{Gray}$a_y$ & \cellcolor{Gray}$0.41$ & \cellcolor{Gray}$0.75$ & \cellcolor{Gray}$1.66$ & \cellcolor{Gray}$0.60$ & \cellcolor{Gray}$0.90$ & \cellcolor{Gray}$0.64$\\
&   & $a_z$ & $0.42$ & $0.60$ & $0.38$ & $0.52$ & $0.70$ & $0.33$\\ \cline{2-9}
& \multirow{3}{*}{\textbf{T2}}
    & $a_x$ & $0.47$ & $0.55$ & $0.86$ & $0.69$ & $0.68$ & $0.81$\\
&   & \cellcolor{Gray}$a_y$ & \cellcolor{Gray}$0.43$ & \cellcolor{Gray}$0.97$ & \cellcolor{Gray}$1.09$ & \cellcolor{Gray}$0.48$ & \cellcolor{Gray}$0.89$ & \cellcolor{Gray}$1.21$\\
&   & $a_z$ & $0.37$ & $0.82$ & $0.54$ & $0.56$ & $0.46$ & $0.53$\\ \cline{1-9}
\multirow{6}{*}{\textbf{S3}} & \multirow{3}{*}{\textbf{T1}}
    & $a_x$ & $0.17$ & $0.17$ & $0.30$ & $0.51$ & $0.55$ & $0.45$\\
&   & \cellcolor{Gray}$a_y$ & \cellcolor{Gray}$0.36$ & \cellcolor{Gray}$0.64$ & \cellcolor{Gray}$0.50$ & \cellcolor{Gray}$0.59$ & \cellcolor{Gray}$0.73$ & \cellcolor{Gray}$1.67$\\
&   & $a_z$ & $0.34$ & $0.41$ & $0.52$ & $1.02$ & $0.42$ & $0.56$\\ \cline{2-9}
& \multirow{3}{*}{\textbf{T2}}
    & $a_x$ & $0.29$ & $0.46$ & $0.38$ & $0.82$ & $0.92$ & $0.51$\\
&   & \cellcolor{Gray}$a_y$ & \cellcolor{Gray}$0.51$ & \cellcolor{Gray}$0.75$ & \cellcolor{Gray}$0.91$ & \cellcolor{Gray}$1.59$ & \cellcolor{Gray}$1.74$ & \cellcolor{Gray}$3.18$\\
&   & $a_z$ & $0.40$ & $0.36$ & $0.51$ & $1.26$ & $1.42$ & $1.40$\\ \cline{1-9}
\multirow{6}{*}{\textbf{S4}} & \multirow{3}{*}{\textbf{T1}}
    & $a_x$ & $0.40$ & $0.64$ & $0.31$ & $0.27$ & $0.36$ & $0.34$\\
&   & \cellcolor{Gray}$a_y$ & \cellcolor{Gray}$0.41$ & \cellcolor{Gray}$0.64$ & \cellcolor{Gray}$0.74$ & \cellcolor{Gray}$0.40$ & \cellcolor{Gray}$0.49$ & \cellcolor{Gray}$0.77$\\
&   & $a_z$ & $0.34$ & $0.54$ & $0.60$ & $0.29$ & $0.33$ & $0.64$\\ \cline{2-9}
& \multirow{3}{*}{\textbf{T2}}
    & $a_x$ & $1.11$ & $1.08$ & $1.24$ & $0.80$ & $1.09$ & $2.05$\\
&   & \cellcolor{Gray}$a_y$ & \cellcolor{Gray}$0.67$ & \cellcolor{Gray}$1.53$ & \cellcolor{Gray}$1.67$ & \cellcolor{Gray}$1.05$ & \cellcolor{Gray}$1.28$ & \cellcolor{Gray}$1.91$\\
&   & $a_z$ & $0.57$ & $1.01$ & $1.05$ & $0.66$ & $0.60$ & $0.73$\\ \cline{1-9}
\multirow{6}{*}{\textbf{S5}} & \multirow{3}{*}{\textbf{T1}}
    & $a_x$ & $0.34$ & $0.51$ & $0.72$ & $0.32$ & $0.30$ & $0.76$\\
&   & \cellcolor{Gray}$a_y$ & \cellcolor{Gray}$0.65$ & \cellcolor{Gray}$0.72$ & \cellcolor{Gray}$1.06$ & \cellcolor{Gray}$0.57$ & \cellcolor{Gray}$0.57$ & \cellcolor{Gray}$1.17$\\
&   & $a_z$ & $0.43$ & $0.33$ & $1.03$ & $0.61$ & $0.41$ & $0.75$\\ \cline{2-9}
& \multirow{3}{*}{\textbf{T2}}
    & $a_x$ & $0.32$ & $0.58$ & $0.87$ & $0.52$ & $0.62$ & $1.03$\\
&   & \cellcolor{Gray}$a_y$ & \cellcolor{Gray}$0.55$ & \cellcolor{Gray}$0.83$ & \cellcolor{Gray}$0.78$ & \cellcolor{Gray}$0.54$ & \cellcolor{Gray}$1.02$ & \cellcolor{Gray}$1.09$\\
&   & $a_z$ & $0.38$ & $0.39$ & $0.69$ & $0.87$ & $0.94$ & $0.95$\\ 
\\
\hline\hline
\\
\end{tabular}
\end{table}

\newpage
\begin{figure}[h]
  \centering
    \includegraphics[width=.29\textwidth]{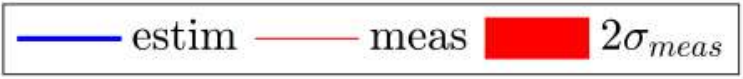}
\end{figure}
\begin{figure}[H]
  \centering
\vspace{-0.7cm}
\includegraphics[width=.9\textwidth]
{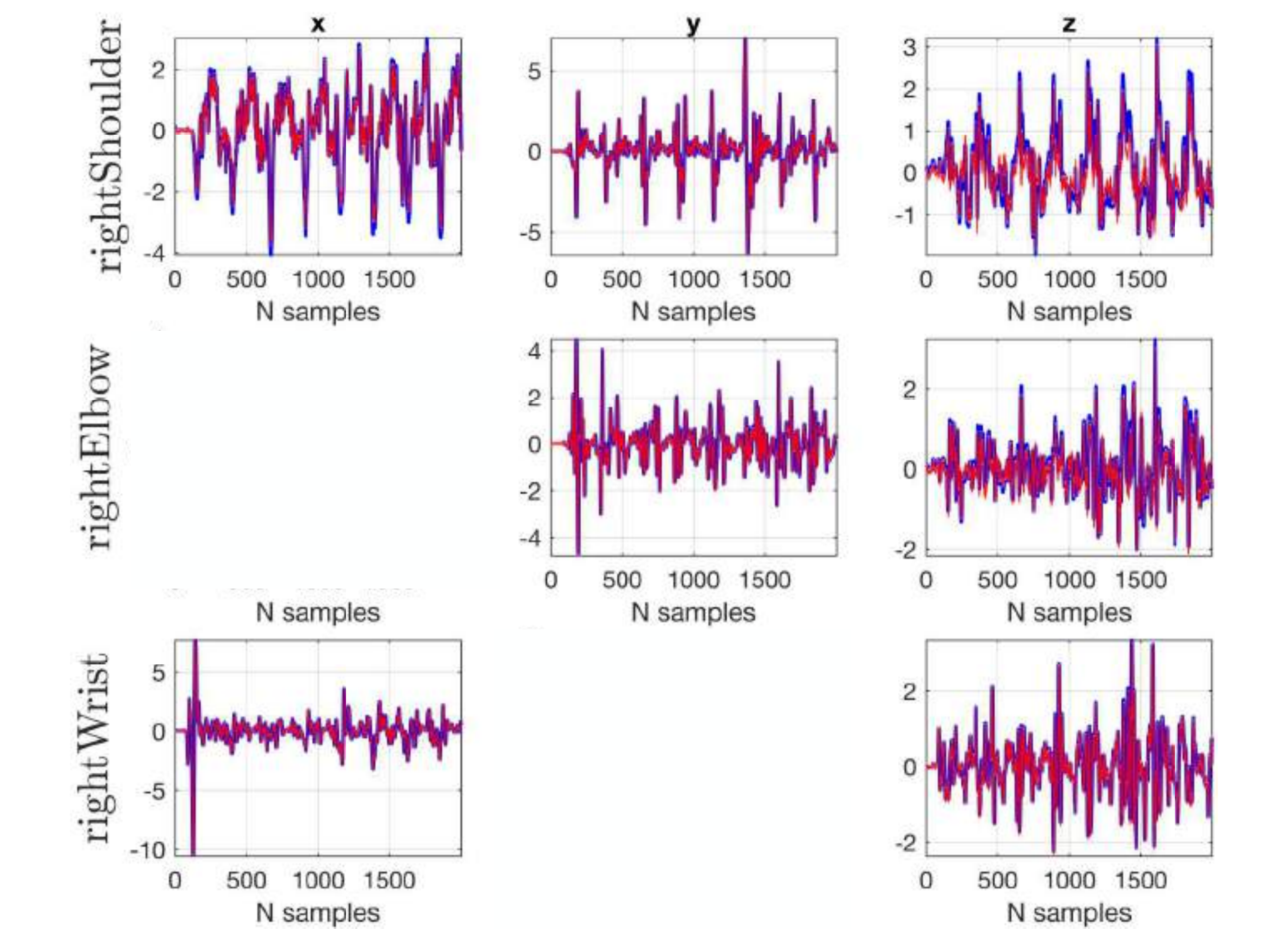}
\end{figure}

\newpage
\begin{figure}[h]
  \centering
    \includegraphics[width=.29\textwidth]{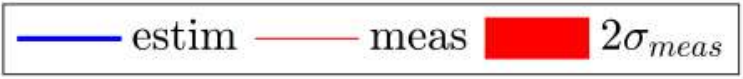}
\end{figure}
\begin{figure}[H]
  \centering
\vspace{-0.7cm}
\includegraphics[width=.9\textwidth]
{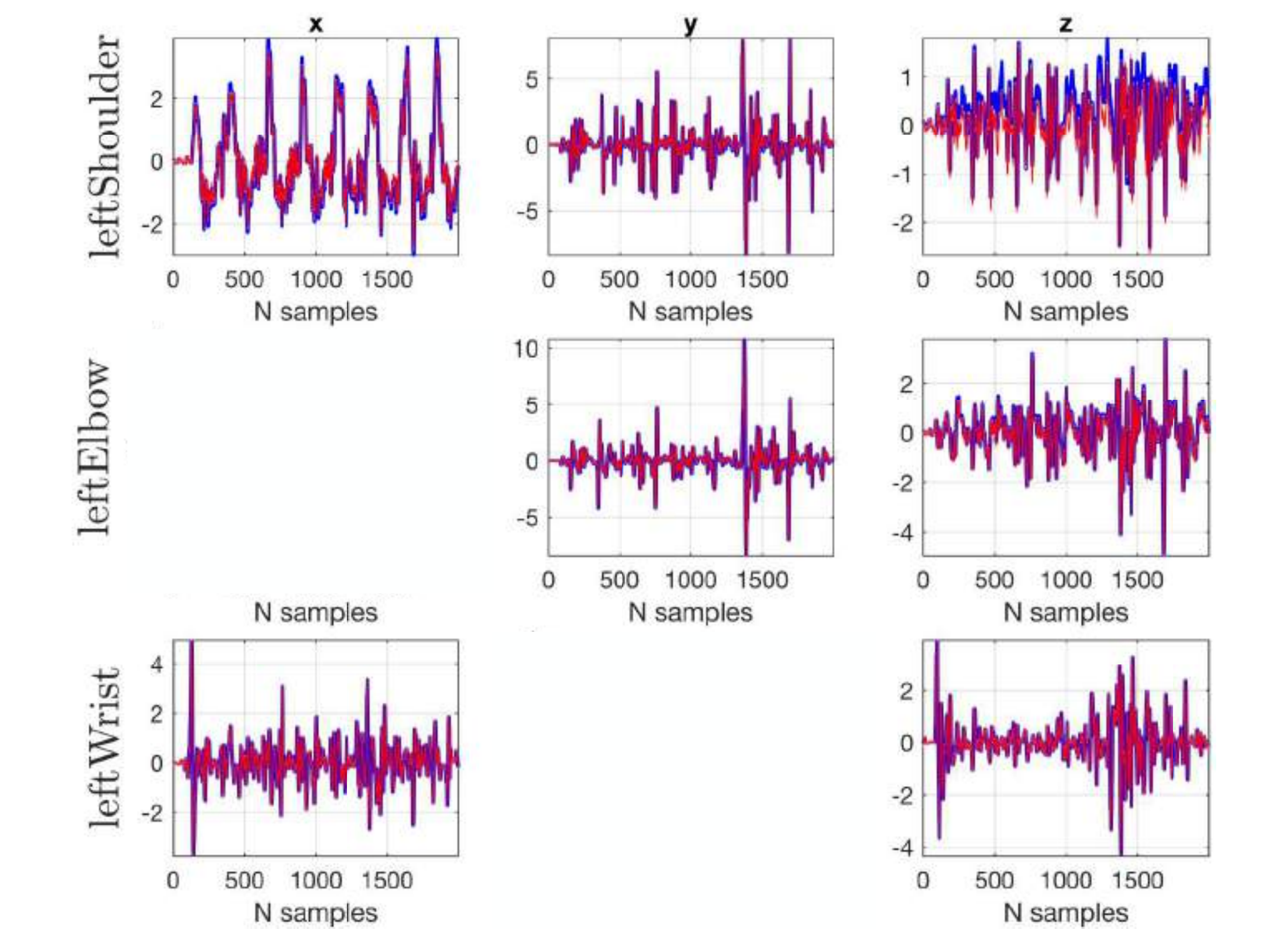}
\end{figure}

\begin{table}[H]
\centering
\caption{RMSE analysis of the joint acceleration
  [\unit{}{\rad\per\second^2}] estimation w.r.t. the related measured
   quantity. (Subjects S1, S2, S3, S4, S5; tasks T1, T2).}
\label{RMSE_MAPestim_DDQ}
\centering
\scriptsize
\begin{tabular}{c|c|ccccccc}
    \\
    \hline\hline
    \\
 {\textbf{Subject}} & {\textbf{Task}}&   & \mcrot{1}{l}{30}{\textbf{rightShoulder}} & \mcrot{1}{l}{30}{\textbf{rightElbow}} &
           \mcrot{1}{l}{30}{\textbf{rightWrist}} 
        & \mcrot{1}{l}{30}{\textbf{leftShoulder}}  & \mcrot{1}{l}{30}{\textbf{leftElbow}}  & \mcrot{1}{l}{30}{\textbf{leftWrist}} \\
\\
\hline
\\
\multirow{6}{*}{\textbf{S1}} & \multirow{3}{*}{\textbf{T1}}
    & $\ddot{q}_x$ & $0.30$ & $    $ & $0.03$ & $0.27$ & $    $ & $0.02$\\
&   & \cellcolor{Gray}$\ddot{q}_y$ & \cellcolor{Gray}$0.09$ & \cellcolor{Gray}$0.11$ & \cellcolor{Gray}$    $ & \cellcolor{Gray}$0.04$ & \cellcolor{Gray}$0.01$ & \cellcolor{Gray}$    $\\
&   & $\ddot{q}_z$ & $0.22$ & $0.30$ & $0.02$ & $0.20$ & $0.14$ & $0.02$\\ \cline{2-9}
& \multirow{3}{*}{\textbf{T2}}
    & $\ddot{q}_x$ & $0.73$ & $    $ & $0.09$ & $0.22$ & $    $ & $0.05$\\
&   & \cellcolor{Gray}$\ddot{q}_y$ & \cellcolor{Gray}$0.53$ & \cellcolor{Gray}$0.20$ & \cellcolor{Gray}$    $ & \cellcolor{Gray}$0.50$ & \cellcolor{Gray}$0.26$ & \cellcolor{Gray}$    $\\
&   & $\ddot{q}_z$ & $0.82$ & $0.39$ & $0.05$ & $0.38$ & $0.38$ & $0.06$\\ \cline{1-9}
\multirow{6}{*}{\textbf{S2}} & \multirow{3}{*}{\textbf{T1}}
    & $\ddot{q}_x$ & $0.13$ & $    $ & $0.02$ & $0.14$ & $    $ & $0.02$\\
&   & \cellcolor{Gray}$\ddot{q}_y$ & \cellcolor{Gray}$0.05$ & \cellcolor{Gray}$0.04$ & \cellcolor{Gray}$    $ & \cellcolor{Gray}$0.04$ & \cellcolor{Gray}$0.03$ & \cellcolor{Gray}$    $\\
&   & $\ddot{q}_z$ & $0.23$ & $0.13$ & $0.05$ & $0.27$ & $0.08$ & $0.01$\\ \cline{2-9}
& \multirow{3}{*}{\textbf{T2}}
    & $\ddot{q}_x$ & $0.12$ & $    $ & $0.02$ & $0.19$ & $    $ & $0.03$\\
&   & \cellcolor{Gray}$\ddot{q}_y$ & \cellcolor{Gray}$0.08$ & \cellcolor{Gray}$0.03$ & \cellcolor{Gray}$    $ & \cellcolor{Gray}$0.13$ & \cellcolor{Gray}$0.06$ & \cellcolor{Gray}$    $\\
&   & $\ddot{q}_z$ & $0.10$ & $0.08$ & $0.03$ & $0.16$ & $0.14$ & $0.03$\\ \cline{1-9}
\multirow{6}{*}{\textbf{S3}} & \multirow{3}{*}{\textbf{T1}}
    & $\ddot{q}_x$ & $0.22$ & $    $ & $0.02$ & $0.36$ & $    $ & $0.06$\\
&   & \cellcolor{Gray}$\ddot{q}_y$ & \cellcolor{Gray}$0.02$ & \cellcolor{Gray}$0.04$ & \cellcolor{Gray}$    $ & \cellcolor{Gray}$0.13$ & \cellcolor{Gray}$0.11$ & \cellcolor{Gray}$    $\\
&   & $\ddot{q}_z$ & $0.24$ & $0.13$ & $0.02$ & $0.19$ & $0.20$ & $0.03$\\ \cline{2-9}
& \multirow{3}{*}{\textbf{T2}}
    & $\ddot{q}_x$ & $0.13$ & $    $ & $0.02$ & $0.44$ & $    $ & $0.10$\\
&   & \cellcolor{Gray}$\ddot{q}_y$ & \cellcolor{Gray}$0.16$ & \cellcolor{Gray}$0.06$ & \cellcolor{Gray}$    $ & \cellcolor{Gray}$0.76$ & \cellcolor{Gray}$0.66$ & \cellcolor{Gray}$    $\\
&   & $\ddot{q}_z$ & $0.38$ & $0.22$ & $0.03$ & $0.53$ & $0.41$ & $0.06$\\ \cline{1-9}
\multirow{6}{*}{\textbf{S4}} & \multirow{3}{*}{\textbf{T1}}
    & $\ddot{q}_x$ & $0.27$ & $    $ & $0.02$ & $0.26$ & $    $ & $0.03$\\
&   & \cellcolor{Gray}$\ddot{q}_y$ & \cellcolor{Gray}$0.02$ & \cellcolor{Gray}$0.03$ & \cellcolor{Gray}$    $ & \cellcolor{Gray}$0.03$ & \cellcolor{Gray}$0.01$ & \cellcolor{Gray}$    $\\
&   & $\ddot{q}_z$ & $0.17$ & $0.07$ & $0.03$ & $0.23$ & $0.12$ & $0.03$\\ \cline{2-9}
& \multirow{3}{*}{\textbf{T2}}
    & $\ddot{q}_x$ & $0.24$ & $    $ & $0.03$ & $0.25$ & $    $ & $0.06$\\
&   & \cellcolor{Gray}$\ddot{q}_y$ & \cellcolor{Gray}$0.27$ & \cellcolor{Gray}$0.06$ & \cellcolor{Gray}$    $ & \cellcolor{Gray}$0.34$ & \cellcolor{Gray}$0.14$ & \cellcolor{Gray}$    $\\
&   & $\ddot{q}_z$ & $0.40$ & $0.31$ & $0.06$ & $0.27$ & $0.28$ & $0.06$\\ \cline{1-9}
\multirow{6}{*}{\textbf{S5}} & \multirow{3}{*}{\textbf{T1}}
    & $\ddot{q}_x$ & $0.29$ & $    $ & $0.04$ & $0.30$ & $    $ & $0.04$\\
&   & \cellcolor{Gray}$\ddot{q}_y$ & \cellcolor{Gray}$0.06$ & \cellcolor{Gray}$0.05$ & \cellcolor{Gray}$    $ & \cellcolor{Gray}$0.07$ & \cellcolor{Gray}$0.07$ & \cellcolor{Gray}$    $\\
&   & $\ddot{q}_z$ & $0.31$ & $0.15$ & $0.03$ & $0.38$ & $0.13$ & $0.02$\\ \cline{2-9}
& \multirow{3}{*}{\textbf{T2}}
    & $\ddot{q}_x$ & $0.14$ & $    $ & $0.03$ & $0.17$ & $    $ & $0.04$\\
&   & \cellcolor{Gray}$\ddot{q}_y$ & \cellcolor{Gray}$0.14$ & \cellcolor{Gray}$0.04$ & \cellcolor{Gray}$    $ & \cellcolor{Gray}$0.32$ & \cellcolor{Gray}$0.11$ & \cellcolor{Gray}$    $\\
&   & $\ddot{q}_z$ & $0.17$ & $0.10$ & $0.02$ & $0.30$ & $0.24$ & $0.03$\\
\\
\hline\hline
\\
\end{tabular}
\end{table}

\newpage
Hereafter, the human joint torque estimated by the MAP algorithm.  No direct
 comparison with the related measured quantity is possible in this case.

\begin{figure}[h]
  \centering
    \includegraphics[width=.1\textwidth]{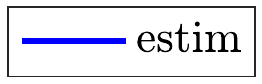}
\end{figure}
\begin{figure}[H]
  \centering
\vspace{-0.7cm}
\includegraphics[width=.84\textwidth]
{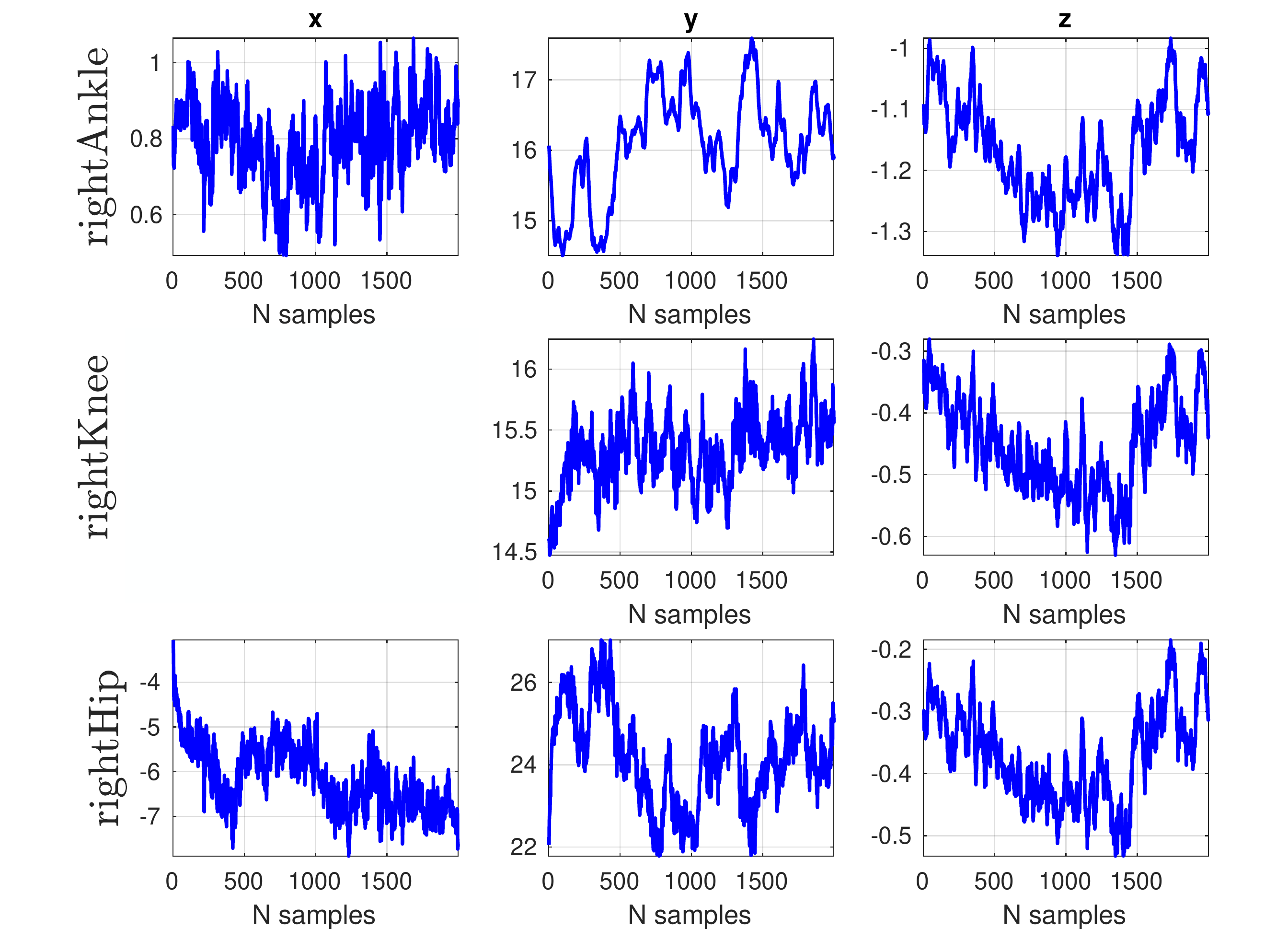}
\end{figure}

\begin{figure}[H]
  \centering
 \vspace{-0.2cm}
\includegraphics[width=.84\textwidth]
{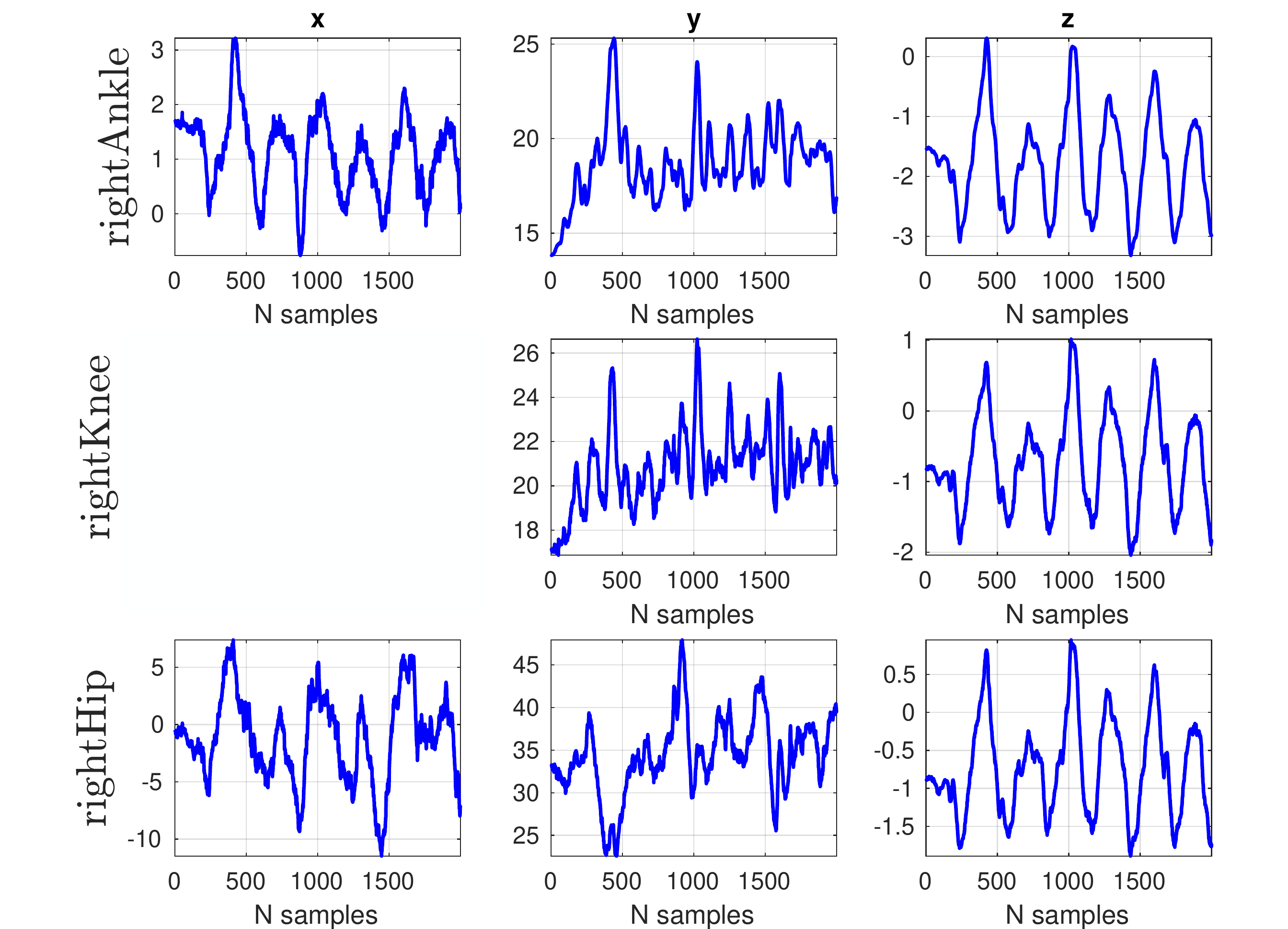}
\caption{Joint torque [\unit{}{\newton\meter}] estimation for the S2 right
 ankle, knee and hip joints, respectively, in the tasks T1 (top) and T2
  (bottom).}
  \label{fig:Figs_dx_leg_estim}
\end{figure}
\newpage
\begin{figure}[h]
  \centering
    \includegraphics[width=.1\textwidth]{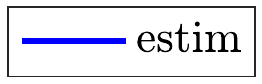}
\end{figure}
\begin{figure}[H]
  \centering
\vspace{-0.7cm}
\includegraphics[width=.86\textwidth]
{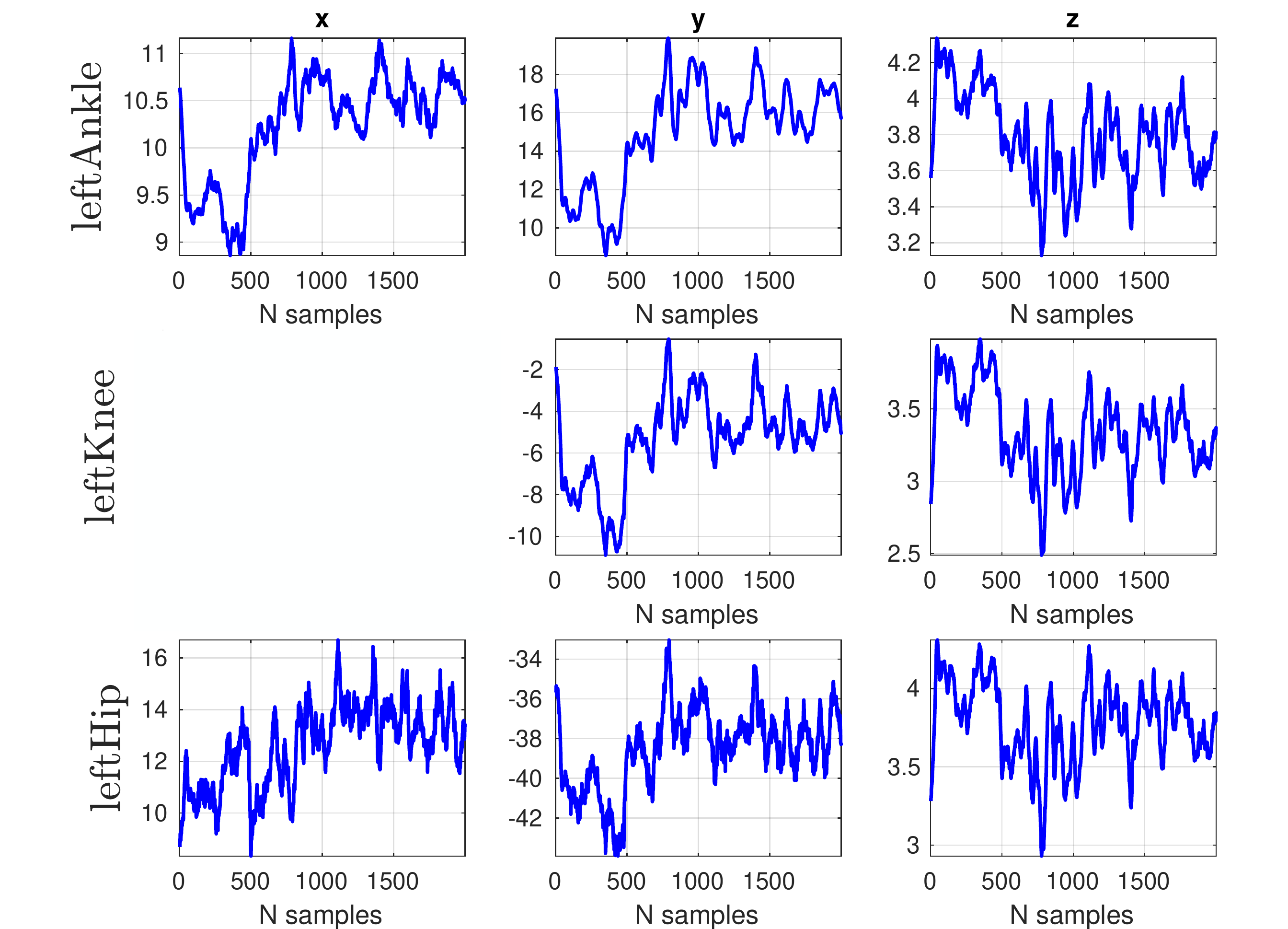}
\end{figure}

\begin{figure}[H]
  \centering
 \vspace{-0.2cm}
\includegraphics[width=.86\textwidth]
{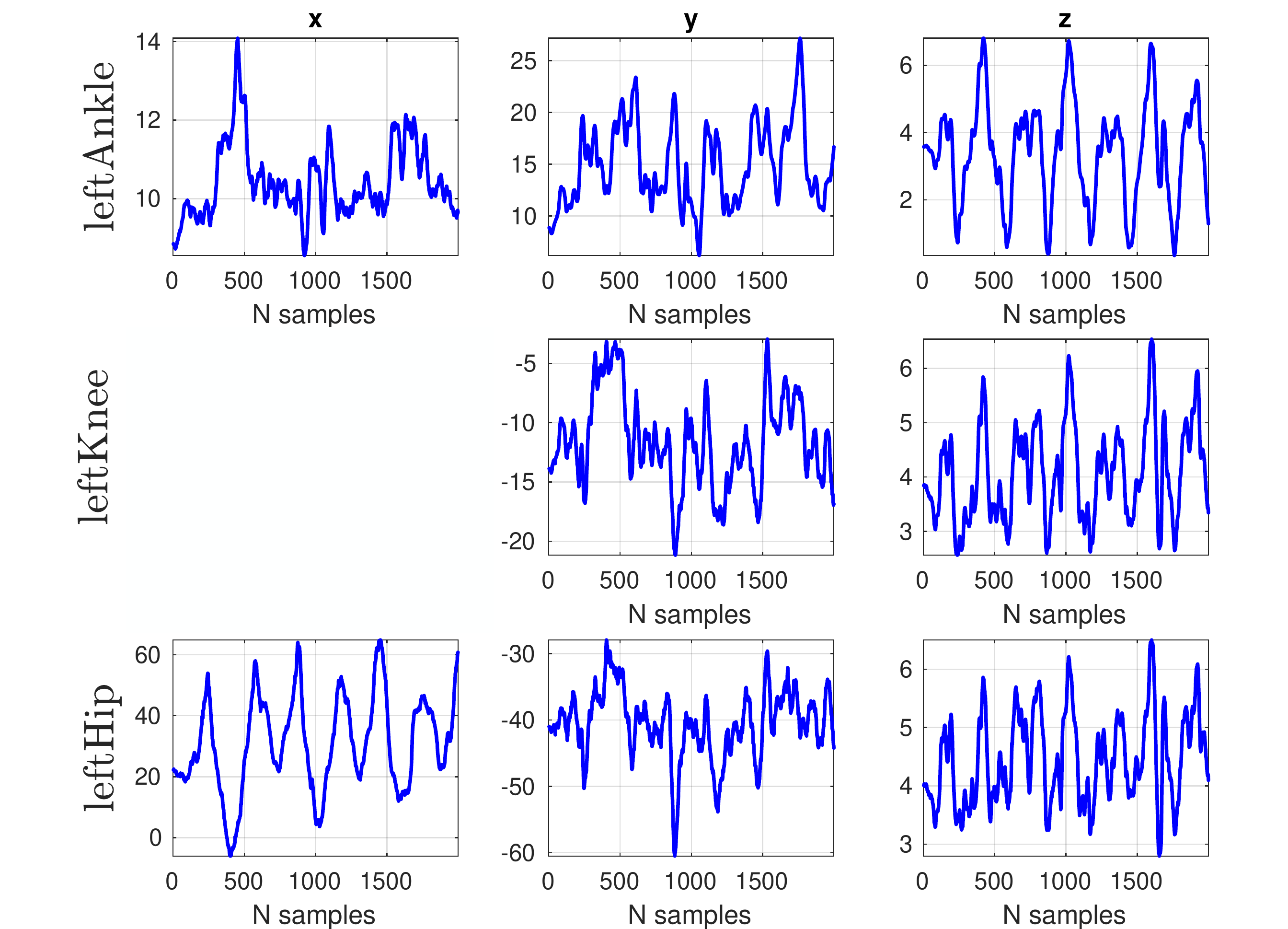}
\caption{Joint torque [\unit{}{\newton\meter}] estimation for the S2 left
 ankle, knee and hip joints, respectively, in the tasks T1 (top) and T2
  (bottom).}
  \label{fig:Figs_sx_leg_estim}
\end{figure}
\newpage
\begin{figure}[h]
  \centering
    \includegraphics[width=.1\textwidth]{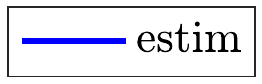}
\end{figure}
\begin{figure}[H]
  \centering
\vspace{-1.5cm}
\includegraphics[width=.84\textwidth]
{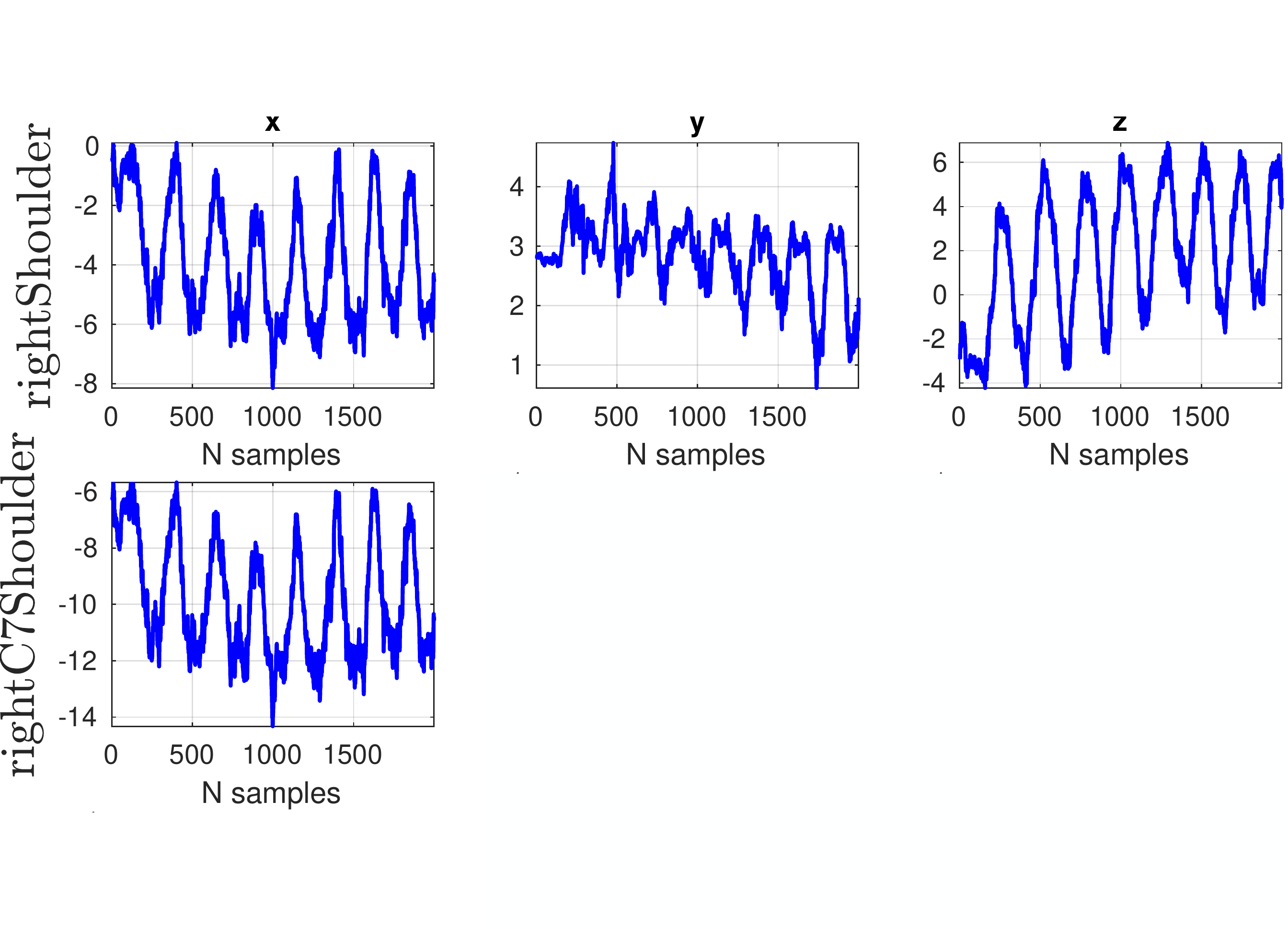}
\end{figure}

\begin{figure}[H]
  \centering
  \vspace{-2cm}
\includegraphics[width=.84\textwidth]
{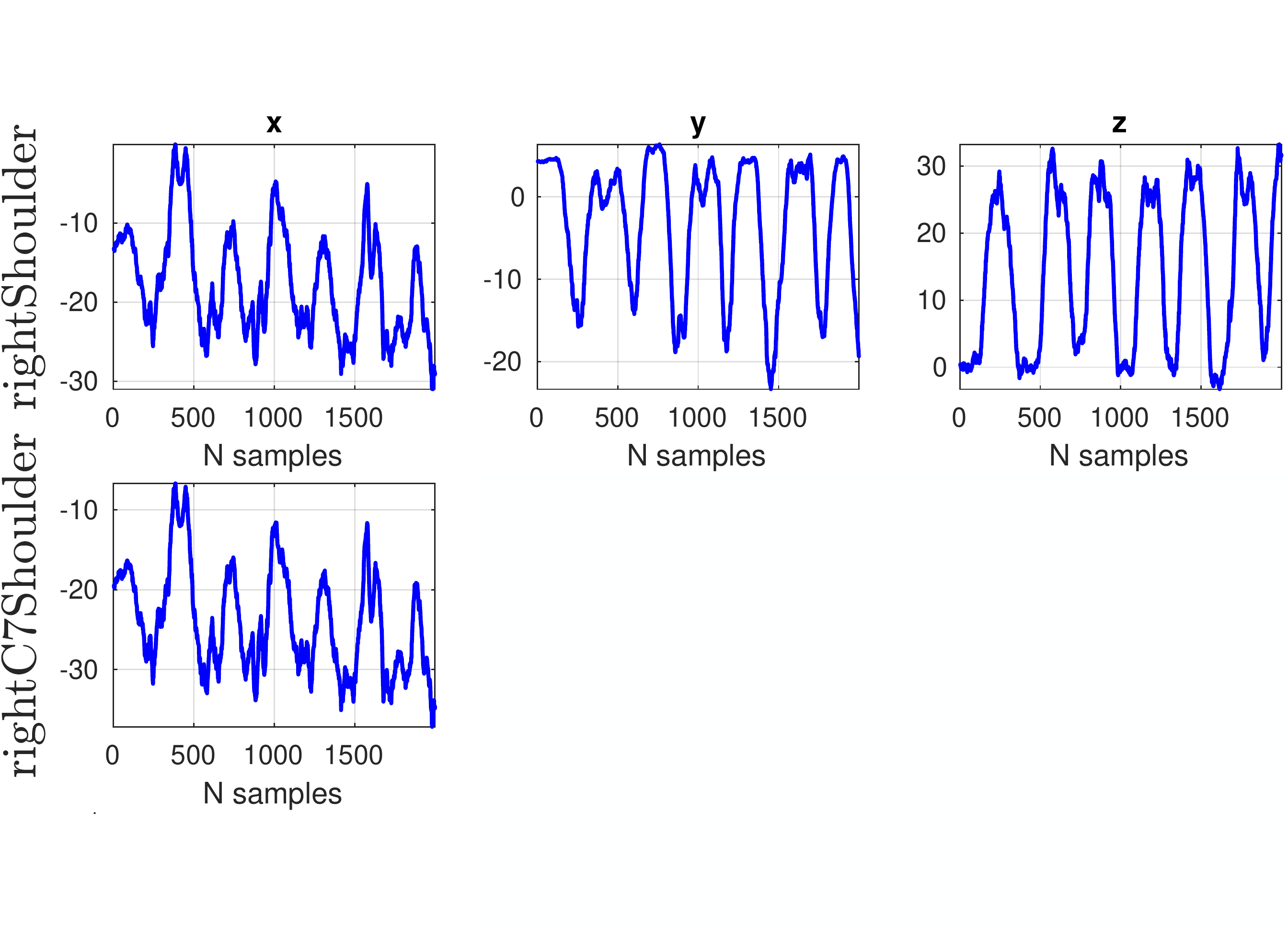}
\vspace{-0.5cm}
\caption{Joint torque [\unit{}{\newton\meter}] estimation for the S2 right
 shoulder and C7 shoulder joints, respectively, in the tasks T1 (top) and T2
  (bottom).}
  \label{fig:Figs_dx_shoulder_estim}
\end{figure}
\newpage
\begin{figure}[h]
  \centering
    \includegraphics[width=.1\textwidth]{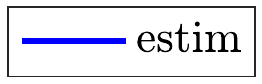}
\end{figure}
\begin{figure}[H]
  \centering
\vspace{-1.5cm}
\includegraphics[width=.84\textwidth]
{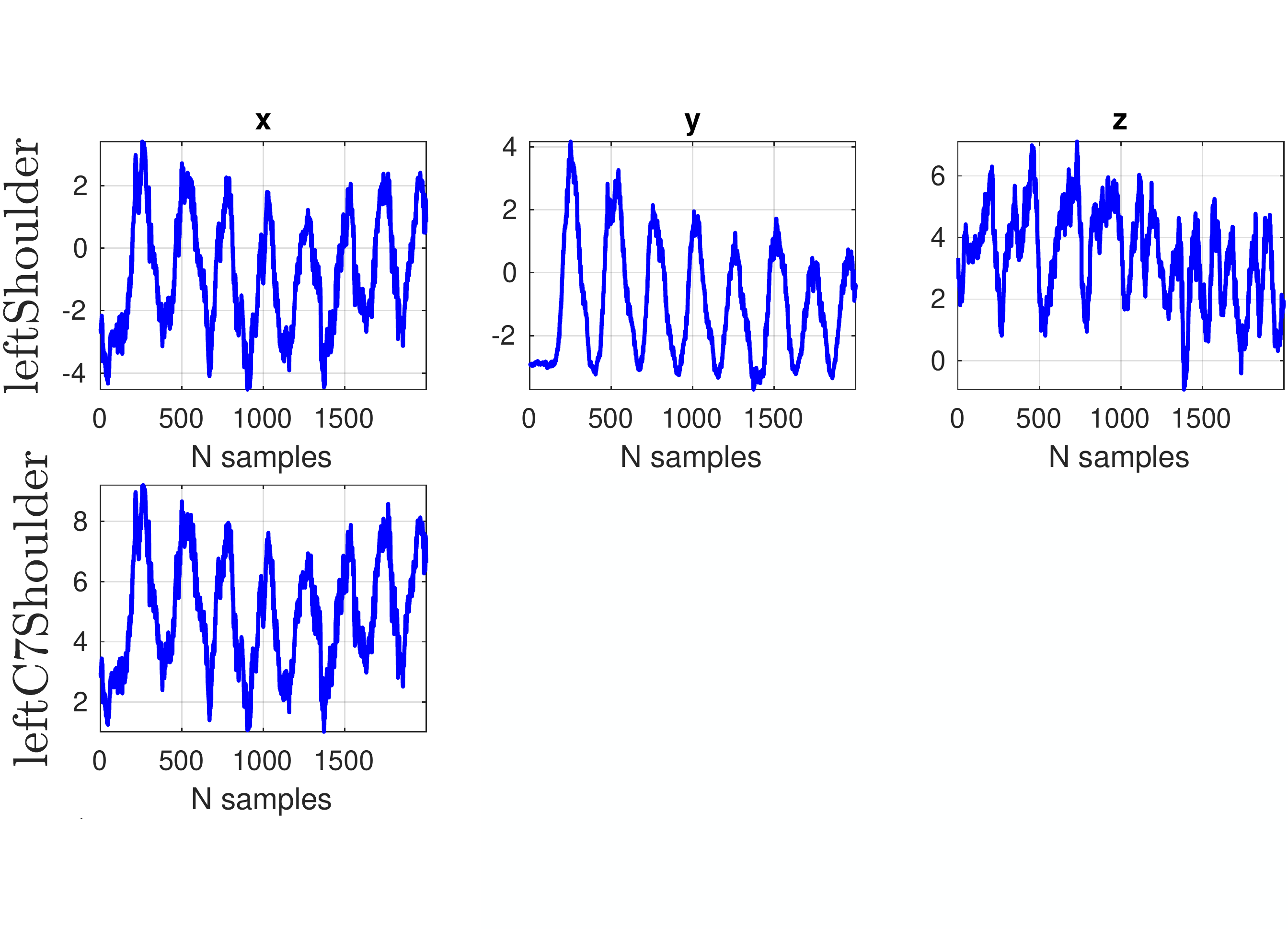}
\end{figure}

\begin{figure}[H]
  \centering
\vspace{-2.1cm}
\includegraphics[width=.84\textwidth]
{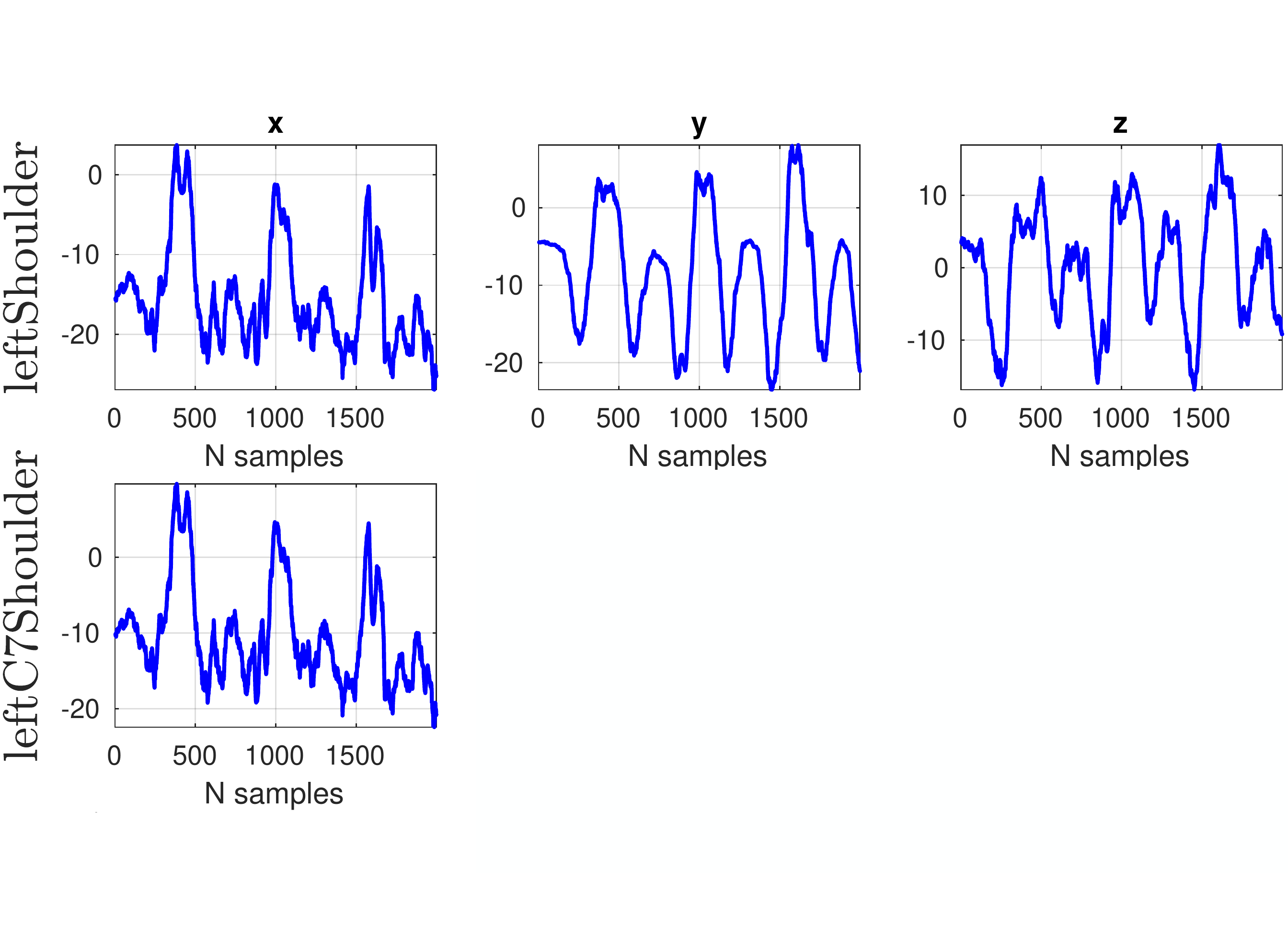}
\vspace{-0.5cm}
\caption{Joint torque [\unit{}{\newton\meter}] estimation for the S2 left
 shoulder and C7 shoulder joints, respectively, in the tasks T1 (top) and T2
  (bottom).}
  \label{fig:Figs_sx_shoulder_estim}
\end{figure}

 Figures from \ref{fig:Figs_dx_leg_estim} to \ref{fig:Figs_sx_shoulder_estim}
  show the joint torque estimation for the right and left ankle, knee,
   hip, shoulder and C7 shoulder joints, respectively.

\newpage
\subsection{Incremental Sensor Fusion Analysis}
 \label{incrementalSensorAnalysis_UW}

As previously said, the novelty of the MAP framework consists in replacing the
 classical RNEA boundary conditions with readings coming from different types of
  sensors.  In this Section, we discuss the benefits of the multi-sensor data
   fusion for solving the  estimation problem by characterizing the effects of
    the data fusion on the covariance of the estimator.
The benefit of the MAP is such that the more sensors we use in the estimation,
 the better the estimation itself will be.  Let us consider writing Equations
  \eqref{eq:sigmaDbar} and \eqref{eq:sigma_dgiveny} in a more compact form,
   such as
\vspace{-0.2cm}
\begin{eqnarray}\label{compactMAPsol}
     \bm{\Sigma_{d|y}}= \left(\bm D^\top \bm{\Sigma}_D^{-1} \bm D + \bm
      {\Sigma}_d^{-1} + \bm Y^\top \bm
{\Sigma}_{y|d}^{-1}\bm Y\right)^{-1} ~.
\end{eqnarray}
Let us assume $m$ multiple statistically independent measurements
\begin{subequations}\label{multipleMeas}
\begin{eqnarray*}
\bm y_1 &=& \bm Y_1 \bm d + \bm b_{Y_1}~,
\\
\bm y_2 &=& \bm Y_2 \bm d + \bm b_{Y_2}~,
\\
&\vdots 
\\
\bm y_m &=& \bm Y_m \bm d + \bm b_{Y_m}~,
\end{eqnarray*}
\end{subequations}
this yields to a diagonal structure for the matrix $\bm {\Sigma}_{y|d}^{-1}$.
Thus, we have:
\begin{eqnarray}
\bm Y^\top \bm {\Sigma}_{y|d}^{-1}\bm Y &=&\begin{bmatrix} \bm Y_1^\top &
\bm Y_2^\top & \dots & \bm Y_m^\top \end{bmatrix}
\begin{bmatrix}
\bm {\Sigma}_{y_1|d}^{-1} & \bm 0 & \dots & \bm 0\\
\bm 0 & \bm {\Sigma}_{y_2|d}^{-1} & \dots & \bm 0\\
\vdots & \vdots & \ddots &\vdots\\
\bm 0 & \bm 0 & \dots & \bm {\Sigma}_{y_m|d}^{-1}
\end{bmatrix}
\begin{bmatrix} \bm Y_1 \\ \bm Y_2\\ \vdots \\ \bm Y_m \end{bmatrix}
\notag
\\
&=&  \bm Y_1^\top \bm{\Sigma}_{y_1|d}^{-1}\bm Y_1 +
\bm Y_2^\top \bm{\Sigma}_{y_2|d}^{-1}\bm Y_2 +
\dots + \bm Y_m^\top \bm {\Sigma}_{y_m|d}^{-1}\bm Y_m~.
\end{eqnarray}
With an abuse of notation, let $\bm d|\bm y_m$ be the estimator
 which exploits all the measurements from $\bm y_1$ up to $\bm y_m$. The
  addition of each measurement induces changes into the associated covariance
matrix according to the following recursive equation:
\vspace{-0.2cm}
\begin{eqnarray}\label{eq:sigma_d|ym}
     \bm {\Sigma}_{d|y_{m}}^{-1} = \bm
{\Sigma}_{d|y_{m-1}}^{-1} + \bm Y_m^\top \bm{\Sigma}_{y_m}^{-1}\bm Y_m~,
 \end{eqnarray}
where, for ${m} = 1$ , the initial condition is
\vspace{-0.2cm}
\begin{eqnarray}\label{eq:sigma_d|ym_init}
\bm {\Sigma}_{d|y_{0}}^{-1}=
 \bm D^\top \bm{\Sigma}_D^{-1} \bm D + \bm {\Sigma}_d^{-1}~.
\end{eqnarray}

\begin{figure}[H]
  \centering
  \begin{subfigure}[b]{0.50\textwidth}
    \includegraphics[width=\textwidth]
    {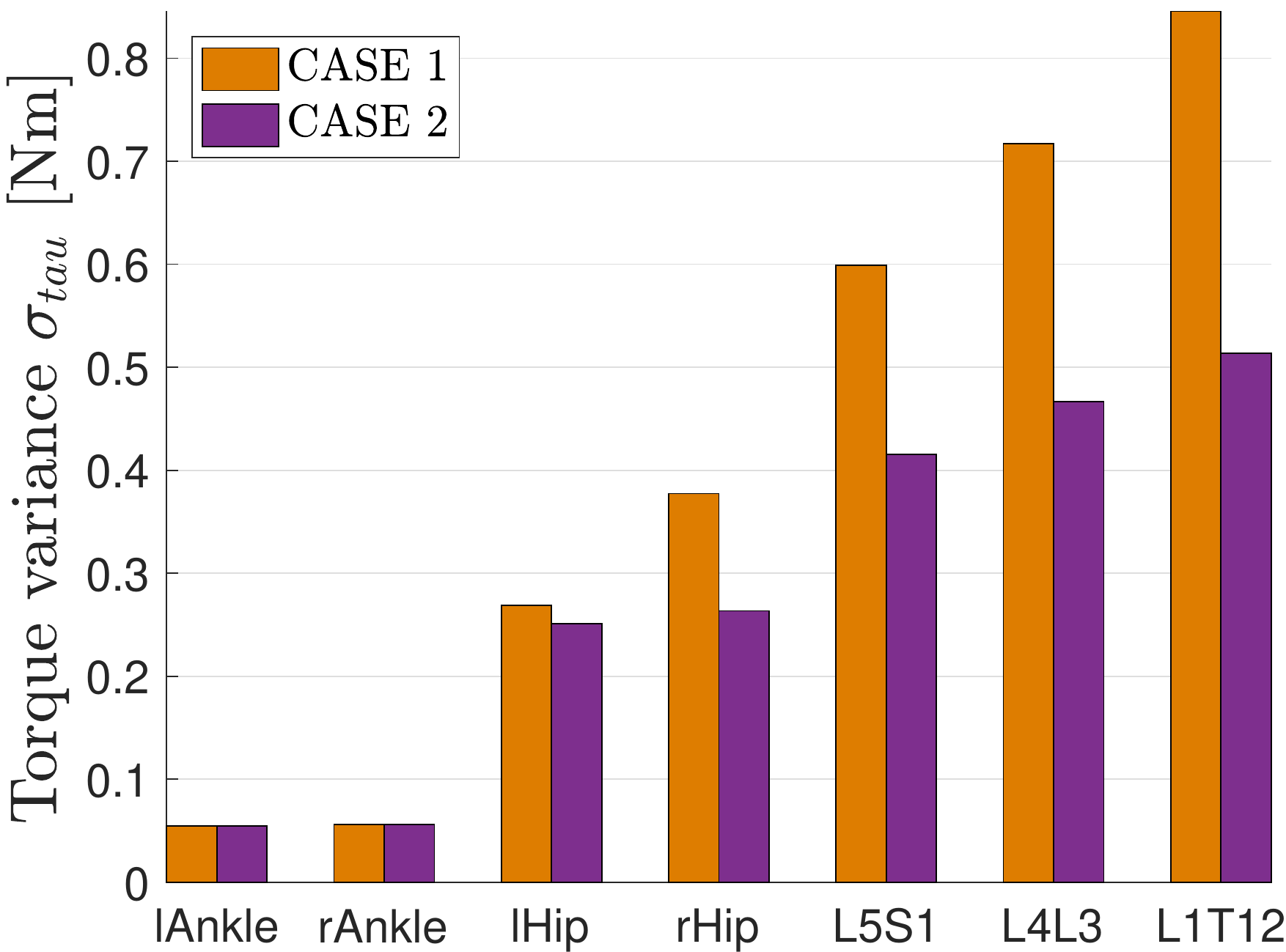}
  \caption{}
\label{fig:incremSens_T1}
        \end{subfigure}
~
  \begin{subfigure}[b]{0.47\textwidth}
    \includegraphics[width=\textwidth]
    {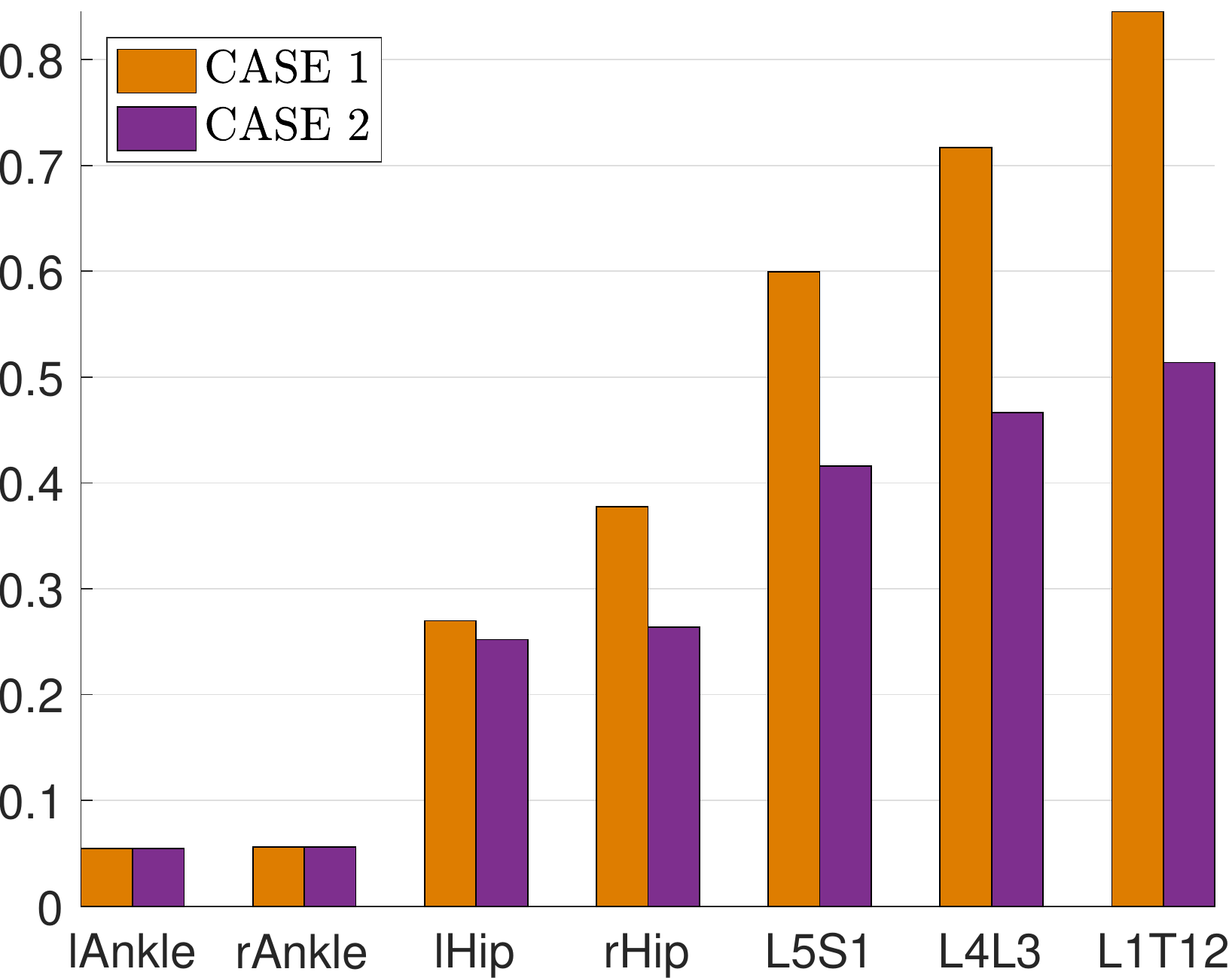}
  \caption{}
\label{fig:incremSens_T2}
      \end{subfigure}
\caption{$5$-subjects mean torque variance [\unit{}{\newton\meter}] by using two
 different versions of the measurements equation: \emph{CASE 1} (in
  orange) and \emph{CASE 2} (in violet), for both left and
  right ankle, hip joints, respectively and L5S1, L4L3, L1T12 joints.  Plots
   are referred to tasks (\subref{fig:incremSens_T1}) T1 and
    (\subref{fig:incremSens_T2}) T2.} 
\label{incremAnalysis_UW}
\end{figure}

A situation of two sets of measurements equations is here evaluated
for the subjects in Table \ref{Subjects_for_analysis_UW}, for tasks T1 and T2.
In particular, the measurements equation \eqref{eq:measEquation} is built for
 two different cases:
\vspace{0.25cm}
\begin{subequations}\label{eq:casesForIncrementalAnalisys}
\begin{eqnarray}
\label{case1_noIMUS}
\mbox{\emph{CASE 1}} \quad \quad 
  \bm y &=& \begin{bmatrix}
  {\bm y}_{\ddot{\bm q}} &
  {\bm y}_{{f}_{{ftShoe,fb}}^x} &
  {\bm y}_{{f}^x}
   \end{bmatrix}^\top \in \mathbb R^{342}~,
\\
\label{case2_allSens}
\mbox{\emph{CASE 2}} \quad \quad 
  \bm y &=& \begin{bmatrix}
  \bm y_{IMUs} &
  {\bm y}_{\ddot{\bm q}} &
  {\bm y}_{{f}_{{ftShoe,fb}}^x} &
  {\bm y}_{{f}^x}
   \end{bmatrix}^\top \in \mathbb R^{390}~.
\end{eqnarray}
\end{subequations}
The MAP algorithm is performed twice by including as set of sensors firstly
 Equation \eqref{case1_noIMUS} and then Equation \eqref{case2_allSens}.
The attempt is to prove that, by adding (e.g., increasing) progressively
 different sensors data, the variance associated to the estimated dynamic
  variables consequently decreases at each MAP computation, by making the
   estimation more reliable.
By passing progressively from \emph{CASE 1} to \emph{CASE 2}
the variance associated to the torques decreases.  Figure
 \ref{incremAnalysis_UW} shows the decreasing behaviour of the $5$-subjects
  mean torque variances from \emph{CASE 1} (in orange) to \emph{CASE 2} (in
   violet) for both left and right ankle, hip joints, respectively and L5S1,
    L4L3, L1T12 joints.  The two plots are for the tasks T1
     (\ref{fig:incremSens_T1}) and T2 (\ref{fig:incremSens_T2}).

The variance values at the ankles do not change significantly among the two
 cases since the ankle torque estimation depends mostly on the readings of the
  ftShoes that are included in both of them.  Conversely, an ever-growing
   decreasing behaviour is present starting from the values associated to the
    hips towards the L1T12 joint on the torso.
This because as we are considering joints always further apart from the feet,
 the ftShoes contribution to the estimation has a lower weight than at the feet
  while the contribution of the IMUs becomes more important (i.e, weights
   more). In Table \ref{incremSensAnalysis} detailed values of the analysis per
    each subject are reported.

\begin{table}[H]
\centering
\caption{Torque variance $\sigma_{tau}$ [\unit{}{\newton\meter}] for both
 \emph{CASE 1} and  \emph{CASE 2}. (Subjects S1, S2, S3, S4, S5; tasks T1, T2).}
\label{incremSensAnalysis}
\centering
\scriptsize
\begin{tabular}{c|c|c|ccccccc}
    \\
    \hline\hline
    \\
{\textbf{Subject}} & {\textbf{Task}} & {\textbf{CASE}} & {\textbf{lAnkle}} &
 {\textbf{rAnkle}} &
                                     {\textbf{lHip}}   & {\textbf{rHip}}   &
                                     {\textbf{L5S1}}   & {\textbf{L4L3}}   &
                                     {\textbf{L1T12}} \\
\\
\hline
\\
\multirow{4}{*}{\textbf{S1}}
& \multirow{2}{*}{\textbf{T1}} & $1$ & $0.0544$ & $0.0560$ & $0.2608$ & $0.3643$ 
                              & $0.5696$ & $0.6938$ & $0.8240$ \\
&  & \cellcolor{Gray}$2$ & \cellcolor{Gray}$0.0544$ & \cellcolor{Gray}$0.0559$ & \cellcolor{Gray}$0.2446$ & \cellcolor{Gray}$0.2568$ 
                              & \cellcolor{Gray}$0.4009$ & \cellcolor{Gray}$0.4572$ & \cellcolor{Gray}$0.5064$ \\ \cline{2-10}
& \multirow{2}{*}{\textbf{T2}} & $1$  & $0.0544$ & $0.0561$ & $0.2613$ & $0.3649$ 
                              & $0.5703$ & $0.6937$ & $0.8244$ \\
&  & \cellcolor{Gray}$2$ & \cellcolor{Gray}$0.0544$ & \cellcolor{Gray}$0.0560$ & \cellcolor{Gray}$0.2449$ & \cellcolor{Gray}$0.2574$ 
                              & \cellcolor{Gray}$0.4015$ & \cellcolor{Gray}$0.4570$ & \cellcolor{Gray}$0.5064$ \\ 
                              \hline 
\multirow{4}{*}{\textbf{S2}}
& \multirow{2}{*}{\textbf{T1}} & $1$ & $0.0543$ & $0.0562$ & $0.2495$ & $0.3289$
                              & $0.5255$ & $0.6079$ & $0.6978$ \\
&  & \cellcolor{Gray}$2$ & \cellcolor{Gray}$0.0543$ & \cellcolor{Gray}$0.0561$ & \cellcolor{Gray}$0.2378$ & \cellcolor{Gray}$0.2483$
                              & \cellcolor{Gray}$0.3924$ & \cellcolor{Gray}$0.4315$ & \cellcolor{Gray}$0.4682$ \\  \cline{2-10}
& \multirow{2}{*}{\textbf{T2}} & $1$ & $0.0544$ & $0.0562$ & $0.2515$ & $0.3244$
                              & $0.5227$ & $0.6072$ & $0.6971$ \\
&  & \cellcolor{Gray}$2$ & \cellcolor{Gray}$0.0544$ & \cellcolor{Gray}$0.0561$ & \cellcolor{Gray}$0.2395$ & \cellcolor{Gray}$0.2455$
                              & \cellcolor{Gray}$0.3911$ & \cellcolor{Gray}$0.4317$ & \cellcolor{Gray}$0.4684$ \\
                               \hline
\multirow{4}{*}{\textbf{S3}}
& \multirow{2}{*}{\textbf{T1}} & $1$ & $0.0544$ & $0.0561$ & $0.2791$ & $0.3979$
                              & $0.6468$ & $0.7696$ & $0.9056$ \\
&  & \cellcolor{Gray}$2$ & \cellcolor{Gray}$0.0544$ & \cellcolor{Gray}$0.0560$ & \cellcolor{Gray}$0.2594$ & \cellcolor{Gray}$0.2716$
                              & \cellcolor{Gray}$0.4361$ & \cellcolor{Gray}$0.4866$ & \cellcolor{Gray}$0.5351$ \\  \cline{2-10}
& \multirow{2}{*}{\textbf{T2}} & $1$ & $0.0544$ & $0.0561$ & $0.2796$ & $0.3984$
                              & $0.6473$ & $0.7697$ & $0.9058$ \\
&  & \cellcolor{Gray}$2$ & \cellcolor{Gray}$0.0544$ & \cellcolor{Gray}$0.0560$ & \cellcolor{Gray}$0.2598$ & \cellcolor{Gray}$0.2722$
                              & \cellcolor{Gray}$0.4366$ & \cellcolor{Gray}$0.4868$ & \cellcolor{Gray}$0.5354$ \\
                               \hline
\multirow{4}{*}{\textbf{S4}}
& \multirow{2}{*}{\textbf{T1}} & $1$ & $0.0544$ & $0.0563$ & $0.3013$ & $0.4558$
                              & $0.7161$ & $0.8658$ & $1.0300$ \\
&  & \cellcolor{Gray}$2$ & \cellcolor{Gray}$0.0544$ & \cellcolor{Gray}$0.0561$ & \cellcolor{Gray}$0.2742$ & \cellcolor{Gray}$0.9172$
                              & \cellcolor{Gray}$0.4567$ & \cellcolor{Gray}$0.5149$ & \cellcolor{Gray}$0.5702$ \\  \cline{2-10}
& \multirow{2}{*}{\textbf{T2}} & $1$ & $0.0544$ & $0.0564$ & $0.3013$ & $0.4559$
                              & $0.7153$ & $0.8642$ & $1.0288$ \\
&  & \cellcolor{Gray}$2$ & \cellcolor{Gray}$0.0544$ & \cellcolor{Gray}$0.0561$ & \cellcolor{Gray}$0.2743$ & \cellcolor{Gray}$0.2924$
                              & \cellcolor{Gray}$0.4567$ & \cellcolor{Gray}$0.5141$ & \cellcolor{Gray}$0.5693$ \\
                               \hline
\multirow{4}{*}{\textbf{S5}}
& \multirow{2}{*}{\textbf{T1}} & $1$ & $0.0544$ & $0.0561$ & $0.2530$ & $0.3401$
                              & $0.5779$ & $0.6491$ & $0.7700$ \\
&  & \cellcolor{Gray}$2$ & \cellcolor{Gray}$0.0544$ & \cellcolor{Gray}$0.0560$ & \cellcolor{Gray}$0.2394$ & \cellcolor{Gray}$0.2487$
                              & \cellcolor{Gray}$0.3913$ & \cellcolor{Gray}$0.4422$ & \cellcolor{Gray}$0.4894$ \\  \cline{2-10}
& \multirow{2}{*}{\textbf{T2}} & $1$ & $0.0544$ & $0.0562$ & $0.2554$ & $0.3439$
                              & $0.5423$ & $0.6495$ & $0.7702$ \\
&  & \cellcolor{Gray}$2$ & \cellcolor{Gray}$0.0544$ & \cellcolor{Gray}$0.0560$ & \cellcolor{Gray}$0.2416$ & \cellcolor{Gray}$0.2518$
                              & \cellcolor{Gray}$0.3950$ & \cellcolor{Gray}$0.4425$ & \cellcolor{Gray}$0.4894$ \\
\\
\hline\hline
\\
\end{tabular}
\end{table}

\vspace{0.8cm}
\section{An Experiment with the iCub}

The analysis, performed in the previous section, is here shown in a variant that
 encompasses the robot iCub.  Data were collected at Istituto Italiano di
  Tecnologia, on a $48$-DoF human model ($\bm d \in \mathbb{R}^{1248}$).
The experimental setup encompassed $i)$ the Xsens suit for the motion tracking,
$ii)$ two standard AMTI OR$6$ force platforms to acquire the ground reaction
 wrenches, $iii)$ the F/T sensors of the robot arms.  ftShoes were not used in
  the experiment since they did not exist at that time.
Kinematic data were acquired at a frequency of \unit{240}{\hertz}.  Each
 platform acquired a sample at a frequency of \unit{1}{\kilo\hertz} by using
  AMTI acquisition units.  

 Experiments were conducted on the iCub \citep{Metta2010}, a full-body humanoid
robot with $53$ DoFs: $6$ in the head, $16$ in each
 arm, $3$ in the torso and $6$ per each leg. The iCub is endowed with whole-body
  distributed F/T sensors, accelerometers, gyroscopes and tactile
   sensors. Specifically, the limbs are equipped with six F/T sensors
    placed in the upper arms, in the upper legs and in the ankles.  Internal
     joint torques and external wrenches are
    estimated  through a whole-body estimation algorithm \citep{Nori2015icub}.
      Robot data were collected at a frequency of \unit{100}{\hertz}.

\begin{figure}[H]
    \centering
  \begin{subfigure}[b]{0.40\textwidth}
    \includegraphics[width=\textwidth]{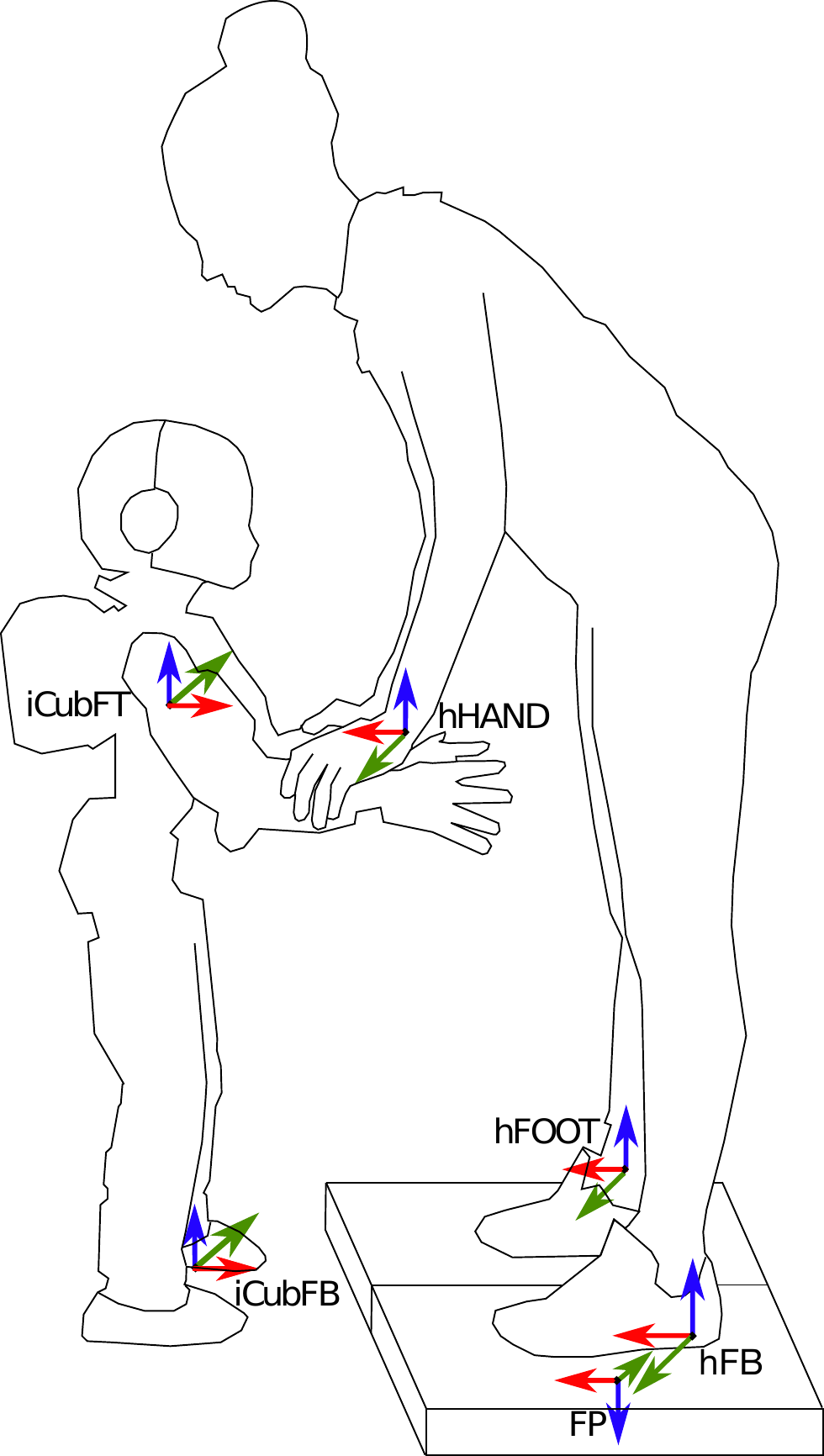}
  \caption{}
\label{fig:interaction_lateral_pHRI}
\end{subfigure}
~\quad
  \begin{subfigure}[b]{0.43\textwidth}
      \centering
    \includegraphics[width=\textwidth]{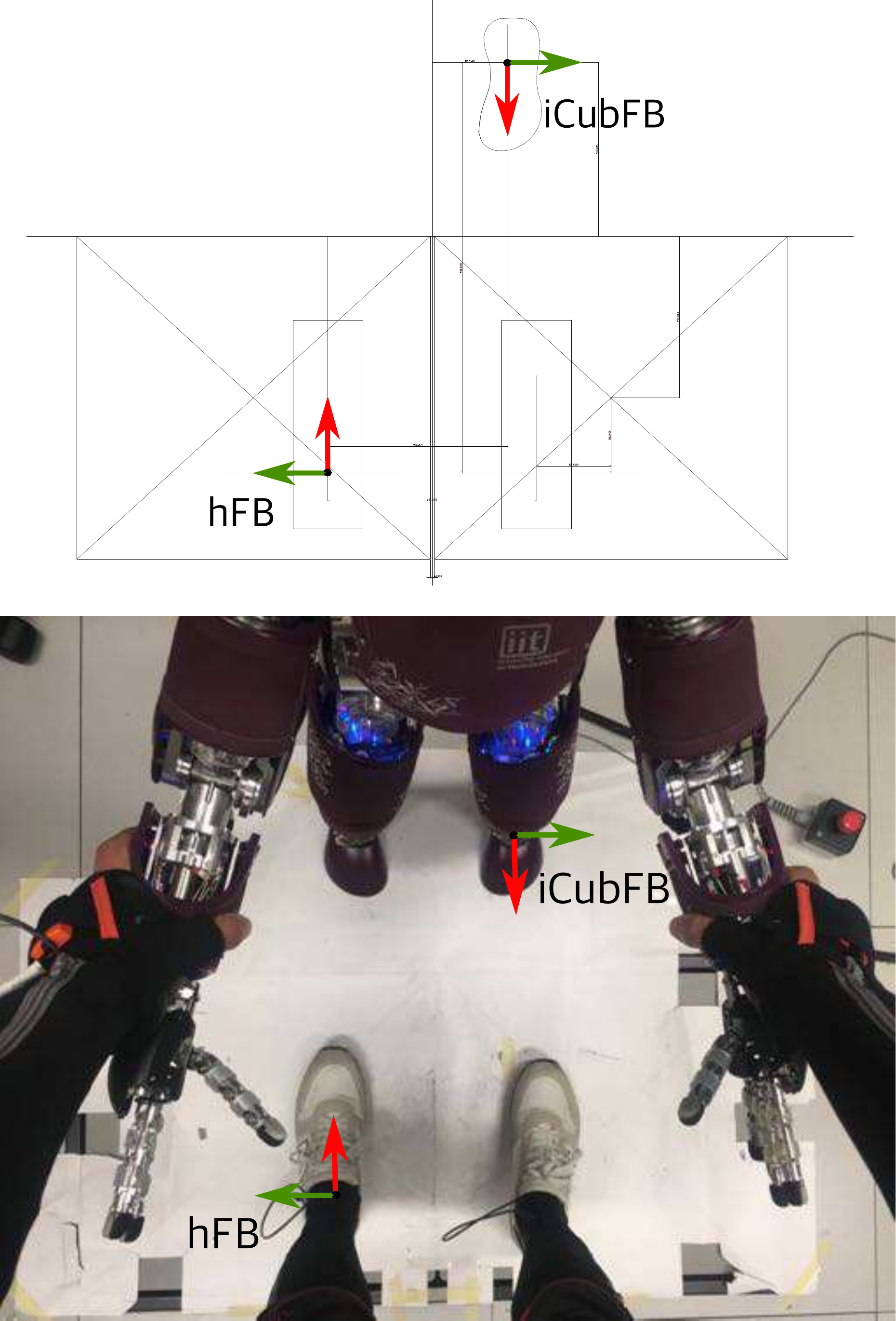}
  \caption{}
\label{fig:interaction_top_pHRI}
\end{subfigure}
\caption{(\subref{fig:interaction_lateral_pHRI}) Human subject
that grasps and pushes down the robot arms.  The figure shows the reference
 frames for the F/T sensor of the robot (iCubFT), the robot fixed base
  (iCubFB), the force plate (FP), the human fixed base (hFB), the human foot
   and hand (hFOOT, hHAND), respectively.  (\subref{fig:interaction_top_pHRI})
    The mutual feet position is defined by a fixture located under both the
     feet.}
\end{figure}

\begin{figure}[H]
  \centering
    \includegraphics[width=1\textwidth]{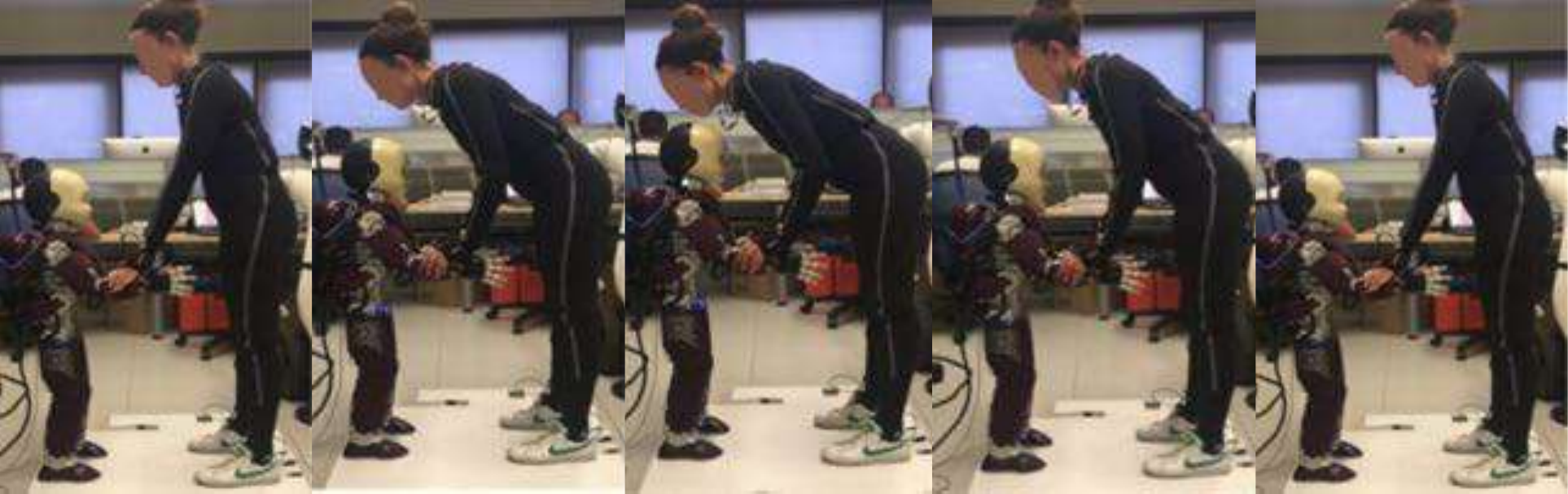}
    \vspace{-0.1cm}
  \caption{Human subject while performing the bowing task with the iCub.}
  \label{fig:Figs_bowingTask_pHRI}
\end{figure}

Ten healthy adult subjects (as in Table \ref{Subjects_for_analysis_PHRI}) have
 been recruited for the experimental session.
Each subject was asked to wear the Xsens suit and to stand on
 the force plates by positioning one foot per platform.  The robot
  was located in front of the subject facing him at a known position w.r.t. the
   human foot location (Figure \ref{fig:interaction_lateral_pHRI}).  The mutual
    feet position was fixed for all the trials and defined by a printed fixture
     located under both feet (Figure \ref{fig:interaction_top_pHRI})  The
      interaction implied
    that the human grasped and pushed down both the robot arms while performing
     a bowing task (Figure \ref{fig:Figs_bowingTask_pHRI}).

\begin{figure}[H]
    \centering
    \vspace{-0.5cm}
    \begin{subfigure}[b]{0.75\textwidth}
        \includegraphics[width=\textwidth]{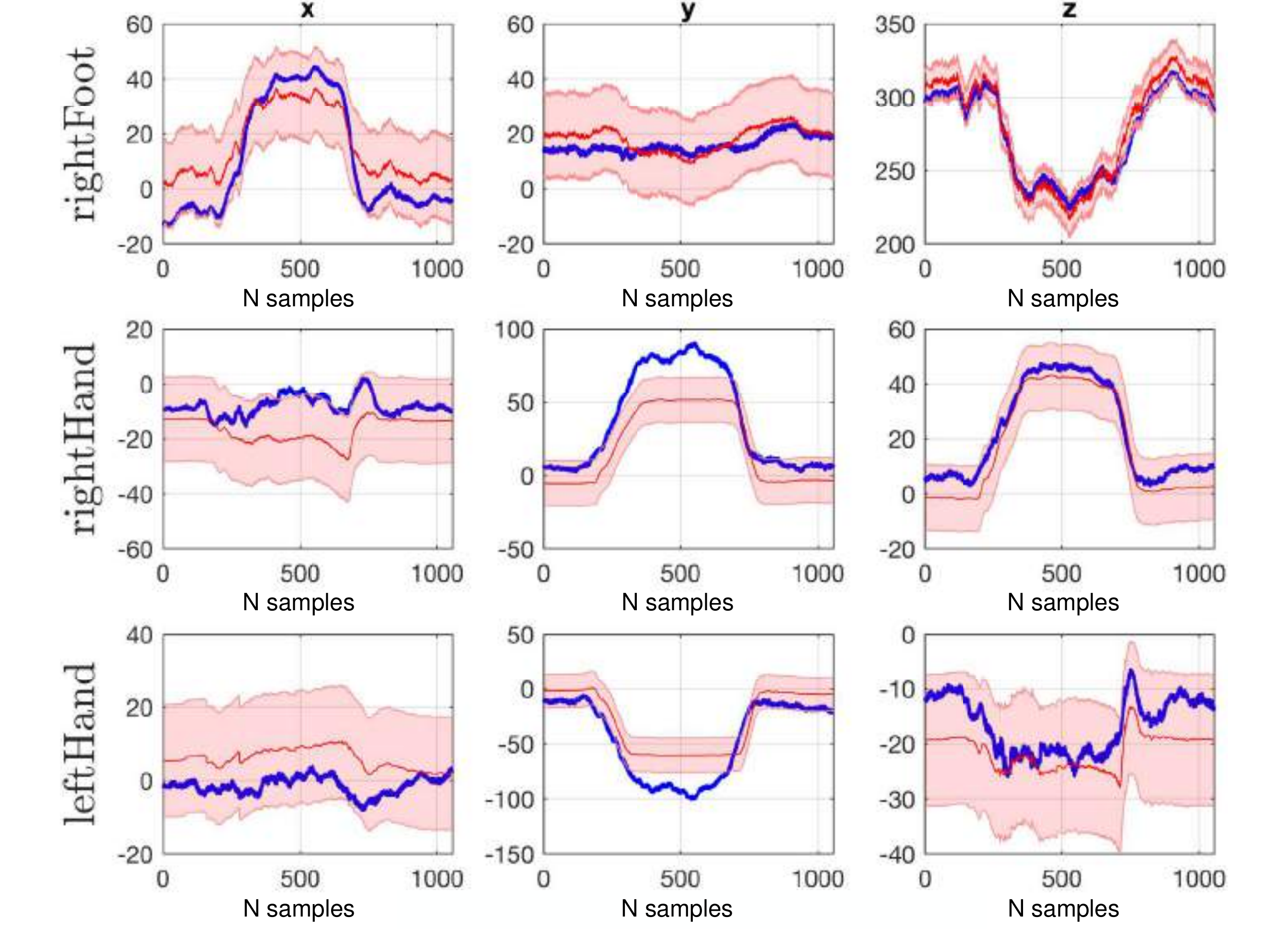}
        \caption{External force [\unit{}{\newton}]}
        \label{fig:fext_comparison}
    \end{subfigure}
    \begin{subfigure}[b]{0.75\textwidth}
                    \vspace{0.7cm}
        \includegraphics[width=\textwidth]{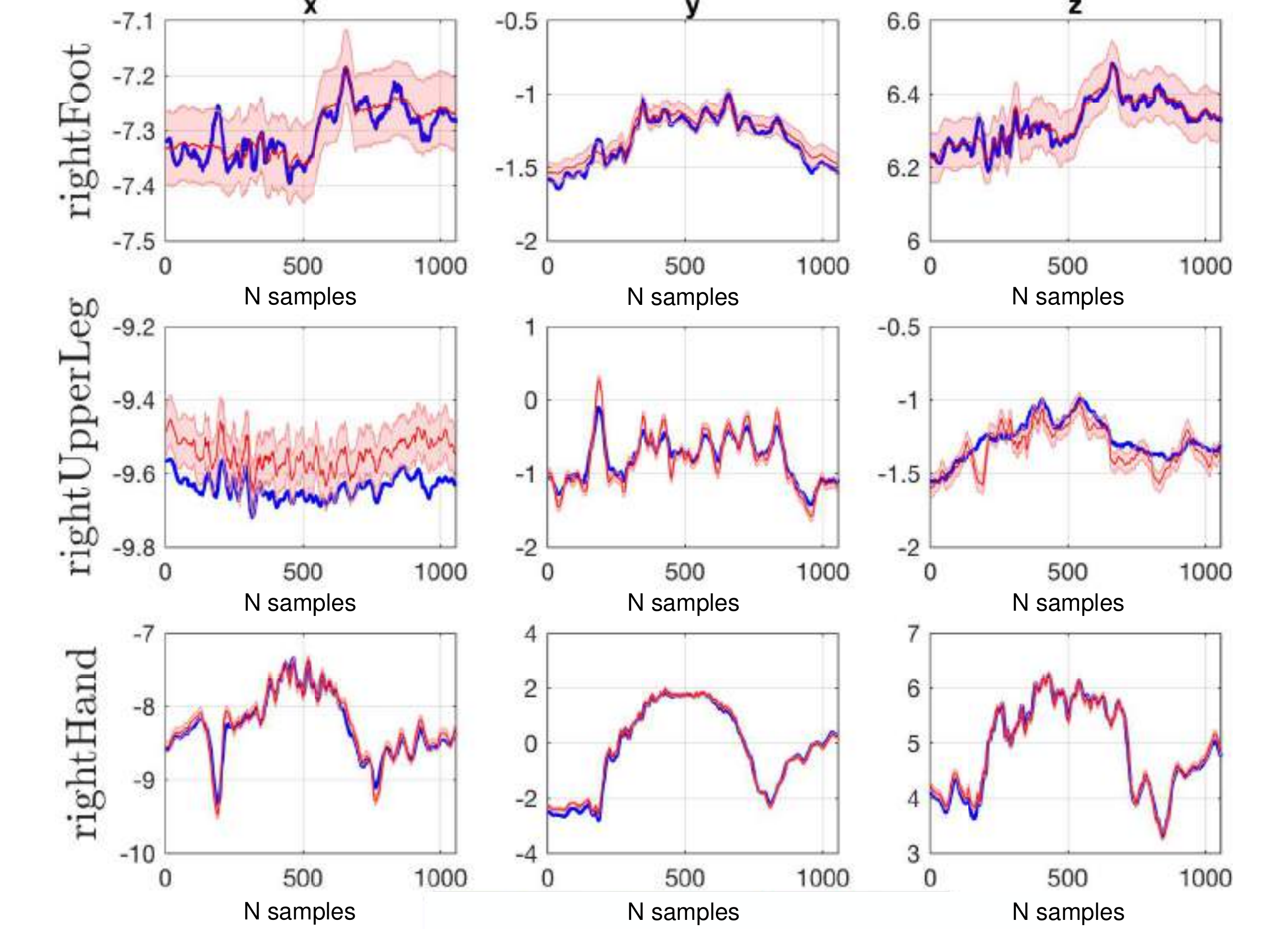}
        \caption{Linear acceleration [\unit{}{\metre\per\second^2}]}
        \label{fig:acc_comparison}
    \end{subfigure}
\caption{Comparison between the variables
 measured (with $2\sigma$ standard deviation, in red) and their MAP estimation
  (in blue) for (\subref{fig:fext_comparison}) the external force  $\bm f^x$
   for the S1 right foot and both the hands links, and
    (\subref{fig:acc_comparison})  the linear acceleration on the S1 right
   foot, upper leg and hand links, respectively.}
\label{fig:pHRI_MAP_accANDddq}
\end{figure}

\begin{figure}[H]
    \centering
        \vspace{-0.5cm}
    \begin{subfigure}[b]{0.75\textwidth}
        \includegraphics[width=\textwidth]{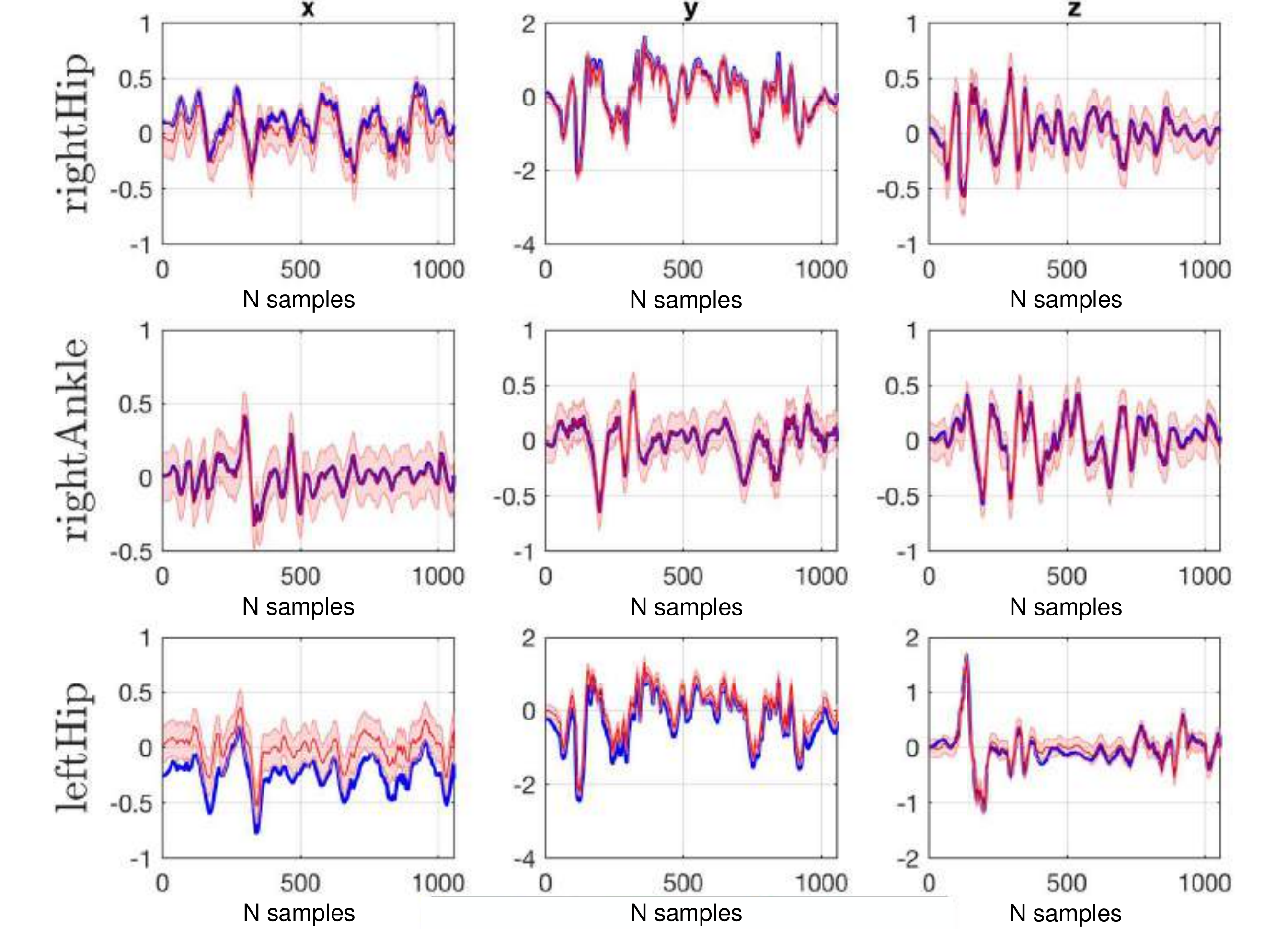}
        \caption{Joint acceleration [\unit{}{\rad\per\second^2}]}
        \label{fig:ddq_comparison}
    \end{subfigure}
    \begin{subfigure}[b]{0.75\textwidth}
            \vspace{0.7cm}
        \includegraphics[width=\textwidth]{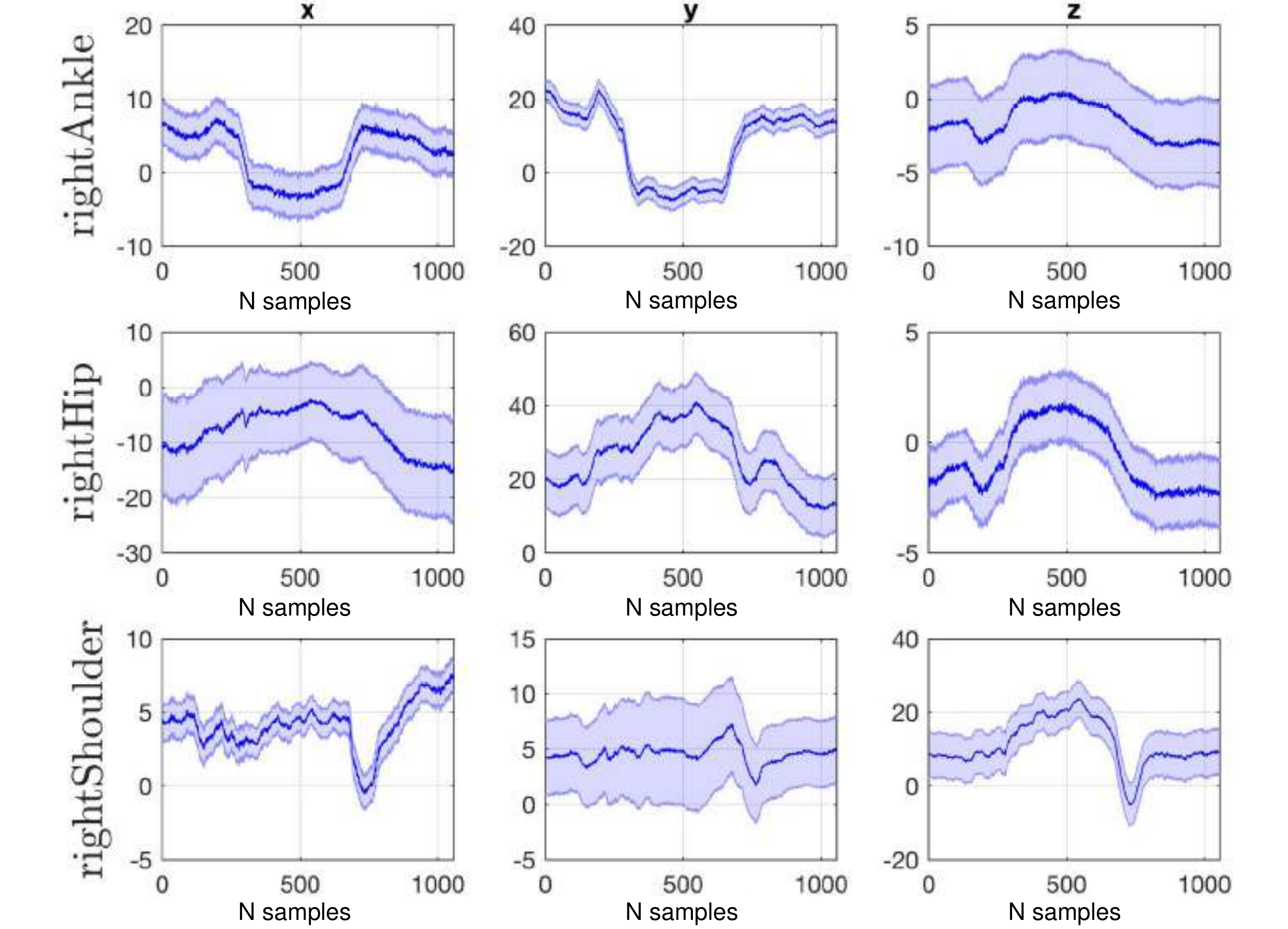}
        \caption{Joint torque [\unit{}{\newton\meter}]}
        \label{fig:tau_estimation}
    \end{subfigure}
\caption{(\subref{fig:ddq_comparison}) Joint acceleration comparison between
 the variables measured (with $2\sigma$ standard deviation, in red) and their
  MAP estimation (in blue) for the S1 hips and right ankle joints.
   (\subref{fig:tau_estimation}) Joint torque MAP
    estimation (with $2\sigma$ estimated standard deviation) for the S1 right
     ankle, hip and shoulder joints, respectively.}
\label{fig:pHRI_MAP_fextANDtau}
\end{figure}

As in Section \ref{MAPanalysis_UW}, even here the MAP algorithm was
 able to estimate $\bm d$ for all the links/joints in the model.  As in the
  previous dataset, the MAP results are here shown for one subject (i.e., S1).
Figure \ref{fig:fext_comparison} shows the comparison for the external
 forces measured (in red) and estimated (in blue) for the right
  foot\footnote{No left foot as previously motivated.} and for both the human
   hands since during the interaction there are forces acting on them.
The same comparison concerning the linear acceleration for the right foot, upper leg
 and hand, respectively, is shown in Figure \ref{fig:acc_comparison}.
Similar analysis in Figure \ref{fig:ddq_comparison} for the joint
 acceleration, for the right ankle and the hips.
Again here, like in the previous case, the MAP algorithm is able to give us a
 suitable estimation of the human joint torques, Figure
  \ref{fig:tau_estimation}.

\begin{table}[H]
\centering
\caption{Subjects eligible for the pHRI analysis.  Each subject was provided of
 a written informative consent before starting the experiment.}
\label{Subjects_for_analysis_PHRI}
\centering
\scriptsize
\vspace{-0.1cm}
\begin{tabular}{c|cccc}
\\
\hline\hline
\\
\textbf{Subject} & \textbf{Gender} & \textbf{Age}
                 & \textbf{Height} [\unit{}{\centi\meter}]
                 & \textbf{Mass} [\unit{}{\kilo\gram}]\\
\\
\hline
\\
\textbf{S1}      & F    & $30$  & $165$  & $60.8$    \\
\rowcolor{Gray}
\textbf{S2}      & F    & $32$  & $166$  & $67.3$    \\
\textbf{S3}      & M    & $29$  & $161$  & $54.4$    \\
\rowcolor{Gray}
\textbf{S4}      & M    & $28$  & $172$  & $64.1$    \\
\textbf{S5}      & F    & $25$  & $171$  & $58.7$    \\
\rowcolor{Gray}
\textbf{S6}      & F    & $27$  & $169$  & $65.4$    \\
\textbf{S7}      & F    & $24$  & $170$  & $68.2$    \\
\rowcolor{Gray}
\textbf{S8}      & F    & $24$  & $165$  & $52.8$    \\
\textbf{S9}      & M    & $40$  & $169$  & $65.4$    \\
\rowcolor{Gray}
\textbf{S10}     & M    & $29$  & $158$  & $54.4$    \\
\\
\hline\hline
\\
\end{tabular}
\end{table}

\vspace{1cm}

\newpage
\begin{figure}[H]
  \centering
    \includegraphics[width=.29\textwidth]{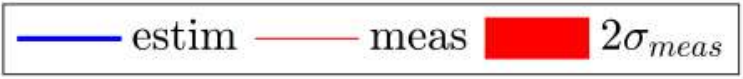}
\end{figure}

\newpage
\begin{figure}[H]
  \centering
    \includegraphics[width=.41\textwidth]{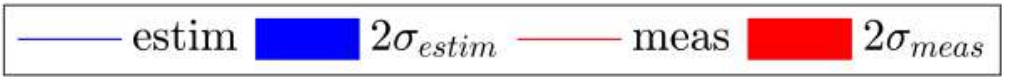}
\end{figure}

\newpage
As in the Section \ref{incrementalSensorAnalysis_UW} analysis, an investigation
 by adding progressively different sensors has been done on the MAP computation.
The set of measurements equation \eqref{eq:measEquation} is here composed as
 follows:
\vspace{0.25cm}
\begin{subequations}\label{eq:casesForIncrementalAnalisys_pHRI}
\begin{eqnarray}
\label{case1_onlyFP_pHRI}
 \mbox{\emph{CASE 1}} \quad \quad 
   \bm y &=& \begin{bmatrix}
   {\bm y}_{\ddot{\bm q}} &
  {\bm y}_{{f}_{{FP,fb}}^x} &
   {\bm y}_{{f}^x}
    \end{bmatrix}^\top \in \mathbb R^{330}~,
\\
\label{case2_noRobot_pHRI}
\mbox{\emph{CASE 2}} \quad \quad 
   \bm y &=& \begin{bmatrix}
   \bm{y}_{IMUs} &
   {\bm y}_{\ddot{\bm q}} &
  {\bm y}_{{f}_{{FP,fb}}^x} &
   {\bm y}_{{f}^x}
    \end{bmatrix}^\top \in \mathbb R^{378}~,
 \\
\label{case3_allSens_pHRI}
\mbox{\emph{CASE 3}} \quad \quad 
   \bm y &=& \begin{bmatrix}
   \bm{y}_{IMUs} &
   {\bm y}_{\ddot{\bm q}} &
  {\bm y}_{{f}_{{FP,fb}}^x} &
   {\bm y}_{{f}^x_{iCubF/T}}
    \end{bmatrix}^\top \in \mathbb R^{390}~.
\end{eqnarray}
\end{subequations}

\begin{figure}[H]
  \centering
    \includegraphics[width=.55\textwidth]{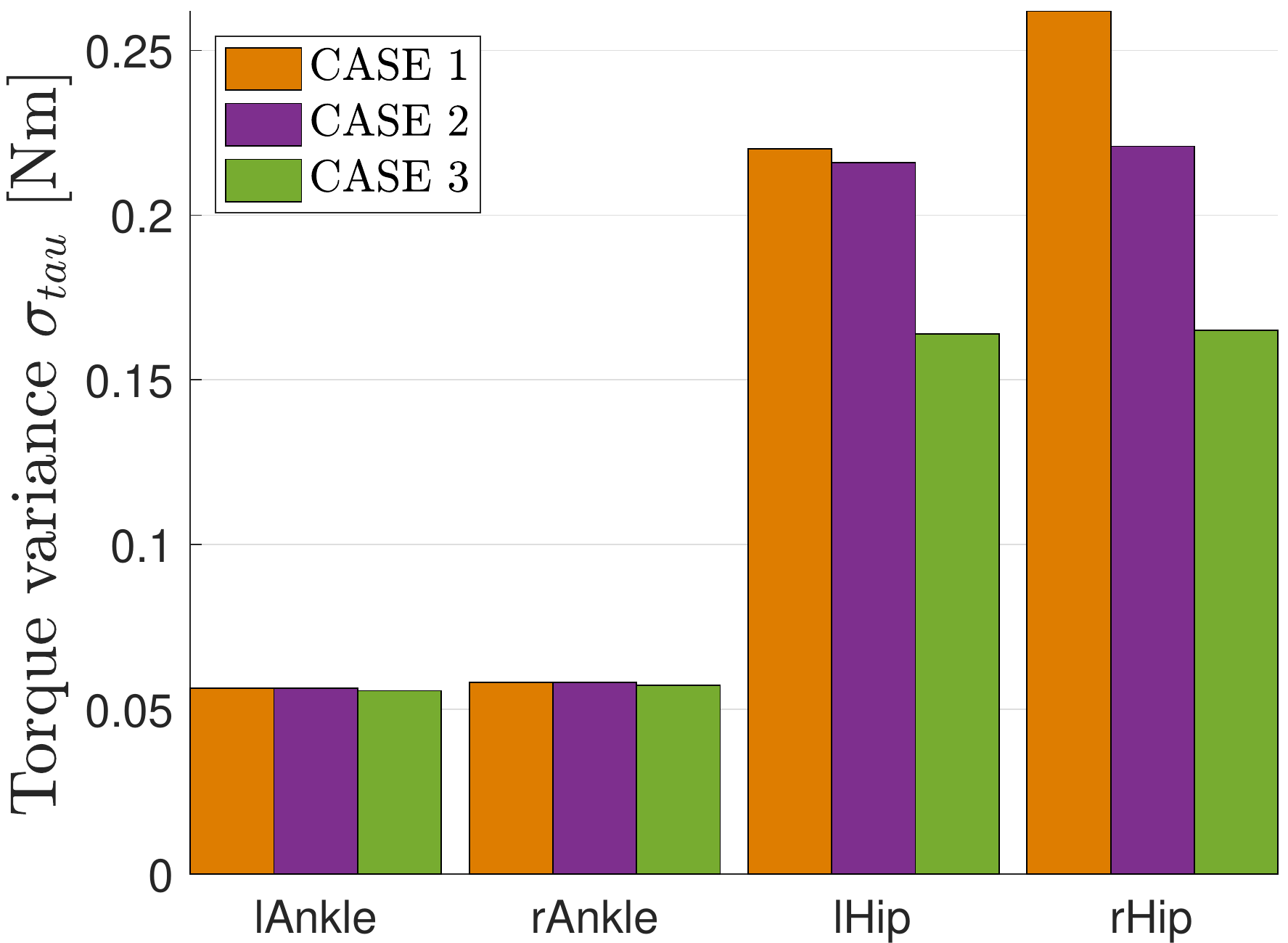}
\caption{$10$-subjects mean torque variance [\unit{}{\newton\meter}] by using
 three different combinations of the measurements equation: \emph{CASE
  1} (in orange), \emph{CASE 2} (in violet) and  \emph{CASE 3} (in green), for
   both left and right ankle, hip joints, respectively, for the pHRI bowing
    task.} 
  \label{fig:Figs_plots_meanTorqueVar_pHRI}
\end{figure}

\vspace{0.25cm}
Even in this case the analysis reveals the important decreasing behaviour as in
Figure \ref{incremAnalysis_UW}.  By passing progressively from \emph{CASE 1} to
 \emph{CASE 3} the variance associated to the $10$-subjects mean torques
  decreases, Figure \ref{fig:Figs_plots_meanTorqueVar_pHRI}.
Again, the variance values on the ankles do not change significantly among the
 three configurations since the ankles torque estimation depends
mostly on the contribution of the force plates that are included in all the
three cases.
And for the same previous reason, the decreasing behaviour becomes to be
 evident at the hips since the torque estimation at the hips is affected
   by the complete set of sensors (i.e., contribution of the force plates
    weights lower, contribution of IMUs + F/T iCub sensors weights more).

\vspace{0.5cm}

\newpage
\subsection{MAP vs. OpenSim Dynamics Estimation}

To test the quality of our estimation w.r.t. a ground-truth tool, 
one subject of Table \ref{Subjects_for_analysis_PHRI} 
 was asked to perform four repetitions of the bowing task in two different
  configurations, i.e., \emph{with} and \emph{without} an additional mass $W$
   of \unit{6}{\kilo}{\gram} roughly positioned in correspondence of the torso
    center of mass. The MAP computation was performed by considering the
     following cases as algorithm inputs (see Table \ref{table:robustness}).

\vspace{0.4cm}
\begin{table}[H]
\caption{Cases for the MAP vs. OpenSim evaluation.}
\label{table:robustness}
\centering
\scriptsize
\begin{tabular}{c|cc}
\\
\hline \hline
\\
 & \textbf{\emph{CASE A}} & \textbf{\emph{CASE B}} \\
\\
\hline
\\
\textbf{URDF model}             & without $W$ & without $W$\\
\rowcolor{Gray}
\textbf{measurements}           & with $W$    & without $W$\\
\textbf{$\bm{\Sigma}_D$}        & $10^{-1}$   & $10^{-4}$  \\
\rowcolor{Gray}
\textbf{$\tau$ estimation}      & $\bm \tau_{(model + \unit{6}{\kilo}{\gram})}$
                                & $\bm \tau_{model}$  \\
\\
\hline\hline
\\
\end{tabular}
\end{table}
In both the cases the analysis is performed with the same URDF model of the
 subject.  In order to highlight a lower reliability for the 
\emph{CASE A}, it is assigned a value to $\bm{\Sigma}_D$ equal to $10^{-1}$
 (different from the value $10^{-4}$ assigned for the \emph{CASE B}).

\begin{figure}[H]
          \centering
          \begin{subfigure}[b]{0.5\textwidth}
              \includegraphics[width=\textwidth]
              {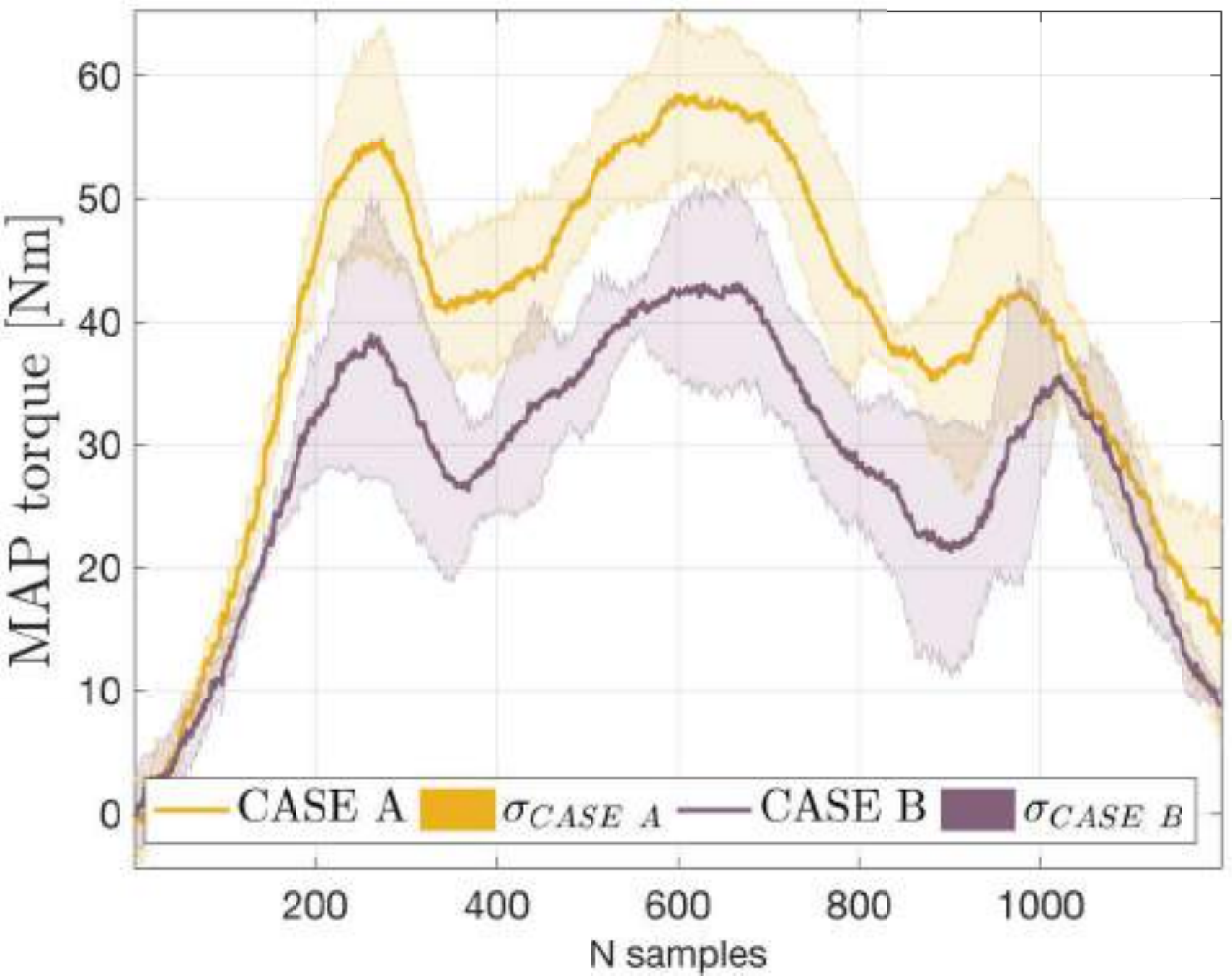}
              \caption{}
              \label{fig:TorqueComparisonMAP}
          \end{subfigure}
          \begin{subfigure}[b]{0.5\textwidth}
              \vspace{0.3cm}
              \includegraphics[width=\textwidth]
              {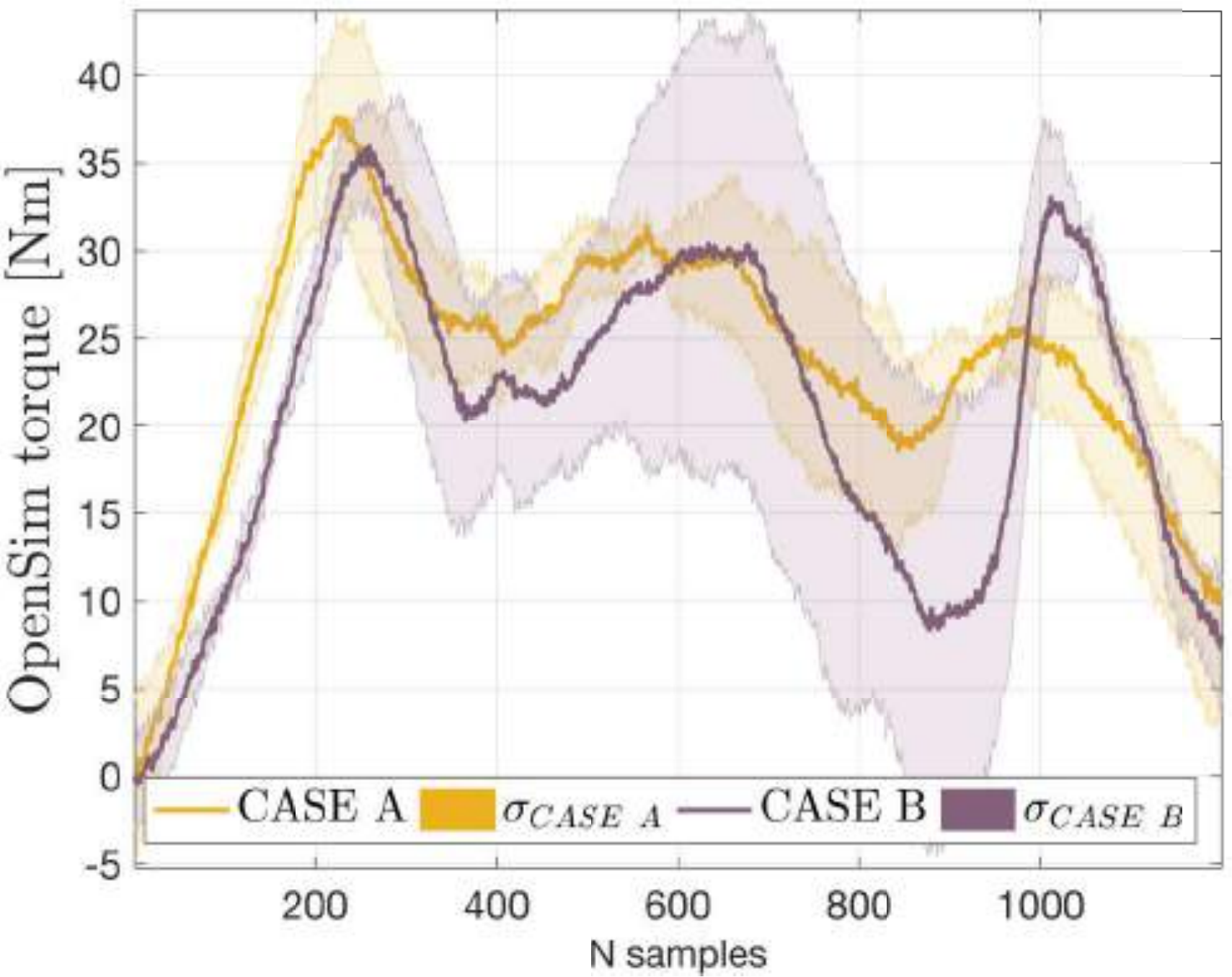}
              \caption{}
              \label{fig:TorqueComparisonOPENSIM}
          \end{subfigure}
\caption{Mean and $\sigma$ standard deviation of the hips torque estimation
among four repetitions of the bowing task performed by a subject with
 (\subref{fig:TorqueComparisonMAP}) the MAP algorithm
 and (\subref{fig:TorqueComparisonOPENSIM}) the OpenSim ID toolbox.
 The analysis has been done for the case with the weight $W$ (\emph{CASE A})
  and the case without $W$ (\emph{CASE B}).}
\label{MAPvsOPENSIM}
 \end{figure}

By exploiting the linearity property of the system we consider 
the following expression for the torques:
\begin{eqnarray} \label{tauEq}
	\bm \tau_{(model + \unit{6}{\kilo}{\gram})} - \bm \tau_{model} =
	\bm \tau_{\unit{6}{\kilo}{\gram}}~,
\end{eqnarray}
where $\bm \tau_{\unit{6}{\kilo}{\gram}}$ is the theoretical torque due to the
 additional $W$ positioned on the torso\footnote{We consider a simple $2$-DoF
  system (see \citep{LatellaSensors2016}) in which the position of $W$ and the
   hip joint angle are known.}.
Figure \ref{MAPvsOPENSIM} shows the mean and the $\sigma$ standard deviation
 of the torque estimation at the hips (i.e., the sum of the torque estimated at
  each hip) by means of the MAP algorithm
(Figure \ref{fig:TorqueComparisonMAP}) and the OpenSim ID toolbox
(Figure \ref{fig:TorqueComparisonOPENSIM}), respectively. 

Given Equation \eqref{tauEq}, it is possible to
    retrieve the error ${\varepsilon}_{\bm{\tau}}$ on the $\bm \tau$
     estimation due to the presence of the weight $W$:
\begin{eqnarray} \label{TorquesError}
{\varepsilon}_{\bm{\tau}} = |\bm \tau_{(model + \unit{6}{\kilo}{\gram})}
     -\bm \tau_{model}| - \bm \tau_{\unit{6}{\kilo}{\gram}}~.
\end{eqnarray}
Equation \eqref{TorquesError} has been computed for both the estimators in
 order to evaluate its effectiveness w.r.t the modelling errors.  The error is
  higher in OpenSim ID computation (6.45 \unit{}{\newton\meter}) than in the
   MAP estimation (3.69 \unit{}{\newton\meter}) since
   OpenSim does not offer the possibility of setting the model reliability in
    the computation.  This highlights that the MAP algorithm is a method more
     robust to
     the modelling errors since it gives the possibility of weighting the
     reliability of the model by properly tuning the related covariance matrix.

\vspace{0.7cm}

\clearpage\null

\chapter{Towards the Real-Time Human Dynamics Estimation }  %
\label{chapter_towardsTheRealTime}

\begin{quotation}
\noindent\emph{
The Scientist must set in order. Science is built up with facts, as a house is
 with stones. But a collection of facts is no more a science than a heap of
  stones is a house.
}
    \begin{flushright}
        Henri Poincaré
    \end{flushright}
\end{quotation}

\vspace{1cm}
\noindent
In this Chapter, a first attempt to aim at advancing the current state of the
 art in pHRI is presented through the design of an estimation tool for
  monitoring the dynamics of the human in a real-time domain.
From a theoretical point of view, the online design does not differ from the
 version implemented in the previous chapter.  However the logic in which input
  data have to be acquired and the way in which they have to be processed are
   completely different in the real-time context.
Here the idea is to exploit a middleware already developed for the robot iCub
 to perform the online human dynamics estimation.  To this scope the 
  software implementation in Figure
   \ref{fig:Figs_schemeAlgorithm_generic_pipeline} was revisited and
    modified accordingly.

All the software related to this Chapter has been released as an open
 source-code and it is hosted on GitHub in the Robotology organization
  \citep{HDErepo}.

\section{YARP Middleware for the Human Framework} \label{YARP_description}

\cite{yarp_site} (Yet Another Robot Platform) is a middleware developed at
 Istituto Italiano di Tecnologia \citep{yarp2006}. It is a C++ platform
  embedding libraries, protocols and tools, written for interfacing 
   humanoid robots.
YARP is mainly used to minimize the effort in the infrastructure-level software
 development by facilitating the code reusability and modularity. 

YARP allows to split the algorithm for the estimation of
 the human dynamics into different modules by ensuring $i)$ communication and
  $ii)$ connection between them but preserving, in the meanwhile, their
 independence \citep{Romano2017codyco}.  The algorithm, represented in
  Figure \ref{fig:Figs_schemeAlgorithm_generic_pipeline} for offline
   evaluations, is hereafter structured into modules (in Figure
    \ref{fig:Figs_realtimeScheme_allSens}) in order to fit the online
     requirements.

\subsubsection{The \emph{human-state-provider} Module}

As suggested by the name, the module has to provide in real-time the human
 state $(\bm{q}, \dot{\bm{q}})$ .  A pre-built URDF model of the human and
  the Xsens motion capture data (coming here from a YARP driver) are the inputs
   of the module.
The information coming from the motion capture system has to be converted in a
 representation compatible with our human dynamic model represented in Equation
  \eqref{eq:systemEq}, as follows.

Let $i$ and $k$ be  two links coupled by a joint.  If ${}^\mathcal{I} \bm
  {T}_i$, ${}^\mathcal{I} \bm {T}_k$ $\in SE(3)$ are the pose of the links
   w.r.t. a generic inertial frame $\mathcal{I}$ in the original human model
    (i.e., the Xsens model) and ${}^\mathcal{I} {\widehat{\bm T}}_i(\bm q)$,
     ${}^\mathcal{I} {\widehat{\bm T}}_k(\bm q)$ $\in SE(3)$ the pose of the
      same quantities in our dynamic model, we can use IK to map the links 
       pose to the configuration $\bm{q}$.
The relative pose is 
${}^i {\bm T}_k = {}^i{\bm T}_{\mathcal{I}}~{}^\mathcal{I} \bm {T}_k$ for the 
 original model,
${}^i {\widehat{\bm T}}_k(\bm q) = {}^i{\widehat{\bm T}}_{\mathcal{I}}(\bm
 q)~{}^\mathcal{I} {\widehat{\bm T}}_k(\bm q)$ for our model. The module solves
  a nonlinear optimization problem, such as
 \begin{equation}\label{optPB_q}
     \min_{\bm q}~ error \big({}^i {\bm T}_k, {}^i {\widehat{\bm T}}_k(\bm q)
      \big)~, \quad {\bm q}_{min} \leq {\bm q} \leq {\bm q}_{max}~.
 \end{equation}
The solution of the problem \eqref{optPB_q} is that value of ${\bm q}$ that
 minimizes the error function $error : SE(3) \rightarrow
  \mathbb{R}$ in the joint limits range $[{\bm q}_{min}, {\bm q}_{max}]$.

To compute the velocity $\dot{\bm{q}}$, the module reads from the Xsens the
 angular velocities $\bm \omega_i$, $\bm \omega_k$ and computes the relative
  angular velocity of the two links ${}^i \bm \omega_k$.  Our model velocity
is then computed by inverting the following relation:
\begin{equation} \label{jacobian_vel_humanModel}
    ^{i}{\widehat{\bm \omega}_k} = ^{i}{\bm {\mathrm{J}}}_k (\bm
     {{q}})\dot{\bm{q}}~,
\end{equation}
where $^{i}{\bm {\mathrm{J}}}_k$ is the relative Jacobian of the link $k$
 w.r.t. the link $i$ and $^{i}{\widehat{\bm \omega}_k}$ is the angular velocity
  of the link $k$ w.r.t. the link $i$ of our dynamic model.  Since in general
   $^{i}{\widehat{\bm \omega}_k} \neq {}^i \bm \omega_k$, Equation
    \eqref{jacobian_vel_humanModel} is solved in the least-squares sense, such
     that
\begin{equation} \label{LS_jointVel}
\dot{\bm{q}}^\ast = \arg\min_{\dot{\bm{q}}}~\big\|^{i}{\widehat{\bm \omega}_k} -
 ^{i}{\bm {\mathrm{J}}}_k (\bm {{q}})\dot{\bm{q}}\big\|^{2}~.
\end{equation}

For each pair of coupled links $(i,k)$, the \emph{human-state-provider} is in
 charge of computing the mapping procedure
$({}^i {\bm T}_k, {}^i \bm \omega_k) \rightarrow ({\bm{q}},\dot{\bm{q}}^\ast)$.
This formulation is quite generic and allows to handle models of different
 complexity.

\subsubsection{The \emph{human-forces-provider} Module}
 The module is composed of two interfaces.  The first interface is in charge of
  reading forces coming from different YARP-based devices (force plates or
   ftShoes) and from YARP ports (the iCub, in case of pHRI).
The second interface transforms force readings into $6$D force vectors
 expressed in human reference frames, as required from the vector of
  measurements $\bm y$.  The proper force transformation requires that the pose
   of the human w.r.t. the force plates and the robot is a known quantity.

\vspace{-0.2cm}
\subsubsection{The \emph{human-dynamics-estimator} Module}
 The module provides the final estimation for the vector $\bm d$ given
  as inputs the previous two modules output data.  In order to proper
   cluster data into $\bm y$ vector, it needs the human state and the forces
    readings both expressed in humans
frames. Then, together with the human state $(\bm{q}, \dot{\bm{q}})$ and some
 information extracted from the URDF model, the module launches the MAP
  algorithm with the Cholesky factorization. 

YARP allows to visualize the real-time human dynamics estimation (whether a pHRI
 is occurring or not) by means of some structured modules and a ROS-based
  visualizer (gray part in the Figure \ref{fig:Figs_realtimeScheme_allSens}).
The \emph{human-viz} module in composed of different YARP submodules that
 are in charge of reading information from the Xsens system, the human state
  $(\bm{q}, \dot{\bm{q}})$, the estimated vector $\bm d$, and of publishing
  information to be sent to the visualization tool in the form of ROS-topic
   messages.
ROS messages (including those coming from the \emph{human-forces-provider}
 module) are visualized in the toolkit RViz \citep{rviz2015}.  The advantage in
  adopting this tool is due to its versatility and agnosticism to the type of
   the data structure or algorithm.

At this preliminary stage, we observed that \unit{100}{\hertz} is a sufficient
 frequency for the outputs of the \emph{human-forces-provider} and the
  \emph{human-dynamics-estimator} modules; \unit{20}{\hertz} for obtaining a
   suitable output of the \emph{human-state-provider} module.  The last
    frequency is strongly dependent on the type of optimization solver used 
    in computing the IK.

\vspace{-0.4cm}
\section{Towards the Online Estimation}

\begin{figure}[H]
  \centering
    \includegraphics[width=\textwidth]{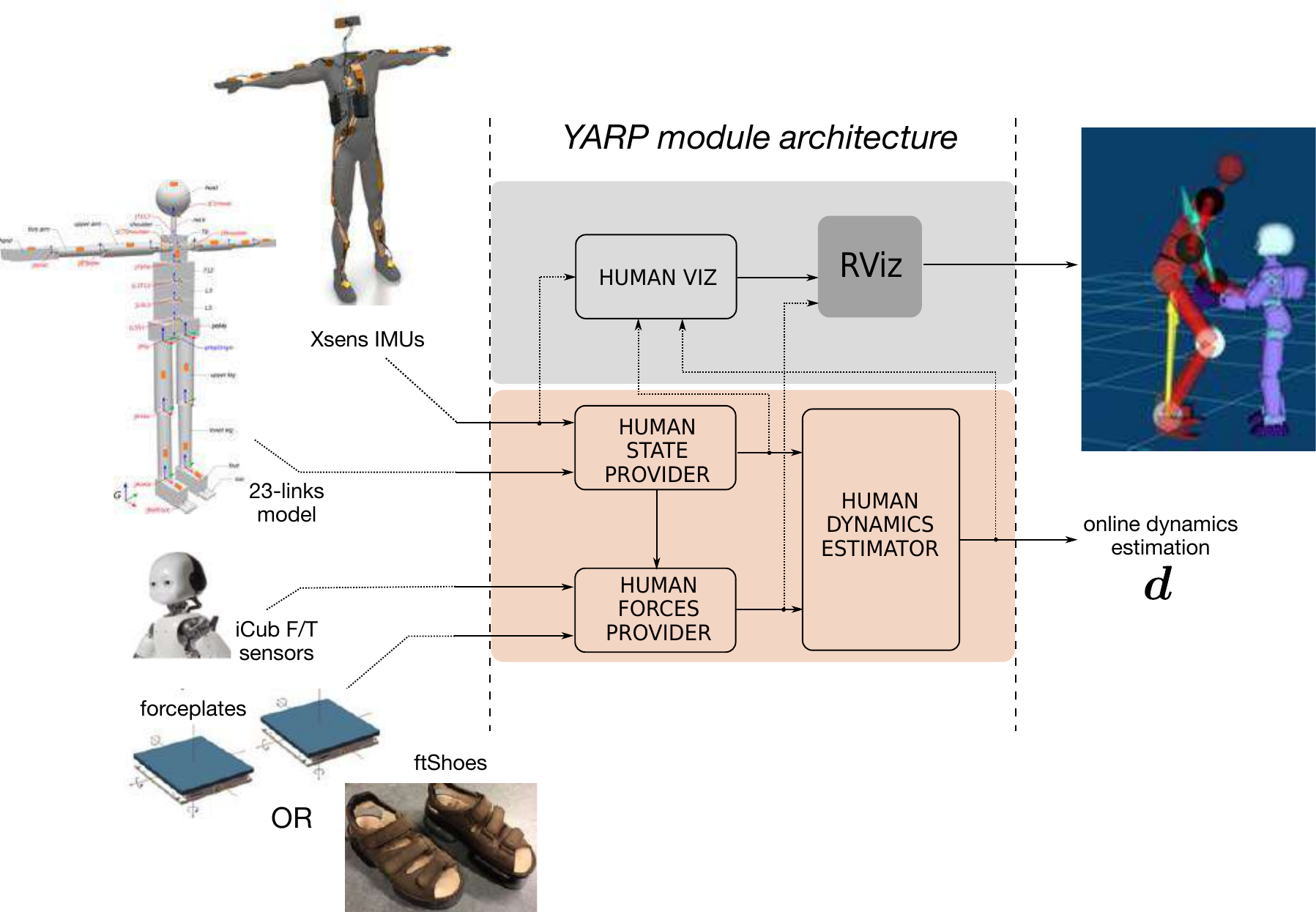}
  \caption{The YARP architecture for estimating and visualizing real-domain
   human dynamics estimation.  The software architecture is able to
     estimate the dynamic variables $\bm d$ for each link/joint in
      the model (pink area) and to visualize the information about the
       kinematics and the dynamics of the human via RViz (gray area)}
  \label{fig:Figs_realtimeScheme_allSens}
\end{figure}

\begin{figure}[H]
    \centering
        \vspace{-0.5cm}
    \begin{subfigure}[b]{0.7\textwidth}
        \includegraphics[width=\textwidth]
        {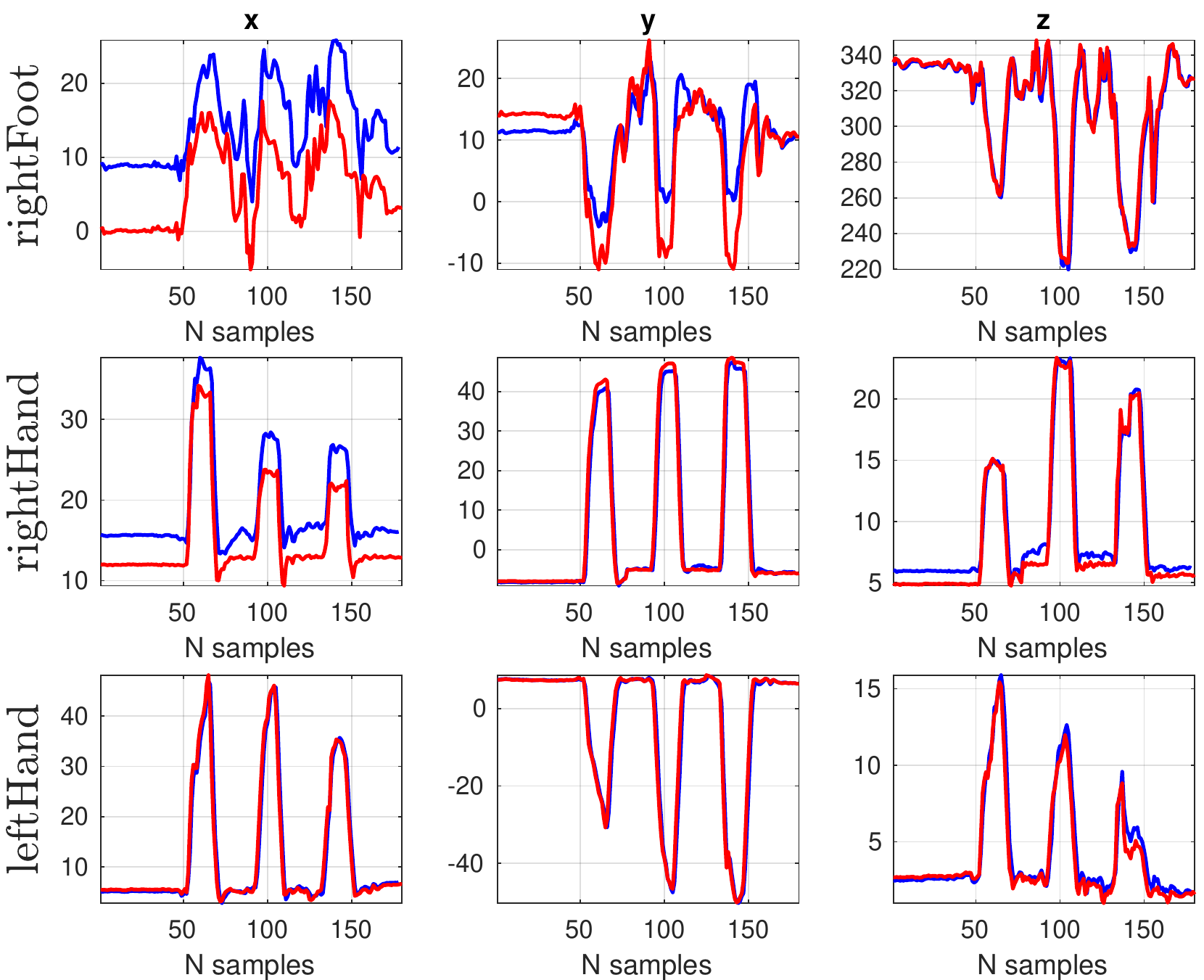}
        \caption{pHRI bowing}
        \label{forcesComparisonFromYarpHDE_bowing}
    \end{subfigure}
    \begin{subfigure}[b]{0.7\textwidth}
            \vspace{0.7cm}
        \includegraphics[width=\textwidth]
        {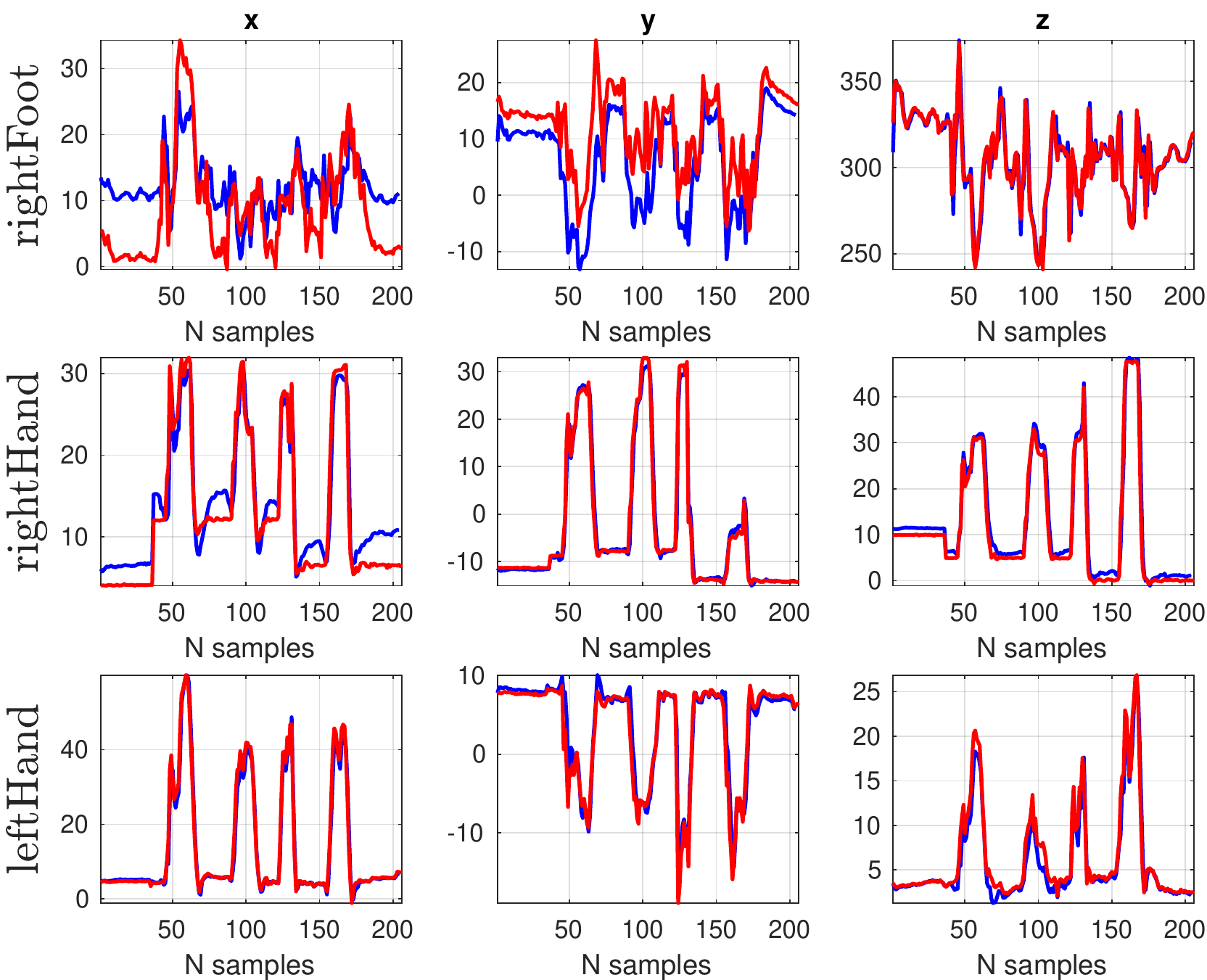}
        \caption{pHRI squat }
        \label{forcesComparisonFromYarpHDE_squat}
    \end{subfigure}
\caption{Comparison between external forces [\unit{}{\newton}]  transformed in
 human frames by the \emph{human-forces-provider} module (in red) and the same
  quantities as estimated from the \emph{human-dynamics-estimator} module (in
   blue), for the pHRI (\subref{forcesComparisonFromYarpHDE_bowing}) bowing and
   (\subref{forcesComparisonFromYarpHDE_squat}) squat tasks.}
  \label{fig:Figs_YARP_HDE_analysis}
\end{figure}

A preliminary investigation on the online algorithm estimation has been done.
  Similarly to the experimental setup in Figure
   \ref{fig:Figs_bowingTask_pHRI}, a subject was asked to wear the suit and the
    ftShoes and to interact with the iCub through a pushing bowing.  The
     estimation vector $\bm d$ was here online computed by the YARP
      infrastructure in Figure \ref{fig:Figs_realtimeScheme_allSens}.
Figure \ref{forcesComparisonFromYarpHDE_bowing} shows very
 preliminary results of three consecutive pHRI bowings.  It shows the
  comparison between the external forces transformed in human frames by the
   \emph{human-forces-provider} module (in red) and the same forces as
    estimated in $\bm d$ by the \emph{human-dynamics-estimator} module (in
     blue).
The same analysis has been done for four repetitions of pHRI squat task, Figure
 \ref{forcesComparisonFromYarpHDE_squat}.
It is important remarking that these preliminary results are relying only on
 the forces measurements (i.e., no IMUs or joint
  accelerations $\ddot {\bm q}$ on vector $\bm y$).  Although they are very
   promising, further investigations are required in the very next future.

\begin{figure}[H]
  \centering
    \includegraphics[width=.6\textwidth]{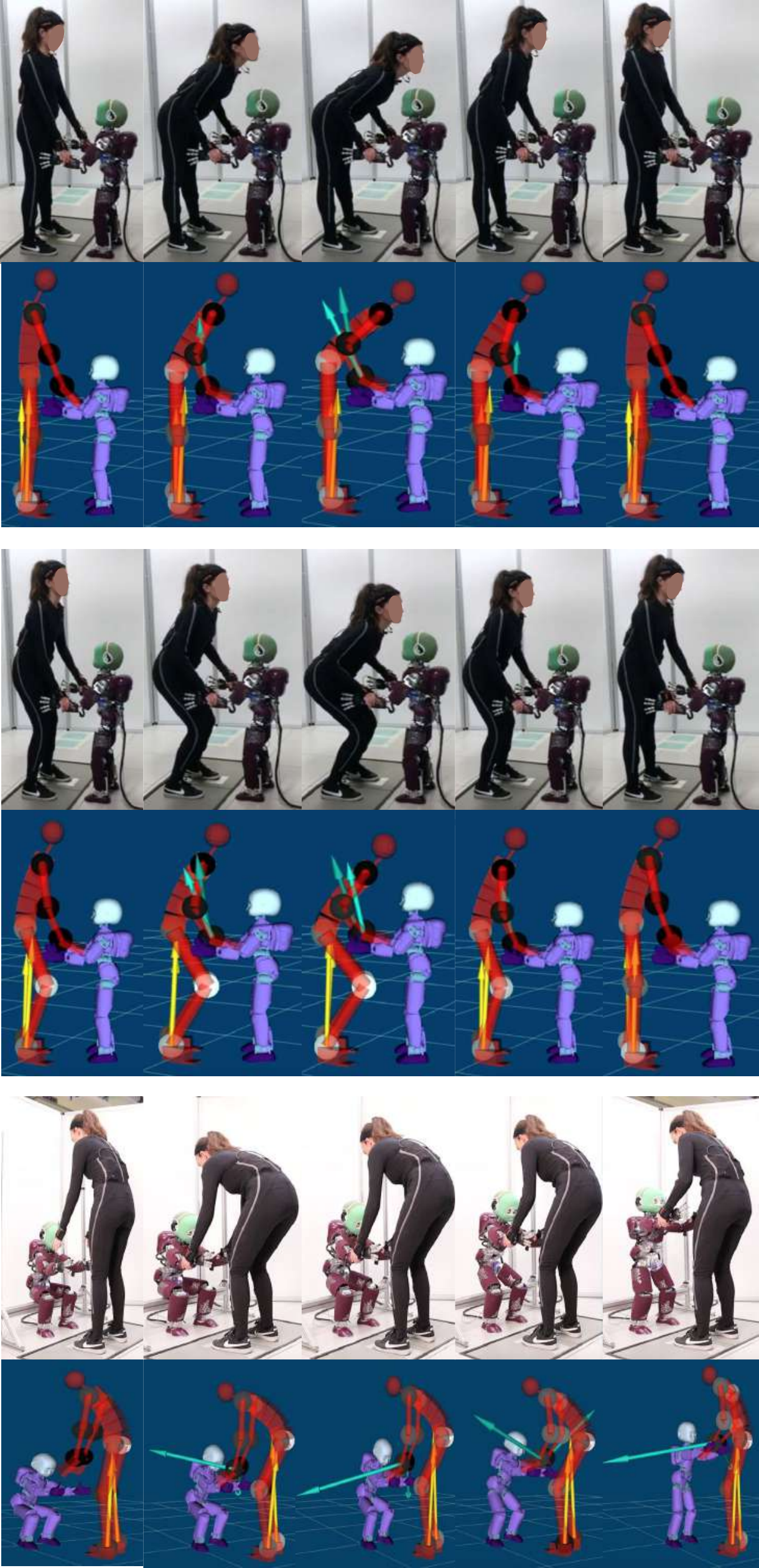}
\caption{Subject performing three different tasks with the related
real-time RViz visualization: a bowing (\emph{on top}), a squat
(\emph{on middle}), a task where the human is helping the iCub to
stand up from a rigid support (\emph{on bottom}).
The visualizer shows the forces expressed in human frames: the ground reaction
 forces (in yellow) and the forces measured by the F/T sensors of the robot at
  the human hands (in light blue).
It visualizes also how much `effort' (in terms of joint torques) the human is
 doing during the pHRI by means of gray-scale spheres placed at the joints: a
  light sphere means a high effort, a dark sphere a minor effort.}
  \label{fig:Figs_realtime_sequenceFP}
\end{figure}

In addition to providing a visual feedback to the estimation, the visualizer
 gives us an important information about how much the human joint effort is.
  The joint torques (estimated via MAP) are represented with spheres whose
   colour is an indicator of the `effort': in a gray scale, a light sphere means
    a high effort, a dark sphere a minor effort.
Figure \ref{fig:Figs_realtime_sequenceFP} shows three different
  pHRI tasks with their real-time visualizations, for a bowing, a squat task,
   a task in which the human is helping the robot to stand up from a rigid
    support, respectively.
At the current stage, any kind of pHRI can be visualized in the
 real-time framework as long as it obeys the initial assumptions, (i.e., fixed
    base and rigid contact conditions).

\newpage
\begin{figure}[h]
  \centering
    \includegraphics[width=.5\textwidth]
    {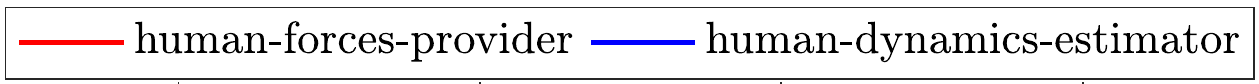}
\end{figure}

\newpage

\chapter{The \emph{human-in-the-loop} Concept }  %
\label{chapter_theHumanInTheLoop}

\begin{quotation}
\noindent\emph{
Without haste, but without rest.
}
    \begin{flushright}
        Wolfgang von Goethe
    \end{flushright}
\end{quotation}

\noindent
This Chapter discusses the possibility to extend the human dynamics estimation
 framework to a new framework that encompasses an \emph{active}
  collaboration with a robot.
The \emph{human-in-the-loop} concept implies that the human agent is
 collaborating with a torque-controlled robot that is able to exploit the human
  collaboration (in term of human dynamics) to achieve its control objective.
The dyadic interaction yields to a bidirectional contribute: 

$i)$ from the human side, it is mandatory to exploit the contact forces
 measured during the interaction through the robot for computing the human
  dynamics;

$ii)$ from the robot side, it is mandatory to exploit the information coming
 from the human dynamics in order to synthesize a new control objective for
  taking advantage from the human help.

\noindent
The partner-aware robot control is out of the scope of this thesis since it
 still has to be developed in the very next future.  However, the Chapter
  introduces the theoretical background on which the control theory will lay
   its foundations.

Within this context, the framework has to be applied to a new coupled
   system composed by the human and the robot.  The two agents maintain their
    own modelling (possibly
     in a  comparable formalism) but new assumptions on the holonomic
      constraints have to be properly done.

\newpage
\section{Coupled System Modelling}

Consider an interaction scenario with two agents: the human and the robot
 (see a generic scenario in Figure \ref{fig:Figs_holonomicConstraintsZoom}).
Consider also to express the two systems with the formalism adopted for humanoid
  robots. The advantage of this choice is straightforward since it allows to
   handle both the systems with the same mathematical tool. In this domain, the
    application of the floating-base formalism in Section
     \eqref{Lagrangian_definition} leads to two sets of motion equations for the
      two systems, such that
\begin{subequations} \label{coupled_system_eq}
\begin{eqnarray}
\label{human_eqMotion}
\bm {\mathrm{M}}(\bm {{q}}) \dot{\bm{\nu}} + \bm {\mathrm{C}}(\bm {{q}},
 \bm{\nu})\bm{\nu} + \bm {\mathrm{G}}(\bm {{q}}) &=& \begin{bmatrix} \bm 0 \\
  {\bm \tau} \end{bmatrix} + \bm {\mathrm{J}}^\top(\bm {{q}}) \bm{f}~,
\\
\label{robot_eqMotion}
\mathbb{M}(\xoverline{\bm q}) \dot{\xoverline{\bm \nu}} +
 \mathbb{C}(\xoverline{\bm q}, \xoverline{\bm
 \nu})\xoverline{\bm \nu} + \mathbb{G}(\xoverline{\bm q}) &=& \begin{bmatrix}
  \bm 0 \\ \xoverline{\bm \tau} \end{bmatrix} + \mathbb{J}^\top(\xoverline{\bm q})
   \xoverline{\bm{ f}}~,
\end{eqnarray} 
\end{subequations}
where the physical meaning of each term is defined in Table
 \ref{table_coupleSystem_variables}. 

Assume that the human is subject to ${k_h}$ external forces $\bm{f} \in
 \mathbb R^{6k_h}$.  These forces are composed of
 two subsets: the forces $\bm{f}^e \in\mathbb R^{6k^e_h}$ applied to the system
  by the \emph{environment} $e$ (e.g., from the ground), and the forces
   $\bm{f}^d \in\mathbb R^{6k^d_h}$ due to the \emph{dyadic} $d$ interaction
    with the other agent, such that
\begin{eqnarray}
\label{ext_forces_onHumanSys}
\bm{f} = \begin{bmatrix}
\bm{f}_1^e &
\bm{f}_2^e &
\hdots &
\bm{f}_{k^e_h}^{e} &
\bm{f}_1^d &
\bm{f}_2^d &
\hdots &
\bm{f}_{k^d_h}^{d}
\end{bmatrix}^\top \in \mathbb R^{6k_h}~.
\end{eqnarray}
Similarly for the robot agent, there are two types of forces
 $\xoverline{\bm{f}}^e \in\mathbb R^{6k^e_r}$ and $\xoverline{\bm{f}}^d
  \in\mathbb R^{6k^e_r}$, thus
\begin{eqnarray}
\label{ext_forces_onRobotSys}
\xoverline{\bm{f}} = \begin{bmatrix}
\xoverline{\bm{f}}_1^e &
\xoverline{\bm{f}}_2^e &
\hdots &
\xoverline{\bm{f}}_{k^e_r}^{e} &
\xoverline{\bm{f}}_1^d &
\xoverline{\bm{f}}_2^d &
\hdots &
\xoverline{\bm{f}}_{k^d_r}^{d}
\end{bmatrix}^\top \in \mathbb R^{6k_r}~.
\end{eqnarray}
Since the two systems are interacting with the environment independently from
 each other and from the mutual interaction, there is not any relation between
  the
  forces that the environment applies to each system (namely, $\bm{f}^e$ and
   $\xoverline{\bm{f}}^e$).  Conversely, when the contact occurs, the
    interaction implies a holonomic relation between  $\bm{f}^d$ and
   $\xoverline{\bm{f}}^d$.

\begin{table}[h]
\centering
\small
\caption{Physical meaning of the Equations \eqref{human_eqMotion} and
 \eqref{robot_eqMotion} terms.  In order to differentiate the same term for
  the two systems, the $\mathbbm{double-bold}$ font is used for the robot
   matrices and the upper-lined notation for the robot variables. $^\ast$$k$ is
   the number of ($6$D) forces that each system is subject to from external
    entities (the other agent and/or the external environment).}
\label{table_coupleSystem_variables}
\begin{tabular}{c|cc}
\\
\hline\hline
\\
\textbf{Terms in Equations \eqref{coupled_system_eq}} & \textbf{Human} &
 \textbf{Robot}\\
\\
\hline
\\
Internal DoFs            & $n_h$  & $n_r$\\
\rowcolor{Gray}
Mass matrix              & $\bm {\mathrm{M}} \in \mathbb
                           {R}^{({n_h+6})\times({n_h+6})}$
                         & $\mathbb{M} \in \mathbb
                           {R}^{({n_r+6})\times({n_r+6})}$\\
Coriolis effects matrix  & $\bm {\mathrm{C}} \in \mathbb
                           {R}^{({n_h+6})\times({n_h+6})}$
                         & $\mathbb{C} \in \mathbb
                           {R}^{({n_r+6})\times({n_r+6})}$\\
\rowcolor{Gray}
Gravity bias             & $\bm {\mathrm{G}} \in \mathbb R^{n_h+6}$
                         & $\mathbb{G} \in \mathbb R^{n_r+6}$\\
Configuration            & $\bm {{q}}\in SE(3)\times\mathbb R^{n_h}$
                         &  $\xoverline{\bm q} \in SE(3)\times\mathbb R^{n_r}$\\
\rowcolor{Gray}
Velocity                 & ${\bm\nu} \in \mathbb R^{n_h+6}$
                         & $\xoverline{\bm \nu} \in \mathbb R^{n_r+6}$\\
Torque                   & $\bm \tau \in \mathbb R^{n_h}$
                         & $\xoverline{\bm \tau} \in \mathbb R^{n_r}$\\
\rowcolor{Gray}
Jacobian                 & $\bm {\mathrm{J}}(\cdot)$ & $\mathbb{J}(\cdot)$\\
Force$^\ast$             & $\bm{f} \in \mathbb R^{6k_h}$
                         & $\xoverline{\bm{f}}  \in \mathbb R^{6k_r}$\\
\\
\hline \hline
\\
\end{tabular}
\end{table}

\section{Rigid Constraints}

Let $\mathcal{I}$ be a generic inertial frame and define a set of frames
\begin{equation} \label{setFrames_human}
\mathcal{C} = \{c_1^e \quad c_2^e \quad \hdots \quad
c_{k^e_h}^{e} \quad c_1^d \quad c_2^d \quad \hdots \quad c_{k^d_h}^{d} \}
\end{equation}
 associated to the human forces $\bm{f}$ of
  \eqref{ext_forces_onHumanSys}, attached to the human links on which the
   forces are acting. More in detail, frames positions correspond to the point
    (on the link) of the force application with a \emph{z}-axis pointing
    on the direction normal to the contact plane.
If we consider a generic element $c_{k} \in \mathcal{C}$ (both for $c^{e}$ or
 $c^{d}$), it is always
 possible to describe the following relation
\begin{equation} \label{jacobian_vel_human}
    ^{\mathcal{I}}{\bm v}_{c_k} = \bm {\mathrm{J}}_{c_k}(\bm {{q}}) {\bm{\nu}}~,
\end{equation}
where the Jacobian is the map between the human floating-base velocity
 ${\bm{\nu}}$ and the velocity of the frame $c_{k}$ attached to the link w.r.t.
 ${\mathcal{I}}$.

Similarly for the robot, Equation \eqref{setFrames_human} becomes
 \begin{equation} \label{setFrames_robot}
\xoverline{\mathcal{C}}
  = \{\xoverline{c}_1^e \quad \xoverline{c}_2^e \quad \hdots \quad
   \xoverline{c}_{k^e_r}^{e} \quad
   \xoverline{c}_1^d \quad \xoverline{c}_2^d \quad \hdots \quad
    \xoverline{c}_{k^d_r}^{d} \}
 \end{equation}
and if we consider the generic $k$-th element ${\xoverline{c}_k} \in
 \xoverline{\mathcal{C}}$,  thus 
\begin{equation} \label{jacobian_vel_robot}
^{\mathcal{I}}{\bm v}_{{\bar{c}}_k} = \mathbb{J}_{{\bar{c}}_k}(\bm
     {{\xoverline{q}}}) {\xoverline{\bm \nu}}~.
\end{equation}

\begin{figure}[H]
  \centering
    \includegraphics[width=1\textwidth]{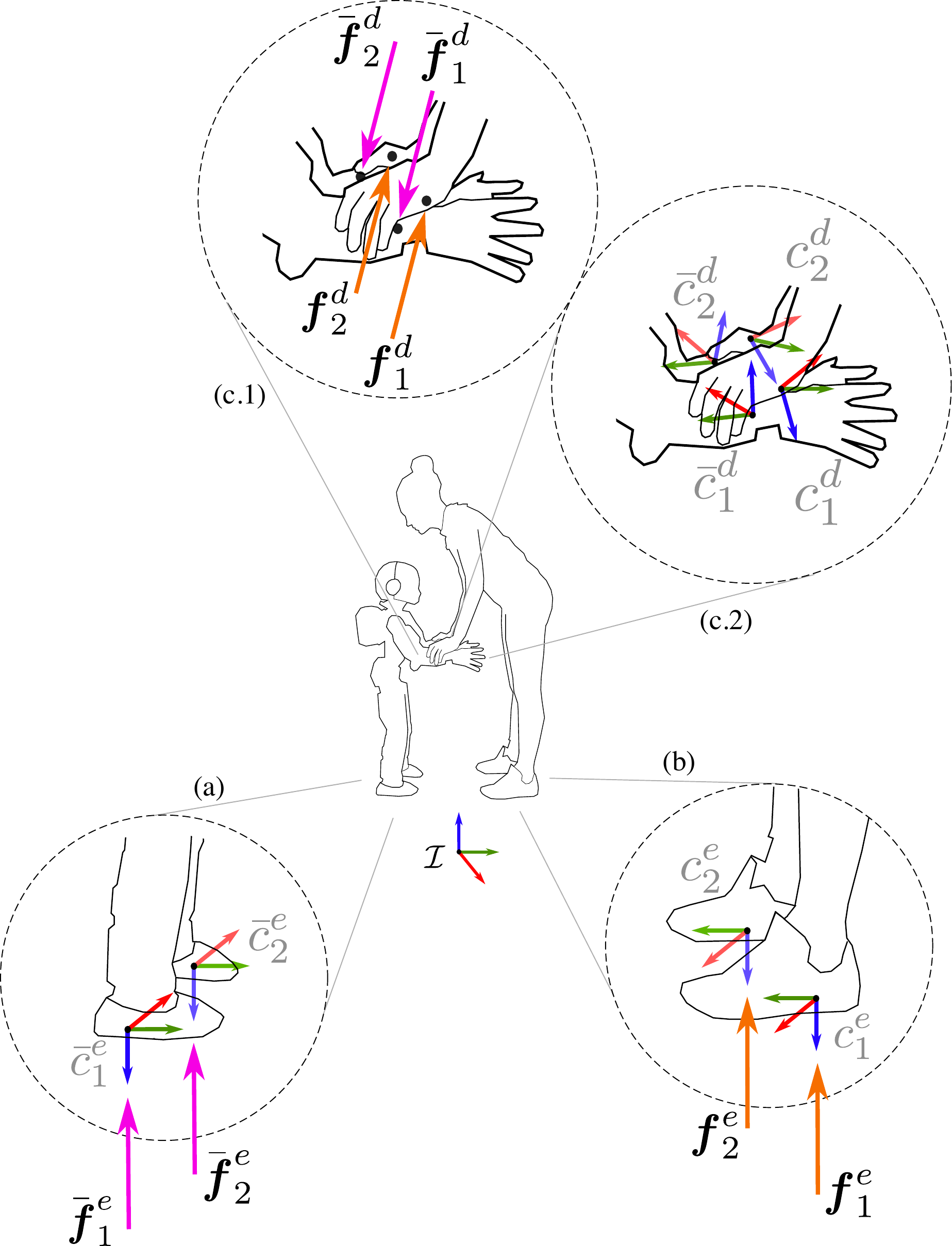}
  \caption{Example of pHRI scenario: both the agents are with their feet fixed
   on the ground (${k^e_h}$ = $2$, ${k^e_r}$ = $2$) while the human is pushing
    down the robot arms (${k^d_h}$ = $2$, ${k^d_r}$ = $2$) . Details (a)
    and (b) show the environment forces acting on the robot and the human,
     respectively.  Both the feet are shown with their associated reference
      frames.
Details (c.1) and (c.2) show the forces exchanged during the interaction and
 their frames, respectively.  The frames are here represented no coincident
  between the human and the robot.}
  \label{fig:Figs_holonomicConstraintsZoom}
\end{figure}

We can now distinguish between two types of rigid constraints occurring in a
 pHRI scenario, as represented in Figure \ref{fig:Figs_holonomicConstraintsZoom}.

\newpage
\vspace{0.6cm}
\noindent
\textbf{1) Environment Holonomic Constraints}
\vspace{0.2cm}

If both the systems are rigidly attached to the ground during the
 interaction (see (a) and (b) in Figure
  \ref{fig:Figs_holonomicConstraintsZoom}), it can be assumed that the
   $k$-th frame in the set  \eqref{setFrames_human}
  associated to $\bm{f}^e \in \bm{f}$ and the $k$-th frame in the set 
   \eqref{setFrames_robot} associated to $\xoverline{\bm{f}}^e \in
    \xoverline{\bm{f}}$ have a constant
    pose w.r.t. $\mathcal{I}$.  For the duration of the contact, they have
     both zero velocity such that
\begin{subequations} \label{vel_zero_envConstraints} 
\begin{eqnarray}
\label{human_vel_zero}
\bm 0 &=& \bm {\mathrm{J}}_{c^e_k}(\bm {{q}}) {\bm{\nu}}~,
\\
\label{robot_vel_zero}
\bm 0 &=& \mathbb{J}_{\bar{c}^e_k}(\bm {{\xoverline{q}}}) {\xoverline{\bm
 \nu}}~.
\end{eqnarray}
\end{subequations}
The differentiation of the Equations \eqref{vel_zero_envConstraints} yields to:
\begin{subequations} \label{holonom_constraint_env}
\begin{eqnarray} \label{holonom_constraint_env_HUMAN}
\bm 0 &=& \bm {\mathrm{J}}_{c^e_k}(\bm {{q}}) \dot{{\bm{\nu}}} +
\dot {\bm {\mathrm{J}}}_{c^e_k}(\bm {{q}}) {{\bm{\nu}}}~,
\\
\label{holonom_constraint_env_ROBOT}
\bm 0 &=& \mathbb{J}_{\bar{c}^e_k}(\bm {{\xoverline{q}}}) \dot{\xoverline{\bm
 \nu}} + \dot{\mathbb{J}}_{\bar{c}^e_k}(\bm {{\xoverline{q}}})
  {\xoverline{\bm \nu}}~,
\end{eqnarray}
\end{subequations}
where \eqref{holonom_constraint_env_HUMAN} represents the holonomic
 constraint of the human with the environment and 
  \eqref{holonom_constraint_env_ROBOT} the holonomic constraint of the robot
   with the environment.

\vspace{0.9cm}
\noindent
\textbf{2) Dyadic Interaction Holonomic Constraints}
\vspace{0.2cm}

When the two systems mutually interact through a physical contact (see (c.1)
 and (c.2) in Figure \ref{fig:Figs_holonomicConstraintsZoom}), the relative
 transformation between the frames ${c^d_k} \in \mathcal{C}$ and ${\bar{c}^d_k}
  \in \xoverline{\mathcal{C}}$ assumes the following form:
\begin{equation} \label{transform_frames_dyadic}
^{c^d_k}{X}_{\bar{c}^d_k}(\bm {{q}}, \bm {{\xoverline{q}}}) =
 ^{{c}^d_k}{X}_{\mathcal{I}}(\bm {{q}})~^{\mathcal{I}}{X}_{\bar{c}^d_k}(\bm
  {{\xoverline{q}}})~,
\end{equation}
where $^{{c}^d_k}{X}_{\mathcal{I}}(\bm {{q}})$ denotes the transformation from
the inertial frame to the human frame ${c^d_k}$ and
 $^{\mathcal{I}}{X}_{\bar{c}^d_k}$ from the robot frame ${\bar{c}^d_k}$ to
${\mathcal{I}}$.
When $^{c^d_k}{X}_{\bar{c}^d_k}(\bm {{q}}, \bm {{\xoverline{q}}})$ is constant,
 it means that the human and the robot are in contact.  The assumption of rigid
  contact yields to the condition that the relative velocity between ${c^d_k}$
   and ${\bar{c}^d_k}$ is zero and that they are moving with the same velocity
    w.r.t. ${\mathcal{I}}$, such as
\begin{equation} \label{rel_framesVelocities_noCoincident}
    ^{\mathcal{I}}{\bm v}_{c^d_k} = ^{c^d_k}{X}_{\bar{c}^d_k}(\bm {{q}}, \bm
     {{\xoverline{q}}})~^{\mathcal{I}}{\bm v}_{\bar{c}^d_k}~.
\end{equation}
Consider, for the sake of simplicity, a situation in which the two frames
 ${c^d_k}$ and ${\bar{c}^d_k}$ are coincident, thus
  $^{c^d_k}{X}_{\bar{c}^d_k}(\bm {{q}}, \bm {{\xoverline{q}}}) = \bm 1_6$ and
   Equation \eqref{rel_framesVelocities_noCoincident} becomes
\begin{equation} \label{rel_framesVelocities_coincident}
    ^{\mathcal{I}}{\bm v}_{c^d_k} =~^{\mathcal{I}}{\bm v}_{\bar{c}^d_k}~.
\end{equation}
By substituting Equations \eqref{jacobian_vel_human} and \eqref{jacobian_vel_robot} in
 \eqref{rel_framesVelocities_coincident}
\begin{equation} \label{velSystems_dyadicConstraints}
\bm {\mathrm{J}}_{c^d_k}(\bm {{q}}) {\bm{\nu}} = \mathbb{J}_{\bar{c}^d_k}(\bm
     {{\xoverline{q}}}) {\xoverline{\bm \nu}}~,
\end{equation}
and differentiating Equation \eqref{velSystems_dyadicConstraints}
\begin{equation} \label{velDiff_dyadicConstraints}
\bm {\mathrm{J}}_{c^d_k}(\bm {{q}}) \dot{{\bm{\nu}}} + \dot {\bm
 {\mathrm{J}}}_{c^d_k}(\bm {{q}}) {{\bm{\nu}}} = 
\mathbb{J}_{\bar{c}^d_k}(\bm {{\xoverline{q}}}) \dot{\xoverline{\bm
 \nu}} + \dot{\mathbb{J}}_{\bar{c}^d_k}(\bm {{\xoverline{q}}})
  {\xoverline{\bm \nu}}~,
\end{equation}
\begin{equation} \label{velDiff_dyadicConstraints_matrixForm}
\begin{bmatrix} 
\bm {\mathrm{J}}_{c^d_k}(\bm {{q}}) & -~\mathbb{J}_{\bar{c}^d_k}(\bm
 {{\xoverline{q}}})
\end{bmatrix}
\begin{bmatrix} \dot{\bm{\nu}} \\ \dot{\bar{\bm \nu}} \end{bmatrix} +
\begin{bmatrix}
\dot {\bm {\mathrm{J}}}_{c^d_k}(\bm {{q}}) & 
-~\dot{\mathbb{J}}_{\bar{c}^d_k}(\bm {{\xoverline{q}}})
\end{bmatrix}
\begin{bmatrix}{\bm{\nu}} \\ {\bar{\bm \nu}} \end{bmatrix} = \bm 0~.
\end{equation}

\vspace{1cm}
In general, the pHRI scenario is fully described by the set of the following
 Equations:
\begin{itemize}
\item{\eqref{coupled_system_eq} for the motion description of the two systems;}
\item{\eqref{holonom_constraint_env} for the holonomic constraints of both the
 systems with the external environment (e.g., the ground in Figure
  \ref{fig:Figs_holonomicConstraintsZoom});}
\item{\eqref{velDiff_dyadicConstraints_matrixForm} for the holonomic constraint
 imposes by the rigid contact interaction between the human and the robot.}
\end{itemize}

A first investigation into the direction of a reactive pHRI has been done in
 \citep{Romano2017codyco}.  The paper attempts to answer the question
  \emph{``How can we predict human intentions so as to synthesize robot
   controllers that are aware of and can react to the human presence?''} by
    considering an interaction scenario between a human (equipped with the
     sensor technology described in the previous chapters) and a human-aware
      iCub robot.
In the paper, the momentum-based balancing controller of the robot has been
 modified (see Section $2$ of the paper) to take into account and exploit the
  human forces.  A task for the robot stand-up from a rigid support has been
   performed with (as at the bottom of the Figure
    \ref{fig:Figs_realtime_sequenceFP} ) and without the human help.
Preliminary results show that the robot needs to provide less torque when
 helped by the human since it is capable to exploit the human assistance.

\chapter{Conclusions and Forthcoming Works}  %
\label{chapter_conclusion}

\begin{quotation}
\noindent\emph{
 Whatever you can do or dream you can, begin it. Boldness has genius, power and
  magic in it. (Begin it!)
}
    \begin{flushright}
(misattributed to) Wolfgang von Goethe
    \end{flushright}
\end{quotation}

\vspace{1.5cm}

The understanding of human dynamics and the way in which its contribution
can be applied to enhance a physical human-robot interaction are two of the
 most promising challenges for the scientific community.  The ever-growing
  interest in the topic, mainly aroused in the last two decades, led us to
   explore and shape the interaction mechanism involved during a physical
    interaction between humans and robotic machines.

Three years ago, the project call of my Ph.D. course stated the following
 sentence:
 \emph{[...] this research proposal aims at developing a prototype of a force
  and motion capture system for humans}.  By bearing in mind the final
   objective of the project, we really developed that prototype.  This thesis
    describes the design of a novel framework for the simultaneous human
     whole-body motion tracking and dynamics estimation.
However, it puts several questions that have to be tackled in the future if the
 human estimation will be integrated in the robot control loop to enhance the
  human-robot interaction.

In this conclusive Chapter, every treated topic in the thesis is recalled
 together with a discussion of the results and the future developments.

\section{Discussion}

In \textbf{Chapter \ref{Chapter_human_modelling}} we apply the estimation
 approach to human models described as articulated multi-body systems
  composed of $1$-DoF revolute joints, by using the classical formalism
   widespread in robotics \citep{Featherstone2008}.
In particular we combined this type of joints to obtain a series of joints with
 a high number of DoFs ($2$ or $3$).  As a straightforward consequence, the
  model is composed by $23$ real and $23$ fake links (see Appendix
   \ref{URDF_human_modelling_appendix}).  This results in a very
    big vector $\bm d$ (i.e., $\bm d \in
    \mathbb R^{1248}$) and affects remarkably the computational time.  The first
     intervention will deal with a solution for removing the fake links from
      the model.

Furthermore, real human joints barely exhibit
  the pure-axial motion, thus our modelling is only a rough approximation of
   the complexity exhibited by real-life biomechanical joints.
 Despite the fact that we chose this joint model for an initial exploration of
  the method,
the proposed algorithm is not limited to this particular choice.  In particular
 the properties of any joint (with an arbitrary number of DoFs)
  can be encapsulated in an interface where the relative position, velocity
 and acceleration of the two bodies connected by it are described by arbitrary
    functions of joint coordinates variables and their derivatives that can
     then be directly inserted in the equations of the system
      \eqref{eq:systemEq}.  In this way, any kind of joint modelling can be
       described under this formalism.
In the future we plan to generalize the method to arbitrarily complex
 musculoskeletal models (see a possible inspiring implementation in
  \citep{seth2010minimal}).

Another important investigation will concern the estimation of the human
 inertial parameters (mass, CoM, inertias).  The use of the anthropometric
  tables is currently the most used tool for estimating such values.
Even though the tables allow to scale the model along with the subject, this
 could be a rough approximation for the non-standard population samples
  (children, elderly, obese, individual with prostheses) and this is pushing
   the scientific community towards the development of new alternative ways to
    estimate these parameters directly from data \citep{Venture2009},
     \citep{Venture2017}.

\vspace{0.3cm}
\textbf{Chapter \ref{Chapter_estimation_problem}} describes the human dynamics
 estimation problem by framing the solution in a probabilistic
  domain by means of a MAP estimator or equivalently a weighted least-squares.
  The main limitation of the current stage
   lies in its fixed-base formulation.   Even if a mathematical formalism
    already exists for the floating-base representation (e.g., Equation
     \eqref{floating_Lagrangian_representation}) the existing software tools
      do not support it yet.
The upgrade to a floating-base model is mandatory if we want to test our
 algorithm in more complex experimental setups where both humans and
  robots could move while interacting.

\vspace{0.3cm}

\textbf{Chapter \ref{chapter_implementationANDvalidation}} presents a software
 implementation for the MAP algorithm specifically tailored for Matlab offline
  validation procedures.  The Chapter shows how the algorithm is able to
   estimate human kinematic and dynamic variables.
The estimation capability of the algorithm is validated through a comparison
 between those variables that can be measured and estimated at the same time.
  Although Section \ref{dataAnalysis} leads to suitable
   estimations, the method questionability lies the over-reliance of the
    solution on the chosen measurement covariances.
Their initial setting was manually tuned by using datasheet values (when
 available) and kept constant for all the setups.  This may have altered the
  goodness of the results.
The problem equally affects the real-time YARP estimation since, at the current
 stage, even there the measurement covariances are defined in a fixed initial
  configuration file. 
The next forthcoming work goes towards the direction of a sort of data-driven
 covariance estimation as it is well-known that the measurement covariances may
  vary when the process is at different operating conditions
   \citep{chapterElsevierBVarianceEstimation}.
The Expectation-Maximization (EM) algorithm is one solution for the data-driven
 estimation \citep{Keller1992} \citep{Chen1997} \citep{Kulathinal2002}.  By
  starting from a known initial covariance of the measurement (e.g., from the
   sensor datasheet), the covariance $\bm {\Sigma}_{y|d}$ (Equation
    \eqref{eq:sigma_dgiveny}) could be optimized until the EM does not increase
     the likelihood anymore.

\vspace{0.3cm}
\textbf{Chapter \ref{chapter_towardsTheRealTime}} endeavours to design a
 C++ based tool for monitoring the real-time dynamics of a human
  being physically involved in a collaborative task with a robot.  This
   Chapter, in a sense, describes the real-time evolution of the previous
offline framework and proposes a preliminary validation analysis with very
 promising results.
The real-time context automatically yields to a different way for retrieving
 the human state and the joint accelerations w.r.t. the offline case.  
 Currently, a straightforward way for computing ${\bm{q}}$ and $\dot{\bm{q}}$
  from data has been already implemented in Section
   \ref{YARP_description}.
However, we are still investigating on the real-time way for obtaining
 $\ddot{\bm{q}}$. Some solutions are suggested in \citep{Bogert2013} and in the
 GitHub repository \citep{RTbiomech_repo}.

For the time being, all the analysis performed in Chapters
 \ref{chapter_implementationANDvalidation} and
  \ref{chapter_towardsTheRealTime} completely disregard $-$at each computation
   step $t$$-$ the contribution of the computation at the previous step
    $t{-}1$.  Thus, the previous state does not influence every new
     computation at all.
This is due to the fact that ${\bm{q}}$ and $\dot{\bm{q}}$ are known and
 without uncertainty.  This assumption is definitively too restrictive since we
  typically have access to a limited set of these
   variables through noisy measurements.  It is well-known that in practical
    acquisitions, the state is affected by statistical noise and other
    inaccuracies.  The Kalman filtering analysis will allow us to obtain a more
     accurate estimation than those based on a single measurement alone (i.e.,
      the current case).
This research is therefore moving towards an extended Kalman filtering (EKF)
 analysis \citep{Lefebvre2004}.  The EKF is a common used tool for improving the
  estimation performances of nonlinear problems (e.g., \citep{Lin2012}) and many
   variants have been developed starting from it (e.g.,\citep{Joukov2017}). 
However, in order to apply the Kalman theory to our estimation problem, several
modifications on the formalism in the system \eqref{eq:systemEq} are
 required.  The first step is to include the estimation of the state $\bm x =
  (\bm q, \dot{\bm q})$ that currently is missing in the problem.
This is not a trivial task and complications arise from the fact that the system \eqref{eq:systemEq} is not linear in $\bm x$.  Appendix 
\ref{dyn&stateEstim} recalls the mathematical computations necessary to modify
 the MAP problem for a simultaneous human dynamics and state estimation.  This
  will be the very first step in order to fit the EKF requirements.

\vspace{0.3cm}
The long-term objective is represented by \textbf{Chapter
 \ref{chapter_theHumanInTheLoop}} where the possibility to extend the human
  dynamics estimation framework to a new framework that encompasses an
   \emph{active} collaboration with a robot is discussed.
It is worth remarking that, at the current stage, the robot is considered as a
 \emph{passive} forces measurer.  In the experimental setups of Figures
  \ref{fig:Figs_bowingTask_pHRI} and \ref{fig:Figs_realtime_sequenceFP}, the
   robot is considered only as a tool for estimating the external forces
    acting on the human hands.
 
However, since the human dynamics is of pivotal importance for a control design
 aimed at considering the \emph{human in the loop}, the forthcoming idea
will be to provide (online) the robot with the human force feedback.  This
 information could be used as a tool for \emph{reactive} human-robot
 collaboration (implying a robot reactive control) and, in a long-term
 perspective, for \emph{predictive} collaboration and enhancing
  remarkably the interaction naturalness.
Thus, in the near future a new robot controller has to be
designed in order that the robot can adapt and adjust the interaction strategy
 accordingly.  This is also the first milestone of the H$2020$ An.Dy project
 that represents the technology outcome of this work and aims at advancing the
  current state of the art in the pHRI field.

\begin{appendices}
\clearpage\null

\begingroup
  \pagestyle{empty}
  \null
  \newpage
\endgroup

\chapter{Top-Down and Bottom-Up Approaches} \label{TopDown_BottomUp}

\begin{figure}[H]
  \centering
 \includegraphics[width=.4\textwidth]{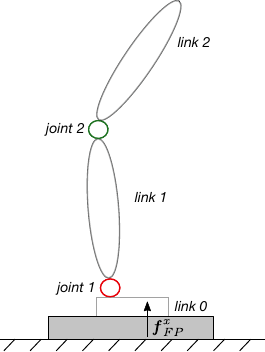}
  \caption{Representation of a $2$-DoF fixed-base model standing on a force
   plate (in grey).}
  \label{illustrativeExample_Appendix}
\end{figure}

Consider a generic $2$-DoF model (Figure
 \ref{illustrativeExample_Appendix}).  The model is represented as a
 kinematic tree with $N_B$ = $2$ moving links numbering from $1$ to $N_B$.  
 Assume that the model is standing on a force plate (FP), in rigid contact with
   link $0$ (i.e., the fixed base), that provides a measurement of the
   force that the model is exchanging with the ground.  No external forces are
    acting on links $1$ and $2$.  Let
$\bm {\underline g} =
\begin{bmatrix}
   0 & 0 & -9.81 & 0 & 0 & 0
\end{bmatrix}^T$
be the gravitational spatial acceleration vector expressed in the body frame
 $0$.

\noindent
The kinematics of the system is described by the following equations:
\begin{eqnarray}
\label{A:boundary conditions_vel}
\bm {\underline v}_0 &=& \bm 0 \\
\label{A:boundary conditions_accGrav}
\bm {\underline a}_0 &=& -\bm {\underline g}\\
\vspace{1cm} \notag\\
\bm {\underline v}_1 &=&\prescript{1}{}{\bm X_{0}} \bm {\underline
 v}_{0} + \bm {\bar S}_1 \dot {\bm q}_1\\
\bm {\underline v}_{J1} &=& \bm {\bar S}_1 \dot {\bm q_1}\\
\label{A:kinematics_link1}
\bm {\underline a}_1 &=& \prescript{1}{}{\bm X_{0}} \bm {\underline
 a}_{0} + \bm {\bar S}_1 \ddot {\bm q}_1 + \bm {\underline
  v}_1{\times}~ \bm {\underline v}_{J1}\\
\vspace{1cm} \notag\\
\bm {\underline v}_2 &=& \prescript{2}{}{\bm X_{1}} \bm {\underline
 v}_{1} + \bm {\bar S}_2 \dot {\bm q}_2\\
\bm {\underline v}_{J2} &=& \bm {\bar S}_2 \dot {\bm q_2}\\
\label{A:kinematics_link2}
\bm {\underline a}_2 &=& \prescript{2}{}{\bm X_{1}} \bm {\underline
 a}_{1} + \bm {\bar S}_2 \ddot {\bm q}_2 + \bm {\underline
  v}_2{\times}~ \bm {\underline v}_{J2}
\end{eqnarray}
The common mathematical approach to compute the dynamics addresses to
 recursively solving the Newton-Euler equations for each link of the body.  The
  recursive algorithm yields to different results depending on the choice of
   the starting point of the propagation.
If the algorithm starts from link $0$ going upward to link $2$, the recursion
 is called \emph{bottom-up}.  Reversely, from link $2$ to link $0$, it is a
  \emph{top-down} recursion.  Examine, in the our model specific case, what is
   the difference of the two approaches.

\subsection*{Top-Down}

\begin{eqnarray}
\bm {\underline f}^B_2 &=&  \bm {\mathrm{\underline I}}_2 \bm {\underline a}_2
 + \bm
 {\underline v}_2{\times^*}~ \bm {\mathrm{\underline I}}_2 \bm {\underline
  v}_2\\
\label{A:TD_dyn_link2}
\bm {\underline f}_2 &=& \bm {\underline f}^B_2 - \bm {\underline f}_2^x = \bm
 {\underline f}^B_2\\
\vspace{1cm} \notag\\
\bm {\underline f}^B_1 &=&  \bm {\mathrm{\underline I}}_1 \bm {\underline a}_1
 + \bm
 {\underline v}_1{\times^*}~ \bm {\mathrm{\underline I}}_1 \bm {\underline v}_1\\
\label{A:TD_dyn_link1}
\bm {\underline f}_1 &=& \bm {\underline f}^B_1 - \bm {\underline f}_1^x +
 \prescript{1}{}{\bm X_{2}^*} \bm {\underline f}_{2} = \bm {\underline f}^B_1 +
  \prescript{1}{}{\bm X_{2}^*} \bm {\underline f}_{2}\\
\vspace{1cm} \notag\\
\bm {\underline f}^B_0 &=&  \bm {\mathrm{\underline I}}_0 \bm {\underline a}_0
 + \bm
 {\underline v}_0{\times^*}~ \bm {\mathrm{\underline I}}_0 \bm {\underline v}_0
  = 
 \bm {\mathrm{\underline I}}_0 \bm {\underline a}_0 \\
\label{A:TD_link0}
\bm {\underline f}_0 & =& \bm {\underline f}^B_0 - \bm {\underline f}_0^x +
 \prescript{0}{}{\bm X_{1}^*} \bm {\underline f}_{1}
\end{eqnarray}

\noindent
Since $\bm {\underline f}_0 = \bm 0$ and $\bm {\underline f}_0^x =
 \prescript{0}{}{\bm X_{FP}^*} \bm
 {\underline f}_{FP} $, thus
\begin{eqnarray}
\bm 0 &=& \bm {\mathrm{\underline I}}_0 \bm {\underline a}_0 -
 \prescript{0}{}{\bm
 X_{FP}^*}\bm {\underline f}_{FP} + \prescript{0}{}{\bm X_{1}^*} \bm
 {\underline f}_{1} \\
 \Rightarrow
 \prescript{0}{}{\bm X_{1}^*} \bm {\underline f}_{1}
  &=& \bm {\mathrm{\underline I}}_0 \bm {\underline g} + \prescript{0}{}{\bm
  X_{FP}^*}\bm {\underline f}_{FP} \\
\label{A:TD_overdetermidSolution}
\Rightarrow
 \bm {\underline f}_{1} &=& \prescript{1}{}{\bm X_{0}^*} \big(\bm
  {\mathrm{\underline I}}_0 \bm {\underline g} + \prescript{0}{}{\bm X_{FP}^*}
   \bm {\underline f}_{FP}\big)
\end{eqnarray}
The overdeterminancy of the system yields to a physical inconsistency: $\bm
 {\underline f}_{1}$ is defined from both Equations \eqref{A:TD_dyn_link1} and
  \eqref{A:TD_overdetermidSolution}.

\subsection*{Bottom-Up}

\begin{eqnarray}
\bm {\underline f}_0 & =& \bm {\underline f}^B_0 - \bm {\underline f}_0^x +
 \prescript{0}{}{\bm X_{1}^*} \bm {\underline f}_{1} \\
\Rightarrow
 \bm {\underline f}_{1} &=& \prescript{1}{}{\bm X_{0}^*} \big(\bm
  {\mathrm{\underline I}}_0 \bm {\underline g} + \prescript{0}{}{\bm X_{FP}^*}
   \bm {\underline f}_{FP}\big) \\
\vspace{1cm} \notag\\
\bm {\underline f}_1 &=& \bm {\underline f}^B_1 +
  \prescript{1}{}{\bm X_{2}^*} \bm {\underline f}_{2}\\
\Rightarrow
\prescript{1}{}{\bm X_{2}^*} \bm {\underline f}_{2} &=& \bm {\underline f}_1 -
 \bm {\underline f}^B_1\\
\label{A:BU_solution}
\Rightarrow
 \bm {\underline f}_{2} &=& \prescript{2}{}{\bm X_{1}^*}
  \big(\bm {\underline f}_1 - \bm {\underline f}^B_1 \big)\\
\vspace{1cm} \notag\\
\label{A:BU_overdetermidSolution}
\bm {\underline f}_2 &=& \bm {\underline f}^B_2 - \bm {\underline f}_2^x = \bm
 {\underline f}^B_2
\end{eqnarray}
Again, as in the top-down approach, there is a physical inconsistency for the
 force $\bm {\underline f}_2$ that is represented by both \eqref{A:BU_solution}
  and \eqref{A:BU_overdetermidSolution}.

\subsection*{Solution Criterion}

When an additional measurement (e.g., $ {\underline f}_{FP}$) is added into the
 computation, the system becomes overdetermined.  It is evident at the top-most
  segment (in the case of bottom-up) or at the
   bottom-most segment (in the case of top-down) where the physics condition
    are not satisfied anymore.
The solution is obtained by discarding one set of $6$ equations (e.g.,
 \eqref{A:TD_dyn_link1} or \eqref{A:TD_overdetermidSolution} for the top-down,
  \eqref{A:BU_solution} or \eqref{A:BU_overdetermidSolution} for the bottom-up)
   by strongly conditioning the final result of the computation.
An important drawback of this criterion is that all the other variables (both
 forces and kinematic data) that entered in the discarded set of equations will
  remain unused in the computation.

\begingroup
  \pagestyle{empty}
  \null
  \newpage
\endgroup

\chapter{URDF Human Modelling} \label{URDF_human_modelling_appendix}

The URDF is an XML specification describing
 the kinematic and dynamic properties of a robot in \cite{urdf_ros}.  It is
  composed of \texttt{link} and \texttt{joint} elements.  
For the \texttt{link} several
   attributes are specified: the name, the inertial properties (mass, CoM
    origin, inertias), the visual properties (boxes origin and geometry).
Hereafter an example for the RightLowerLeg element.

\vspace{0.5cm}
\begin{lstlisting}[language=xml]
<linkName ="RightLowerLeg">
    <inertial>
         <massValue ="RIGHTLOWERLEGMASS"/>
         <!--COM origin w.r.t. jRightKnee-->
         <originxyz ="RIGHTLOWERLEG_COM_ORIGIN" rpy ="0 0 0" />
         <inertia ixx="RIGHTLOWERLEGINERTIAIXX" iyy="RIGHTLOWERLEGINERTIAIYY"
          izz="RIGHTLOWERLEGINERTIAIZZ" ixy="0" ixz="0" iyz="0"/>
    </inertial>
    <visual>
         <!--box origin w.r.t. jRightKnee-->
         <origin xyz="RIGHTLOWERLEG_BOX_ORIGIN" rpy="0 0 0" />
         <geometry>
                <cylinder length="RIGHTLOWERLEGHEIGHT"
                 radius="RIGHTLOWERLEGRADIUS"/>
         </geometry>
    </visual>
</link>
\end{lstlisting}

\begin{figure}[H]
  \centering
    \includegraphics[width=0.9\columnwidth]{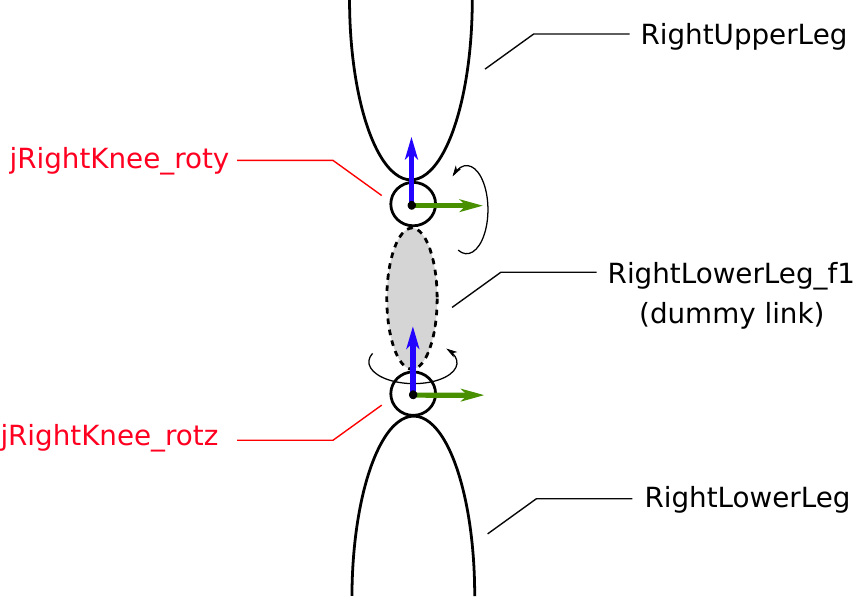}
     \caption{The dummy link connects the $1$-DoF joint jRightKnee (rotation
      along $y$) and the $1$-DoF joint jRightKnee (rotation along $z$), in order
to make possible the creation of a $2$-DoF joint.} 
  \label{dummy_link}
\end{figure}

For the \texttt{joint} attributes : the name and the type, the origin, the
 parent and child
 link, the axis along which the motion is allowed.  Since the URDF does not
  support spherical joint design,  we combine joints of revolute type to obtain
   a series of joints with a high number of DoFs.
The limitation on the \texttt{joint} automatically implied an important
 modification of the entire structure in the tree.  Since the URDF supports
  only a structure where each link is connected with $1$-DoF joint (a
   condition of two consecutive joints is not supported), thus it had to be
    necessary the creation of a `dummy' \texttt{link}.
Here follow the XML tag for the creation of the dummy \texttt{link} for the
 RightLowerLeg (see also Figure \ref{dummy_link}).

\vspace{0.5cm}
\begin{lstlisting}[language=xml]
<link name="RightLowerLeg_f1">
    <inertial>
         <mass value="DUMMYMASS"/>
         <origin xyz="0 0 0" rpy="0 0 0" />
         <inertia ixx="DUMMYININERTIA" iyy="DUMMYININERTIA"
          izz="DUMMYININERTIA" ixy="0" ixz="0" iyz="0"/>
    </inertial>
</link>
\end{lstlisting}
\vspace{0.5cm}

\newpage
Hereafter, it is showed the XML tag for defining
 \texttt{joint} elements.  In the specific case of the right knee, it is
  specified the lower joint limit (\unit{0}{\degree} for the flexion along $y$,
 \unit{-40}{\degree} for the lateral rotation along $z$) and the upper joint
  limit (\unit{~135}{\degree} for the flexion along $y$, \unit{30}{\degree} for
   the lateral rotation along $z$).  Note that the limit effort and velocity are
    safety attributes specific for controllers settings.  Their values are not
     used in the case of human models. 

\vspace{0.2cm}
\begin{lstlisting}[language=xml]
<joint name="jRightKnee_roty" type="revolute">
    <origin xyz="jRightKnee_ORIGIN" rpy="0 0 0"/>
    <parent link="RightUpperLeg"/>
    <child link="RightLowerLeg_f1"/>
    <dynamics damping="0.0" friction="0.0"/>
    <limit effort="30" velocity="1.0" lower="0" upper="2.35619"/>
    <axis xyz="0 1 0"/>
</joint>
<joint name="jRightKnee_rotz" type="revolute">
    <origin xyz="0 0 0" rpy="0 0 0"/>
    <parent link="RightLowerLeg_f1"/>
    <child link="RightLowerLeg"/>
    <dynamics damping="0.0" friction="0.0"/>
    <limit effort="30" velocity="1.0" lower="-0.698132" upper="0.523599"/>
    <axis xyz="0 0 1"/>
</joint>
\end{lstlisting}

The final URDF file is a XML list of \texttt{link} and \texttt{joint} elements
 entirely describing the model.  Hereafter are listed, in two different XML
  windows, all the links and the joints of the $48$-DoF URDF template.
 
\vspace{0.2cm}
\begin{lstlisting}[language=xml]
<!--URDF MODEL 48 DoFs-->
<robot name="XSensStyleModel_template">
<!--LINKS-->
    <!--Link base (1)-->
    <link name="Pelvis">
    <!--Chain from (2) to (7)-->
    <link name="L5_f1">
    <link name="L5">
    <link name="L3_f1">
    <link name="L3">
    <link name="T12_f1">
    <link name="T12">
    <link name="T8_f1">
    <link name="T8_f2">
    <link name="T8">
    <link name="Neck_f1">
    <link name="Neck_f2">
    <link name="Neck">
    <link name="Head_f1">
    <link name="Head">
    <!--Chain from (8) to (11)-->
    <link name="RightShoulder">
    <link name="RightUpperArm_f1">
    <link name="RightUpperArm_f2">
    <link name="RightUpperArm">
    <link name="RightForeArm_f1">
    <link name="RightForeArm">
    <link name="RightHand_f1">
    <link name="RightHand">
    <!--Chain from (12) to (15)-->
    <link name="LeftShoulder">
    <link name="LeftUpperArm_f1">
    <link name="LeftUpperArm_f2">
    <link name="LeftUpperArm">
    <link name="LeftForeArm_f1">
    <link name="LeftForeArm">
    <link name="LeftHand_f1">
    <link name="LeftHand">
    <!--Chain from (16) to (19)-->
    <link name="RightUpperLeg_f1">
    <link name="RightUpperLeg_f2">
    <link name="RightUpperLeg">
    <link name="RightLowerLeg_f1">
    <link name="RightLowerLeg">
    <link name="RightFoot_f1">
    <link name="RightFoot_f2">
    <link name="RightFoot">
    <link name="RightToe">
    <!--Chain from (20) to (23)-->
    <link name="LeftUpperLeg_f1">
    <link name="LeftUpperLeg_f2">
    <link name="LeftUpperLeg">
    <link name="LeftLowerLeg_f1">
    <link name="LeftLowerLeg">
    <link name="LeftFoot_f1">
    <link name="LeftFoot_f2">
    <link name="LeftFoot">
    <link name="LeftToe">
<!--JOINTS-->
    <!--Chain from (2) to (7)-->
    <joint name="jL5S1_rotx" type="revolute">
    <joint name="jL5S1_roty" type="revolute">
    <joint name="jL4L3_rotx" type="revolute">
    <joint name="jL4L3_roty" type="revolute">
    <joint name="jL1T12_rotx" type="revolute">
    <joint name="jL1T12_roty" type="revolute">
    <joint name="jT9T8_rotx" type="revolute">
    <joint name="jT9T8_roty" type="revolute">
    <joint name="jT9T8_rotz" type="revolute">
    <joint name="jT1C7_rotx" type="revolute">
    <joint name="jT1C7_roty" type="revolute">
    <joint name="jT1C7_rotz" type="revolute">
    <joint name="jC1Head_rotx" type="revolute">
    <joint name="jC1Head_roty" type="revolute">
    <!--Chain from (8) to (11)-->
    <joint name="jRightC7Shoulder_rotx" type="revolute">
    <joint name="jRightShoulder_rotx" type="revolute">
    <joint name="jRightShoulder_roty" type="revolute">
    <joint name="jRightShoulder_rotz" type="revolute">
    <joint name="jRightElbow_roty" type="revolute">
    <joint name="jRightElbow_rotz" type="revolute">
    <joint name="jRightWrist_rotx" type="revolute">
    <joint name="jRightWrist_rotz" type="revolute">
    <!--Chain from (12) to (15)-->
    <joint name="jLeftC7Shoulder_rotx" type="revolute">
    <joint name="jLeftShoulder_rotx" type="revolute">
    <joint name="jLeftShoulder_roty" type="revolute">
    <joint name="jLeftShoulder_rotz" type="revolute">
    <joint name="jLeftElbow_roty" type="revolute">
    <joint name="jLeftElbow_rotz" type="revolute">
    <joint name="jLeftWrist_rotx" type="revolute">
    <joint name="jLeftWrist_rotz" type="revolute">
    <!--Chain from (16) to (19)-->
    <joint name="jRightHip_rotx" type="revolute">
    <joint name="jRightHip_roty" type="revolute">
    <joint name="jRightHip_rotz" type="revolute">
    <joint name="jRightKnee_roty" type="revolute">
    <joint name="jRightKnee_rotz" type="revolute">
    <joint name="jRightAnkle_rotx" type="revolute">
    <joint name="jRightAnkle_roty" type="revolute">
    <joint name="jRightAnkle_rotz" type="revolute">
    <joint name="jRightBallFoot_roty" type="revolute">
    <!--Chain from (20) to (23)-->
    <joint name="jLeftHip_rotx" type="revolute">
    <joint name="jLeftHip_roty" type="revolute">
    <joint name="jLeftHip_rotz" type="revolute">
    <joint name="jLeftKnee_roty" type="revolute">
    <joint name="jLeftKnee_rotz" type="revolute">
    <joint name="jLeftAnkle_rotx" type="revolute">
    <joint name="jLeftAnkle_roty" type="revolute">
    <joint name="jLeftAnkle_rotz" type="revolute">
    <joint name="jLeftBallFoot_roty" type="revolute">
\end{lstlisting}

\begin{figure}[H]
    \centering
    \vspace{0.2cm}
      \includegraphics[width=1\columnwidth]{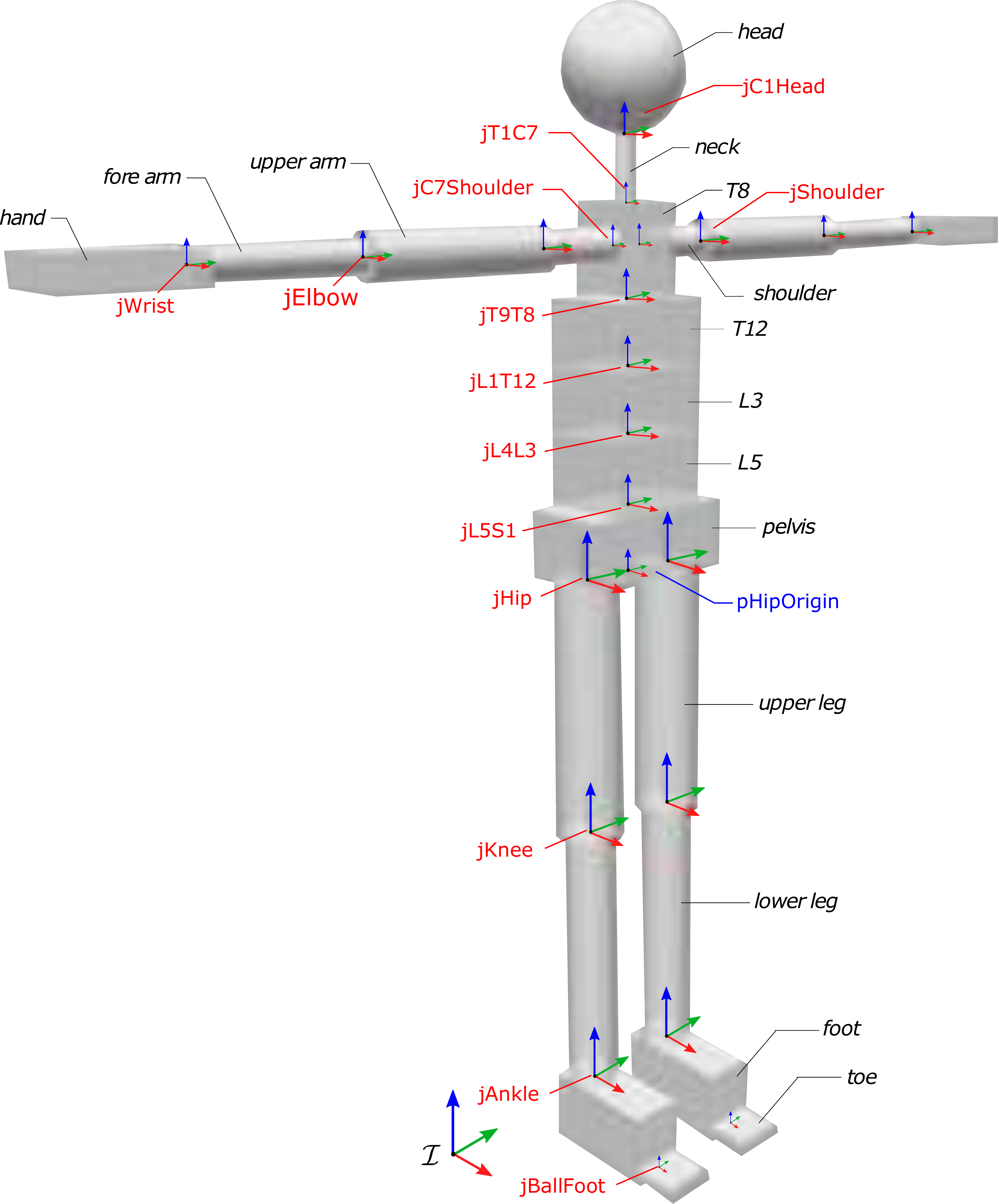}
       \caption{URDF human body model with labels for links
  and joints. 
  The Gazebo simulator is used for the model visualization (DUMMYMASS $=$
   \unit{0.0001}{\kilo}{\gram}, DUMMYININERTIA $=$
    \unit{0.0003}{\kilo\gram~\metre\squared}).}
    \label{human_urdfModel_noSens}
  \end{figure}

\begin{figure}[H]
    \includegraphics[width=\columnwidth]{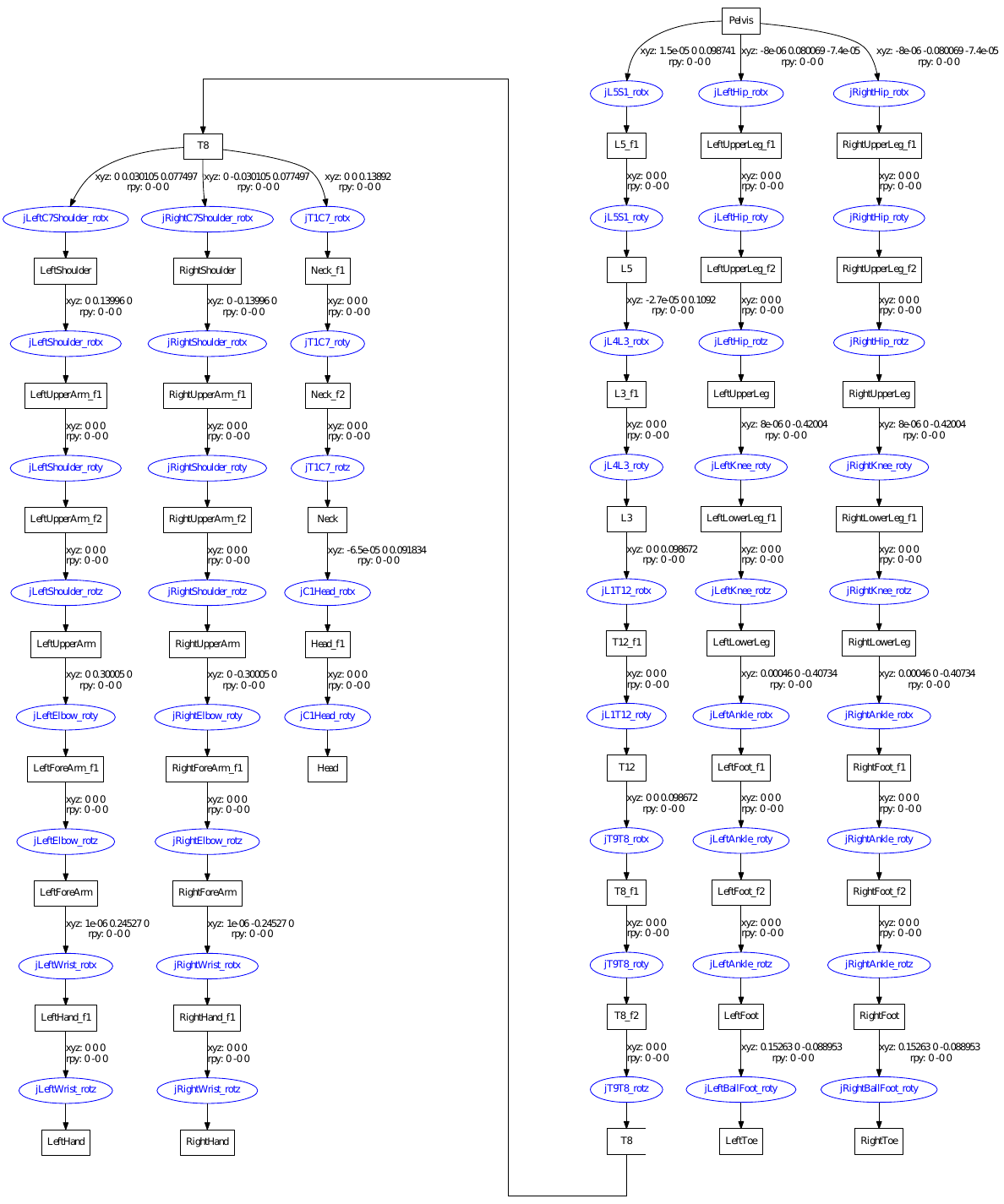}
  \caption{Graphviz diagram visualization \citep{urdf_ros} of the URDF
   model hierarchy.  Black boxes are the links, blue shapes are the joints.}
  \label{fig:Figs_hierarchicalURDF}
\end{figure}

\newpage
The \cite{gazeboSim} simulator
  is used to visualize the model in Figure \ref{human_urdfModel_noSens}.  The
   Graphviz (ROS) visualizer represents the URDF model hierarchy (Figure
    \ref{fig:Figs_hierarchicalURDF}).


\newpage
\section{The Non-Standard URDF Extension} \label{nonStandardURDF}
  
The URDF semantics does not support officially the possibility to insert sensor
 information on the model.
In this thesis it is used a non-standard extension to the format where each
 sensor is encoded in the XML file as  a new type of element \texttt{sensor}.
  For loading sensor information, the name and type of sensor, the link to
   which each sensor is attached and its position in the link have to be
    specified.  Hereafter an example of XML tag for the sensor attached on the
     RightLowerLeg.

\vspace{0.2cm}
\begin{lstlisting}[language=xml]
<!-- Gyroscope-->
<sensor name="RightLowerLeg_gyro" type="gyroscope">
    <parent link="RightLowerLeg"/>
    <!-- sensor pose w.r.t. RightLowerLeg -->
    <origin xyz="RIGHTLOWERLEG_S_POS" rpy="RIGHTLOWERLEG_S_RPY"/>
</sensor>
<!--Accelerometer -->
<sensor name="RightLowerLeg_accelerometer" type="accelerometer">
    <parent link="RightLowerLeg"/>
    <!-- sensor pose w.r.t. RightLowerLeg -->
    <origin xyz="RIGHTLOWERLEG_S_POS" rpy="RIGHTLOWERLEG_S_RPY"/>
</sensor>
\end{lstlisting}
The sensors (IMUs) attached to the human body links are shown in the following
 XML list, extracted from the URDF template:
\vspace{0.2cm}
\begin{lstlisting}[language=xml]
<!--SENSORS-->
   <!-- Sensor  1-->
   <sensor name="Pelvis_gyro" type="gyroscope">
   <sensor name="Pelvis_accelerometer" type="accelerometer">
   <!-- Sensor  2-->
   <sensor name="T8_gyro" type="gyroscope">
   <sensor name="T8_accelerometer" type="accelerometer">
   <!-- Sensor  3-->
   <sensor name="Head_gyro" type="gyroscope">
   <sensor name="Head_accelerometer" type="accelerometer">
   <!-- Sensor  4-->
   <sensor name="RightShoulder_gyro" type="gyroscope">
   <sensor name="RightShoulder_accelerometer" type="accelerometer">
   <!-- Sensor  5-->
   <sensor name="RightUpperArm_gyro" type="gyroscope">
   <sensor name="RightUpperArm_accelerometer" type="accelerometer">
   <!-- Sensor  6-->
   <sensor name="RightForeArm_gyro" type="gyroscope">
   <sensor name="RightForeArm_accelerometer" type="accelerometer">
   <!-- Sensor  7-->
   <sensor name="RightHand_gyro" type="gyroscope">
   <sensor name="RightHand_accelerometer" type="accelerometer">
   <!-- Sensor  8-->
   <sensor name="LeftShoulder_gyro" type="gyroscope">
   <sensor name="LeftShoulder_accelerometer" type="accelerometer">
   <!-- Sensor  9-->
   <sensor name="LeftUpperArm_gyro" type="gyroscope">
   <sensor name="LeftUpperArm_accelerometer" type="accelerometer">
   <!-- Sensor  10-->
   <sensor name="LeftForeArm_gyro" type="gyroscope">
   <sensor name="LeftForeArm_accelerometer" type="accelerometer">
   <!-- Sensor  11-->
   <sensor name="LeftHand_gyro" type="gyroscope">
   <sensor name="LeftHand_accelerometer" type="accelerometer">
   <!-- Sensor  12-->
   <sensor name="RightUpperLeg_gyro" type="gyroscope">
   <sensor name="RightUpperLeg_accelerometer" type="accelerometer">
   <!-- Sensor  13-->
   <sensor name="RightLowerLeg_gyro" type="gyroscope">
   <sensor name="RightLowerLeg_accelerometer" type="accelerometer">
   <!-- Sensor  14-->
   <sensor name="RightFoot_gyro" type="gyroscope">
   <sensor name="RightFoot_accelerometer" type="accelerometer">
   <!-- Sensor  15-->
   <sensor name="LeftUpperLeg_gyro" type="gyroscope">
   <sensor name="LeftUpperLeg_accelerometer" type="accelerometer">
   <!-- Sensor  16-->
   <sensor name="LeftLowerLeg_gyro" type="gyroscope">
   <sensor name="LeftLowerLeg_accelerometer" type="accelerometer">
   <!-- Sensor  17-->
   <sensor name="LeftFoot_gyro" type="gyroscope">
   <sensor name="LeftFoot_accelerometer" type="accelerometer">
\end{lstlisting} %

\begingroup
  \pagestyle{empty}
  \null
  \newpage
\endgroup

\chapter{Offline Inverse Kinematics Computation}
 \label{IK}

\begin{figure}[H]
  \centering
    \includegraphics[width=1\textwidth]{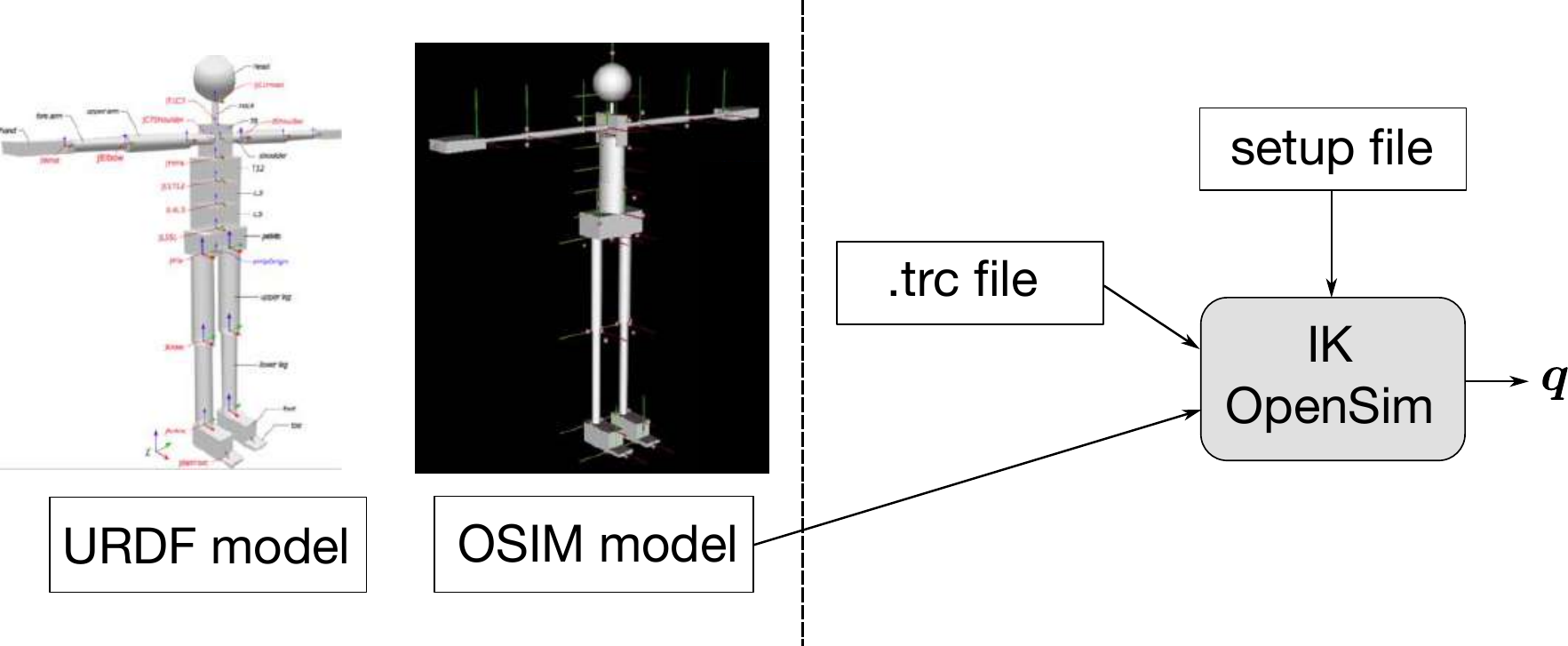}
  \caption{Pipeline for the OpenSim IK.}
  \label{fig:Figs_IK_offlineStep}
\end{figure}

Within the context of the \emph{offline} algorithm validation (see Chapter
   \ref{chapter_implementationANDvalidation}), the Inverse Kinematics (IK) for
    the human model has been performed by means of the OpenSim IK toolboox
     \citep{opensim_site}, through an OpenSim API for Matlab.
Figure \ref{fig:Figs_IK_offlineStep} shows the elements involved in the IK
 computation.

The primary OpenSim IK toolbox inputs are the following files.
\begin{itemize}
\item {A trajectories (.trc) file containing the experimental marker trial
 trajectories of the human subject acquired from a motion capture system.  In
  our specific case, trajectories of $64$ anatomical bony landmarks have been
   provided by Xsens, Figure \ref{fig:Figs_XsensMarkers}.}
\begin{figure}[H]
 \centering
   \includegraphics[width=1\textwidth]{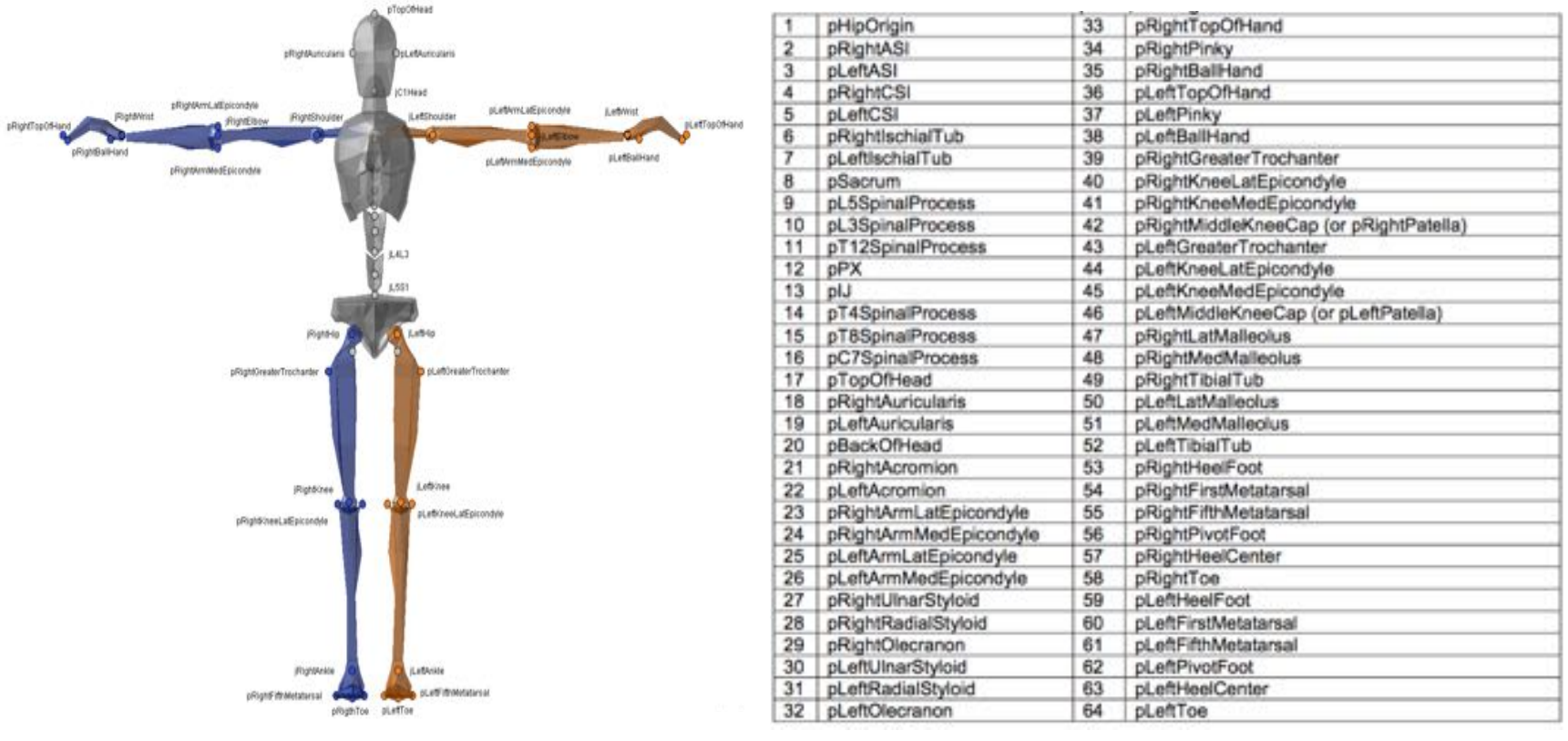}
 \caption{Xsens anatomical bony landmarks.}
\source{Xsens, MVN User Manual, 2005.}
 \label{fig:Figs_XsensMarkers}
\end{figure}
\item {The subject-specific OSIM model created by $i)$ matching
 the same structure of the URDF model (see Appendix
  \ref{URDF_human_modelling_appendix})} and $ii)$ matching the position of
   the $64$ anatomical markers (provided by Xsens) as listed in Figure
    \ref{fig:Figs_XsensMarkers}.
    Here following the markers section in a OSIM model.
\vspace{0.2cm}
\begin{lstlisting}[language=xml]
<!--OSIM MODEL 48 DoFs-->
</OpenSimDocument>
<Model name="XSensStyleModel">
<!--Markers in the model.-->
    <MarkerSet>
      <!--Pelvis markers-->
      <Marker name="pHipOrigin"></Marker>
      <Marker name="pRightASI"></Marker>
      <Marker name="pLeftASI"></Marker>
      <Marker name="pRightCSI"></Marker>
      <Marker name="pLeftCSI"></Marker>
      <Marker name="pRightIschialTub"></Marker>
      <Marker name="pLeftIschialTub"></Marker>
      <Marker name="pSacrum"></Marker>
      <!--L5 markers-->
      <Marker name="pL5SpinalProcess"></Marker>
      <!--L3 markers-->
      <Marker name="pL3SpinalProcess"></Marker>
      <!--T12 markers-->
      <Marker name="pT12SpinalProcess"></Marker>
      <!--T8 markers-->
      <Marker name="pPX"></Marker>
      <Marker name="pIJ"></Marker>				
      <Marker name="pT4SpinalProcess"></Marker>
      <Marker name="pT8SpinalProcess"></Marker>
      <Marker name="pC7SpinalProcess"></Marker>
      <!--Head markers-->	
      <Marker name="pTopOfHead"></Marker>
      <Marker name="pRightAuricularis"></Marker>
      <Marker name="pLeftAuricularis"></Marker>
      <Marker name="pBackOfHead"></Marker>
      <!--RightShoulder markers-->	
      <Marker name="pRightAcromion"></Marker>
      <!--RightUpperArm markers-->
      <Marker name="pRightArmLatEpicondyle"></Marker>
      <Marker name="pRightArmMedEpicondyle"></Marker>
      <!--RightForeArm markers-->
      <Marker name="pRightUlnarStyloid"></Marker>
      <Marker name="pRightRadialStyloid"></Marker>
      <Marker name="pRightOlecranon"></Marker>
      <!--RightHand markers-->
      <Marker name="pRightTopOfHand"></Marker>
      <Marker name="pRightPinky"></Marker>
      <Marker name="pRightBallHand"></Marker>
      <!--LeftShoulder markers-->	
      <Marker name="pLeftAcromion"></Marker>
      <!--LeftUpperArm markers-->
      <Marker name="pLeftArmLatEpicondyle"></Marker>
      <Marker name="pLeftArmMedEpicondyle"></Marker>
      <!--LeftForeArm markers-->
      <Marker name="pLeftUlnarStyloid"></Marker>
      <Marker name="pLeftRadialStyloid"></Marker>
      <Marker name="pLeftOlecranon"></Marker>
      <!--LeftHand markers-->
      <Marker name="pLeftTopOfHand"></Marker>
      <Marker name="pLeftPinky"></Marker>
      <Marker name="pLeftBallHand"></Marker>
      <!--RightUpperLeg markers-->
      <Marker name="pRightGreaterTrochanter"></Marker>
      <Marker name="pRightPatella"></Marker>
      <!--RightLowerLeg markers-->
      <Marker name="pRightKneeLatEpicondyle"></Marker>
      <Marker name="pRightKneeMedEpicondyle"></Marker>
      <Marker name="pRightLatMalleolus"></Marker>
      <Marker name="pRightMedMalleolus"></Marker>
      <Marker name="pRightTibialTub"></Marker>
      <!--RightFoot markers-->
      <Marker name="pRightHeelFoot"></Marker>
      <Marker name="pRightFirstMetatarsal"></Marker>
      <Marker name="pRightFifthMetatarsal"></Marker>
      <Marker name="pRightPivotFoot"></Marker>
      <Marker name="pRightHeelCenter"></Marker>
      <!--RightToe markers-->
      <Marker name="pRightToe"></Marker>
      <!--LeftUpperLeg markers-->
      <Marker name="pLeftGreaterTrochanter"></Marker>
      <Marker name="pLeftPatella"></Marker>
      <!--LeftLowerLeg markers-->
      <Marker name="pLeftKneeLatEpicondyle"></Marker>
      <Marker name="pLeftKneeMedEpicondyle"></Marker>
      <Marker name="pLeftLatMalleolus"></Marker>
      <Marker name="pLeftMedMalleolus"></Marker>
      <Marker name="pLeftTibialTub"></Marker>
      <!--LeftFoot markers-->
      <Marker name="pLeftHeelFoot"></Marker>
      <Marker name="pLeftFirstMetatarsal"></Marker>
      <Marker name="pLeftFifthMetatarsal"></Marker>
      <Marker name="pLeftPivotFoot"></Marker>
      <Marker name="pLeftHeelCenter"></Marker>
      <!--LeftToe markers-->
      <Marker name="pLeftToe"></Marker>
    </MarkerSet>
  </Model>
</OpenSimDocument>
    \end{lstlisting}
\vspace{-0.4cm}
\item {A setup XML file containing all the setting information for the IK
 computation (including the markers weight, the time range, the acceptable
  accuracy).}
\end{itemize}

The IK tool solves a weighted least-squares problem by means of a general
 quadratic programming solver.  The solver minimizes, for each timestamp in the
  task range, the difference between the position of the markers on the model
   and the experimental data.
The Matlab API yields to

\vspace{0.2cm}
\begin{lstlisting}[style=nonumbers]
    import org.opensim.modeling.*  %
    osimModel = Model(filenameOsimModel);
    osimModel.initSystem();
    ikTool = InverseKinematicsTool(setupFile);
    ikTool.setModel(osimModel);
    ikTool.setMarkerDataFileName(filenameTrc);
    ikTool.setOutputMotionFileName(outputMotionFilename);
    ikTool.run();
\end{lstlisting}
The output of the system is a motion (.mot) file (i.e., outputMotionFilename)
 containing the joint angles $\bm q$ of the model.

\chapter{Simultaneous Human Dynamics and State Estimation}
 \label{dyn&stateEstim}

Consider the system \eqref{eq:systemEq} in Chapter
 \ref{Chapter_estimation_problem}:
\begin{eqnarray} \label{eq:systemEq_Appendix}
    \begin{bmatrix}
    \bm Y(\bm{q}, \dot{\bm{q}}) \\ \bm D(\bm{q}, \dot{\bm{q}}) \\
     \end{bmatrix} \bm d +
     \begin{bmatrix} \bm b_Y(\bm{q}, \dot{\bm{q}})
    \\ \bm b_D(\bm{q}, \dot{\bm{q}})
    \end{bmatrix} =
    \begin{bmatrix} \bm y\\
     \bm 0
     \end{bmatrix}~, \qquad
rank\left(\begin{bmatrix}
    \bm Y(\bm{q}, \dot{\bm{q}}) \\ \bm D(\bm{q}, \dot{\bm{q}}) \\
     \end{bmatrix}\right) = d~,
\end{eqnarray}
whose MAP Gaussian solution is represented by Equations \eqref{MAP_solution},
 such that
\begin{subequations} \label{MAP_solution_Appendix}
\begin{eqnarray}
\label{eq:sigma_dgiveny_Appendix}
\bm {\Sigma}_{d|y} &=& \left(\xoverline{ \bm {\Sigma}}_D^{-1} + \bm Y^\top \bm
 {\Sigma}_{y|d}^{-1}\bm Y\right)^{-1}~, 
\\
\label{eq:mu_dgiveny_Appendix} 
\bm d &=& \bm {\Sigma}_{d|y} \left[ \bm Y^\top \bm
 {\Sigma}_{y|d}^{-1} (\bm y-\bm b_Y) + \xoverline{\bm {\Sigma}}_D^{-1}
  \xoverline{\bm {\mu}}_D\right]~.
\end{eqnarray} 
\end{subequations}
The solution \eqref{eq:mu_dgiveny_Appendix} is obtained by assuming the $\bm q$
 and $\dot{\bm q}$ without uncertainty.  It is well-known that in practical
  acquisitions, the state is affected by statistical noise and other
   inaccuracies.  Thus, this Appendix deals with the way to include the state
    $\bm x =(\bm q, \dot{\bm q})$ in the estimation problem.
This is not a trivial task and complications arise from the fact that 
 \eqref{eq:systemEq_Appendix} is not linear in $\bm x$.  
If $\bar {\bm d}$ and $\xoverline {\bm x}$ are the mean on the vector $\bm d$
 and $\bm x$, respectively, thus the first order approximation around these
  quantities of  \eqref{eq:systemEq_Appendix} is such that

\begin{eqnarray} \label{eq:systemEq_EKF_tmp} 
\begin{bmatrix}
\bm b_Y(\bar {\bm x}) \\ 
\bm b_D(\bar {\bm x})
\end{bmatrix}
+
\begin{bmatrix}
\bm Y(\bar {\bm x})  \\
\bm D(\bar {\bm x}) 
\end{bmatrix} \bm d 
+
\begin{bmatrix}
\partial \bm b_Y \left(\bar {\bm d}, \bar {\bm x}\right) \\
\partial \bm b_D \left(\bar {\bm d}, \bar {\bm x}\right)
\end{bmatrix} (\bm x - \bar {\bm x})
  = \begin{bmatrix}
\bm y \\ 
\bm 0
\end{bmatrix}~,
\end{eqnarray}
where
\begin{eqnarray}
\label{deriv_bY}
\bm b_Y  \big(\bar {\bm x}\big) &=& \frac{\partial} {\partial {\bm x}}
  \bm b_Y( {\bm x}) \Big|_{\bm x = \bar {\bm x}}~,\\
\label{deriv_bD}
\bm b_D  \big(\bar {\bm x}\big) &=& \frac{\partial} {\partial {\bm x}}
  \bm b_D( {\bm x}) \Big|_{\bm x = \bar {\bm x}}~,\\
\label{deriv_Y}
\bm Y  \big(\bar {\bm x}\big) &=& \frac{\partial} {\partial {\bm x}}
  \bm Y( {\bm x}) \Big|_{\bm x = \bar {\bm x}}~,\\
\label{deriv_D}
\bm D  \big(\bar {\bm x}\big) &=& \frac{\partial} {\partial {\bm x}}
  \bm D( {\bm x}) \Big|_{\bm x = \bar {\bm x}}~,\\
\label{deriv_bY_double}
\partial \bm b_Y \left(\bar {\bm d}, \bar {\bm x}\right) &=& \frac{\partial}
 {\partial {\bm x}} \Big[  \bm Y( {\bm x}) \bar {\bm d} + \bm b_Y( {\bm x})
  \Big]\Big|_{\bm x = \bar {\bm x}}~,\\
\label{deriv_bD_double}
\partial \bm b_D \left(\bar {\bm d}, \bar {\bm x}\right) &=& \frac{\partial}
 {\partial {\bm x}} \Big[\bm D( {\bm x}) \bar {\bm d} + \bm b_D( {\bm x})
  \Big]\Big|_{\bm x = \bar {\bm x}}~.
\end{eqnarray}
The system \eqref{eq:systemEq_EKF_tmp} could be rearranged in the same
 structure of system \eqref{eq:systemEq_Appendix}:
\begin{eqnarray} 
\begin{bmatrix} \label{eq:systemEq_EKF}
\bm Y(\bar {\bm x})  & \partial \bm b_Y \left(\bar {\bm d}, \bar {\bm x}\right)
 \\
\bm D(\bar {\bm x})  & \partial \bm b_D \left(\bar {\bm d}, \bar {\bm x}\right)
\end{bmatrix} 
\begin{bmatrix}
\bm d  \\
\bm x
\end{bmatrix} 
+ \begin{bmatrix}
\bm b_Y(\bar {\bm x}) - \partial \bm b_Y \left(\bar {\bm d}, \bar {\bm
 x}\right) \bar {\bm
 x} \\
\bm b_D(\bar {\bm x}) - \partial \bm b_D \left(\bar {\bm d}, \bar {\bm
 x}\right) \bar {\bm
 x}
\end{bmatrix}
  = \begin{bmatrix}
\bm y \\ 
\bm 0
\end{bmatrix}~.
\end{eqnarray}
The system \eqref{eq:systemEq_EKF} becomes the new system to be solved in the
 MAP domain.  The mean and the covariance of the PDF $p(\bm d,\bm x|\bm y)$ can
  be obtained as in \eqref{MAP_solution} by replacing 
  
\begin{eqnarray*}
\bm Y (\bm q, \dot{\bm q}) & \leftrightarrow & \begin{bmatrix}
\bm Y(\bar {\bm x})  & \partial \bm b_Y \left(\bar {\bm d}, \bar {\bm x}\right)
\end{bmatrix}~, \\
\bm D (\bm q, \dot{\bm q}) & \leftrightarrow & \begin{bmatrix}
\bm D(\bar {\bm x})  & \partial \bm b_D \left(\bar {\bm d}, \bar {\bm x}\right)
\end{bmatrix}~, \\
\bm b_Y(\bm q, \dot{\bm q}) & \leftrightarrow & \begin{bmatrix}
\bm b_Y(\bar {\bm x}) - \partial \bm b_Y \left(\bar {\bm d}, \bar {\bm
 x}\right) \bar {\bm x} \end{bmatrix}~, \\
\bm b_D(\bm q, \dot{\bm q}) &\leftrightarrow &\begin{bmatrix}
\bm b_D(\bar {\bm x}) - \partial \bm b_D \left(\bar {\bm d}, \bar {\bm
 x}\right) \bar {\bm x}
\end{bmatrix}~, \\
\bm \mu_d &\leftrightarrow & \begin{bmatrix}\bm \mu_d & \bm
 \mu_x\end{bmatrix}^T~,\\
\bm \Sigma_d &\leftrightarrow & \begin{bmatrix} \bm \Sigma_d & \bm 0 \\ \bm 0 &
  \bm \Sigma_x\end{bmatrix}~,
\end{eqnarray*}
being $\bm x \sim \mathcal N(\bm \mu_{\bm x}, \bm \Sigma_{\bm x})$ the Gaussian
 distribution for the state $\bm x$.
 Details on how to compute \eqref{deriv_bY_double} and \eqref{deriv_bD_double}
  derivatives are shown in Section $5$A of \citep{NoriNav2015}.

\clearpage\null
\clearpage\null
\end{appendices}

\begin{spacing}{0.9}
    
\bibliography{References/ms} %
\bibliographystyle{plainnat}

\vspace{2cm}
\begin{quotation}
\noindent\emph{
The man who doesn't read good books has no advantage over the man who can't
 read them.
}
    \begin{flushright}
        Mark Twain
    \end{flushright}
\end{quotation}
\end{spacing}

\end{document}